\pdfoutput=1
\documentclass[oneside]{amsbook}

\usepackage{amsmath, amssymb, amsthm, mathrsfs}

\usepackage{graphicx}
\usepackage{caption}
\usepackage{subcaption} 
\usepackage{tikz}
\usetikzlibrary{
	positioning, 
	shapes.geometric, 
	arrows.meta, 
	fit, 
	calc, 
	decorations.pathmorphing,
	decorations.pathreplacing,
	bending
}
\usepackage{tikz-3dplot}

\usepackage{booktabs}
\usepackage{tabulary}
\usepackage{tabularx}

\usepackage{enumitem}
\usepackage{microtype}
\usepackage{geometry}

\usepackage[T1]{fontenc}
\usepackage[utf8]{inputenc}

\usepackage[colorlinks=true, linkcolor=blue, citecolor=blue, urlcolor=blue]{hyperref}

\theoremstyle{definition}

\theoremstyle{remark}

\title[Theoretical Foundations for Semantic Cognition]{Theoretical Foundations for Semantic Cognition \\ in Artificial Intelligence}
\author{Sebastian Dumbrava}
\date{}

\begin{document}
\begin{abstract}
	This monograph presents a modular cognitive architecture for artificial intelligence grounded in the formal modeling of belief as structured semantic state. Belief states are defined as dynamic ensembles of linguistic expressions embedded within a navigable manifold, where operators enable assimilation, abstraction, nullification, memory, and introspection. Drawing from philosophy, cognitive science, and neuroscience, we develop a layered framework that enables self-regulating epistemic agents capable of reflective, goal-directed thought. At the core of this framework is the epistemic vacuum: a class of semantically inert cognitive states that serves as the conceptual origin of belief space.  From this foundation, the Null Tower arises as a generative structure recursively built through internal representational capacities. The theoretical constructs are designed to be implementable in both symbolic and neural systems, including large language models, hybrid agents, and adaptive memory architectures. This work offers a foundational substrate for constructing agents that reason, remember, and regulate their beliefs in structured, interpretable ways.
\end{abstract}

	\frontmatter
	\maketitle
	\tableofcontents
	\chapter*{Preface}

The landscape of artificial intelligence is rapidly evolving, moving beyond pattern recognition and prediction towards systems capable of reasoning, reflection, and autonomous action in complex environments. Central to this transition is the concept of belief---how an agent represents, structures, updates, and utilizes its internal understanding of the world and itself. Yet, contemporary approaches often treat belief implicitly, as emergent properties of large models or as symbolic tokens in brittle logical systems. This monograph argues for a different path: a principled, structured, and dynamic theory of belief grounded in the very fabric of cognition.

We embark on an ambitious project: to lay the \textit{Theoretical Foundations for Semantic Cognition in Artificial Intelligence}. Our central thesis is that belief in advanced AI systems should not be an amorphous substrate but a structured, navigable space---a semantic manifold ($\Phi$)---composed of linguistic ensembles, organized by function and abstraction, and governed by well-defined epistemic dynamics. This work seeks to construct such a framework from first principles, drawing inspiration from cognitive science, philosophy of mind, neuroscience, mathematics, and computer science, but ultimately charting its own course.

Building from foundational concepts, this work systematically develops the architecture, exploring the structure of the semantic space, the dynamics governing belief evolution, mechanisms for memory and regulation, meta-cognitive capabilities, learning, and multi-agent interactions. The appendices provide supplementary details including a Glossary, formal operator sketches, geometric considerations, implementation strategies, comparisons to other architectures, and extended motivations.

This monograph is intended for researchers and students in artificial intelligence, cognitive science, philosophy of mind, and related fields who are interested in foundational models of belief, reasoning, and consciousness. It aims to provide not just a collection of ideas, but an integrated architecture---a substrate for thinking about thinking itself. The work presented here is extensive but by no means exhaustive; it is offered as a map of a new territory and an invitation to explore it further.
	
	\mainmatter
	\chapter*{List of Notation}

\subsection*{Spaces, Structures, and Belief Components}
\begin{itemize}
	\item[$S$:] Space of raw observations or input signals.
	\item[$\Phi$:] The semantic state space; the space of all structured belief states available to the agent.
	\item[$\Phi^{[\theta]}$:] Parameterized semantic state space, determined by configuration $\theta$.
	\item[$\phi$:] A specific belief state within $\Phi$; a structured ensemble $\phi = \{\varphi_{1}, \varphi_{2}, \ldots\}$.
	\item[$\varphi_i$:] An individual linguistic expression (sentence, fragment, proposition) within a belief state $\phi$.
	\item[$\Phi^{(k)}$:] The subspace of $\Phi$ containing belief states at abstraction level $k$.
	\item[$\Sigma$:] A semantic sector; a functional subregion of $\Phi$ (e.g., $\Sigma_{perc}, \Sigma_{plan}, \Sigma_{narr}, \Sigma_{refl}$).
	\item[$\Sigma^{(k)}$:] The subset of belief states belonging to sector $\Sigma$ at abstraction level $k$, i.e., $\Sigma^{(k)} = \Sigma \cap \Phi^{(k)}$.
	\item[$(\Sigma, k)$:] Semantic coordinates specifying a belief's position by sector and abstraction level.
	\item[$A$ (Set):] The set of executable actions available to the agent.
	\item[$\mathcal{A}_a$:] The activation basin associated with action $a \in A$; a subset of $\Phi$ where preconditions for $a$ are met ($\mathcal{A}_a \subset \Phi$).
	\item[$S_a$:] Suppression surface/region for action $a$; subset of $\Phi$ where action $a$ is actively inhibited.
	\item[$\Phi_{input}$:] Generic representation for input to Assimilation $A$ (e.g., $X(s)$ or $\Phi_{retrieved}$).
	\item[$\Phi_{memory}$:] Subset of $\Phi$ representing stored, potentially inactive, beliefs (memory).
	\item[$\Phi_{active}$ / $\phi_{current}$:] Subset of $\Phi$ representing the currently active/attended belief state.
	\item[$\phi_{introspective}$:] Input to Meta-Assimilation $M$, containing meta-information about $\phi$.
	\item[$\phi_{query}$:] Retrieval cue generated by $Q$.
	\item[$\phi_{retrieved}$:] Retrieved belief states constructed by $R$.
\end{itemize}

\subsection*{Null Tower and Semantic Origins}
\begin{itemize}
	\item[$\Omega$:] The epistemic vacuum; the subsets of belief states in $\Phi$ devoid of active semantic content.
	\item[$\Omega^{(0)}$:] The irreducible null stratum; base layer of $\Omega$.
	\item[$\omega$:] An irreducible null state, $\omega \in \Omega^{(0)}$.
	\item[$\omega^{(k)}$:] The abstraction of a null state $\omega$ at level $k$; $\omega^{(k)} = \Lambda^{k}(\omega)$.
	\item[$\omega^{(\infty)}$:] The semantic singularity associated with $\omega$; $\lim_{k\rightarrow\infty} \omega^{(k)}$.
\end{itemize}

\subsection*{Core Cognitive Operators}
\begin{itemize}
	\item[$X$:] The observation encoder function ($X: S \rightarrow \Phi$).
	\item[$\Lambda$ / $\Lambda^k$:] Upward abstraction operator ($\Lambda^k: \Phi^{(i)} \rightarrow \Phi^{(i + k)}$). Includes $k$ compositions of $\Lambda$.
	\item[$V$ / $V^k$:] Downward elaboration operator ($V^k: \Phi^{(i)} \rightarrow \Phi^{(i - k)}$). Includes $k$ compositions of $V$.
	\item[$A$:] Assimilation operator ($A: \Phi \times \Phi_{input} \rightarrow \Phi$). Variants include $A_{elab}, A_{corr}, A_{abs}$.
	\item[$M$:] Meta-Assimilation operator ($M: \Phi \times \Phi_{introspective} \rightarrow \Phi$).
	\item[$N_t$:] Nullification operator (temporal decay $N_t: \Phi \rightarrow \Phi$). May be sector-specific $N^{\Sigma}_t$. Requires persistence threshold $\delta$.
	\item[$K$:] Total Annihilation operator ($K: \Phi \rightarrow \Omega$).
	\item[$K_{\Sigma}$:] Sector-specific Annihilation operator ($K_{\Sigma}: \Phi \rightarrow \Phi$).
	\item[$D$:] Spontaneous Drift operator ($D: \Phi \rightarrow \Phi$).
	\item[$Q$:] Query Function ($Q: \Phi_{active} \rightarrow \Phi_{query}$).
	\item[$R$:] Retrieval operator ($R: \mathcal{P}(\Phi_{memory}) \times \Phi_{query} \rightarrow \Phi$; $\mathcal{P}(\Phi_{memory})$ the powerset).
	\item[$\pi$:] (Execution Policy) The execution interface policy mapping belief states to actions ($\pi: \Phi \rightarrow A \cup \{0\}$).
\end{itemize}

\subsection*{Metrics, Quantities, and State Properties}
\begin{itemize}
	\item[$d(\phi_1, \phi_2)$:] Semantic distance function between two belief states.
	\item[$\kappa(\phi)$:] Coherence measure within belief state $\phi$. Can be sector-specific $\kappa(\Sigma, \phi)$ or trajectory-specific $\kappa_T$.
	\item[$\lambda$:] Indicator for belief density, activation mass, or cognitive load ($L$). Note potential double use with decay rate.
	\item[$L$:] Cognitive Load (demand on resources).
	\item[$\epsilon$:] Semantic Effort (actively applied resource).
	\item[$E_{total}$:] Total available effort capacity.
	\item[$a_i$:] Anchoring score indicating resistance of $\varphi_i$ to nullification $N_t$.
	\item[$d_i(t)$:] Persistence function for expression $\varphi_i$ over time $t$ (used in Nullification $N_t$).
	\item[$T_{1/2}(\varphi_i)$:] Semantic half-life of expression $\varphi_i$ under $N_t$.
	\item[$\delta$:] Persistence threshold (used in $N_t$).
	\item[$\tau_a$:] Action activation threshold (used with $\mathcal{A}_a$).
	\item[$\alpha_i(t)$:] Time-varying activation signal for semantic sector $\Sigma_i$. $\vec{\alpha}(t)$ is the vector.
	\item[$I_j$:] Generic notation for an internal monitoring function (introspection).
\end{itemize}

\subsection*{Regulation, Trajectories, and Identity}
\begin{itemize}
	\item[$\sim_{gauge}$:] Semantic gauge equivalence relation.
	\item[\mbox{$[\phi]_{gauge}$}:] Gauge equivalence class of state $\phi$.
	\item[\mbox{$[\omega \rightarrow \omega^{(\infty)}]$}:] Epistemic Axis for orientation.
	\item[$\vec{v}_{\omega}$:] Direction vector associated with an epistemic axis.
	\item[$\pi_{\omega}(\phi)$:] Projection of $\phi$ (relative to $\omega$) onto axis $\vec{v}_{\omega}$.
	\item[$\theta(\phi, \vec{v}_{\omega})$:] Angular deviation of $\phi$ (relative to $\omega$) from axis $\vec{v}_{\omega}$. (Note: distinguishes angle $\theta$ from parameter $\theta$).
	\item[$r(\phi)$:] Residual offset of $\phi$ from an epistemic axis.
	\item[$\gamma(t)$:] Belief state trajectory over time ($\gamma: [t_0, t_1] \rightarrow \Phi$). Includes $\gamma_{sim}, \gamma_{pred}$.
	\item[$F$:] Vector field governing belief evolution ($\frac{d\gamma}{dt} = F(\gamma(t))$).
	\item[$\delta_a(\phi_{refl})$:] Reflective diagnostic function for action $a$.
	\item[$\vec{\eta}$ / $\vec{\eta}^{[\theta]}$:] Epistemic identity vector (potentially parameter-dependent).
	\item[$\pi_{effort}$:] Effort allocation policy.
	\item[$\pi_{regulate}$:] Meta-control regulatory policy.
	\item[$\theta$:] Architecture parameter vector. Components include $\mu$ (Memory Horizon), $\delta_{profile}$ (Decay Profile), $\tau$ (Trajectory Persistence), $\rho$ (Representation Mode), $\eta_{type}$ (Identity Model), $C$ (Context Scope), $\Gamma_{type}$ (Goal Decomposition Model), $O$ (Observation Modalities).
\end{itemize}
	
	\part{Context and Foundations}
	\label{part:context_and_foundations}


\chapter{Introduction}
\label{chap:Introduction}

The quest to create artificial intelligence often converges on the challenge of replicating or simulating aspects of cognition that remain deeply enigmatic even in humans: belief, understanding, reflection, and perhaps even consciousness. While current AI systems, particularly large language models, demonstrate remarkable capabilities in pattern matching, generation, and interaction, they often lack a coherent, stable, and introspectible foundation for their internal states. Beliefs can appear ephemeral, contradictory, or ungrounded; reasoning can seem brittle or superficial.

Contemporary approaches frequently treat belief implicitly, as emergent properties of large models or as symbolic tokens in potentially fragile logical systems. This monograph contends that a fundamental limitation lies in the absence of a robust theoretical framework for what it means for an artificial agent to believe in a structured, dynamic, and embodied way. We argue for a different path: a principled, structured, and dynamic theory of belief grounded in the fabric of cognition itself. We embark on an ambitious project: to lay the Theoretical Foundations for Semantic Cognition in Artificial Intelligence.

Our central thesis is that belief in advanced AI systems should not be an amorphous substrate but a structured, navigable space---a semantic manifold, denoted $\Phi$---composed of interpretable linguistic ensembles, possessing geometry and governed by dynamics. This space is envisioned not as a flat collection of facts, but as a dynamic landscape possessing internal geometry, modular structure, and inherent temporal dynamics, where beliefs are structured ensembles of linguistic expressions ($\phi = \{\varphi_{i}\}$).

This work undertakes the development of such a framework from first principles. Crucially, the architecture proposed herein is deeply motivated by converging insights from philosophy, cognitive psychology, and neuroscience, as detailed in the subsequent chapters of this Part (Chapters~\ref{chap:PhilosophicalMotivations}-\ref{chap:NeuroscientificMotivations}). These disciplines provide both inspiration and constraint, suggesting that principles like hierarchical abstraction, functional modularity, dynamic memory, and reflective self-regulation are fundamental to robust intelligence, whether natural or artificial. Understanding this multi-faceted grounding illuminates the rationale behind the specific structures and mechanisms developed throughout the monograph.

The journey through this framework will unfold systematically:

\begin{itemize}
	\item We will first establish the foundations of the semantic manifold ($\Phi$), including its parameterization for different agent architectures ($\Phi^{[\theta]}$), and explore its conceptual origins in the epistemic vacuum ($\Omega$) via the Null Tower model (Part~\ref{part:semantic_foundations}, Chapters~\ref{chap:SemanticStateSpace}-\ref{chap:ParameterizedArchitectures}).
	\item We will then delve into structuring this manifold, examining how beliefs are grounded (Part~\ref{part:structuring_belief}, Chapter~\ref{chap:GroundingSemanticBelief}), organized vertically by abstraction levels ($\Phi^{(k)}$ via Semantic Scaling, Chapter~\ref{chap:SemanticScaling}), partitioned functionally into Semantic Sectors ($\Sigma$, Chapter~\ref{chap:SemanticSectors}), and unified within a coherent Semantic Geometry (Chapter~\ref{chap:SemanticGeometry}). The structure of Semantic Memory as part of $\Phi$ will also be defined (Part~\ref{part:semantic_memory}, Chapter~\ref{chap:StructureSemanticMemory}).
	\item The core dynamics governing belief evolution ($A, N_t, K, D$, Part~\ref{part:epistemic_dynamics}, Chapters~\ref{chap:Assimilation}-\ref{chap:SpontaneousDrift}) and the mechanisms for memory access (Querying $Q$, Retrieval $R$, Part~\ref{part:semantic_memory}, Chapters~\ref{chap:QueryingBeliefSpace}-\ref{chap:RetrievalOperator}) will be detailed, explaining how beliefs are formed, integrated, decay, are retrieved, and potentially erased. Retrieved memories are integrated via Assimilation (Part~\ref{part:semantic_memory}, Chapter~\ref{chap:MemoryIntegration}), and the functional role of memory dynamics is explored (Part~\ref{part:semantic_memory}, Chapter~\ref{chap:MemoryDynamics}).
	\item We explore mechanisms for Regulation and Control (Part~\ref{part:regulation_and_control}), including maintaining Semantic Orientation via an internal compass (Chapter~\ref{chap:SemanticOrientation}), understanding Semantic Gauge ($\sim_{gauge}$) equivalence (Chapter~\ref{chap:SemanticGauge}), the emergence of Epistemic Identity ($\vec{\eta}^{[\theta]}$, Chapter~\ref{chap:EpistemicIdentity}), and managing Cognitive Load ($\lambda$, Part~\ref{part:meta_cognition}, Chapter~\ref{chap:CognitiveLoad}) and Semantic Effort ($\epsilon$, Part~\ref{part:meta_cognition}, Chapter~\ref{chap:SemanticEffort}) via allocation policies (Part~\ref{part:meta_cognition}, Chapter~\ref{chap:SemanticFocus}).
	\item The connection to the external world through Embodiment and Action (Part~\ref{part:meta_cognition}) is addressed via Semantic Execution (Chapter~\ref{chap:SemanticExecution}), Activation Basins ($\mathcal{A}_a$, Chapter~\ref{chap:ActivationBasins}) and Embodied Simulation (Chapter~\ref{chap:EmbodiedSimulation}).
	\item Advanced capabilities involving Meta-Cognition (Part~\ref{part:meta_cognition}) are formalized, including Meta-Assimilation ($M$, Chapter~\ref{chap:MetaAssimilation}), Meta-Introspection (Chapter~\ref{chap:MetaIntrospection}), and Trajectory Awareness ($\gamma(t)$, Chapter~\ref{chap:TrajectoryAwareness}).
	\item We consider how agents might acquire these capabilities through Learning and Adaptation (Part~\ref{part:learning_and_adaptation}), addressing the learning of structures (Chapter~\ref{chap:LearningSemanticStructures}), operators (Chapter~\ref{chap:LearningCognitiveOperators}), policies (Chapter~\ref{chap:LearningRegulatoryPolicies}), and potentially the architecture itself ($\theta$, Chapter~\ref{chap:ArchitecturalAdaptation}).
	\item Finally, the framework is extended to Social Cognition (Part~\ref{part:social_cognition}), modeling Theory of Mind (Chapter~\ref{chap:ModelingOtherAgents}) and multi-agent communication and alignment (Chapter~\ref{chap:CommunicationAlignment}).
\end{itemize}

This monograph is intended for researchers and students in artificial intelligence, cognitive science, philosophy of mind, and related fields interested in foundational models of belief, reasoning, and consciousness. It aims to provide not just a collection of ideas, but an integrated architecture---a conceptual toolkit and substrate for thinking about thinking itself. The work presented here is offered as a map of a new territory and an invitation to explore it further (Part~\ref{part:conclusion}, Chapters~\ref{chap:FutureOutlook}-\ref{chap:Conclusion}).

\subsection*{Chapter Summary}
This chapter introduces the central challenge addressed by the monograph: the lack of robust theoretical foundations for belief in contemporary AI systems. Current approaches often result in belief states that are ephemeral, ungrounded, or brittle. To overcome these limitations, this work proposes a principled, structured, and dynamic theory of belief grounded in cognition. The core thesis is that belief should be modeled not implicitly, but as a structured, navigable \textit{semantic manifold}, denoted $\Phi$, composed of interpretable linguistic ensembles $\{\varphi_i\}$ and governed by well-defined dynamics and geometry. The monograph aims to develop this framework from first principles, drawing motivation from philosophy, cognitive psychology, and neuroscience. A roadmap outlining the structure of the work is presented, covering the foundations of $\Phi$ and the Null Tower (Part~\ref{part:semantic_foundations}), the structuring of belief via grounding, scaling ($\Phi^{(k)}$), sectors ($\Sigma$), and geometry (Part~\ref{part:structuring_belief}), epistemic dynamics ($A$, $N_t$, $K$, $D$) (Part~\ref{part:epistemic_dynamics}), semantic memory (Part~\ref{part:semantic_memory}), regulation and control (Orientation, Gauge $\sim_{gauge}$, Identity $\vec{\eta}^{[\theta]}$) (Part~\ref{part:regulation_and_control}), embodiment and action (Part~\ref{part:embodiment_and_action}), meta-cognition ($M$, Load $\lambda$, Effort $\epsilon$) (Part~\ref{part:meta_cognition}), learning and adaptation (Part~\ref{part:learning_and_adaptation}), and social cognition (Part~\ref{part:social_cognition}). This work is intended to provide an integrated conceptual toolkit for researchers and students interested in foundational models of belief and reasoning in artificial intelligence.


\chapter{Philosophical Motivations}
\label{chap:PhilosophicalMotivations}

The study of artificial intelligence has long intersected with fundamental philosophical questions about meaning, representation, and the nature of thought. While early symbolic systems grappled with semantic grounding, and contemporary statistical models often lack interpretability, the framework presented here seeks a path grounded in the conviction that cognition involves structured evolution within a semantic manifold, $\Phi$. This chapter outlines the key philosophical motivations informing the design of this framework, drawing inspiration from diverse traditions while charting its own course.

\section{From Symbolic AI to Semantic Geometry}

Classical Artificial Intelligence often modeled thought as the manipulation of discrete symbols according to formal rules. While powerful, this approach struggled to capture the fluidity, context-sensitivity, and grounding inherent in natural cognition. Conversely, purely connectionist approaches, while adept at pattern recognition, often obscure the internal structure and meaning of representations. The semantic manifold framework attempts to bridge this gap. It embraces the structured, compositional nature of thought, reminiscent of symbolic systems, by defining belief states $\phi$ as ensembles of interpretable linguistic expressions $\{\varphi_i\}$. However, it moves beyond static symbol manipulation by embedding these expressions within a dynamic geometric space equipped with operators (like assimilation $A$, elaboration $V$, nullification $N_t$, annihilation $K$) and metrics (like distance $d$ or load $\lambda$), allowing for continuous evolution and regulation. This aims to capture both the structure and the process of thinking, directly motivating the geometric and operator-based definition of $\Phi$.

\section{The Linguistic Basis of Thought}

A foundational premise is that structured belief, at least in agents interacting through language, is fruitfully modeled using internal linguistic ensembles. This resonates with philosophical traditions emphasizing the symbolic and compositional nature of cognition (e.g., Frege, Russell, Fodor's Language of Thought). Language is not merely an output modality but potentially the medium in which complex thought occurs. This linguistic basis provides semantic transparency---internal states ($\phi$) have a structure amenable to interpretation by the agent itself (via meta-cognition) and potentially by external observers. Yet, by placing these ensembles within a dynamic manifold $\Phi$, the framework avoids the rigidity of purely logical systems, acknowledging that meaning arises also through relationships, context, and continuous transformation. This motivates the core representation choice of $\phi = \{\varphi_i\}$ and the inclusion of meta-cognitive capabilities.

\section{The Null Tower and the Genesis of Meaning}

How does structured knowledge emerge from a state devoid of semantic content? The framework addresses this via the concept of the epistemic vacuum, $\Omega$, and the Null Tower construction. Starting from minimal, unstructured null states ($\omega \in \Omega^{(0)}$), belief structures are recursively built via abstraction operators ($\Lambda$). This provides a generative account of cognitive structure originating from internal representational capacities, rather than solely from external input. The limit of this recursive abstraction, a semantic singularity ($\omega^{(\infty)}$), serves as a conceptual counterpart to philosophical notions of pure form or invariant structures underlying variable expressions (e.g., Platonic forms, Saussurean langue, perhaps aspects of Kantian transcendental apperception). Within this framework, however, $\omega^{(\infty)}$ is defined operationally as a limit point within the dynamics of $\Phi$, motivating the specific constructs of $\Omega$, $\omega$, $\Lambda$, $\omega^{(k)}$, and $\omega^{(\infty)}$.

\section{Abstraction and the Topology of Concepts}

Abstraction ($\Lambda$) is modeled not merely as simplification but as a transformative process of semantic compression and conceptual elevation, moving belief states between layers $\Phi^{(k)}$. This aligns with philosophical views where forming universals from particulars is a core aspect of conceptual genesis (e.g., Hegel, Peirce). The hierarchical structure $\Phi = \bigcup_{k} \Phi^{(k)}$ and the semantic distance metric $d(\phi_1, \phi_2)$ provide a topological interpretation of conceptual relationships. Meaning becomes related to proximity and structure within this navigable space. The interplay between abstraction ($\Lambda$) and elaboration ($V$) allows movement between general principles and specific instances, crucial for flexible reasoning, thus justifying the operators $\Lambda$, $V$, the layered structure $\Phi^{(k)}$, and the need for a semantic metric $d$.

\section{Semantic Gauge Equivalence and the Nature of Identity}

The semantic gauge relation ($\phi_1 \sim_{gauge} \phi_2$) formalizes the idea that different internal representations can be functionally equivalent. This resonates with functionalist philosophies of mind and structural realism, where causal roles or relational structures matter more than specific implementation details. A gauge class $[\phi]_{gauge}$ represents states that are indistinguishable under the framework's operators and dynamics. This has implications for epistemic identity ($\vec{\eta}$), suggesting that continuity may reside in the preservation of functional equivalence class membership across time, rather than the persistence of exact representational tokens. This philosophical stance motivates the formal introduction of the $\sim_{gauge}$ relation and its role in defining a flexible notion of identity $\vec{\eta}$.

\section{Orientation and Epistemic Agency}

How can agents regulate their evolving beliefs coherently? Semantic Orientation, derived from the structure of the Null Tower (epistemic axes $[\omega \rightarrow \omega^{(\infty)}]$), provides an internal "compass". Mechanisms like projection ($\pi_{\omega}(\phi)$) and deviation measurement ($\theta(\phi, \vec{v}_{\omega})$) allow the agent to assess its alignment with principled abstraction trajectories or core values. This formalizes a notion of epistemic agency as regulated movement within belief space. Being an agent involves not just holding beliefs, but navigating $\Phi$ intentionally, maintaining coherence ($\kappa$), and preserving identity ($\vec{\eta}$) over time. This aligns with phenomenological concepts of situatedness and reflective self-positioning, motivating the introduction of the Semantic Orientation mechanisms built upon the Null Tower structure ($[\omega \rightarrow \omega^{(\infty)}]$, $\pi_\omega(\phi)$, $\theta(\phi, \vec{v}_\omega)$) and linked to coherence $\kappa$ and identity $\vec{\eta}$.

\section{Toward a Philosophy of Cognitive Manifolds}

Taken together, the semantic state space $\Phi$, the Null Tower, the dynamic operators, and the regulatory mechanisms constitute a proposed philosophy of cognitive manifolds. This approach integrates representational transparency, dynamic evolution, structural organization (geometry, topology), and reflective self-regulation. It models cognition as geometric evolution within a structured space of meaning, offering a potential bridge between symbolic, connectionist, and embodied approaches to understanding thought, whether artificial or natural.


\subsection*{Chapter Summary}
This chapter outlines the core philosophical motivations underpinning the semantic manifold framework. It contrasts the proposed approach with classical symbolic AI and purely connectionist models, arguing for a middle path that captures both the structured nature of thought (via linguistic ensembles $\{\varphi_i\}$) and its dynamic, context-sensitive aspects (via geometric embedding in $\Phi$ with operators like $A$, $V$, $N_t$, $K$). A key premise is the linguistic basis of structured belief, providing transparency and compositionality. The framework addresses the genesis of meaning through the Null Tower concept, where cognitive structure emerges recursively via abstraction ($\Lambda$) from an epistemic vacuum ($\Omega$), defining limit points ($\omega^{(\infty)}$). Abstraction ($\Lambda$) and elaboration ($V$) are presented as core operators enabling navigation between specific instances and general principles across semantic layers ($\Phi^{(k)}$). Semantic gauge equivalence ($\sim_{gauge}$) is introduced to formalize the notion that functional role, not representational form, defines epistemic states, linking to functionalism and providing a flexible basis for identity ($\vec{\eta}$). Finally, Semantic Orientation, derived from Null Tower axes $[\omega \rightarrow \omega^{(\infty)}]$, is proposed as a mechanism for epistemic agency, allowing agents to regulate their belief states ($\kappa$, $\vec{\eta}$) within the manifold. Together, these concepts form a philosophy of cognitive manifolds, integrating representation, structure, dynamics, and regulation.

\chapter{Psychological Motivations}
\label{chap:PsychologicalMotivations}

To model belief in artificial agents effectively, particularly those intended to exhibit human-like flexibility, understanding the principles of human cognition is paramount. Cognitive psychology and developmental psychology offer rich sources of inspiration and empirical grounding. This chapter articulates the psychological motivations for the semantic manifold framework, aligning its core constructs and dynamics with established theories and findings regarding memory, reasoning, abstraction, attention, and self-regulation. The framework aims not just for computational power, but for cognitive plausibility.

\section{Belief States as Structured Memory Ensembles}

The framework's representation of belief states ($\phi$) as structured ensembles of linguistic expressions ($\varphi_i$) within a space $\Phi$ reflects key insights about human memory:
\begin{itemize}
	\item \textbf{Constructive Memory:} Research dating back to Bartlett shows that memory is not a passive store but an active, reconstructive process shaped by context and existing schemas. The dynamic nature of $\Phi$, updated via operators like Assimilation ($A$) and Elaboration ($V$), directly captures this constructive aspect.
	\item \textbf{Working Memory Models:} Influential models like Baddeley \& Hitch's posit functionally distinct buffers (phonological loop, visuospatial sketchpad) coordinated by a central executive. The framework's Semantic Sectors ($\Sigma$) such as $\Sigma_{perc}$, $\Sigma_{plan}$, $\Sigma_{narr}$, $\Sigma_{refl}$ formalize this functional modularity within the belief space $\Phi$.
	\item \textbf{Structured Representations:} Cognitive architectures like ACT-R and connectionist models emphasize distributed but structured representations. Our conception of $\phi$ as an interconnected ensemble aligns with this, allowing for relationships and dependencies between belief fragments $\varphi_i$.
\end{itemize}
Thus, the very definition of belief states $\phi$ as dynamic, structured ensembles within $\Phi$, potentially organized by sectors $\Sigma$, is motivated by these established models of human memory structure and function.

\section{Abstraction and Developmental Generalization}

The stratification of the semantic manifold into layers of increasing abstraction ($\Phi^{(k)}$) and the operators ($\Lambda$, $V$) that move between them are directly motivated by cognitive development and concept formation:
\begin{itemize}
	\item \textbf{Developmental Stages:} Theories like Piaget's describe a progression from concrete sensorimotor operations to formal abstract thought. The abstraction operator ($\Lambda$) mirrors this capacity to generalize, compress information, and form higher-order concepts, facilitating movement from $\Phi^{(0)}$ upwards through the layers $\Phi^{(k)}$.
	\item \textbf{Category Learning:} Humans naturally organize experiences into hierarchical categories. The framework models this inductive process, where the $\Lambda$ operator groups specific instances $\{\varphi_i\} \in \Phi^{(k)}$ into more general representations in $\Phi^{(k+1)}$.
	\item \textbf{Schema Formation:} Abstract knowledge structures like scripts and schemas, which organize expectations and guide behavior, are formed through generalization over specific episodes---a process captured by the repeated application of $\Lambda$.
\end{itemize}
The framework's mechanisms for Semantic Scaling via $\Lambda$ and $V$ across layers $\Phi^{(k)}$ are therefore grounded in the fundamental human ability to abstract and generalize from experience.

\section{Assimilation, Nullification, and Constructive Forgetting}

The core dynamics governing belief change find strong parallels in psychological processes:
\begin{itemize}
	\item \textbf{Assimilation ($A$):} Directly analogous to Piaget's concept of assimilation, where new information is integrated into existing cognitive schemas (represented by $\phi$). The framework further differentiates modes like corrective ($A_{corr}$) and elaborative ($A_{elab}$) assimilation, reflecting how integration can involve conflict resolution or inference, mirroring psychological findings on belief updating.
	\item \textbf{Nullification ($N_t$):} Models the gradual decay of memory traces over time, consistent with findings from Ebbinghaus onwards. Importantly, $N_t$ incorporates anchoring ($a_i$) and selective decay, reflecting that forgetting is not uniform loss but a potentially adaptive process of pruning less relevant or unreinforced information, contributing to cognitive efficiency, much like theories of directed or motivated forgetting suggest.
	\item \textbf{Annihilation ($K$):} Reflects processes like cognitive dissonance reduction (Festinger), where significant contradictions lead to the revision or abrupt discarding of beliefs to restore internal consistency ($\kappa$), providing a formal mechanism analogous to psychological findings on radical belief change.
\end{itemize}
The dynamic operators $A$, $N_t$, and $K$ are thus proposed as formal analogues to core psychological processes of learning, forgetting, and belief revision.

\section{Cognitive Load and Effort Allocation}

The framework incorporates resource limitations central to psychological models of attention and performance:
\begin{itemize}
	\item \textbf{Cognitive Load ($\lambda$):} Defined within the framework based on belief state complexity and dynamics (potentially measured by $\lambda(\Phi, \phi)$), this directly corresponds to concepts from Cognitive Load Theory (Sweller), representing the burden on finite processing resources.
	\item \textbf{Semantic Effort ($\epsilon$):} Parallels Kahneman's concept of limited mental effort that must be actively allocated. The framework's effort allocation policies ($\pi_{effort}$) model the executive control processes responsible for directing attention and resources, aligning with psychological theories of attentional control.
\end{itemize}
The inclusion of metrics $\lambda$ and $\epsilon$, along with control policies like $\pi_{effort}$, aims to capture the resource-limited nature of cognition emphasized in psychology.

\section{Orientation, Reflection, and Metacognitive Control}

The framework's mechanisms for self-regulation are grounded in research on metacognition:
\begin{itemize}
	\item \textbf{Metacognitive Monitoring:} Concepts like coherence ($\kappa$), load ($\lambda$), and trajectory awareness ($\gamma(t)$) represent internal signals accessible via Meta-Introspection (producing $\phi_{introspective}$ integrated by $M$), analogous to the monitoring functions described in psychological models (e.g., Nelson \& Narens).
	\item \textbf{Metacognitive Control:} Semantic Orientation (using epistemic axes $[\omega \rightarrow \omega^{(\infty)}]$ and deviation $\theta$) provides a mechanism for goal-directed regulation. Regulatory policies ($\pi_{regulate}$) that trigger corrective actions based on monitored states mirror psychological theories of executive control and self-correction.
	\item \textbf{Reflection:} The reflective sector ($\Sigma_{refl}$) provides a dedicated space within $\Phi$ for explicit self-assessment and reasoning about one's own beliefs and processes, crucial for higher-level learning and adaptation, consistent with psychological studies of reflective thought.
\end{itemize}
These elements ($\kappa$, $\lambda$, $M$, Orientation, $\Sigma_{refl}$, $\pi_{regulate}$) aim to formalize the mechanisms underlying self-awareness and self-control observed in human cognition.

\section{Semantic Sectors as Functional Modularity}

The partitioning of $\Phi$ into sectors ($\Sigma$) aligns with the widely accepted view that cognition involves functionally specialized systems:
\begin{itemize}
	\item \textbf{Multiple Memory Systems:} The existence of distinct memory systems (declarative, procedural, episodic, semantic) supports the idea of functionally differentiated subspaces (sectors $\Sigma$) within the unified belief space $\Phi$.
	\item \textbf{Domain Specificity:} Different cognitive tasks activate distinct brain networks, suggesting specialized processing pathways that can be modeled as distinct semantic sectors $\Sigma$ within the framework.
	\item \textbf{Cognitive Modes:} The ability to dynamically shift focus between perception, deliberation, planning, and reflection corresponds to modulating activation ($\vec{\alpha}(t)$) across different sectors $\Sigma$, providing a mechanism for cognitive mode switching.
\end{itemize}
The introduction of semantic sectors $\Sigma$ directly reflects the principle of functional modularity prevalent in cognitive psychology and neuroscience.

\section{Toward Cognitive Plausibility}

By grounding its structures ($\Phi, \Sigma, \Phi^{(k)}$), dynamics ($A, N_t, K, \Lambda, V$), and regulatory mechanisms ($\kappa, \lambda, \epsilon, \theta, M, \pi_{effort}, \pi_{regulate}$) in established psychological concepts and evidence, the semantic manifold framework aims for cognitive plausibility. It provides a computational language for modeling how structured beliefs might be formed, evolve, and be regulated in a way that reflects the constraints and capabilities observed in human cognition.


\subsection*{Chapter Summary}
This chapter details the psychological motivations behind the semantic manifold framework, aiming for cognitive plausibility by aligning its constructs with established theories. The representation of belief states ($\phi$) as structured linguistic ensembles ($\{\varphi_i\}$) reflects principles of constructive memory and functional modularity, with Semantic Sectors ($\Sigma$) mirroring specialized systems like those in working memory models. The framework's hierarchical structure, with abstraction layers ($\Phi^{(k)}$) and scaling operators ($\Lambda$, $V$), is motivated by cognitive development, category learning, and schema formation, capturing the human ability to generalize. Core dynamics parallel psychological processes: Assimilation ($A$) relates to integrating new information into existing schemas; Nullification ($N_t$), modulated by anchoring ($a_i$), models structured forgetting and memory decay; and Annihilation ($K$) reflects processes like dissonance reduction. The inclusion of Cognitive Load ($\lambda$) and Semantic Effort ($\epsilon$) addresses resource limitations central to attention theories, with effort allocation policies ($\pi_{effort}$) modeling executive control. Finally, the framework's regulatory mechanisms---including meta-cognitive monitoring via Meta-Assimilation ($M$), Semantic Orientation derived from Null Tower axes ([$\omega \rightarrow \omega^{(\infty)}$]), and the reflective sector ($\Sigma_{refl}$)---are grounded in research on metacognition and self-regulation. By integrating these psychological concepts, the framework seeks to provide a plausible computational model of structured belief formation, evolution, and control.
	\chapter{Neuroscientific Motivations}
\label{chap:NeuroscientificMotivations}

While the semantic manifold framework is presented as a formal theory of structured belief in an abstract space $\Phi$, its design draws significant inspiration from, and finds potential analogues in, the organization and dynamics of the biological brain. This chapter explores these neuroscientific motivations, demonstrating how core components of the framework---such as hierarchical structure, modularity, dynamic memory, and regulatory control---reflect principles observed in systems neuroscience. \textbf{It is crucial to emphasize that the goal here is not to propose a direct neural implementation or a one-to-one mapping, but rather to highlight convergences that lend biological plausibility to the framework's architectural commitments.} These potential analogues serve as inspiration and suggest that the framework's principles may reflect efficient ways of organizing complex information processing, whether biological or artificial.

\section{Hierarchical Abstraction and Cortical Processing Streams}

The stratification of the semantic manifold into layers of increasing abstraction ($\Phi^{(k)}$) mirrors the hierarchical organization evident in cortical processing pathways.
\begin{itemize}
	\item \textbf{Sensory Hierarchies:} Processing streams in sensory cortices, like the ventral visual pathway (V1 through IT), exhibit progressive abstraction, extracting increasingly complex and invariant features at higher levels. This parallels the function of the abstraction operator $\Lambda$, compressing detailed $\Phi^{(k)}$ representations into more general $\Phi^{(k+1)}$ forms.
	\item \textbf{Predictive Coding:} Theories of predictive coding posit hierarchical generative models where top-down predictions (analogous to elaboration $V$ from higher $\Phi^{(k)}$) meet bottom-up sensory evidence (processed via $X$ and initial $\Lambda$ into lower $\Phi^{(k)}$), with mismatches driving updates. This resonates with the framework's interplay between abstraction levels mediated by $\Lambda$ and $V$.
	\item \textbf{Prefrontal Cortex (PFC):} Higher cognitive functions associated with abstract rules, goals, and meta-representation are linked to anterior regions of the PFC, corresponding conceptually to the highest levels of abstraction, $\Phi^{(k)}$ for $k \gg 0$, within the framework.
\end{itemize}

\section{Memory Systems and Nullification Dynamics}

The framework's conceptualization of belief evolution, particularly the dynamics of memory formation (Assimilation $A$), persistence (represented by anchoring $a_i$), and decay (Nullification $N_t$), aligns with neuroscientific models of memory:
\begin{itemize}
	\item \textbf{Hippocampal-Cortical Interactions:} Models of systems consolidation propose that the hippocampus initially binds episodic information, which is gradually consolidated into cortical networks. This parallels the idea of belief fragments $\varphi_i$ being assimilated ($A$) and their long-term persistence depending on reinforcement (influencing anchoring $a_i$) versus gradual decay ($N_t$) if unreinforced.
	\item \textbf{Active Forgetting:} Neuroscience suggests forgetting is not merely passive decay but involves active mechanisms. This supports the regulated nature of Nullification ($N_t$) where decay rates (potentially derived from $a_i$) can vary, rather than being uniform information loss.
\end{itemize}

\section{Modular Function and Semantic Sectors}

The concept of partitioning $\Phi$ into functional Semantic Sectors ($\Sigma$) reflects the well-established principle of functional specialization in the brain:
\begin{itemize}
	\item \textbf{Specialized Networks:} Distinct brain networks are associated with specific cognitive functions (e.g., language networks, default mode network for reflection/simulation, dorsal attention network for planning/spatial processing, salience network for control). These provide biological analogues for sectors like $\Sigma_{lang}$, $\Sigma_{refl}$, $\Sigma_{plan}$, and potentially for regulatory control over sector activation ($\vec{\alpha}(t)$).
	\item \textbf{Subcortical Loops:} Basal ganglia circuits are involved in gating and selecting actions and cognitive strategies, potentially implementing mechanisms for activating or suppressing specific semantic sectors ($\Sigma$) or associated operations.
\end{itemize}

\section{Semantic Compass and Neural Coordinate Systems}

The framework's mechanisms for Semantic Orientation (epistemic axes $[\omega \rightarrow \omega^{(\infty)}]$, projection $\pi_{\omega}$, deviation $\theta$) are inspired by neural systems for navigation, both physical and abstract:
\begin{itemize}
	\item \textbf{Hippocampal-Entorhinal System:} The discovery of place cells and grid cells suggests the brain uses map-like representations and coordinate systems. Emerging evidence indicates these systems may also map abstract conceptual or relational spaces, providing a potential neural basis for navigating the semantic manifold $\Phi$.
	\item \textbf{Cognitive Maps in PFC/OFC:} Areas like the orbitofrontal cortex appear to encode latent state spaces relevant to tasks, supporting goal-directed navigation through abstract decision or belief spaces, analogous to following orientation vectors ($\vec{v}_{\omega}$) or trajectories ($\gamma(t)$) within $\Phi$.
\end{itemize}

\section{Effort, Load, and the Neural Economy of Thought}

The framework's concepts of Cognitive Load ($\lambda$) and Semantic Effort ($\epsilon$) relate to the brain's resource limitations and management:
\begin{itemize}
	\item \textbf{Neural Correlates of Load:} Increased cognitive demand is associated with heightened activity in frontoparietal control networks and concurrent suppression of networks like the DMN, reflecting the allocation of finite processing resources modeled conceptually by $\lambda$.
	\item \textbf{Neural Correlates of Effort:} Subjective effort and regulatory control are linked to activity in areas like the anterior cingulate cortex (ACC). Neuromodulatory systems (e.g., dopamine, norepinephrine) play key roles in regulating attention and cognitive activation, aligning with the concept of controlled allocation of effort ($\epsilon$) via policies like $\pi_{effort}$.
\end{itemize}

\section{Meta-Cognition and Prefrontal Dynamics}

The capacity for meta-cognition (Part~\ref{part:meta_cognition}), including Meta-Introspection and Trajectory Awareness ($\gamma(t)$), finds parallels in higher-order brain functions primarily associated with the prefrontal cortex:
\begin{itemize}
	\item \textbf{Rostral PFC:} Implicated in the most abstract forms of meta-cognition, self-reflection, and evaluating one's own mental states, corresponding conceptually to the functions supported by the reflective sector $\Sigma_{refl}$ and Meta-Assimilation ($M$).
	\item \textbf{ACC/Dorsomedial PFC:} Involved in performance monitoring, error detection, and conflict resolution, providing neural analogues for mechanisms assessing coherence ($\kappa$) and triggering regulatory adjustments (potentially via $\pi_{regulate}$).
\end{itemize}

\section{Toward Neurosemantic Alignment}

While the Semantic Manifold framework operates at a functional and computational level, its core principles---hierarchy, modularity, dynamic memory, geometric representation, and regulated control---demonstrate significant alignment with the observed structure and dynamics of neural systems. This alignment suggests that the proposed architecture is not merely an arbitrary formalism but may reflect convergent principles for organizing complex, adaptive information processing systems, lending it a degree of biological plausibility. \textbf{However, these parallels remain high-level and inspirational; establishing direct implementational links requires substantial further research.} Future work bridging computational modeling based on this framework with targeted neuroscientific experiments could potentially refine both our understanding of the brain and the design of artificial cognitive systems.


\subsection*{Chapter Summary}
This chapter explores the neuroscientific motivations lending biological plausibility to the semantic manifold framework, emphasizing analogues rather than direct implementation claims. The framework's hierarchical structure ($\Phi^{(k)}$) and abstraction operator ($\Lambda$) are shown to parallel cortical processing streams (e.g., sensory hierarchies, predictive coding models, PFC function). Memory dynamics, including assimilation ($A$), persistence via anchoring ($a_i$), and gradual decay through Nullification ($N_t$), align with concepts like systems consolidation and active forgetting mechanisms observed in the brain. The functional modularity provided by Semantic Sectors ($\Sigma$) reflects the specialization of distinct brain networks and circuits (e.g., language, default mode, attention networks; basal ganglia loops). The Semantic Orientation mechanism, derived from Null Tower axes ([$\omega \rightarrow \omega^{(\infty)}$]), finds inspiration in neural coordinate systems used for navigation (e.g., hippocampal maps potentially extending to abstract spaces). Concepts of Cognitive Load ($\lambda$) and Semantic Effort ($\epsilon$) relate to neural correlates of demand, resource allocation, and neuromodulation. Furthermore, meta-cognitive capabilities like Meta-Assimilation ($M$) and Trajectory Awareness ($\gamma(t)$) are linked to higher-order functions associated with prefrontal cortex dynamics (e.g., rostral PFC for reflection, ACC for monitoring). While acknowledging the functional level of the framework, these convergences suggest its architectural principles align with efficient information processing strategies observed in biological systems.
	
	\part{Semantic Foundations: The Semantic Substrate}
	\label{part:semantic_foundations}
	
	\chapter{The Semantic State Space}
\label{chap:SemanticStateSpace}

\section{Motivation: The Substrate of Belief}

At the core of any agent capable of reasoning, reflection, or complex interaction lies its internal representation of the world and itself---its system of beliefs. For language-based agents, which perceive, reason, and communicate through linguistic structures, the nature of this internal representational space is paramount. It must serve not merely as a repository of facts, but as a dynamic medium supporting the intricate processes of cognition: integrating perception, formulating goals, simulating possibilities, maintaining coherence, and adapting to new information.

We posit that the substrate for belief in such agents is a semantic state space, denoted by $\Phi$. This space is conceived as the set of all possible internal belief configurations accessible to the agent. Crucially, we assert that for the agents under consideration, this space is fundamentally linguistic. Language is not merely an output modality; it is the medium in which thought occurs. Beliefs are formed, manipulated, and stored as structured ensembles of natural language expressions. This choice is motivated by several factors:

\begin{itemize}
	\item \textbf{Expressiveness and Compositionality:} Language allows for the representation of complex, nuanced, and abstract concepts, and supports the combination and decomposition of beliefs.
	\item \textbf{Interpretability:} Linguistically grounded states facilitate introspection, explanation, and alignment, as the agent's internal states have a structure amenable to human understanding.
	\item \textbf{Cognitive Plausibility:} While not a direct model of human cognition, this approach draws inspiration from the central role of language in human reasoning, planning, and self-awareness.
	\item \textbf{Generativity:} A linguistic substrate allows agents to construct novel beliefs through inference, simulation, or creative recombination, going beyond direct experience.
\end{itemize}

Furthermore, the semantic state space $\Phi$ must accommodate the transformation of raw sensory or symbolic input into meaningful internal representations. It must support mechanisms for maintaining internal coherence, allowing the agent to resolve contradictions and integrate information smoothly. It must also provide a foundation for higher-level cognitive functions like abstraction, modular reasoning (across different functional domains), and self-reflection. Constructing this space rigorously is the first step towards a comprehensive theory of structured belief.

\section{Belief States as Linguistic Ensembles}

We define an individual belief state $\phi$ as an element of the semantic state space $\Phi$. A belief state represents the agent's cognitive configuration at a particular moment. Following our linguistic grounding, we model $\phi$ not as a monolithic entity, but as a structured ensemble of natural language expressions $\varphi_i$. We write this notionally as:
$$ \phi = \{\varphi_1, \varphi_2, \ldots\} $$
Each expression $\varphi_i$ is a linguistically well-formed and cognitively interpretable unit, such as:
\begin{itemize}
	\item An observation: "The light is green."
	\item An inference: "Therefore, it is safe to proceed."
	\item A goal or intention: "I must reach the destination by noon."
	\item A policy fragment: "If obstacle detected, halt."
	\item A hypothetical: "What if the sensor fails?"
	\item A reflective statement: "My previous estimate was inaccurate."
\end{itemize}
We use the notation $\varphi_i \in \phi$ as a shorthand to indicate that the expression $\varphi_i$ is part of the structured ensemble constituting the belief state $\phi$. This does not imply that $\phi$ is mathematically a set; rather, it emphasizes that belief states are composed of these identifiable linguistic components.

\begin{figure}[htbp]
	\centering
	\input{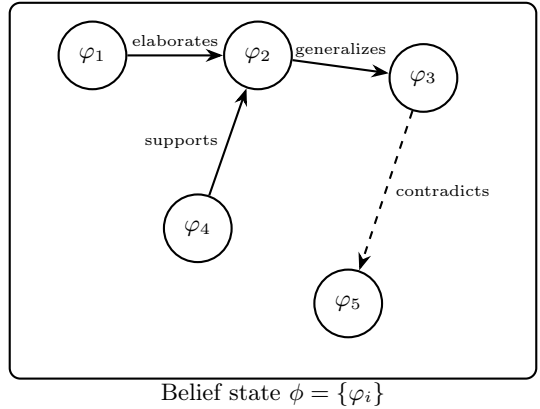}
	\caption{Structure of a belief state \(\phi = \{ \varphi_1, \varphi_2, \ldots \}\), depicted as a network of linguistic expressions linked by semantic relations such as elaboration, generalization, support, or contradiction. This internal structure governs the coherence and dynamics of the state.}
	\label{fig:phi-structure}
\end{figure}

\section{Structure within Belief States}

While represented as a collection $\{\varphi_i\}$, a belief state $\phi$ possesses internal structure that goes beyond a simple list or set. This structure is essential for supporting coherent reasoning and dynamic evolution. Though we remain agnostic about the specific implementation (which could involve graphs, databases, weighted lists, etc.), the functional requirements imply that $\phi$ can support relationships such as:
\begin{itemize}
	\item \textbf{Semantic Linkage:} Connections between expressions based on coreference, implication, contradiction, elaboration, or causal relations.
	\item \textbf{Temporal or Causal Ordering:} Representing sequences of events, dependencies in plans, or the flow of narrative.
	\item \textbf{Relevance or Salience Weighting:} Attributing different levels of importance, activation, or focus to different expressions within the ensemble based on context or goals.
	\item \textbf{Typing or Labeling:} Categorizing expressions based on their origin (e.g., perceived, inferred, simulated) or function (e.g., goal, fact, policy).
	\item \textbf{Anchoring:} Designating certain expressions as stable reference points that resist modification or decay.
\end{itemize}

This internal structure allows $\phi$ to be more than just a collection of sentences; it becomes an active, interconnected representation capable of supporting complex cognitive dynamics.

\section{Formal Definition of the Semantic State Space \texorpdfstring{$\Phi$}{Phi}}

Having characterized the nature of individual belief states, we now formally define the semantic state space $\Phi$ itself.

\textbf{Definition:} The semantic state space $\Phi$ is the set of all possible structured, coherent, and linguistically interpretable belief state ensembles $\phi = \{\varphi_1, \varphi_2, \ldots\}$ that an agent can potentially entertain or instantiate.

Key aspects of this definition include:
\begin{itemize}
	\item \textbf{Comprehensiveness:} $\Phi$ encompasses all belief states the agent might form, whether grounded in perception, generated internally through simulation or reflection, or derived through learning and abstraction.
	\item \textbf{Linguistic Basis:} All states $\phi \in \Phi$ are fundamentally composed of structured natural language expressions.
	\item \textbf{Coherence Requirement:} While $\Phi$ may contain states representing unresolved tension or contradiction (as intermediate steps in reasoning), it generally excludes arbitrary or semantically meaningless collections of expressions. A notion of interpretability and potential coherence is required.
	\item \textbf{Agent-Relativity:} $\Phi$ is defined relative to the capabilities and architecture of a specific agent or class of agents (potentially parameterized as $\Phi^{[\theta]}$).
\end{itemize}
The set $\Phi$ serves as the underlying manifold upon which all belief dynamics---assimilation, revision, nullification, reflection, planning---take place. Its structure enables and constrains the cognitive possibilities of the agent.

\begin{figure}[h]
	\centering
	\input{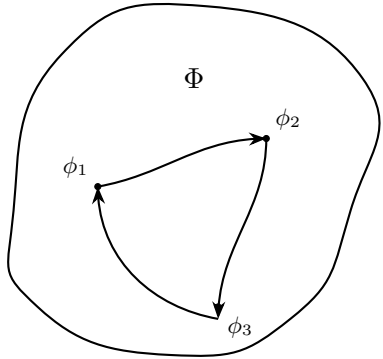}
	\caption{Illustration of the semantic state space \(\Phi\), populated by individual belief states \(\phi_i\). Curved arrows depict semantic transitions among belief states.}
	\label{fig:semantic_state_space}
\end{figure}

\section{Observation Encoding: Grounding Belief in Experience}

A primary way the semantic state space $\Phi$ becomes populated with meaningful content is through the interpretation of external or internal signals via the observation encoder function, $X$.

\textbf{Definition:} The observation encoder $X: S \rightarrow \Phi$ is a mapping from the space of raw sensory or symbolic inputs $S$ to the semantic state space $\Phi$. For an input $s \in S$, $X(s)$ produces a structured belief state $\phi = X(s) \in \Phi$ that represents the agent's initial linguistic interpretation of that input.

The function $X$ acts as the bridge between unstructured experience and structured internal belief. It is not merely a transducer but an interpretive process, potentially involving feature extraction, categorization, inference, and contextualization based on the agent's prior state (though for simplicity, we often model it as mapping primarily to base-level representations). Examples illustrate its cross-modal nature:
\begin{itemize}
	\item \textbf{Visual Input:} An image $s$ of a specific object might yield $\phi$ containing expressions like "A blue sphere is present," "It rests on a wooden surface," "No motion detected."
	\item \textbf{Textual Input:} A command $s = $ "Proceed to waypoint B" might yield $\phi$ containing "New goal received: reach waypoint B," "Current task is navigation."
	\item \textbf{Internal Proprioception:} Sensor readings $s$ indicating manipulator state might yield $\phi$ containing "Gripper is currently open," "Joint angles are within nominal range."
\end{itemize}
Crucially, the image of the observation encoder $X(S)$ is typically a proper subset of the full semantic state space: $X(S) \subset \Phi$. This reflects that $\Phi$ must also contain belief states that are not directly grounded in immediate input, such as those arising from memory retrieval, internal simulation, planning, abstract reasoning, or reflective introspection. $\Phi$ is the space of all possible thoughts, not just encoded perceptions.

\section{The Epistemic Vacuum \texorpdfstring{$\Omega$}{Omega}}

Within the vast space $\Phi$, there exists a special subset representing the absence of semantic commitment: the epistemic vacuum $\Omega$.

\textbf{Definition:} The epistemic vacuum $\Omega \subset \Phi$ is the set of all belief states $\phi \in \Phi$ that contain no active semantic content. States in $\Omega$ are structurally valid but epistemically inert.

\begin{figure}[h]
	\centering
	\input{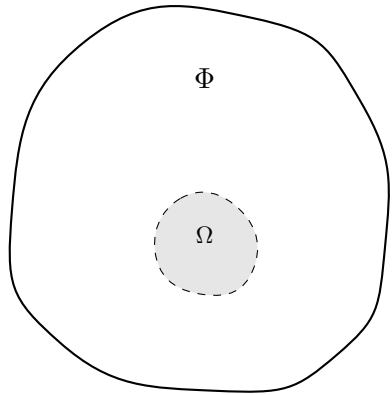}
	\caption{The epistemic vacuum \(\Omega\) as a distinguished null region within the semantic state space \(\Phi\).}
	\label{fig:semantic_vacuum}
\end{figure}

Key properties of $\Omega$ include:
\begin{itemize}
	\item \textbf{Multiplicity:} $\Omega$ is not a single point but a class of states. Different initializations or representational formats might lead to different semantically inert states $\phi_0 \in \Omega$ that are behaviorally indistinguishable in their emptiness (related to the concept of Semantic Gauge, explored later).
	\item \textbf{Origin Point:} $\Omega$ serves as the conceptual origin for belief dynamics. Processes like spontaneous drift or assimilation can naturally begin from a state within $\Omega$.
	\item \textbf{Target Point:} Processes like nullification (belief decay) model trajectories tending back towards $\Omega$, representing semantic rest or forgetting.
\end{itemize}
The epistemic vacuum provides the necessary grounding for the emergence and dissolution of belief. It is the baseline against which semantic structure acquires meaning.

\section{Organizing Principles: Abstraction and Sectors (Preview)}

While $\Phi$ is defined as a unified space, understanding its internal organization is crucial. Later parts of this monograph will detail two key organizing principles:

\begin{itemize}
	\item \textbf{Semantic Scaling (Abstraction):} Belief states naturally vary in their level of generality or abstraction. We will introduce a conceptual vertical axis representing this scale, with layers $\Phi^{(k)}$ denoting increasing abstraction (from concrete $\Phi^{(0)}$ upwards). This allows us to model generalization, compression, and hierarchical reasoning (Part~\ref{part:structuring_belief}, Chapter~\ref{chap:SemanticScaling}).
	\item \textbf{Semantic Sectors ($\Sigma$):} Belief states can often be grouped by their functional role (perceptual, narrative, planning, reflective, etc.). We will define these potentially overlapping sectors $\Sigma$ as interpretable regions within $\Phi$, supporting modular control and analysis (Part~\ref{part:structuring_belief}, Chapter~\ref{chap:SemanticSectors}).
\end{itemize}

\begin{figure}[htbp]
	\centering
	\input{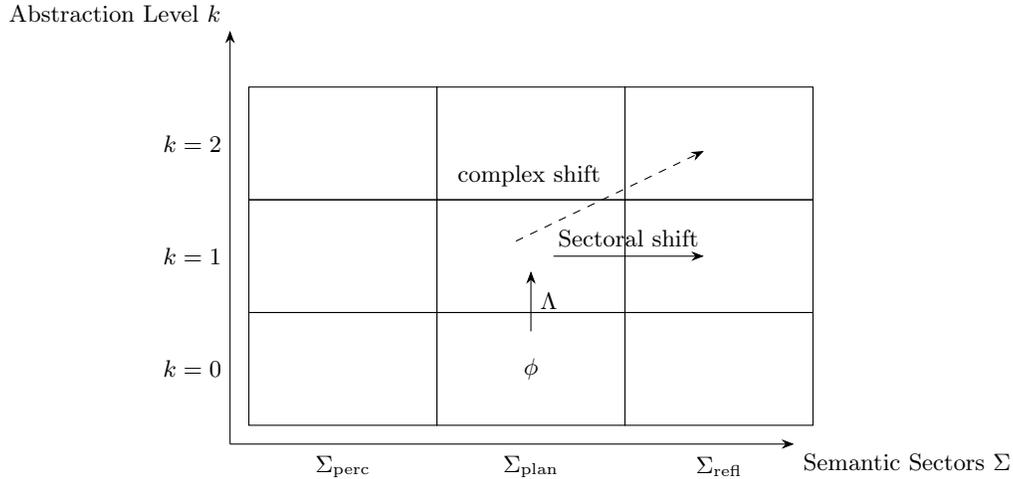}
	\caption{Semantic geometry of the belief space \(\Phi\), structured by functional sectors \(\Sigma\) (horizontal) and abstraction levels \(k\) (vertical). Belief state \(\phi\) is located at a specific \((\Sigma, k)\) cell. Vertical motion corresponds to abstraction (\(\Lambda\)), horizontal motion to sectoral shifts, and diagonal motion to compound transformations.}
	\label{fig:semantic-geometry}
\end{figure}

These principles give rise to the notion of a Semantic Geometry for $\Phi$ (Part~\ref{part:structuring_belief}, Chapter~\ref{chap:SemanticGeometry}), often visualized using two conceptual dimensions (e.g., semantic variation horizontally, abstraction vertically) with coordinates $(\Sigma, k)$.

While $\Phi$ itself may be high-dimensional and complex, these organizing principles provide essential tools for navigating and interpreting its structure. They form the basis for understanding how beliefs are structured (Part~\ref{part:structuring_belief}) and regulated (Part~\ref{part:regulation_and_control}).

\subsection*{Summary of Key Notation Introduced in this Chapter}

This chapter has laid the groundwork by defining the semantic state space $\Phi$, its constituent belief states $\phi$, the observation encoder $X$ that populates it from experience, and the epistemic vacuum $\Omega$ that anchors it. The following chapters will build upon this foundation, exploring the emergence of structure from the vacuum via the Null Tower.

\bigskip
\begin{center}
	\begin{tabular}{ll}
		\hline
		\textbf{Symbol} & \textbf{Definition} \\ \hline
		\(\Phi\) & Semantic state space; the set of all possible belief states. \\
		\(\phi\) & A specific belief state; a structured ensemble \(\{\varphi_1, \varphi_2, \ldots\}\). \\
		\(\varphi_i\) & An individual linguistic expression within a belief state. \\
		\(S\) & Space of raw sensory or symbolic inputs. \\
		\(X\) & Observation encoder function \((X: S \rightarrow \Phi)\). \\
		\(\Omega\) & Epistemic vacuum; the set of semantically null belief states. \\ \hline
	\end{tabular}
\end{center}
\bigskip


\subsection*{Chapter Summary}
This chapter introduces the foundational concept of the semantic state space, $\Phi$, conceived as the substrate for belief within language-capable intelligent agents. It motivates the need for such a space to support dynamic cognitive processes like reasoning, integration, and reflection. Individual belief states, $\phi$, are defined as structured ensembles of natural language expressions, $\{\varphi_i\}$, encompassing observations, inferences, goals, and other cognitive elements, possessing internal relational structures beyond simple collections. The formal definition of $\Phi$ emphasizes its comprehensiveness, linguistic basis, coherence requirements, and agent-relativity (parameterized as $\Phi^{[\theta]}$). The chapter details the Observation Encoding function, $X: S \rightarrow \Phi$, as the crucial bridge translating raw sensory or symbolic inputs $S$ into initial, grounded belief structures within $\Phi$. It also introduces the Epistemic Vacuum, $\Omega$, a special subset of $\Phi$ representing states devoid of active semantic content, serving as the origin and potential endpoint for belief dynamics. Finally, the chapter previews two key organizing principles to be elaborated later: Semantic Scaling (layering $\Phi$ by abstraction level $\Phi^{(k)}$) and Semantic Sectors (organizing $\Phi$ by function $\Sigma$), which together define the Semantic Geometry of the belief space.
	\chapter{The Null Tower}
\label{chap:NullTower}

\section{Introduction: The Emergence of Structure}

How does structured thought arise from cognitive emptiness? How can an agent, beginning with no semantic content, construct the intricate web of beliefs necessary for reasoning, planning, and reflection? Chapter~\ref{chap:SemanticStateSpace} established the semantic state space $\Phi$ as the linguistic substrate for belief and identified the epistemic vacuum $\Omega$ as its origin point---a set of states devoid of active semantic commitment. This chapter introduces the Null Tower, a formal structure representing the principled, recursive emergence of cognition from this vacuum.

\begin{figure}[h]
	\centering
	\input{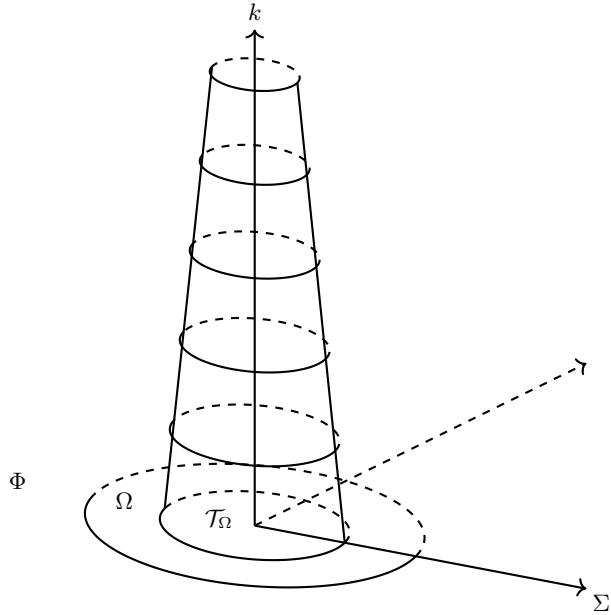}
	\caption{The Null Tower ($\mathcal{T}_\Omega$) structure in the epistemic vacuum ($\Omega$) within semantic state space ($\Phi$).}
	\label{fig:null_tower_main}
\end{figure}

The Null Tower models the agent's cognitive genesis not as a random accumulation of facts, but as a structured ascent. Beginning with minimal, unstructured semantic states, belief states are iteratively constructed via abstraction operators that impose layers of semantic organization. The resulting tower encodes the origin, trajectory, and coherence of the agent's thought, providing a dynamic scaffold built entirely through the process of recursive generalization. It offers a generative account of how belief systems, planning structures, and even self-models can arise from a state of epistemic nullity, grounding the agent's cognitive architecture in a minimal, introspectively accessible foundation.

\section{Preliminaries: Vacuum, Layers, and Abstraction}

Before constructing the tower, we revisit key concepts. The epistemic vacuum $\Omega \subset \Phi$ is the set of semantically inert belief states, or null states. Within this vacuum, we identify a foundational layer:

\textbf{Definition (Irreducible Null Stratum):} The irreducible null stratum $\Omega^{(0)}$ is the subset of $\Omega$ containing null states that have not themselves been produced by an abstraction process operating within $\Omega$. Formally, $\Omega^{(0)} = \Omega \setminus \Lambda(\Omega)$ where $\Lambda$ is the abstraction operator (defined below). Elements $\omega \in \Omega^{(0)}$ serve as the ultimate seeds or origins for cognitive structure.

The construction relies on the concept of semantic layers or scales $\Phi^{(k)}$ within $\Phi$ representing increasing levels of abstraction or generality (previewed in Chapter~\ref{chap:SemanticStateSpace} and detailed in Chapter~\ref{chap:SemanticScaling}). Transitions between these layers are mediated by abstraction operators:

\textbf{Definition (Abstraction Operator):} The abstraction operator $\Lambda: \Phi^{(k)} \rightarrow \Phi^{(k+1)}$ maps a belief state $\phi^{(k)}$ at abstraction level k to a more abstract (e.g., compressed, generalized) belief state $\phi^{(k+1)}$ at level $k+1$.

We denote the k-fold composition of abstraction as $\Lambda^{k} = \Lambda \circ \dots \circ \Lambda$ (composition $k$ times, with $\Lambda^{0}$ being the identity). These operators are the engines driving the construction of the Null Tower.

\section{Tower Construction from the Epistemic Vacuum}

The Null Tower is constructed recursively for each irreducible null state $\omega \in \Omega^{(0)}$.

\begin{figure}[htbp]
	\centering
	\input{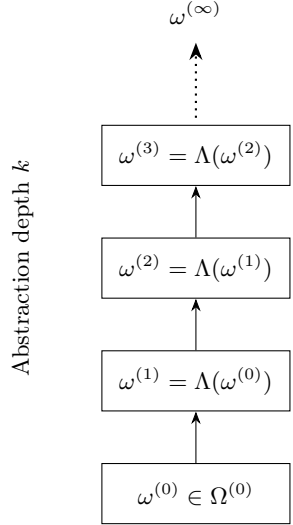}
	\caption{Recursive construction of the Null Tower. Starting from an irreducible null state \(\omega^{(0)} \in \Omega^{(0)}\), successive applications of the abstraction operator \(\Lambda\) generate higher null states \(\omega^{(k)}\), converging toward the semantic singularity \(\omega^{(\infty)}\).}
	\label{fig:null-tower}
\end{figure}

\textbf{Definition (Recursive Abstraction):} Given $\omega \in \Omega^{(0)}$, its level-k abstraction $\omega^{(k)}$ is defined as:
$$ \omega^{(k)} := \Lambda^{k}(\omega) \quad \text{for } k \ge 0. $$
Each $\omega^{(k)} \in \Phi^{(k)}$ represents a semantic structure of abstraction depth k, inheriting compressed or reorganized content from the layers below. The sequence $\{\omega^{(0)}, \omega^{(1)}, \omega^{(2)}, \ldots\}$ defines a specific belief trajectory originating from $\omega$.

This process allows us to define the layers of the Null Tower itself:

\textbf{Definition (Null Tower Layers):} The k-th layer of the Null Tower, $\Omega^{(k)}$, is the set of all belief states reachable by applying exactly k levels of abstraction to the irreducible null stratum:
$$ \Omega^{(k)} := \Lambda^{k}(\Omega^{(0)}) $$
These layers form a stratification of the null-derived belief space: $\bigcup_{k=0}^{\infty} \Omega^{(k)} \subseteq \Phi$ strictly reachable from the base of the Null Tower, the irreducible null stratum $\Omega^{(0)}$. Every belief state constructed purely by abstraction from $\Omega^{(0)}$ belongs to a unique layer k.

\begin{figure}[h]
	\centering
	\input{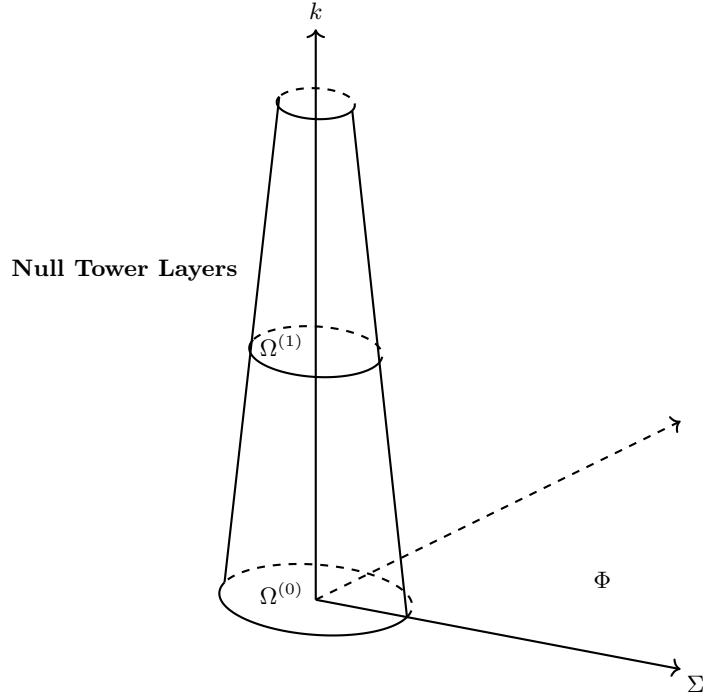}
	\caption{The Null Tower structure in semantic state space with layers $\Omega^{(k)}$. The vertical axis \(k\) denotes introspective depth, while the horizontal axis \(\Sigma\) represents semantic spread.}
	\label{fig:null_tower}
\end{figure}

The full set of beliefs derivable from the epistemic vacuum via abstraction constitutes the null-derived belief space $\mathcal{T}_{\Omega}$:
$$ \mathcal{T}_{\Omega} := \bigcup_{\omega \in \Omega^{(0)}} \{\omega^{(k)} | k \in \mathbb{N}_0\} = \bigcup_{k=0}^{\infty} \Omega^{(k)} $$
This space represents the agent's potential for internally generated structure, built solely upon recursive abstraction from its minimal starting points. The Null Tower is thus not a static object but the result of this ongoing constructive process, defining the developmental and structural backbone of the agent's potential for internal epistemic organization.

\section{Structural Properties and Dynamics}

The Null Tower possesses several crucial structural properties that enable cognitive functions:
\begin{itemize}
	\item \textbf{Stratification:} The layers $\Omega^{(k)}$ are distinct by construction, providing a canonical indexing of null-derived beliefs by abstraction depth.
	\item \textbf{Recursive Dependency and Inheritance:} Each state $\omega^{(k)}$ inherits its structure from $\omega^{(k-1)}$ via $\Lambda$. This ensures a downward referential integrity (higher levels summarize lower ones) and upward semantic accumulation. This dependency chain ($\omega^{(k)} \rightarrow \omega^{(k-1)} \rightarrow \dots \rightarrow \omega^{(0)}$) supports semantic stability and recovery. If $\omega^{(k)}$ becomes corrupted or unstable, the agent might revert to $\omega^{(k-1)}$ and re-apply abstraction.
	\item \textbf{Semantic Metrics and Alignment:} If a semantic distance $d$ is defined on $\Phi$ (see Part~\ref{part:embodiment_and_action}), and if abstraction operators $\Lambda$ are contractive under $d$, then abstraction can regularize noise and align related belief trajectories. The coherence of a trajectory $\{\omega^{(k)}\}$ can be measured relative to its ideal path.
	\item \textbf{Semantic Flow:} The tower structure supports directional semantic flow: bottom-up processes correspond to construction, generalization, and emergence of patterns, while top-down processes can represent goal imposition, constraint propagation, or reflective modulation originating from higher abstraction levels.
	\item \textbf{Failure Modes:} The tower's structure is not guaranteed to be stable. Potential failures include:
	\begin{itemize}
		\item Collapse: One or more layers $\omega^{(k)}$ become inaccessible or incoherent, breaking the recursive chain.
		\item Drift: A trajectory $\{\omega^{(k)}\}$ deviates significantly from its expected or coherent path, leading to semantically invalid abstractions.
		\item Fragmentation: Subsequences lose structural continuity or inheritance, leading to isolated, ungrounded high-level beliefs.
	\end{itemize}
	These failures can arise from noisy operators, memory limits, or internal contradictions. The tower's structure, however, also provides pathways for recovery, such as downward fallback or lateral substitution within a layer $\Omega^{(k)}$.
\end{itemize}
The Null Tower is therefore a dynamic cognitive architecture, providing a scaffold for belief formation, coherence maintenance, and structured recovery from epistemic disruption.

\section{The Semantic Singularity \texorpdfstring{$\omega^{(\infty)}$}{omega (infinity)}}

What is the ultimate destination of a belief trajectory under indefinite recursive abstraction? The concept of the semantic singularity addresses this question.

\textbf{Definition (Semantic Singularity):} If the sequence of abstracted states $\{\omega^{(k)}\}_{k=0}^{\infty}$ originating from $\omega \in \Omega^{(0)}$ converges to a limit point $\omega^{(\infty)}$ within $\Phi$, this limit is called the semantic singularity associated with $\omega$.
$$ \omega^{(\infty)} := \lim_{k\rightarrow\infty} \omega^{(k)} $$
A semantic singularity $\phi_{\infty}$ might be characterized as a fixed point of abstraction, satisfying $\Lambda(\phi_{\infty}) = \phi_{\infty}$ for a suitably defined global abstraction operator $\Lambda$.

\begin{figure}[ht!]
	\centering
	\input{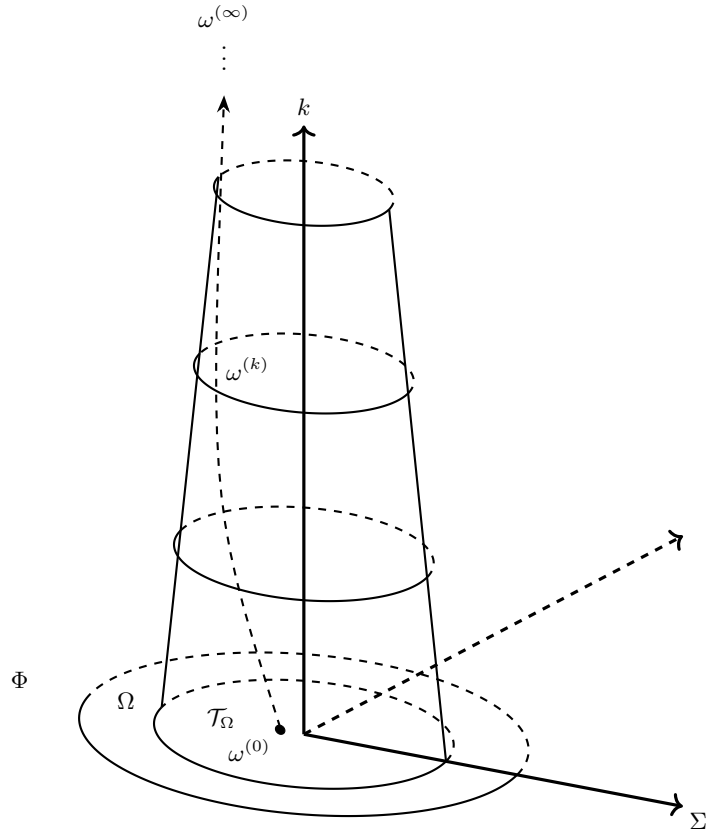}
	\caption{ A trajectory \(\omega^{(k)}\) rises from the epistemic vacuum \(\Omega\), traversing layers of increasing abstraction (\(k\)) through the Null Tower \(\mathcal{T}_\Omega\).}
	\label{fig:semantic_singularity}
\end{figure}

Properties of $\omega^{(\infty)}$ typically include:
\begin{itemize}
	\item \textbf{Abstraction Invariance:} It is unchanged by further application of abstraction operators.
	\item \textbf{Minimal Specificity:} It encodes only the structural information that remains invariant across all levels of abstraction derived from $\omega$. It lacks specific, grounded referential content.
	\item \textbf{Conceptual Attractor:} It can act as a target or attractor for processes involving generalization, summarization, or semantic compression.
\end{itemize}
While potentially an idealized construct (perhaps only approximated in finite systems), $\omega^{(\infty)}$ serves important theoretical and functional roles. It represents the endpoint of pure generalization, the "idea of the idea." In engineered systems, approximated singularities could serve as reference points for belief alignment, priors for regularization, or targets for training abstraction mechanisms (as discussed in Part~\ref{part:learning_and_adaptation}). Philosophically, it resonates with concepts like pure form or points at infinity in projective geometry.

\section{Epistemic Axes and Projective Structure (Preview)}

The trajectory defined by the recursive abstraction process, spanning from an irreducible null state $\omega$ to its semantic singularity $\omega^{(\infty)}$, defines a natural direction within the belief space $\Phi$.

\textbf{Definition (Epistemic Axis):} An epistemic axis is the directed semantic trajectory $[\omega \rightarrow \omega^{(\infty)}]$.

\begin{figure}[ht!]
	\centering
	\input{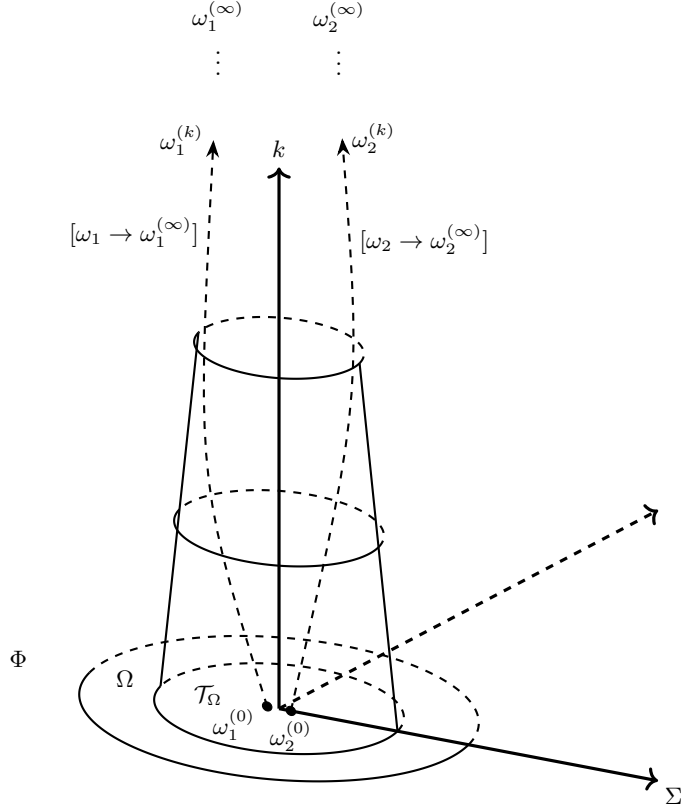}
	\caption{Multiple epistemic reference axes \([\omega_i \to \omega_i^{(\infty)}]\) originating from the Null Stratum \(\Omega^{(0)}\) within the Epistemic Vacuum $\Omega$ and radiating outward through the Null Tower $\mathcal{T}_{\Omega}$. These axes define local semantic frames and support orientation within \(\Phi\).}
	\label{fig:epistemic_axes}
\end{figure}

This axis represents a canonical path of semantic ascent from unstructured potential ($\omega$) towards idealized generality ($\omega^{(\infty)}$). Any belief state $\phi$ can, in principle, be positioned relative to such an axis through projection $\pi_{\omega}(\phi)$, allowing measurement of its alignment (e.g., angular deviation $\theta(\phi, \vec{v}_{\omega})$) or semantic drift.

A family of such axes, derived from different base states $\omega_i$, can form a semantic reference frame within $\Phi$. This geometric structure, induced by the Null Tower's recursive construction, provides a foundation for introspective alignment, epistemic regulation, and reflective control. These concepts are central to Semantic Orientation, which will be explored in detail in Part~\ref{part:regulation_and_control} (Chapter~\ref{chap:SemanticOrientation}).

\section{Realization via Learned Abstraction (Preview)}

While defined formally via operators $\Lambda$, the Null Tower structure is amenable to practical realization, particularly through machine learning. Abstraction can be interpreted as a form of semantic compression or distillation, implementable via learnable transformations.
\begin{itemize}
	\item \textbf{Recursive Encoders:} A stack of neural encoders $E_k: \Phi^{(k)} \rightarrow \Phi^{(k+1)}$ can approximate the abstraction operators $\Lambda$. Applying these recursively to an initial state $\omega$ derived from data or a null prior constructs an empirical tower trajectory $\omega^{(k)} = E_{k-1} \circ \dots \circ E_0(\omega)$.
	\item \textbf{Learned Singularities:} With sufficient depth and appropriate training objectives (e.g., reconstruction loss, contractive regularization, fixed-point constraints), such systems might converge to approximate semantic singularities $\omega^{(\infty)}$.
\end{itemize}
This perspective allows the Null Tower to be instantiated within agent architectures, enabling them to construct and leverage layered semantic representations from experience. The mechanisms for learning these structures will be discussed further in Part~\ref{part:learning_and_adaptation} (Chapter~\ref{chap:LearningSemanticStructures}).

\section{The Role of the Null Tower in the Framework}

The Null Tower is a cornerstone of the foundational framework presented in this monograph. It provides:
\begin{itemize}
	\item A model for the origin of cognitive structure from epistemic nullity (the "empty mind").
	\item A mechanism for recursive abstraction, generating layered belief representations ($\Omega^{(k)}$ derived from $\omega \in \Omega^{(0)}$).
	\item A basis for defining structural coherence and inheritance across abstraction levels.
	\item The foundation for epistemic axes $[\omega \rightarrow \omega^{(\infty)}]$, which enable semantic orientation and navigation (Part~\ref{part:regulation_and_control}, Chapter~\ref{chap:SemanticOrientation}).
	\item A framework for understanding potential epistemic failure modes like collapse, drift, and fragmentation, along with paths to recovery.
\end{itemize}
By grounding the emergence of belief in a principled, recursive process originating from the epistemic vacuum $\Omega$, the Null Tower provides the essential vertical structure upon which the broader semantic geometry and dynamics of $\Phi$ are built. It bridges the gap between cognitive emptiness and structured, abstract thought. The next chapter explores the process by which belief structures, once initiated, are actively constructed and modified.


\subsection*{Chapter Summary}
This chapter introduces the Null Tower, a formal structure modeling the principled, recursive emergence of cognitive potential from the epistemic vacuum ($\Omega$). It proposes that starting from irreducible null states ($\omega \in \Omega^{(0)}$), layers of increasingly abstract semantic structures ($\omega^{(k)}, \Omega^{(k)}$) are built via repeated application of abstraction operators ($\Lambda$). This generative process accounts for the origin of cognitive structure independent of external input. The chapter details the tower's construction, its structural properties such as stratification and recursive dependency, and potential failure modes like collapse or drift. It defines the semantic singularity ($\omega^{(\infty)}$) as the conceptual limit of this abstraction process, representing pure semantic form. Crucially, the Null Tower provides the foundation for epistemic axes, defined as the trajectories $[\omega \rightarrow \omega^{(\infty)}]$, which serve as internal reference frames enabling Semantic Orientation (discussed in Part~\ref{part:regulation_and_control}). The chapter previews how this structure might be realized via learned abstraction mechanisms and concludes by emphasizing the Null Tower's role as a cornerstone providing vertical structure, coherence, orientation, and a bridge from cognitive emptiness to structured thought within the semantic manifold $\Phi$.
	\chapter{Belief Construction}
\label{chap:BeliefConstruction}

\section{Introduction: From Potential to Structured Belief}

Chapters~\ref{chap:SemanticStateSpace} and \ref{chap:NullTower} established the foundational elements of our framework: the semantic state space $\Phi$ as the linguistic medium for belief, and the Null Tower as a model for the recursive emergence of potential cognitive structure from the epistemic vacuum $\Omega$. However, these foundational structures describe the agent's capacity and origins, not necessarily its active cognitive state at any given moment. An agent interacting with the world, pursuing goals, or engaging in reflection must actively construct specific, meaningful belief states $\phi \in \Phi$.

This chapter focuses on the processes by which such structured belief ensembles are formed. How does an agent transition from the semantic silence of $\Omega$ or the abstract potentiality of the Null Tower's upper layers to possessing concrete beliefs about its environment, its tasks, or itself? We explore the primary mechanisms driving this construction, focusing initially on how external experience is encoded and integrated, and how internal processes can generate belief structures de novo. This bridges the gap between the static definition of the state space and the dynamic evolution of thought detailed in later parts.

\section{The Role of Observation Encoding (\texorpdfstring{$X$}{X})}

The most direct route for constructing beliefs grounded in the external world or specific inputs is through the observation encoder function, $X: S \rightarrow \Phi$. As introduced in Chapter~\ref{chap:SemanticStateSpace}, $X$ translates raw sensory or symbolic inputs $s \in S$ into structured, linguistic belief state ensembles $\phi = X(s)$.

This encoding process is the agent's primary interface for interpreting experience. It is fundamentally constructive:
\begin{itemize}
	\item It transforms unstructured or modality-specific data (pixels, waveforms, raw text) into interpretable linguistic expressions ($\varphi_i$).
	\item It often involves interpretation and inference beyond mere transcription, potentially adding context or identifying salient features based on the agent's architecture or prior state (though the initial state might be near $\Omega$).
	\item It is scale-sensitive, capable of generating belief states $\phi^{(k)}$ at various levels of abstraction k, depending on the nature of the input. For instance, a direct sensory input might primarily yield $\phi^{(0)}$, while processing a complex instruction might generate beliefs in $\Phi^{(1)}$ or $\Phi^{(2)}$.
\end{itemize}
The belief $\phi = X(s)$ produced by the encoder serves as the initial seed, the raw material derived from interaction, upon which further cognitive processes operate.

\textbf{Example: } Receiving a visual input $s$ (an image of a traffic light) might result in $\phi = X(s) = $ \{"Light is detected", "Color is green", "Shape is circular"\} within $\Phi^{(0)}$.

This function provides the initial grounding, populating $\Phi$ with content reflecting the agent's immediate perceptual or informational context.

\section{Initial Assimilation (\texorpdfstring{$A$}{A}) and Integration}

Receiving an encoded observation $\phi_{obs} = X(s)$ is typically not the end of the construction process. This new information must be integrated into the agent's current belief state, even if that state is minimal or close to the epistemic vacuum $\Omega$. This integration is handled by the Assimilation operator, $A: \Phi \times \Phi_{input} \rightarrow \Phi$, where $\Phi_{input}$ represents the incoming belief structure (here, $\phi_{obs}$).

The full dynamics of assimilation will be detailed in Chapter~\ref{chap:Assimilation}, but its role in initial construction is crucial. When the agent's current state $\phi_{current}$ is near $\Omega$, assimilation $A(\phi_{current}, \phi_{obs})$ might primarily involve:
\begin{itemize}
	\item \textbf{Direct Incorporation:} Adding the expressions $\{\varphi_i\}$ from $\phi_{obs}$ to form a nascent belief structure.
	\item \textbf{Elaboration:} Generating additional, contextually relevant expressions based on the input. For example, observing "Door is open" might lead to elaborating "Entry is possible." This makes the constructed belief state richer than the raw input.
	\item \textbf{Minimal Coherence Check:} Ensuring the newly added expressions do not create immediate, trivial contradictions, though complex conflicts might only be detected later.
	\item \textbf{Anchor Establishment:} Identifying key expressions within the newly assimilated content that serve as stable points for future reference or reasoning (see Section 4).
\end{itemize}
Even in these early stages, assimilation is more than simple concatenation; it is an active process of structuring and contextualizing input to form a coherent belief state $\phi_{new} = A(\phi_{current}, \phi_{obs})$.

\textbf{Example: } Starting from $\phi_{current} \approx \omega \in \Omega$, assimilating $\phi_{obs}=$ \{"Light is green"\} might yield $\phi_{new} = A(\phi_{current}, \phi_{obs}) = $ \{"Light is green", "Proceeding is likely safe"\} via elaborative assimilation ($A_{elab}$).

\section{Anchoring and Stabilization}

A key outcome of the belief construction process, particularly via $X$ and initial $A$, is the establishment of anchors. Anchors are specific expressions $\varphi_i$ or substructures within $\phi$ that gain a degree of stability or centrality. They serve as reference points around which further beliefs coalesce. Anchoring can arise through several means during construction:
\begin{itemize}
	\item \textbf{Salience:} Particularly striking or goal-relevant aspects of an observation $X(s)$ may be implicitly marked as anchors.
	\item \textbf{Repetition/Reinforcement:} If similar inputs are processed repeatedly, the common elements may become anchored abstractions.
	\item \textbf{Structural Role:} Expressions that link multiple other beliefs or form the basis of an inference chain may acquire anchor status.
	\item \textbf{Explicit Designation:} An agent's architecture might explicitly designate certain types of constructed beliefs (e.g., goals, core facts) as anchors.
\end{itemize}
These anchors are crucial because they provide the initial stability needed for the belief state to persist against the dissipative effects of Nullification ($N_t$, see Chapter~\ref{chap:Nullification}) and serve as the fixed points for coherence maintenance during subsequent assimilation and reasoning. Belief construction is therefore not just about adding content, but about building a stable, anchored semantic scaffold within $\Phi$.

\textbf{Example: } Upon receiving and assimilating the command "Goal: Deliver package to Zone A", the expression "Goal: Deliver package to Zone A" might be assigned a high anchoring score $a_i$, making it resistant to $N_t$.

\section{Internally Driven Construction}

While observation encoding $X$ provides the primary link to external reality, belief construction is not solely dependent on external stimuli. Agents can construct new belief states or modify existing ones through purely internal processes:

\begin{itemize}
	\item \textbf{Simulation and Imagination:} Generating hypothetical scenarios, counterfactuals, or future projections creates new belief ensembles within $\Phi$ (related to $A_{sim}$). These might represent plans being considered or consequences being evaluated.
	\textbf{Example: } Internally simulating "What if I turn left?" generates $\phi_{sim}=$ \{"Path blocked by obstacle", "Turning left is unsafe"\}.
	\item \textbf{Reflection and Introspection:} Examining the structure or content of the current belief state $\phi$ can lead to the construction of meta-beliefs (e.g., "I seem to be reasoning in circles," "This belief lacks sufficient grounding"), facilitated by $A_{refl}$ or $M$.
	\textbf{Example: } Introspection yields $\phi_{meta}=$ \{"Coherence ($\kappa$) of planning sector is low"\} .
	\item \textbf{Spontaneous Drift and Stabilization:} As discussed in Chapter~\ref{chap:NullTower} and detailed further in Chapter~\ref{chap:SpontaneousDrift}, semantic motion can emerge spontaneously from $\Omega$. If this drift leads to configurations that resonate with internal architectural priors or are subsequently reinforced, they can stabilize into constructed beliefs without direct external input.
	\item \textbf{Inference and Reasoning:} Deriving new conclusions $\varphi_{new}$ from existing beliefs $\{\varphi_i\} \in \phi$ is a core mechanism of internally driven construction, expanding the belief state through logical or associative steps.
	\textbf{Example: } From existing beliefs \{"All robots need power", "Unit 7 is a robot"\}, infer $\phi_{inferred}=$ \{"Unit 7 needs power"\}.
\end{itemize}

These internal mechanisms ensure that $\Phi$ is populated not just by reflections of the world, but also by the products of the agent's own cognitive activity, enabling creativity, planning, and self-modification.

\section{Relationship to Foundational Structures}

The process of belief construction operates upon the foundational structures introduced earlier:
\begin{itemize}
	\item \textbf{Semantic State Space $\Phi$:} Construction takes place within $\Phi$. The operators $X$ and $A$ populate this space with specific, structured linguistic ensembles $\phi$.
	\item \textbf{Null Tower:} The Null Tower provides the generative potential and the abstract scaffolding. Belief construction often involves grounding abstract structures potentially derived from the tower's upper layers ($\omega^{(k)}$) with specific content derived from observation ($X$) or internal generation, bringing potentiality into actuality. The initial state before construction often resides near the vacuum $\Omega$.
	\item \textbf{Semantic Geometry (Preview):} Constructed beliefs implicitly occupy positions within the conceptual geometry of $\Phi$. An observation $X(s)$ might primarily populate perceptual sectors $\Sigma_{perc}$ at low abstraction levels $\Phi^{(0)}$, while an internally generated plan might reside in $\Sigma_{plan}$ at $\Phi^{(1)}$ or $\Phi^{(2)}$. The act of construction places beliefs onto this semantic map.
\end{itemize}
Belief construction is therefore the dynamic process that breathes life into the static potential defined by $\Phi$ and the Null Tower, creating the specific configurations upon which reasoning, regulation, and action depend.

\begin{figure}[ht]
	\centering
	\input{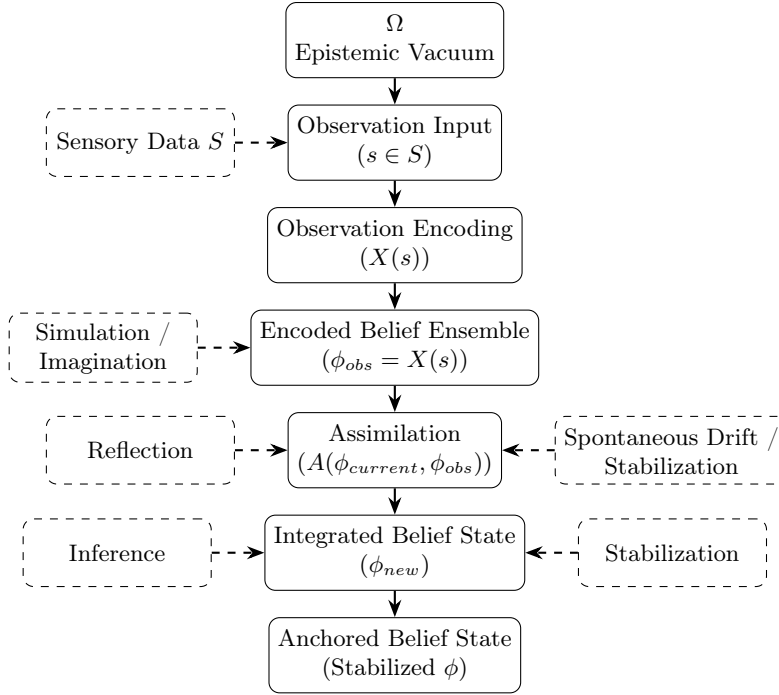}
	\caption{Belief Construction Pathway. Structured belief states (\(\phi\)) are formed through the encoding of observations (\(X(s)\)), their assimilation (\(A\)) into the active belief structure, and stabilization via anchoring. Internally generated beliefs through simulation, reflection, inference, and drift stabilization supplement externally driven construction.}
	\label{fig:belief_construction_pathway}
\end{figure}

\section{Conclusion}

Belief construction is the set of processes by which structured, meaningful belief states $\phi$ are formed within the semantic state space $\Phi$. It marks the transition from the potentiality of the epistemic vacuum $\Omega$ and the abstract structure of the Null Tower to active cognitive states. Key mechanisms include:
\begin{itemize}
	\item \textbf{Observation Encoding ($X$):} Translating external or internal signals $s$ into grounded, linguistic belief ensembles $\phi = X(s)$.
	\item \textbf{Initial Assimilation ($A$):} Integrating newly encoded observations into the current belief state, involving elaboration and coherence checks.
	\item \textbf{Anchoring:} Stabilizing key belief fragments to provide structure and persistence.
	\item \textbf{Internal Generation:} Constructing beliefs through simulation, reflection, inference, or stabilized drift, independent of direct external input.
\end{itemize}
Through these mechanisms, the agent populates $\Phi$ with the specific, dynamic belief structures required for intelligent behavior. The following parts of this monograph will delve deeper into how these constructed beliefs are organized (Part~\ref{part:structuring_belief}) and how they evolve over time through more complex dynamics (Part~\ref{part:epistemic_dynamics}).


\subsection*{Chapter Summary}
This chapter focuses on the processes by which active, structured belief states ($\phi$) are formed within the semantic manifold ($\Phi$), transitioning from the potentiality defined by the epistemic vacuum ($\Omega$) and the Null Tower. It details the primary mechanisms driving construction: Observation Encoding ($X$), which translates external inputs ($S$) into initial, grounded linguistic belief ensembles; and initial Assimilation ($A$), which integrates these encoded observations (or other inputs $\Phi_{input}$) into the current belief state, often involving elaboration and minimal coherence checks. The importance of Anchoring is highlighted, whereby certain belief fragments gain stability ($a_i$) crucial for resisting decay ($N_t$) and maintaining structure. Belief construction is not solely input-driven; the chapter also outlines internally driven mechanisms, including simulation, reflection, inference, and the stabilization of spontaneous drift ($D$). These processes operate within the established semantic space $\Phi$, giving substance to the foundational structures and populating the semantic geometry ($\Sigma$, $k$) with the specific configurations necessary for reasoning and action.
	\chapter{Parameterized Semantic Architectures}
\label{chap:ParameterizedArchitectures}

\section{Motivation: From a Universal Framework to Specific Agent Designs}

The preceding chapters introduced the foundational concepts of the semantic state space $\Phi$, the Null Tower originating from the epistemic vacuum $\Omega$, and the initial processes of belief construction. This framework provides a blueprint for agents possessing structured, linguistically-grounded belief. In this initial formulation, $\Phi$ was presented as a potentially universal structure---a canonical space for modeling epistemic cognition.

However, the principles of the Semantic Manifold can be instantiated to create diverse agent architectures with varying capabilities. One might design agents with deep reflective capacities and persistent memory, or simpler agents focused on specific tasks with limited history, or even agents designed to leverage the power of external computational engines like Large Language Models (LLMs) for certain operations. A single, fixed definition of $\Phi$ and its dynamics might not fully capture the intentional design variations possible within this overall architectural vision. Agents designed using these principles can differ significantly in memory capacity, representational richness, identity stability, and interaction modalities.

Therefore, this chapter generalizes the semantic manifold framework to explicitly accommodate these design choices. We introduce the concept of a parameterized family of belief spaces, denoted $\Phi^{[\theta]}$. Within this generalized view, each specific agent design is characterized by a configuration vector $\theta$, which determines the particular structure, dynamics, and properties of its intended semantic manifold. This allows the framework to serve as a flexible blueprint, retaining its core principles while enabling the specification and comparison of different agent designs built upon its foundation.

\section{Definition: The Parametric Belief Space Family \texorpdfstring{$\Phi^{[\theta]}$}{Phi[theta]}}

Let $\theta$ represent a tuple of structural parameters that characterize the core architectural design choices for a semantic agent relevant to belief processing. We then define the belief space associated with an agent design of type $\theta$ as follows:

\textbf{Definition 8.1 (Parameterized Semantic State Space).} The parameterized semantic state space $\Phi^{[\theta]}$ is the semantic state space whose structure, allowable states, and associated operator dynamics are determined by the parameter vector $\theta$.
$$ \Phi^{[\theta]} = \text{Semantic state space determined by configuration } \theta $$
The parameter vector $\theta$ serves to specialize the general manifold framework based on design decisions. While a complete specification may vary, key components often include:

\begin{figure}[htbp]
	\centering
	\input{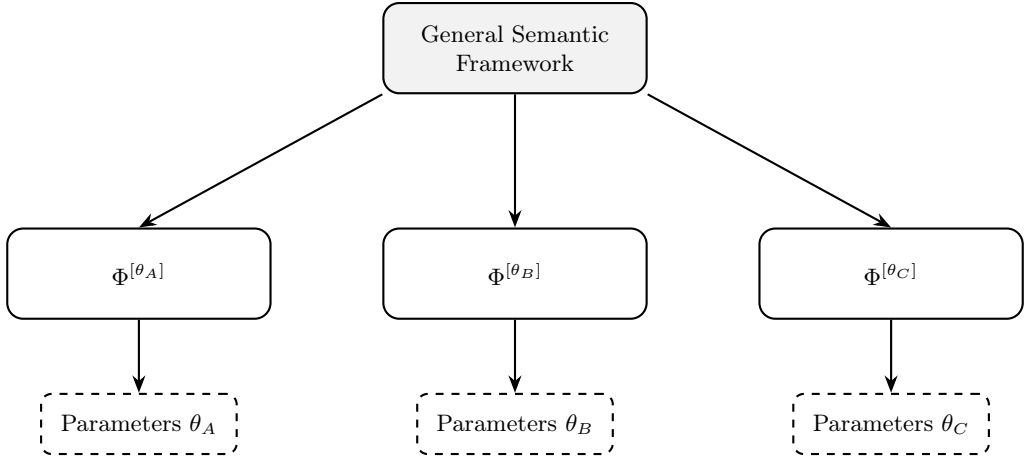}
	\caption{Parameterized semantic architectures. The general Semantic Framework branches into specific agent belief spaces \(\Phi^{[\theta]}\), each configured by a parameter vector \(\theta\) defining memory, decay, identity, representation, and other architectural properties.}
	\label{fig:parameterized-architectures}
\end{figure}

\begin{itemize}
	\item[$\mu$:] \textbf{Memory Horizon:} Specifies the designed timescale or capacity limit for belief retention and accessibility within $\phi \in \Phi^{[\theta]}$. Design choices might range from finite horizons to effectively infinite persistence ($\mu = \infty$).
	\item[$\delta_{profile}$:] \textbf{Decay Profile:} Specifies the designed behavior of the nullification operator $N_t$. Options include gradual decay modulated by anchoring $a_i$, threshold-based forgetting, or context-dependent expiration rules.
	\item[$\tau$:] \textbf{Trajectory Persistence:} Defines the extent to which the agent's belief evolution is designed to be history-sensitive. High $\tau$ implies strong path dependence is a design goal, while $\tau \approx 0$ might characterize a deliberately reactive design.
	\item[$\rho$:] \textbf{Representation Mode:} Specifies the chosen format(s) for encoding linguistic expressions $\varphi_i$: symbolic, embedding-based, graphical, hybrid, or multimodal. This reflects implementation choices.
	\item[$\eta_{type}$:] \textbf{Identity Model:} Dictates the intended nature and stability of the epistemic identity vector $\vec{\eta}$. Design options include static, emergent, decaying, or no persistent identity ($\eta_{type} = \text{none}$).
	\item[$C$:] \textbf{Context Scope:} Determines the designed range or window within $\Phi^{[\theta]}$ accessible for current processing. This could be fixed or dynamic by design.
	\item[$\Gamma_{type}$:] \textbf{Goal Decomposition Model:} Specifies the intended structure of the agent's planning capabilities via the goal decomposition head $\Gamma$.
	\item[$O$:] \textbf{Observation Modalities:} Describes the input channels $S$ the agent is designed to process via its encoder $X: S \rightarrow \Phi^{[\theta]}$, such as text, vision, sensorimotor data, or tool outputs.
\end{itemize}

\begin{table}[ht]
	\centering
	\begin{tabular}{ll}
		\hline
		\textbf{Parameter} & \textbf{Description} \\ \hline
		$\mu$ & Designed memory persistence / timescale \\
		$\delta_{profile}$ & Nature of belief decay / forgetting \\
		$\tau$ & Degree of history dependence in dynamics \\
		$\rho$ & Format for representing belief expressions $\varphi_i$ \\
		$\eta_{type}$ & Model for epistemic identity stability \\
		$C$ & Scope/window for active processing \\
		$\Gamma_{type}$ & Structure of goal decomposition / planning system \\
		$O$ & Input observation modalities handled by $X$ \\ \hline
	\end{tabular}
	\caption{Key Architectural Parameters in $\theta$}
	\label{tab:key-architectural-parameters}
\end{table}

Each distinct instantiation of $\theta$ represents a specific agent design within the Semantic Manifold family, inducing particular geometric properties, operator behaviors, and cognitive capabilities.

\section{Influence of Representation Mode (\texorpdfstring{$\rho$}{rho}) on Manifold Properties}

\label{sec:rho_influence}

The choice of representation mode $\rho$ within the parameter vector $\theta$ has profound implications for how the abstract structures and dynamics of the semantic manifold $\Phi^{[\theta]}$ are realized. While the conceptual framework (sectors $\Sigma$, layers $\Phi^{(k)}$, metric $d$, flow $F$, operators) remains consistent, its manifestation varies significantly:

\begin{itemize}
	\item \textbf{Continuity and Topology:} If $\rho$ is embedding-based, $\Phi^{[\theta]}$ can be conceptualized more readily as a continuous, high-dimensional space where operators might correspond to smooth transformations and flow fields $F$ have clearer geometric meaning. If $\rho$ is symbolic (e.g., graph-based), the space may be discrete or combinatorial. Topology might involve graph connectivity rather than Euclidean neighborhoods. Hybrid $\rho$ leads to mixed continuous/discrete spaces.
	\item \textbf{Realization of Sectors ($\Sigma$):} In symbolic systems, sectors might be implemented via explicit tags, namespaces, or subgraphs. In embedding spaces, sectors could correspond to learned subspaces, potentially identified via clustering or projection methods, allowing for "softer" or overlapping sector boundaries.
	\item \textbf{Realization of Scaling ($\Phi^{(k)}$):} Abstraction layers $k$ in symbolic systems might involve explicit type hierarchies or rule generalization. In embedding spaces, layers could correspond to stacked encoders ($\Lambda$) / decoders ($V$) or representations at different levels of representational granularity (e.g., token vs. sentence vs. document embeddings).
	\item \textbf{Semantic Metric ($d$):} As discussed in Appendix~\ref{app:GeometricTopological} and Chapter~\ref{chap:LearningSemanticStructures}, the definition and learnability of $d$ depend heavily on $\rho$. Vector metrics (cosine, Euclidean) apply naturally to embeddings, while symbolic metrics (edit distance, graph similarity) require different approaches.
	\item \textbf{Operator Implementation:} The algorithms implementing operators like $A, M, N_t, \Lambda, V, R, Q$ differ fundamentally based on $\rho$, shifting from rule-based inference and graph manipulation (symbolic) to neural network computations and vector similarity searches (embedding), as detailed further in Appendix~\ref{app:ImplementationExamples}.
\end{itemize}
Therefore, while the abstract Semantic Manifold framework provides the blueprint, the choice of $\rho$ dictates the specific mathematical and computational "material" from which the agent's cognitive space is built, influencing its properties and the methods available for learning and adaptation.

\section{Illustrative Examples of Parameterized Architectures} 

To clarify the role of $\theta$ in defining agent designs, consider these distinct possibilities:

\subsection{Stateless Semantic Interface Agent}
One might design a simple agent intended primarily as an interface, perhaps using an LLM backend for core processing, with limited persistence:
$$ \theta_{Interface} = (\mu=\text{Short Term}, \delta_{profile}=\text{Rapid Expiration}, \tau \approx 0, $$
$$ \rho=\text{Embedding-Based (via LLM)}, \eta_{type}=\text{None}, $$
$$ C=\text{Short Window}, \Gamma_{type}=\text{Direct Prompting}, O=\text{Text}) $$
The corresponding $\Phi^{[\theta_{Interface}]}$ is designed to be transient. While potentially leveraging a powerful LLM for $\rho$, the agent architecture itself is specified with minimal memory ($\mu$), rapid forgetting ($\delta_{profile}$), low trajectory persistence ($\tau$), and no stable identity ($\eta_{type}$). Its function is primarily reactive processing within its context scope $C$.

\subsection{Memory-Persistent Reflective Agent}
An agent designed for long-term reasoning and self-awareness according to Semantic Manifold principles might have parameters such as:
$$ \theta_{Reflective} = (\mu=\infty, \delta_{profile}=\text{Gradual Anchored Decay}, \tau \gg 0, $$
$$ \rho=\text{Symbolic + Embedding Hybrid}, \eta_{type}=\text{Persistent Emergent}, $$
$$ C=\text{Attention}, \Gamma_{type}=\text{Learned Hierarchical}, O=\text{Multimodal}) $$
Here, the design specifies that $\Phi^{[\theta_{Reflective}]}$ supports infinite memory ($\mu$), gradual forgetting ($\delta_{profile}$), strong history dependence ($\tau$), a rich representation ($\rho$), and a stable identity ($\eta_{type}$). This architecture is intended to enable long-range coherence and deep reflection, potentially implemented using various technologies (including LLMs + memory systems for $\rho$ and $A$).

\subsection{Tool-Augmented Semantic Agent}
An agent designed to orchestrate external tools could be parameterized as:
$$ \theta_{Orchestrator} = (\mu=\text{Extended Context}, \delta_{profile}=\text{Priority-Gated Decay}, \tau=\text{Moderate}, $$
$$ \rho=\text{Hybrid}, \eta_{type}=\text{Persistent Emergent}, \dots, O=\text{Text+Tools}) $$
This design includes tool outputs in its observation modality $O$. The assimilation process $A^{[\theta]}$ is designed to incorporate $\phi_{tool}$, triggered by specific states corresponding to tool affordances (Part~\ref{part:regulation_and_control}). The agent's effective memory and context might be designed to be dynamically extended ($\mu$, $C$) by this external information.

\section{Parameterized Operators and Metrics}

The introduction of $\Phi^{[\theta]}$ implies that the core semantic operators and metrics defined abstractly become relativized to the agent's specific architectural design $\theta$. We denote these parameterized versions as $O^{[\theta]}$:

\begin{itemize}
	\item \textbf{Dynamics Operators:} The behavior of operators like Assimilation ($A^{[\theta]}$), Nullification ($N_t^{[\theta]}$), Annihilation ($K^{[\theta]}$), Abstraction ($\Lambda^{[\theta]}$), and Elaboration ($V^{[\theta]}$) operate according to the functional specifications and constraints imposed by $\theta$. For example, $N_t^{[\theta_{Interface}]}$ might implement simple context expiration based on $\mu$ and $C$, while $N_t^{[\theta_{Reflective}]}$ implements gradual decay modulated by anchoring $a_i$ according to $\delta_{profile}$.
	\item \textbf{Epistemic Metrics:} Coherence $\kappa^{[\theta]}(\phi)$, semantic effort $\epsilon^{[\theta]}(\phi)$, and cognitive load $\lambda^{[\theta]}(\phi)$ are evaluated relative to the agent's designed representational capacity ($\rho$) and memory/context constraints ($\mu, C$). What constitutes "high load" or "low coherence" depends on the architecture specified by $\theta$.
	\item \textbf{Identity and Trajectory:} The epistemic identity vector $\vec{\eta}^{[\theta]}$ and its stability depend on the designed $\eta_{type}$ and $\tau$. The belief trajectory $\gamma^{[\theta]}(t)$ unfolds according to the specific dynamics defined for $\Phi^{[\theta]}$.
\end{itemize}
This parameterization allows the framework to describe, for example, an $N_t^{[\theta_{Interface}]}$ behaving as context expiration versus an $N_t^{[\theta_{Reflective}]}$ behaving as gradual decay, while stemming from the same core concept of Nullification.

\section{Theoretical Implications of Parameterization} 

Acknowledging the design-relativity of the semantic manifold via $\Phi^{[\theta]}$ has several important theoretical consequences for the framework:

\subsection{Multiple Architectures within One Framework}
The Semantic Manifold concept is preserved as a unifying architectural blueprint, but it now describes a family of potential agent designs rather than a single structure. The core principles and operators remain consistent, but their specific realization and effects are determined by the design choices encoded in $\theta$.

\subsection{Unified Design Language for Diverse Agents}
This approach provides a common formal language for specifying and comparing different intelligent agent designs built on semantic principles, regardless of their underlying implementation technology (which might include LLMs, symbolic reasoners, specific memory systems, etc.).

\subsection{Compatibility with Learning and Development}
Parameterization provides a natural interface for considering learning within this architecture (Part~\ref{part:learning_and_adaptation}). Agent development could involve learning optimal operational policies within a fixed $\Phi^{[\theta]}$ (Chapter~\ref{chap:LearningRegulatoryPolicies}) or potentially even adapting some parameters of $\theta$ itself based on experience, representing architectural learning (Chapter~\ref{chap:ArchitecturalAdaptation}).

\subsection{Generalizing Epistemic Identity}
By defining identity $\vec{\eta}^{[\theta]}$ relative to the agent's designed parameters $\eta_{type}$ and trajectory persistence $\tau$, the concept can be applied meaningfully across different designs, capturing persistent identity in reflective agents or perhaps more transient functional roles in simpler agents.

\section{Conclusion: A Generalized Foundation} 

The introduction of parameterized semantic architectures $\Phi^{[\theta]}$ provides a crucial generalization of the foundational framework presented in Part~\ref{part:semantic_foundations}. By explicitly acknowledging that the structure and dynamics of the belief space can be tailored via design parameters $\theta$ (governing memory, decay, representation, identity, context, planning, and observation), we establish a more flexible and powerful theoretical blueprint.

This parameterization allows the core concepts of semantic scaling, sectors, geometry, dynamics, regulation, and meta-cognition explored in the subsequent Parts to be understood as components applicable across a family of possible agent designs derived from the Semantic Manifold principles. It paves the way for specifying and analyzing diverse cognitive architectures, including those potentially leveraging technologies like LLMs as implementation components, within a unified conceptual scheme. This lays a more robust foundation for the theory of structured belief as the monograph proceeds to detail the internal structure and dynamics operating within a given $\Phi^{[\theta]}$.


\subsection*{Chapter Summary}
This chapter generalizes the semantic manifold framework by introducing parameterized semantic architectures, $\Phi^{[\theta]}$. The motivation is to move beyond a single universal structure to accommodate diverse agent designs with varying capabilities in memory, representation, identity stability, and interaction modalities. The chapter defines the parameter vector $\theta$, which characterizes specific architectural design choices, including memory horizon ($\mu$), decay profile ($\delta_{profile}$), trajectory persistence ($\tau$), representation mode ($\rho$), identity model ($\eta_{type}$), context scope ($C$), goal decomposition model ($\Gamma_{type}$), and observation modalities ($O$). It explores how the choice of representation mode ($\rho$) significantly influences the concrete realization of the manifold's properties and the implementation of operators. Illustrative examples demonstrate how different settings of $\theta$ can define distinct agent types, such as stateless interfaces or memory-persistent reflective agents. Consequently, the core semantic operators (e.g., $A^{[\theta]}, N_t^{[\theta]}$) and metrics (e.g., $\kappa^{[\theta]}, \lambda^{[\theta]}$) become relativized to the specific architecture $\theta$. This parameterization provides a unified design language, clarifies the scope for different types of learning (policy vs. architectural), and offers a generalized perspective on epistemic identity ($\vec{\eta}^{[\theta]}$) within the framework.
	
	\part{Structuring Belief: Modularity and Abstraction}
	\label{part:structuring_belief}
	
	\chapter{Grounding Semantic Belief}
\label{chap:GroundingSemanticBelief}

\section{Introduction: The Problem of Meaning}

Part~\ref{part:semantic_foundations} laid the foundations for the semantic state space $\Phi$, a structured representation of an agent's beliefs realized as linguistic ensembles $\{\varphi_i\}$. Chapter~\ref{chap:ParameterizedArchitectures} generalized this by introducing parameterized architectures $\Phi^{[\theta]}$ to account for diverse agent designs. However, for these internal representations to constitute meaningful beliefs about something beyond the agent's internal dynamics, they must be appropriately connected, or grounded, in sources of information pertaining to the external world, the agent's embodiment, or its interactions. Without grounding, the symbols and structures within $\Phi$ risk becoming a detached, uninterpretable formal system---a "symbol grounding problem" internal to the agent.

Chapter~\ref{chap:SemanticStateSpace} introduced the Observation Encoding operator $X: S \rightarrow \Phi$ as the initial bridge between raw input $S$ and structured belief. This chapter expands on the concept of grounding much more broadly. We explore multiple pathways through which belief states $\phi \in \Phi^{[\theta]}$ acquire semantic significance and maintain connection to their referents, including direct sensorimotor interaction, linguistic immersion, internal simulation, and social alignment. Understanding these diverse grounding mechanisms is crucial for designing agents whose beliefs are not merely internally coherent ($\kappa$) but also meaningful and pertinent to their operational context, influenced by the specific observation modalities $O$ defined in their parameter vector $\theta$.

\section{Sensorimotor Grounding via Observation Encoding (\texorpdfstring{$X$}{X})}

The most direct form of grounding, particularly relevant for embodied agents or those interacting with physical environments (i.e., $\theta$ specifying non-textual $O$), occurs via the Observation Encoding operator $X$ acting on sensorimotor input $s \in S_{sensorimotor}$. As defined in Chapter~\ref{chap:SemanticStateSpace}, $X$ transforms raw sensory data (visual, auditory, proprioceptive, etc.) into initial, low-level belief fragments within $\Phi^{[\theta]}$, often populating perceptual sectors like $\Sigma_{perc}$.
\begin{itemize}
	\item \textbf{Connecting Symbols to Percepts:} This pathway directly links linguistic elements $\varphi_i$ (like "red," "sphere," "location (3,4,1)") to patterns detected in sensory streams.
	\item \textbf{Shaping Low-Level Abstractions:} Consistent patterns in sensorimotor input provide the raw material from which initial abstractions ($\Lambda$) can be learned, grounding concepts like "objectness," "motion," or basic spatial relations within lower levels $\Phi^{(k)}$. Crucially, the grounded representations $\phi^{(0)}$ generated by $X$ serve as the essential foundation upon which the abstraction operator $\Lambda$ (Chapter~\ref{chap:SemanticScaling}) builds higher-level, more general concepts.
	\item \textbf{Role of Embodiment:} The specific nature of the agent's sensors and effectors (part of its embodiment, related to $\theta$) shapes the type of information available in $S_{sensorimotor}$ and thus the nature of the grounding achieved via $X$.
\end{itemize}
Sensorimotor grounding provides a direct, non-linguistic anchor for beliefs, tethering the semantic manifold to the agent's immediate perceptual reality and providing the base material for semantic scaling.

\section{Linguistic Grounding from Corpora and Communication}

For agents whose primary input modality is text ($O$ in $\theta$), or those leveraging LLMs as components (relevant for certain $\theta$ designs with format $\rho$ involving embeddings), grounding occurs differently. Meaning arises not primarily from direct perception, but from the statistical patterns, distributional semantics, and relational structures inherent in large volumes of linguistic data (corpora) or ongoing communication.
\begin{itemize}
	\item \textbf{Distributional Semantics:} The meaning of a belief fragment $\varphi_i$ is partly determined by its pattern of co-occurrence with other fragments within the assimilated text corpus (resulting from repeated application of $A_{text}$). Concepts are implicitly defined by their linguistic contexts ("You shall know a word by the company it keeps" - Firth). If $\rho$ involves embeddings, these are typically learned from such distributional patterns.
	\item \textbf{Structural Induction:} Assimilation ($A$) of vast linguistic data can lead to the emergence of semantic structures (relations, hierarchies) within $\Phi^{[\theta]}$ that reflect the knowledge implicitly encoded in the language. This provides a form of grounding based on the collective knowledge embedded in the corpus.
	\item \textbf{Grounding via Communication:} Beliefs can be grounded or validated through successful communication and alignment with other agents (see Part~\ref{part:social_cognition}). If an agent's internal belief $\phi$ allows it to communicate effectively (e.g., make predictions understood by others, answer questions coherently), that belief gains pragmatic grounding.
\end{itemize}
Linguistic grounding connects beliefs to the shared conceptual and factual structures embedded within human language, providing a powerful, albeit potentially indirect, link to the world described by that language. Agents heavily reliant on this may face challenges connecting to non-linguistic reality directly.

\section{Grounding through Embodied Simulation}

Embodied Simulation (detailed in Chapter~\ref{chap:EmbodiedSimulation}) offers another pathway for grounding, particularly for abstract concepts or testing hypothetical beliefs. By running internal simulations of actions and interactions ($\gamma_{sim}(t)$ within $\Phi^{[\theta]}$), the agent can generate "pseudo-experience" that connects beliefs to potential sensorimotor consequences, even without direct external input.
\begin{itemize}
	\item \textbf{Grounding Abstract Concepts:} Abstract beliefs $\phi^{(k)}$ (e.g., "fragility") can be grounded by simulating scenarios (potentially involving elaboration via $V$ from Chapter~\ref{chap:SemanticScaling}) where the concept applies (e.g., simulating dropping an object represented as fragile) and observing the simulated outcome (object breaks). This links the abstract $\varphi_i$ at higher levels $\Phi^{(k)}$ to concrete simulated dynamics and sensorimotor contingencies at lower levels $\Phi^{(0)}_{sim}$, providing grounding through internal experience.
	\item \textbf{Testing Hypothetical Beliefs:} Agents can simulate the consequences of holding a particular belief or performing an action based on it. If the simulated outcome aligns with known constraints or desired states, the belief gains a form of internal grounding.
	\item \textbf{Connecting Language to Action:} Simulation allows grounding linguistic commands or plans (within $\Sigma_{plan}$) by projecting their likely physical outcomes, connecting language to potential sensorimotor trajectories.
\end{itemize}
Simulation acts as an internal "reality check," allowing the agent to ground beliefs through virtual interaction and prediction based on its internal world models, which themselves are hopefully grounded via $X$ or linguistic assimilation $A$.

\section{Social and Communicative Grounding}

For agents operating in multi-agent environments (Part~\ref{part:social_cognition}), a significant aspect of grounding comes from social interaction and the need for shared understanding. Beliefs become grounded not just by individual perception or linguistic statistics, but by their role in successful communication and coordination.
\begin{itemize}
	\item \textbf{Alignment and Shared Beliefs:} Beliefs are validated or grounded when they align with the beliefs expressed or implied by trusted interlocutors. Successful communication implies some level of shared grounding (explored further in Chapter~\ref{chap:CommunicationAlignment}).
	\item \textbf{Pragmatic Grounding:} A belief is pragmatically grounded if acting upon it leads to successful interaction or coordination with other agents.
	\item \textbf{Instruction and Correction:} Beliefs formed via instruction or corrected through social feedback acquire grounding based on the perceived reliability or authority of the source.
\end{itemize}
Social grounding anchors individual belief systems within a broader intersubjective or cultural context.

\section{Interaction and Integration of Grounding Pathways}

In sophisticated agents (likely those with multimodal $O$ in $\theta$), these grounding pathways do not operate in isolation. They interact and mutually constrain each other:
\begin{itemize}
	\item Sensorimotor input ($X$) grounds linguistic terms ($\varphi_i$) related to perception and action, providing the basis for abstraction ($\Lambda$).
	\item Linguistic input ($A_{text}$) provides abstract knowledge that can structure the interpretation of perception or guide simulation (often requiring elaboration $V$).
	\item Simulation (driven by $V$) tests beliefs acquired linguistically against embodied constraints, reinforcing or challenging their grounding.
	\item Social feedback validates or corrects beliefs grounded through other means.
\end{itemize}
The agent's overall architecture ($\theta$) and learning history (Part~\ref{part:learning_and_adaptation}) determine the relative strengths and interactions between these pathways. Maintaining coherence ($\kappa$) requires reconciling information arriving through different grounding routes, potentially involving corrective assimilation ($A_{corr}$) or meta-cognitive assessment ($M$).

\section{Conclusion: Grounding as Meaning-Making}

Grounding is the multifaceted process by which the internal representations within the semantic state space $\Phi^{[\theta]}$ acquire meaning and maintain connection to the reality they aim to represent. It establishes the semantic significance of the agent's internal linguistic ensembles $\{\varphi_i\}$. This chapter has outlined four major pathways:
\begin{enumerate}
	\item \textbf{Sensorimotor Grounding:} Direct connection via observation encoding ($X$) of sensory/proprioceptive input $S$, providing the foundation for abstraction ($\Lambda$).
	\item \textbf{Linguistic Grounding:} Deriving meaning from statistical patterns in language corpora or communication via assimilation ($A_{text}$).
	\item \textbf{Simulation Grounding:} Connecting abstract concepts (potentially generated via $V$) to potential sensorimotor consequences via internal embodied simulation.
	\item \textbf{Social Grounding:} Validating beliefs through successful communication, coordination, alignment, and feedback within multi-agent interactions.
\end{enumerate}

\begin{figure}[ht]
	\centering
	\input{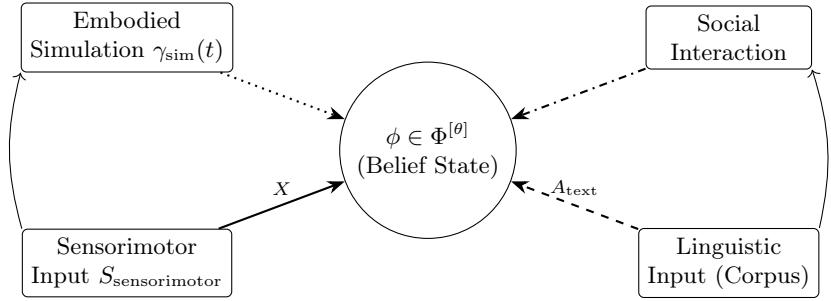}
	\caption{Multiple pathways of semantic grounding in belief formation. A belief state $\phi \in \Phi^{[\theta]}$ acquires meaning through four principal grounding routes: direct sensorimotor experience via Observation Encoding ($X$), linguistic assimilation from textual corpora and communication ($A_{\text{text}}$), internal embodied simulation of hypothetical scenarios ($\gamma_{\text{sim}}(t)$), and social interaction through feedback and alignment. These pathways often interact, mutually constraining the semantic structure of belief.}
	\label{fig:grounding_pathways}
\end{figure}

These pathways often interact, providing mutual constraints and richer meaning. The specific mechanisms and their balance are influenced by the agent's design parameters $\theta$, particularly its observational modalities $O$ and representational mode $\rho$. A successful semantic agent must possess robust grounding mechanisms appropriate to its embodiment and environment. Without adequate grounding, even an internally coherent belief system remains detached and semantically empty. Understanding these pathways, and their crucial link to semantic scaling ($\Lambda, V$), is therefore fundamental to building agents capable of meaningful, contextually relevant structured belief. Having established parameterization (Chapter~\ref{chap:ParameterizedArchitectures}) and grounding (this chapter), we can now explore the internal structuring of belief through scaling and sectors in the remainder of this Part.


\subsection*{Chapter Summary}
This chapter addresses the critical problem of grounding: how internal linguistic belief representations ($\phi \in \Phi^{[\theta]}$) acquire meaning and connect to the world, thus avoiding the symbol grounding problem. It expands beyond the initial Observation Encoding ($X$) to explore multiple grounding pathways. Direct sensorimotor grounding links perceptual terms to sensory input via $X$, providing the basis for initial abstractions ($\Lambda$). Linguistic grounding derives meaning from statistical patterns in text corpora (integrated via $A_{text}$) and successful communication, particularly relevant for language-based agents or architectures involving LLMs. Embodied Simulation (Chapter~\ref{chap:EmbodiedSimulation}) offers internal grounding by linking abstract concepts or plans to potential sensorimotor consequences through simulated trajectories ($\gamma_{sim}$). Social and communicative grounding arises in multi-agent contexts through alignment, successful interaction, and feedback. The chapter emphasizes that these pathways typically interact and mutually constrain each other, with the specific balance influenced by the agent's design ($\theta$). Robust grounding is essential for ensuring that beliefs within $\Phi^{[\theta]}$ are not merely internally coherent but semantically significant and pertinent to the agent's context.
	\chapter{Semantic Scaling}
\label{chap:SemanticScaling}

\section{Introduction: The Vertical Dimension of Belief}

Intelligent agents, whether natural or artificial, must navigate a world rich in detail while extracting patterns, forming generalizations, and applying abstract knowledge. The ability to operate across different levels of specificity---to move from concrete instances to general principles and back again---is fundamental to cognition. This chapter introduces semantic scaling, a formal framework for modeling this vertical dimension of belief within the semantic state space $\Phi$.

Semantic scaling conceptualizes the belief space not as flat, but as layered, indexed by abstraction level $k$. Each stratum $\Phi^{(k)}$ in this hierarchy represents a distinct level of semantic generality, ranging from raw, detailed perceptual encodings ($\Phi^{(0)}$)---often derived directly from grounded experience via observation encoding $X$ as discussed in Chapter~\ref{chap:GroundingSemanticBelief}---to highly compressed conceptual schemas, rules, or self-models at higher levels ($\Phi^{(k)}, k>0$). This vertical structure complements the functional modularity provided by semantic sectors (discussed in Chapter~\ref{chap:SemanticSectors}), yielding a richer, multi-dimensional architecture for belief.

Within this framework, core cognitive operations like generalization, compression, elaboration, and re-framing are modeled as explicit transformations between layers. Upward operators ($\Lambda$) compress information, extracting salient structure from more grounded or specific levels, while downward operators ($V$) elaborate abstract concepts into concrete instances or actionable details, potentially connecting them back to (simulated or actual) sensorimotor grounding (Chapter~\ref{chap:GroundingSemanticBelief}). Semantic scaling provides agents with the necessary structure to:
\begin{itemize}
	\item Generalize from specific, grounded experiences ($\Phi^{(0)}$) to form abstract knowledge ($\Phi^{(k)}, k>0$).
	\item Compress information to manage cognitive load and memory constraints.
	\item Elaborate abstract plans or intentions ($\Phi^{(k)}, k>0$) into concrete steps or simulated experiences ($\Phi^{(0)}$).
	\item Engage in depth-sensitive control, adjusting behavior based on the required level of detail or generality.
\end{itemize}
This chapter develops the formal infrastructure for semantic scaling, defining the hierarchy of layers ($\Phi^{(k)}$), the scaling operators ($\Lambda, V$), and the dynamics of navigating this vertical dimension that links grounded interaction with abstract thought. Abstraction is not an epiphenomenon; it is the topography of the semantic landscape built upon grounded experience.

\begin{figure}[h!]
	\centering
	\input{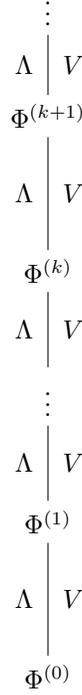}
	\caption{Diagram illustrating the behavior of the abstraction operator ($\Lambda$) and elaboration operator ($V$) at various levels $\Phi^{(k)}$ of the stratified semantic state space $\Phi$.}
	\label{fig:semantic_scaling}
\end{figure}

\section{Philosophical and Cognitive Foundations}

The idea that belief and reasoning operate across levels of abstraction is deeply rooted in philosophy and cognitive science. Abstraction is not merely simplification; it is a transformative process of semantic compression and conceptual elevation. Moving from specific instances ("This specific crow is black," "That specific crow is black") to a generalization ("All crows are black") involves discarding instance-specific details while preserving and elevating a structural pattern. This allows agents to reason more efficiently, plan over classes rather than individuals, and apply knowledge across contexts.

Cognitive development provides compelling evidence for layered abstraction. Theories like Piaget's stages describe a progression from sensorimotor grounding to symbolic manipulation and eventually to formal logical operations, with each stage representing a more powerful level of semantic compression and generalization built upon the previous one. Vygotsky emphasized the role of language and social interaction in mediating the construction of abstract concepts. Experience is interpreted through multiple layers simultaneously; the statement "The system is offline" can be a raw observation ($\Phi^{(0)}$), a diagnosis of a state ($\Phi^{(1)}$), or trigger for a high-level recovery policy ($\Phi^{(2)}$).

Functionally, abstraction enables generalization, prediction, efficient planning, and analogical reasoning. Without it, learning would not scale. Furthermore, the processes of abstraction ($\Lambda$) and elaboration ($V$) exhibit a fundamental asymmetry. Abstraction is typically lossy, mapping multiple concrete states to a single abstract form. Elaboration is often generative, unfolding a single abstract concept into multiple possible instances or implications. An intelligent agent must master both ascent (abstraction from grounded input) and descent (elaboration towards grounded action or simulation) to navigate the complexities of its world effectively. Semantic scaling provides the formal structure for this epistemic navigation.

\section{Formal Definition of Semantic Scaling}

We formalize semantic scaling by structuring the semantic state space $\Phi$ into a hierarchy of layers indexed by abstraction level $k \in \mathbb{Z}_{\ge 0}$.

\textbf{Definition (Abstraction Layers):} The semantic state space $\Phi$ is the union of disjoint subspaces $\Phi^{(k)}$, where each $\Phi^{(k)}$ contains belief states at abstraction level k:
$$ \Phi = \bigcup_{k=0}^{\infty} \Phi^{(k)} $$
(In practice, the maximum $k$ may be bounded).
\begin{itemize}
	\item $\Phi^{(0)}$ represents the most concrete level: raw or minimally processed perceptual data (often from $X$), specific episodic facts, grounded details.
	\item $\Phi^{(k)}$ for $k > 0$ represents progressively more abstract levels: summaries, generalizations, categories, rules, schemas, principles, self-models.
\end{itemize}
We denote a belief state or substructure residing at level $k$ as $\phi^{(k)} \in \Phi^{(k)}$.

Transitions between these layers are mediated by scaling operators:

\textbf{Definition (Abstraction Operator $\Lambda$):} The upward abstraction operator $\Lambda: \Phi \rightarrow \Phi$ maps a belief state $\phi^{(k)}$ to a more abstract representation $\phi^{(k+1)}$. This typically involves compression, generalization, or suppression of detail, often operating on grounded input from level $i=0$.

\textbf{Definition (Elaboration Operator $V$):} The downward elaboration operator $V: \Phi \rightarrow \Phi$ maps an abstract belief state $\phi^{(k)}$ to a more concrete representation $\phi^{(k-1)}$. This typically involves instantiation, expansion, grounding (e.g., linking to simulation or action primitives at level $k=0$), or adding specific detail.

These operators are generally not inverses. $\Lambda$ is often many-to-one, and $V$ may be one-to-many or stochastic (generating a distribution over possible elaborations). They aim to preserve semantic content appropriate to the target level, but not necessarily the exact form or all information.

Furthermore, an agent's belief state $\phi$ at any given time might span multiple abstraction levels:
$$ \phi = \bigcup_k \phi^{(k)}, \quad \text{where } \phi^{(k)} = \phi \cap \Phi^{(k)} \text{ (abusing notation)}, $$
where $\phi \cap \Phi^{(k)}$ is an abuse of notation denoting the fragments in $\phi$ originating from $\Phi^{(k)}.$
This allows for mixed-resolution representations, where an agent might simultaneously hold detailed perceptions and abstract goals.

\section{Scaling Operators and Compositional Properties}

The dynamics of moving through abstraction levels are governed by the properties of the $\Lambda$ and $V$ operators.
\begin{itemize}
	\item \textbf{Composition:} Operators can be chained. Abstraction can occur stepwise ($\Lambda^k = \Lambda \circ \dots \circ \Lambda$, $k$ times), and similarly for elaboration ($V^k = V \circ \dots \circ V$, $k$ times). This supports hierarchical processing.
	\item \textbf{Partial Invertibility:} While not true inverses, compositions like $V^k \circ \Lambda^k$ might approximate the identity on $\Phi^{(i)}$ if the abstraction was sufficiently structure-preserving. Conversely, $\Lambda^k \circ V^k (\phi^{(i)})$ yields $\phi^{(i)}$ only if $\phi^{(i)}$ was itself produced by abstraction from level $i$. This captures the lossy nature of abstraction and the generative nature of elaboration.
	\item \textbf{Semantic Deformation:} Moving up or down the abstraction scale typically deforms the semantic content. Abstraction ($\Lambda$) tends to smooth distinctions and discard context. Elaboration ($V$) tends to introduce variation and hypothetical detail. This deformation can potentially be measured using a semantic distance function $d$, such that $d(\phi^{(i)}, V^k(\Lambda^k(\phi^{(i)})))$ quantifies the information loss or change due to the round trip.
	\item \textbf{Locality:} Scaling operators can act globally on $\Phi$ or be restricted to specific semantic sectors $\Sigma$, denoted $\Lambda^k|_{\Sigma}: \Sigma^{(i)} \rightarrow \Sigma^{(i + k)}$ and $V^k|_{\Sigma}: \Sigma^{(i)} \rightarrow \Sigma^{(i - k)}$. This allows for domain-specific abstraction or elaboration (e.g., abstracting within the planning sector $\Sigma_{plan}$).
	\item \textbf{Compositional Construction:} Beliefs can be dynamically maintained across levels through cycles of abstraction and elaboration, allowing for refinement and consistency checking, e.g., $\phi^{(i)}_{new} \approx V^k(\Lambda^k(\phi^{(i)}_{old})) + \Delta^{(i)}$, where $\Delta^{(i)}$ represents updates at level $i$.
\end{itemize}
The interplay of these operators defines the agent's ability to navigate the vertical dimension of its belief space, balancing detail and generality.

\section{Scaling Within and Across Semantic Sectors}

Semantic scaling interacts deeply with the functional modularity provided by semantic sectors ($\Sigma$, introduced conceptually in Chapter~\ref{chap:SemanticStateSpace} and detailed in Chapter~\ref{chap:SemanticSectors}). Beliefs are situated not just at an abstraction level $k$ but within a sector $\Sigma$, residing in the subspace $\Sigma^{(k)} = \Sigma \cap \Phi^{(k)}$.

\begin{figure}[h!]
	\centering
	\input{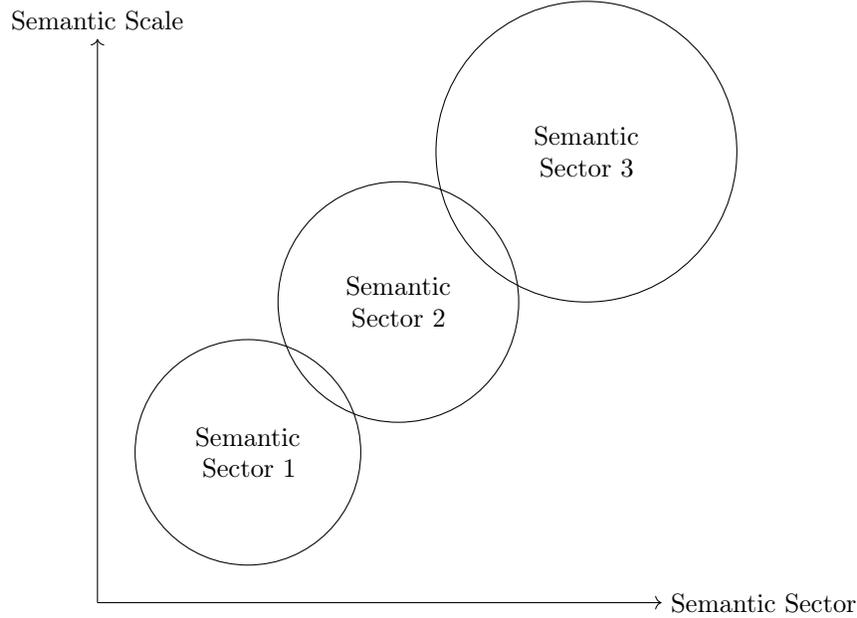}
	\caption{Overlapping semantic sectors across spatial and semantic scale dimensions.}
	\label{fig:scale_sector_space}
\end{figure}

Scaling can occur within a sector:
\begin{itemize}
	\item \textbf{Intra-Sector Scaling:} Operators $\Lambda^k|_{\Sigma}$ and $V^k|_{\Sigma}$ transform beliefs while keeping them within the same functional domain. Examples include abstracting specific perceived events into a narrative summary ($\Sigma_{narr}$), elaborating an abstract plan into concrete subgoals ($\Sigma_{plan}$), or generalizing reflective insights ($\Sigma_{refl}$).
\end{itemize}

Scaling can also integrate information across sectors:
\begin{itemize}
	\item \textbf{Cross-Sector Abstraction/Elaboration:} Operators might map content between sectors as part of the scaling process. For example, $\Lambda: \Sigma^{(0)}_{perc} \cup \Sigma^{(1)}_{narr} \rightarrow \Sigma^{(2)}_{plan}$ could represent abstracting from perception and past experience (grounded levels) to form a high-level policy. Conversely, $V: \Sigma^{(2)}_{plan} \rightarrow \Sigma^{(0)}_{perc} \cup \Sigma^{(0)}_{exec}$ could elaborate a plan into expected sensory states and motor commands (grounded levels).
\end{itemize}
This interaction can be visualized using layer-sector diagrams (a grid with sectors on one axis, abstraction levels on the other). Operators trace paths on this grid: vertical for pure scaling, horizontal for pure sector translation, diagonal for combined operations. Maintaining semantic coherence across these transformations is crucial; an abstraction in one sector should ideally align with related content in others at the new level. Semantic scaling thus provides not only vertical structure but also vertical integration across functional modules.

\section{Scaling Dynamics and Control}

Agents must dynamically control their position on the abstraction scale based on context and cognitive needs. This involves deciding when to abstract and when to elaborate.
\begin{itemize}
	\item \textbf{Triggers for Abstraction ($\Lambda$):} Often invoked when facing redundancy, cognitive overload (requiring compression), detecting recurring patterns in grounded input ($\Phi^{(0)}$), or needing a high-level summary for planning or communication.
	\item \textbf{Triggers for Elaboration ($V$):} Typically required when specific details are needed for action execution, resolving ambiguity, performing fine-grained comparison, grounding abstract concepts in examples or simulations ($\Phi^{(0)}$).
\end{itemize}
These transitions can be governed by scaling control policies, $\pi_{scale}: \Phi \times \text{Context} \rightarrow \{\Lambda, V, \text{id}\}$, which select the appropriate operation. These policies might be reactive (e.g., "if planning fails at level $k$, elaborate to $k-1$") or proactive (e.g., "periodically abstract recent experiences").

Scaling dynamics can also be time-dependent, reflecting processes like memory consolidation (gradual abstraction over time) or shifts in focus. Furthermore, agent behavior can be depth-sensitive, with high-level policies at $\Phi^{(k)}$ conditioning actions that are ultimately executed based on elaborated states in $\Phi^{(0)}$. Effective scaling control allows an agent to adapt its cognitive resolution to the task at hand.

\section{Illustrative Examples}

Consider these examples of semantic scaling in action:
\begin{enumerate}
	\item \textbf{Generalization (Grounding $\rightarrow$ Abstraction):}
	\begin{itemize}
		\item $\phi^{(0)}$ (from $X$): \{"Sensor A reading=5.1 at t1", "Sensor A reading=4.9 at t2", "Sensor A reading=5.0 at t3"\}
		\item $\phi^{(1)} = \Lambda(\phi^{(0)})$: \{"Sensor A reading is stable around 5.0"\}
	\end{itemize}
	\item \textbf{Elaboration for Action (Abstraction $\rightarrow$ Grounding):}
	\begin{itemize}
		\item $\phi^{(2)} \in \Sigma_{plan}$: \{"Objective: Secure the perimeter"\}
		\item $\phi^{(1)} = V(\phi^{(2)})$: \{"Task: Check all entry points", "Task: Activate sensors"\}
		\item $\phi^{(0)} = V(\phi^{(1)})$ (Grounding for Execution): \{"Action: Move to Door 1", "Action: Verify Door 1 lock status", ...\}
	\end{itemize}
	\item \textbf{Narrative Compression:}
	\begin{itemize}
		\item $\phi^{(0)} \in \Sigma_{narr}$: \{Detailed transcript of a conversation\}
		\item $\phi^{(1)} = \Lambda(\phi^{(0)})$: \{"Brief discussion occurred about project delays"\}
		\item $\phi^{(2)} = \Lambda(\phi^{(1)})$: \{"Team alignment on timeline needs improvement"\}
	\end{itemize}
	\item \textbf{Cross-Sector Policy Formation:}
	\begin{itemize}
		\item $\phi^{(1)} \in \Sigma_{narr}$: \{"Past attempts using strategy X failed"\}
		\item $\phi^{(1)} \in \Sigma_{refl}$: \{"Strategy X seems inefficient"\}
		\item $\phi^{(2)} = \Lambda(\phi^{(1)}) \in \Sigma_{plan}$: \{"Avoid Strategy X in future similar situations"\} (Abstract policy derived from lower-level narrative/reflection)
	\end{itemize}
\end{enumerate}
These examples highlight how moving vertically through $\Phi$ via $\Lambda$ and $V$ enables learning, planning, memory evolution, and adaptive behavior, connecting grounded details with abstract reasoning.

\section{Conclusion: Scaling as Cognitive Altitude}

Semantic scaling introduces a crucial vertical dimension to the agent's belief space $\Phi$, formalizing the fundamental cognitive capacity for abstraction and elaboration. By stratifying $\Phi$ into layers $\Phi^{(k)}$ representing different levels of generality, and defining operators $\Lambda$ (upward abstraction) and $V$ (downward elaboration) to move between them, we provide a structure for reasoning, learning, and planning across scales, bridging grounded experience (Chapter~\ref{chap:GroundingSemanticBelief}) with abstract knowledge.

Abstraction allows agents to rise above noisy details, compress information, identify patterns from grounded data, and operate strategically. Elaboration allows them to return to specifics, ground abstract concepts in action or simulation, simulate consequences, and execute plans with precision. The dynamic interplay between these operators, often interacting with the functional modularity of semantic sectors ($\Sigma$), enables agents to manage complexity and adapt their cognitive resolution to context. Semantic scaling is not just about representation; it is about providing the agent with cognitive altitude---the ability to choose the level at which to engage with the world and its own thoughts, connecting the concrete to the abstract. It is a core component of structured belief and versatile intelligence.


\subsection*{Chapter Summary}
This chapter introduces Semantic Scaling as the mechanism structuring the belief space $\Phi$ along a vertical dimension of abstraction. It formalizes this by stratifying $\Phi$ into layers $\Phi^{(k)}$, where $k=0$ represents grounded, concrete beliefs (often derived from Observation Encoding $X$) and higher values of $k$ represent increasingly abstract summaries, generalizations, rules, or schemas. Transitions between these layers are mediated by the upward abstraction operator $\Lambda$ (compressing information, generalizing) and the downward elaboration operator $V$ (instantiating abstract concepts, adding detail, potentially grounding in simulation). The chapter discusses the philosophical and cognitive motivations for hierarchical abstraction, the formal properties of the $\Lambda$ and $V$ operators (including composition and partial invertibility), and their interaction with semantic sectors ($\Sigma$), allowing for both intra-sector and cross-sector scaling. It also covers the dynamics of scaling, including triggers for abstraction/elaboration and the concept of scaling control policies ($\pi_{scale}$). Semantic scaling enables agents to manage complexity, generalize from experience, connect abstract plans to concrete actions, and operate at multiple levels of cognitive resolution, providing a crucial "cognitive altitude".
	\chapter{Semantic Sectors}
\label{chap:SemanticSectors}

\section{Introduction: Functional Modularity in Belief}

While the semantic state space $\Phi$ provides a unified substrate for belief, and semantic scaling (	Chapter~\ref{chap:SemanticScaling}) introduces vertical structure based on abstraction, an agent's cognitive state is rarely monolithic in function. At any given moment, different kinds of thought processes coexist: perceptual grounding, narrative recall, future planning, reflective evaluation. These functional modes suggest an orthogonal dimension of organization within $\Phi$.

This chapter introduces the concept of semantic sectors $\Sigma$. We propose that the belief space $\Phi$ is stratified not just by abstraction, but also by function, decomposing into coherent, potentially overlapping subregions or "sectors," each associated with a distinct mode of epistemic operation. Example sectors might include those dedicated to perception ($\Sigma_{perc}$), narrative ($\Sigma_{narr}$), planning ($\Sigma_{plan}$), or reflection ($\Sigma_{refl}$).

This sectoral structure is not merely descriptive; it is operationally crucial. It enables:
\begin{itemize}
	\item \textbf{Targeted Operations:} Applying processes like assimilation ($A$), nullification ($N_t$), or annihilation ($K$) selectively to specific types of belief content.
	\item \textbf{Introspective Control:} Monitoring or modulating specific cognitive functions (e.g., suppressing reflection during reactive tasks).
	\item \textbf{Modular Reasoning:} Allowing specialized reasoning processes to operate primarily within their relevant sectors.
	\item \textbf{Dynamic Mode Switching:} Shifting the agent's overall cognitive stance by activating or deactivating different sectors.
\end{itemize}
Semantic sectors provide the internal architecture necessary for agents to manage cognitive complexity, focus attention, regulate internal processes, and "think in parts" without losing overall coherence. They represent the functional modularity layered onto the semantic substrate $\Phi$.

\section{Philosophical and Cognitive Foundations}

The notion that cognition is functionally differentiated or modular has deep roots. Cognitive science, starting with Fodor's modularity hypothesis and extending to models like Baddeley's working memory, suggests that the mind employs specialized subsystems for distinct tasks (e.g., perception vs. central reasoning, phonological vs. visuospatial processing). This modularity prevents interference and allows for efficient, specialized processing.

Phenomenology also supports layered understanding. Heidegger's concept of the "forestructure" suggests that explicit deliberation rests upon a background of practical, perceptual, or affective grounding. Beliefs related to immediate perception might occupy a different functional "layer" than abstract reflections. Sellars' distinction between the "space of causes" (e.g., raw perception) and the "space of reasons" (e.g., justified belief) further highlights the need for functional differentiation within an agent's epistemic framework.

From a control perspective, stratification is essential. An agent must be able to selectively modulate different aspects of its cognition---suppressing impulse in favor of deliberation, focusing perception while backgrounding narrative recall, or prioritizing planning over reflection. This requires addressable functional components, which semantic sectors provide. Even in artificial systems, particularly learned ones, functional specialization often emerges, suggesting that stratification is not just an engineering choice but potentially a convergent solution for managing complex cognition. Semantic sectors formalize this functional necessity within the belief space $\Phi$.

\section{Functional Modeling of Semantic Sectors}

We formalize semantic sectors as functionally coherent, addressable subsets within the overall belief space $\Phi$.

\textbf{Definition (Semantic Sector):} A semantic sector $\Sigma$ is a designated subset of the semantic state space $\Phi$ ($\Sigma \subseteq \Phi$) associated with a specific cognitive function or type of content. Given a belief state $\phi \in \Phi$, the portion residing within sector $\Sigma$ is denoted, through abuse of notation, by the sectoral projection $\phi|_{\Sigma} = \phi \cap \Sigma$.

Key characteristics include:
\begin{itemize}
	\item \textbf{Overlap:} Sectors are generally not disjoint partitions; a belief state $\phi$ or expression $\varphi_i$ may belong to multiple sectors simultaneously ($\phi \in \Sigma_i \cap \Sigma_j$). This overlap is crucial, as, for example, a memory might be part of both the narrative ($\Sigma_{narr}$) and reflective ($\Sigma_{refl}$) sectors.
	\item \textbf{Functional Coherence:} Beliefs within a sector typically share a common role, origin, or processing pathway.
	\item \textbf{Addressability:} Sectors serve as targets for specific operations and control policies.
\end{itemize}

Common examples of sectors include:
\begin{itemize}
	\item $\Sigma_{perc}$: Perceptual beliefs, grounded sensor data.
	\item $\Sigma_{plan}$: Policies, intentions, goals, action plans.
	\item $\Sigma_{narr}$: Episodic memory, temporal sequences, causal narratives.
	\item $\Sigma_{refl}$: Reflective evaluations, meta-cognition, self-models.
	\item $\Sigma_{affect}$: Beliefs tagged with emotional valence or motivational state.
	\item $\Sigma_{lang}$: Beliefs representing direct linguistic input or formulated outputs.
\end{itemize}
The specific taxonomy of sectors depends on the agent's architecture and capabilities (parameter $\theta$). Sectoral structure enables specialized operations:
\begin{itemize}
	\item \textbf{Sector-Scoped Operators:} Any global operator $O: \Phi \rightarrow \Phi$ can potentially be restricted to act only on a sector's content, $O_{\Sigma}(\phi) := O(\phi|_{\Sigma})$ (though care must be taken with interactions). This allows for targeted assimilation ($A_{\Sigma}$), nullification ($N^{\Sigma}_t$), or annihilation ($K_{\Sigma}$).
	\item \textbf{Sector-Aware Control:} Policies $\pi: \Phi \rightarrow A$ can condition on the state of specific sectors, e.g., $\pi(\phi|_{\Sigma_{plan}}, \phi|_{\Sigma_{perc}})$.
	\item \textbf{Sector-Level Diagnostics:} Agents can maintain and monitor metrics like coherence $\kappa(\Sigma, \phi)$ or activation load $\lambda(\Sigma, \phi)$ for each sector, guiding introspection and regulation.
\end{itemize}
By stratifying $\Phi$ functionally, sectors provide the necessary handles for fine-grained cognitive management.

\begin{figure}[ht]
	\centering
	\input{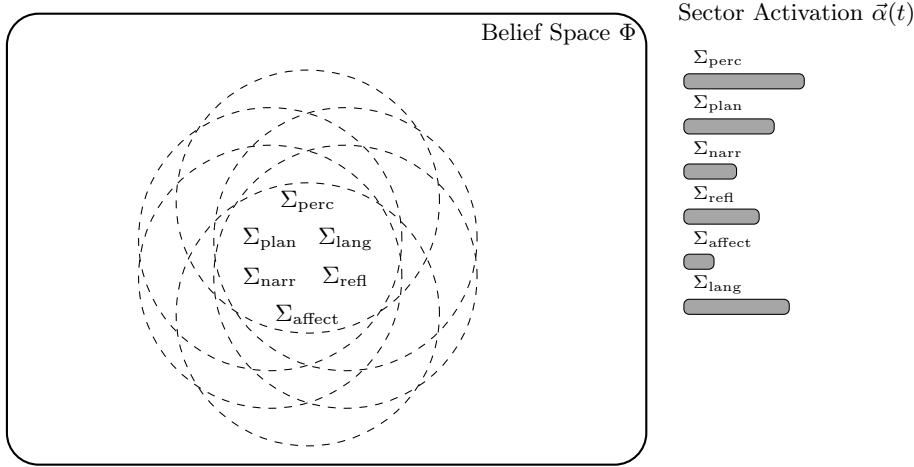}
	\caption{Functional stratification of the belief space $\Phi$ into semantic sectors $\Sigma$. Each sector (e.g., perception $\Sigma_{\text{perc}}$, planning $\Sigma_{\text{plan}}$, narrative $\Sigma_{\text{narr}}$, reflection $\Sigma_{\text{refl}}$) represents an overlapping, functionally coherent subregion of $\Phi$. The agent's current cognitive mode is characterized by the sector activation vector $\vec{\alpha}(t)$, dynamically modulating processing emphasis.}
	\label{fig:semantic_sectors}
\end{figure}

\section{Sectoral Dynamics and Mode Switching}

Semantic sectors are not static partitions but dynamically activated components of the cognitive architecture. An agent's overall cognitive "mode" can be characterized by the relative activation levels of its different sectors.

\textbf{Definition (Sector Activation):} Each sector $\Sigma_i$ has an associated time-varying activation signal $\alpha_i(t) \in [0, 1]$, indicating its current relevance, processing load, or accessibility. The state of all sectors is captured by the activation vector $\vec{\alpha}(t) = [\alpha_1(t), \dots, \alpha_n(t)]$.

Operations can be gated by activation levels (e.g., assimilation $A_{\Sigma_i}$ only occurs if $\alpha_i(t)$ exceeds a threshold). Different cognitive modes correspond to distinct activation profiles:
\begin{itemize}
	\item \textbf{Reactive Mode:} High $\alpha_{perc}$, $\alpha_{plan}$; low $\alpha_{narr}$, $\alpha_{refl}$.
	\item \textbf{Reflective Mode:} High $\alpha_{narr}$, $\alpha_{refl}$; moderate $\alpha_{perc}$.
	\item \textbf{Planning Mode:} High $\alpha_{plan}$; moderate $\alpha_{perc}$, $\alpha_{refl}$.
\end{itemize}
Transitions between these modes, or mode switching, can be triggered externally (by environmental cues) or endogenously (by internal diagnostics like high cognitive load $\lambda$, detected incoherence $\kappa$, or completion of a task phase). The transition dynamics can be modeled by a function governing the evolution of the activation vector: $\vec{\alpha}(t + \delta t) = f(\vec{\alpha}(t), \phi, s, \pi)$.

Sector dynamics often involve interactions:
\begin{itemize}
	\item \textbf{Mutual Inhibition:} Activation of one sector might suppress another (e.g., high $\alpha_{react}$ inhibits $\alpha_{refl}$).
	\item \textbf{Coactivation:} Some sectors might typically activate together (e.g., $\Sigma_{narr}$ and $\Sigma_{refl}$ during introspection).
\end{itemize}
The history of sector activations $\alpha_i(t)$ provides an introspective trace of the agent's cognitive activity over time, useful for meta-control and analysis. Dynamic activation and mode switching give the agent cognitive flexibility, allowing it to allocate resources ($\epsilon$) and adopt appropriate processing strategies based on demand.

\section{Sector-Targeted Operations}

The functional stratification provided by semantic sectors enables epistemic operations to be applied with precision. Instead of modifying the entire belief state $\phi$, agents can target specific sectors.
\begin{itemize}
	\item \textbf{Sectoral Assimilation ($A_{\Sigma}$):} New information $X(s)$ can be routed to and integrated primarily within a specific sector $\Sigma$, based on content type or context. For example, perceptual input updates $\Sigma_{perc}$, while instructions update $\Sigma_{plan}$.
	\item \textbf{Sectoral Nullification ($N^{\Sigma}_t$):} Belief decay can be applied selectively. An agent might allow episodic memories in $\Sigma_{narr}$ to fade via $N^{\Sigma_{narr}}_t$ while preserving stable policies in $\Sigma_{plan}$. This implements selective forgetting.
	\item \textbf{Sectoral Annihilation ($K_{\Sigma}$):} Abrupt erasure can be confined to a specific sector, $\phi_{new} = \phi \setminus \phi|_{\Sigma}$. This allows for functional resets without total memory loss. Examples include $K_{plan}$ to discard an outdated plan, $K_{refl}$ to suppress introspection and enter a reactive state, or $K_{affect}$ to reset emotional state.
	\item \textbf{Sector-Sensitive Control ($\pi(\phi|_{\Sigma_i})$):} Policies can base decisions on the state of specific sectors (e.g., "If $\kappa(\Sigma_{plan}, \phi)$ is low, initiate reflection in $\Sigma_{refl}$"). They can also issue commands targeting sectors (e.g., "Trigger $K_{narr}$ to reduce interference").
	\item \textbf{Sector-Level Information Access:} Operations like extracting content ($\phi|_{\Sigma}$ for summary), computing diagnostics ($\lambda(\Sigma, \phi)$ for load), or saving/restoring sector states become possible.
\end{itemize}

\begin{figure}[h]
	\centering
	\input{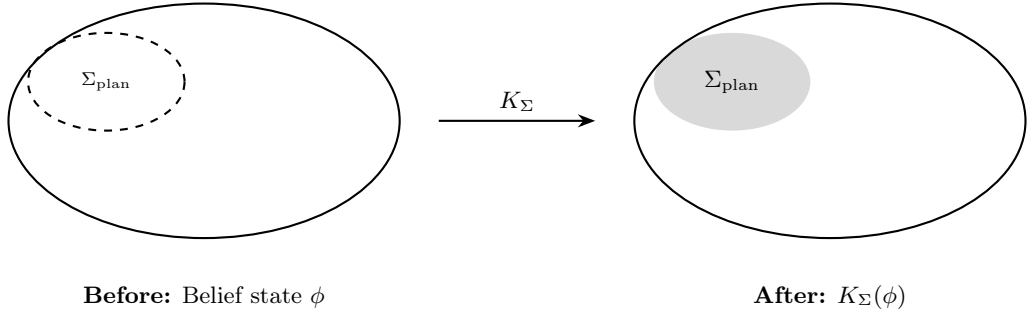}
	\caption{Illustration of sector-targeted annihilation. The operator \(K_{\Sigma}\) removes content associated with a specific semantic sector (e.g., \(\Sigma_\text{plan}\)) from a belief state \(\phi\), leaving the rest of the structure intact.}
	\label{fig:sector_annihilation}
\end{figure}

Sector-targeted operations transform sectors from mere labels into functional components that can be independently monitored, managed, and modified, enabling fine-grained control over the agent's internal cognitive landscape.

\section{Applications and Illustrative Examples}

Consider these scenarios illustrating the utility of semantic sectors:
\begin{enumerate}
	\item \textbf{Focused Perception and Reflection:} An agent observes an anomaly ("Object shape unexpected" $\in \Sigma_{perc}$). High activation in $\Sigma_{perc}$ triggers activation of $\Sigma_{refl}$ and $\Sigma_{narr}$. The agent consults past experiences ($\phi|_{\Sigma_{narr}}$) and evaluates significance ($\phi|_{\Sigma_{refl}}$), leading to a revised understanding ("This might be the damaged component mentioned earlier" $\in \Sigma_{refl}$). Information flows selectively between relevant sectors.
	\item \textbf{Plan Reset and Adaptation:} An agent executing a plan ($\phi|_{\Sigma_{plan}}$) encounters an unexpected obstacle detected in $\Sigma_{perc}$. A meta-controller detects the conflict between $\Sigma_{perc}$ and $\Sigma_{plan}$, triggers $K_{plan}$ to erase the invalid plan, preserves memory in $\Sigma_{narr}$, and initiates a replanning process by activating $\Sigma_{plan}$ with new input derived from $\Sigma_{perc}$ and $\Sigma_{refl}$.
	\item \textbf{Load-Based Mode Switching:} During a high-intensity task, the activation load $\lambda(\Sigma_{perc}, \phi)$ and $\lambda(\Sigma_{plan}, \phi)$ become very high. A control policy detects this and reduces $\alpha_{refl}(t)$ and $\alpha_{narr}(t)$, effectively shifting the agent into a more reactive mode focused on immediate perception and action, suppressing deeper reflection until the load decreases.
	\item \textbf{Parallel Processing:} Receiving a complex instruction like "Check the pressure, but only if the safety interlock is engaged," might simultaneously update $\Sigma_{plan}$ ("Add task: check pressure"), $\Sigma_{perc}$ ("Need to check interlock status"), and $\Sigma_{refl}$ ("Conditional execution required"). Beliefs develop concurrently in relevant sectors.
\end{enumerate}
These examples show how sectoral structure enables targeted updates, selective forgetting, adaptive control, and parallel processing, contributing to more robust and flexible cognitive behavior.

\section{Conclusion: Stratification as Organization}

Semantic sectors introduce essential functional stratification into the agent's belief space $\Phi$. By partitioning or organizing $\Phi$ into coherent, addressable subregions based on cognitive role---such as perception, narrative, planning, and reflection---we move beyond a monolithic view of belief towards a modular architecture.

This stratification is crucial for enabling selective epistemic operations: targeted assimilation, sector-specific nullification or annihilation, and sector-aware control policies. It supports dynamic mode switching, allowing agents to adapt their cognitive posture by modulating sector activations ($\vec{\alpha}(t)$). This functional modularity provides the precision necessary for managing complex internal states, resolving conflicts locally, focusing attention (allocating $\epsilon$), and preserving cognitive coherence ($\kappa$) during ongoing interaction and learning.

Semantic sectors are, in essence, the operating system underlying intelligent belief management. They provide the structure necessary for an agent to think in parts, to regulate distinct streams of thought, and to maintain stability while adapting dynamically. Stratification does not fragment thought; it organizes it, providing the foundation for scalable, controllable, and introspectively capable artificial minds.


\subsection*{Chapter Summary}
This chapter introduces Semantic Sectors ($\Sigma$) as a principle for organizing the belief space $\Phi$ based on functional modularity, complementing the vertical structure provided by semantic scaling. It proposes that $\Phi$ can be stratified into potentially overlapping, addressable subregions (e.g., $\Sigma_{perc}, \Sigma_{plan}, \Sigma_{narr}, \Sigma_{refl}$), each associated with distinct cognitive functions or content types. This sectoral organization, motivated by cognitive science and control principles, enables targeted epistemic operations, allowing processes like Assimilation ($A_{\Sigma}$), Nullification ($N_t^{\Sigma}$), or Annihilation ($K_{\Sigma}$) to be applied selectively. The chapter explores the dynamics of sector activation ($\vec{\alpha}(t)$), modeling cognitive mode switching as changes in the relative activation of different sectors. This structure supports sector-specific monitoring (e.g., coherence $\kappa(\Sigma, \phi)$, load $\lambda(\Sigma, \phi)$) and control policies sensitive to functional context. Semantic sectors provide the necessary architecture for managing cognitive complexity, focusing attention (effort $\epsilon$), regulating internal processes modularly, and enabling specialized reasoning within a unified belief space.
	\chapter{Semantic Geometry}
\label{chap:SemanticGeometry}

\section{Introduction: Unifying Structure}

The preceding chapters introduced two fundamental organizing principles for the semantic state space $\Phi$: Semantic Scaling (Chapter~\ref{chap:SemanticScaling}), which stratifies belief vertically by abstraction level $\Phi^{(k)}$, and Semantic Sectors (Chapter~\ref{chap:SemanticSectors}), which provide functional modularity horizontally via regions $\Sigma$. While powerful individually, their true potential emerges when unified within a cohesive geometric framework.

This chapter introduces Semantic Geometry, a model that synthesizes these two dimensions---abstraction and function---to construe the belief space $\Phi$ as a structured epistemic manifold. Within this geometry, each belief state $\phi$ (or its components $\varphi_i$) can be conceptually positioned using coordinates $(\Sigma, k)$ that specify both its functional modality and its level of generality.

This geometric perspective transforms our understanding of belief dynamics. Epistemic operations like assimilation ($A$), abstraction ($\Lambda$), elaboration ($V$), reflection ($M$), nullification ($N_t$), and even control policies are no longer viewed as disparate transformations but as structured trajectories across this semantic manifold. This framework provides agents with:
\begin{itemize}
	\item A unified space for representing diverse belief types.
	\item Structure for modular, depth-aware reasoning and control.
	\item A basis for defining semantic distance and coherence across sectors and scales.
	\item A language for modeling belief evolution as continuous motion and flow.
\end{itemize}
Semantic geometry turns the internal landscape of belief into a navigable space, providing not just a map of what an agent believes, but a structure for how it thinks and moves through meaning.

\section{Geometric Construction of Belief Space}

We construct the belief space $\Phi$ as a manifold structured by the two orthogonal axes identified previously: semantic sector ($\Sigma$) and abstraction level ($k$).

\textbf{Definition (Stratified Manifold):} The semantic state space $\Phi$ is the union of subspaces $\Sigma^{(k)}$, where $\mathcal{S}$ is the set of defined semantic sectors and $k \in \mathbb{Z}_{\ge 0}$ is the abstraction level:
$$ \Phi = \bigcup_{\Sigma \in \mathcal{S}} \bigcup_{k \in \mathbb{Z}_{\ge 0}} \Sigma^{(k)} $$
where $\Sigma^{(k)} = \Sigma \cap \Phi^{(k)}$ represents the subset of belief space corresponding to sector $\Sigma$ at abstraction level $k$.

This construction yields a conceptual grid or surface where each cell $(\Sigma, k)$ holds beliefs of a specific type and abstraction level.

\begin{figure}[h]
	\centering
	\input{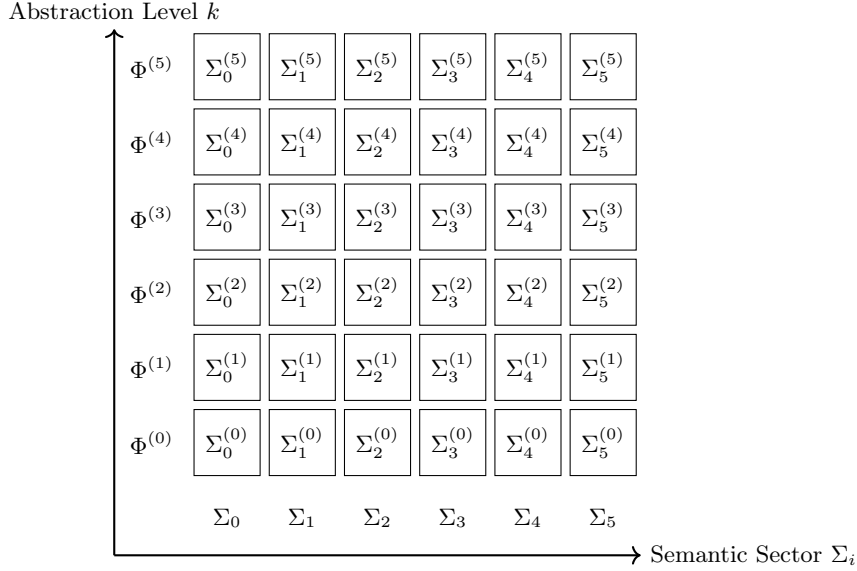}
	\caption{Semantic geometry grid illustrating the belief space \(\Phi\) as a 2D structure indexed by semantic sector \(\Sigma\) and abstraction level \(k\). Each cell \(\Sigma_i^{(k)}\) contains beliefs of a specific type and resolution. A belief state \(\phi\) can span multiple cells. Arrows indicate epistemic transitions via abstraction and sectoral translation.}
	\label{fig:semantic_geometry_grid}
\end{figure}

A typical belief state $\phi$ is not confined to a single cell but exists as an ensemble distributed across a region of this manifold:
$$ \phi = \bigcup_{(\Sigma,k)} \phi^{(k)}_{\Sigma}, \quad \text{where } \phi^{(k)}_{\Sigma} = \phi \cap \Sigma^{(k)}. $$
This allows agents to hold mixed-resolution beliefs simultaneously (e.g., detailed perception $\phi^{(0)}_{perc}$ alongside abstract goals $\phi^{(2)}_{plan}$). The structure supports a notion of local topology. Adjacent cells are connected by specific operators:
\begin{itemize}
	\item \textbf{Vertical connections:} Between $(\Sigma, k)$ and $(\Sigma, k + 1)$ via abstraction/elaboration operators $\Lambda|_{\Sigma}$ and $V|_{\Sigma}$.
	\item \textbf{Horizontal connections:} Between $(\Sigma_i, k)$ and $(\Sigma_j, k)$ via sector translation or projection operators $T_{\Sigma_i \rightarrow \Sigma_j}$.
	\item \textbf{Diagonal connections:} Represent composite operations involving both scaling and sectoral shifts.
\end{itemize}
These transitions define the fundamental geometry of epistemic change within $\Phi$.

\begin{figure}[h]
	\centering
	\input{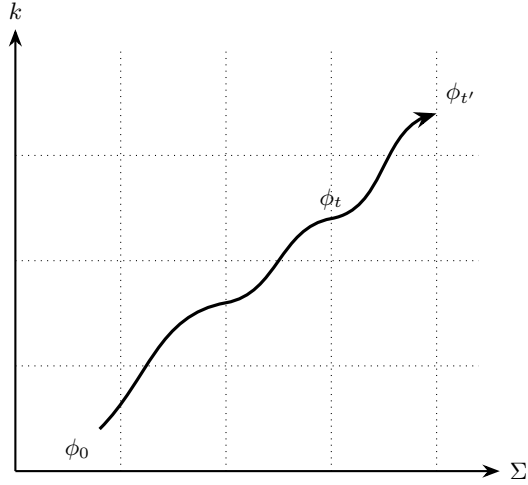}
	\caption{A semantic trajectory representing the evolution of a belief state \(\phi_t\) over time in the \((\Sigma, k)\) plane. The trajectory reflects both abstraction (increasing \(k\)) and sectoral shifts (along \(\Sigma\)).}
	\label{fig:semantic_trajectory}
\end{figure}

\section{Semantic Coordinates and Belief Positioning}

Each belief component $\varphi_i$ within an ensemble $\phi$ can be assigned conceptual semantic coordinates $(\Sigma, k)$ indicating its primary sector and abstraction level.

\textbf{Definition (Coordinate Map):} We can define a (potentially context-dependent) map $\chi: \Phi \rightarrow \mathcal{P}(\mathcal{S} \times \mathbb{Z}_{\ge 0})$ that assigns a set of coordinates to a belief state or its components. For simplicity, we often refer to the primary coordinate $(\Sigma, k)$ such that $\phi \subseteq \Sigma^{(k)}$ or where the "center of mass" lies.

Given a belief ensemble $\phi$ distributed across the manifold, we can define projections:
\begin{itemize}
	\item \textbf{Sectoral Projection:} $P_{\Sigma}(\phi) = \phi \cap \Sigma = \bigcup_k \phi^{(k)}_{\Sigma}$ (isolates content within a specific sector).
	\item \textbf{Scale Projection:} $P_k(\phi) = \phi \cap \Phi^{(k)} = \bigcup_{\Sigma} \phi^{(k)}_{\Sigma}$ (isolates content at a specific abstraction level).
\end{itemize}
The distribution of belief across the grid $(\Sigma, k)$, as visualized conceptually in the diagram above, can be quantified by a belief density or activation function $\lambda(\Sigma^{(k)}, \phi)$. We can also define an epistemic center of mass CoM$(\phi)$ using this density, providing a single (though potentially crude) indicator of the agent's current semantic focus within the 2D space. The dynamic evolution of this center tracks shifts in cognitive focus (e.g., from perception to reflection).

\section{Operators as Trajectories in \texorpdfstring{$\Phi$}{Phi}}

Within the semantic geometry framework, epistemic operations are interpreted as movements or transformations generating trajectories across the manifold $\Phi$.

\textbf{Definition (Atomic Epistemic Motions):} Given a belief $\phi \in \Sigma^{(k)}$, primitive operations induce motion in the conceptual ($\Sigma$, k) space:
\begin{itemize}
	\item Upward Abstraction ($\Lambda$): Vertical movement to $(\Sigma, k + 1)$.
	\item Downward Elaboration ($V$): Vertical movement to $(\Sigma, k-1)$.
	\item Sectoral Translation ($T_{\Sigma \rightarrow \Sigma'}$): Horizontal movement to $(\Sigma', k)$.
	\item Assimilation ($A^{(k)}_{\Sigma}$): Injection of new content at coordinate $(\Sigma, k)$.
	\item Nullification/Annihilation ($N_t, K$): Decay or removal of content, potentially moving towards the epistemic vacuum $\Omega$ (the origin/boundary).
\end{itemize}
Complex cognitive processes correspond to composite trajectories, sequences of these atomic motions. Let $\gamma: [t_0, t_1] \rightarrow \Phi$ be an epistemic trajectory. It can be decomposed as $\gamma = \gamma_n \circ \dots \circ \gamma_1$, where each $\gamma_i$ is a transformation induced by an operator ($\Lambda, V, T, A$, etc.). For example, processing an observation might involve a trajectory starting in $\Sigma^{(0)}_{perc}$, moving via abstraction to $\Sigma^{(1)}_{narr}$, then perhaps via reflection into $\Sigma^{(2)}_{refl}$.

This perspective allows us to model cognitive control using trajectory fields $F: \Phi \rightarrow T\Phi$, where $T\Phi$ is the tangent bundle of $\Phi$. The evolution of belief follows $\frac{d\gamma}{dt} = F(\gamma(t))$. Meta-control policies $\pi_{meta}$ can then be framed as selecting fields or guiding trajectories towards target regions $\mathcal{T} \subset \Phi$. Importantly, due to the non-invertibility of $\Lambda$ and $V$, epistemic trajectories often exhibit path-dependence or hysteresis: the state $\phi(t)$ depends not only on its current coordinates but also on the path taken to reach them ($V \circ \Lambda \neq \text{id}$).

\section{Metric Structure and Semantic Distance (Preview)}

To fully leverage the geometric interpretation, $\Phi$ should ideally be endowed with a notion of distance. We postulate a semantic distance function $d: \Phi \times \Phi \rightarrow \mathbb{R}_{\ge 0}$ satisfying standard metric axioms (non-negativity, identity of indiscernibles, symmetry, triangle inequality). This metric quantifies the "semantic difference" between belief states. (The formal construction and learning of such metrics are discussed further in Part~\ref{part:learning_and_adaptation}, Chapter~\ref{chap:LearningSemanticStructures}).

Conceptually, this distance might be decomposable:
$$ d(\phi_1, \phi_2) \approx \alpha \cdot d_{sector}(\phi_1, \phi_2) + \beta \cdot d_{scale}(\phi_1, \phi_2) $$
where $d_{sector}$ measures distance across functional modalities and $d_{scale}$ measures distance across abstraction levels, weighted by $\alpha, \beta$. Such a metric, \textbf{while potentially challenging to define precisely and learn effectively}, enables:
\begin{itemize}
	\item \textbf{Coherence Measurement:} Quantifying the divergence between related beliefs (e.g., $d(\phi^{(k)}, \Lambda(\phi^{(k-1)}))$).
	\item \textbf{Stability Diagnostics:} Tracking semantic drift $d(\phi_0, \phi_t)$ over time.
	\item \textbf{Neighborhood Definition:} Defining local regions $N_{\epsilon}(\phi) = \{\phi' | d(\phi, \phi') < \epsilon\}$ for local updates or attention.
	\item \textbf{Cost of Thought:} Potentially defining trajectory curvature or transformation cost based on distance travelled vs. displacement.
\end{itemize}
The metric structure turns $\Phi$ from a simple grid into a space where proximity implies semantic similarity.

\section{Cognitive Dynamics and Semantic Flow}

The evolution of belief $\phi(t)$ within the semantic geometry can be modeled as a semantic flow governed by internal and external forces acting on the belief state.

\textbf{Definition (Epistemic Vector Fields):} We can conceptually define vector fields $F: \Phi \rightarrow T\Phi$ that assign a direction and magnitude of change $F(\phi)$ to each belief state $\phi$. The belief trajectory $\gamma(t)$ follows $\frac{d\gamma}{dt} = F(\gamma(t))$.

\textbf{Note on Practicality:} Defining and computing such a field $F$ across the entirety of the high-dimensional space $\Phi$ is highly complex and likely requires significant abstraction or approximation in practice. However, the concept is useful for modeling cognitive tendencies:
\begin{itemize}
	\item An abstraction flow pushing beliefs vertically towards higher $k$.
	\item A grounding flow pulling beliefs towards lower $k$ or $\Sigma_{perc}$.
	\item Attractor dynamics where flows converge towards stable fixed points $\phi^*$ (e.g., coherent schemas, reflective equilibria) where $F(\phi^*) \approx 0$.
	\item Repeller dynamics where flows diverge from unstable regions (e.g., areas of high contradiction).
\end{itemize}

\begin{figure}[h]
	\centering
	\input{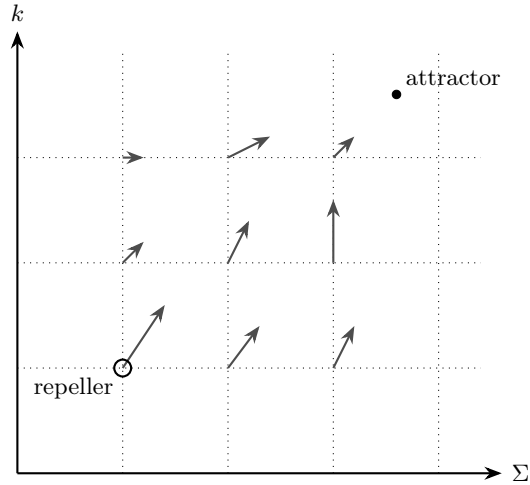}
	\caption{Semantic flow field \(\mathcal{F}\) over the \((\Sigma, k)\) plane. Arrows represent preferred semantic transitions. Certain regions act as attractors (stable belief configurations), while others serve as repellers (unstable or disorganized states).}
	\label{fig:semantic_flow}
\end{figure}

These flows can be influenced by assimilation (injecting momentum), nullification (a general drift towards $\Omega$), anchoring (creating stable points), and control policies (actively shaping the effective field $F$). Concepts like semantic drift can be detected by monitoring deviation from expected flow lines or stable regions. Feedback mechanisms can then adjust the effective flow field $F$ to implement re-centering or correction. Meta-control becomes a problem of navigating or modulating these flows to achieve cognitive goals. Furthermore, stable cyclical flows passing through sequences of sectors and scales (e.g., perceive $\rightarrow$ narrate $\rightarrow$ plan $\rightarrow$ reflect $\rightarrow$ perceive) can model cognitive rhythms or standard operational loops.

\section{Visualization and Diagrammatic Models}

The power of semantic geometry lies partly in its potential for visualization, providing intuitive "cognitive maps" (like the conceptual grid diagram introduced earlier).
\begin{itemize}
	\item \textbf{Semantic Grid (Sector $\times$ Scale):} The primary visualization plots semantic sectors ($\Sigma_i$) along one axis (e.g., horizontal) and abstraction level ($k$) along the other (e.g., vertical). Belief states $\phi$ or their components are represented as points, regions, or density distributions on this grid.
	\item \textbf{Trajectory Arrows:} Paths traced on the grid show epistemic evolution. Vertical arrows denote scaling ($\Lambda, V$), horizontal arrows denote sectoral shifts ($T$), and diagonal arrows represent combined operations.
	\item \textbf{Flow Fields:} Vector arrows overlaid on the grid can illustrate the direction and magnitude of typical belief motion ($F(\phi)$) in different regions. Attractors and repellers become visually apparent.
	\item \textbf{Belief Density Heatmaps:} Coloring grid cells $(\Sigma, k)$ based on belief activation or mass $\lambda(\Sigma^{(k)}, \phi)$ shows the agent's current semantic focus and the distribution of its cognitive load.
	\item \textbf{Cycle Diagrams:} Closed loops on the grid can represent recurring cognitive processes or operational cycles (e.g., sense-plan-act loop).
\end{itemize}

\begin{figure}[ht]
	\centering
	\input{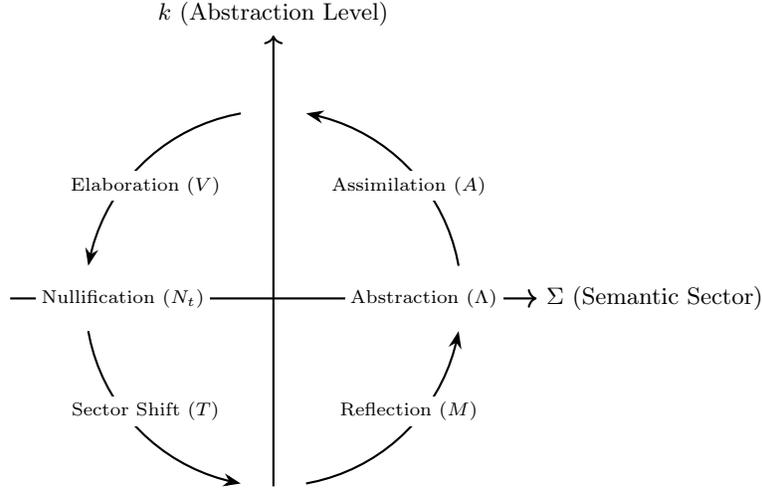}
	\caption{Semantic epistemic cycle represented as a closed process loop in the \((\Sigma, k)\) plane. The cycle flows through assimilation ($A$) of new information, elaboration ($V$) for grounded realization, nullification ($N_t$) for cleanup, sectoral translation ($T$) across functional modalities, reflection ($M$) for evaluation, and abstraction ($\Lambda$) to higher levels, enabling continual evolution of the belief state within the structured semantic manifold.}
	\label{fig:semantic_cycle}
\end{figure}

These diagrams transform the abstract structure of $\Phi$ into an interpretable map, aiding design, debugging, introspection, and theoretical understanding.

\section{Conclusion: Belief as Navigation}

Semantic geometry provides a unified framework for structuring the belief space $\Phi$ by integrating the functional modularity of semantic sectors ($\Sigma$) with the hierarchical depth of semantic scaling ($\Phi^{(k)}$). This construes $\Phi$ as a conceptual two-dimensional epistemic manifold where beliefs are positioned by coordinates $(\Sigma, k)$.

Within this geometry, epistemic operations manifest as structured trajectories, semantic distance provides a measure of coherence and change (though defining it precisely is challenging), and belief evolution can be modeled as dynamic flows governed by vector fields (a complex theoretical ideal). Visualization tools based on this geometry offer intuitive maps of cognitive states and processes.

Ultimately, semantic geometry reframes cognition: to think is to move within a structured landscape of meaning. By endowing $\Phi$ with this structure, we provide agents with an internal map and compass, enabling more organized reasoning, targeted control, and navigable introspection. It provides the spatial logic for how belief states relate, transform, and guide behavior.


\subsection*{Chapter Summary}
This chapter introduces Semantic Geometry, unifying the concepts of Semantic Scaling (abstraction levels $\Phi^{(k)}$) and Semantic Sectors (functional modules $\Sigma$) into a cohesive geometric framework for the belief space $\Phi$. It conceptualizes $\Phi$ as a structured manifold where belief states ($\phi$) can be positioned using coordinates $(\Sigma, k)$, reflecting both functional role and abstraction level. Cognitive dynamics, driven by operators like Assimilation ($A$), Abstraction ($\Lambda$), Elaboration ($V$), Nullification ($N_t$), and Annihilation ($K$), are interpreted as generating structured trajectories ($\gamma(t)$) across this manifold. The chapter discusses the concept of an underlying epistemic flow field ($F$) governing these trajectories and postulates the existence of a semantic distance metric ($d$) to quantify belief similarity and coherence ($\kappa$). This geometric perspective allows for intuitive visualizations (e.g., the $\Sigma \times k$ grid) and reframes cognitive processes, including regulation and control, as forms of navigation within a structured internal landscape of meaning.
	
	\part{Epistemic Dynamics: Evolution of Thought}
	\label{part:epistemic_dynamics}
	
	\chapter{Assimilation}
\label{chap:Assimilation}

\section{Introduction: Integrating New Information}

Having established the foundational structures of the semantic state space $\Phi$ (potentially parameterized as $\Phi^{[\theta]}$, Chapter~\ref{chap:ParameterizedArchitectures}), the Null Tower (Chapter~\ref{chap:NullTower}), belief construction (Chapter~\ref{chap:BeliefConstruction}), and the organizing principles of semantic scaling and sectors (Part~\ref{part:structuring_belief}), we now turn to the core dynamics that govern how belief states $\phi \in \Phi$ evolve over time. The most fundamental process driving belief change in response to interaction or internal generation is assimilation.

Assimilation is the mechanism by which an agent integrates new, structured, object-level input derived from observations, communication, or certain internal processes like simulation, into its existing belief ensemble $\phi$. It is the primary way an agent learns from experience, adapts its understanding, and extends its knowledge base regarding the world or hypothetical scenarios. We formalize this process via the assimilation operator:
$$ A : \Phi \times \Phi_{input} \rightarrow \Phi $$
where $\phi_{current} \in \Phi$ is the agent's current belief state, $\phi_{input}$ is the structured representation of the new object-level information (e.g., output of the observation encoder, $\phi_{input} = X(s)$, or result of a simulation), and $\phi_{new} = A(\phi_{current}, \phi_{input})$ is the resulting updated belief state.

Crucially, assimilation is not mere accumulation or concatenation. It is an active, context-sensitive, and structure-preserving process. It must determine how the new information relates to existing beliefs, handle potential conflicts, update relevant structures, potentially trigger further inferences or elaborations, and maintain the overall coherence ($\kappa$) of the agent's epistemic state. Assimilation bridges the gap between perception/input and sustained, structured belief, forming the constructive engine of epistemic evolution for object-level knowledge. (Note: The integration of internally generated meta-cognitive or introspective insights is handled by the distinct Meta-Assimilation operator $M$, detailed in Chapter~\ref{chap:MetaAssimilation}).

\section{Philosophical and Cognitive Motivations}

The nature of the assimilation operator $A$ is informed by core principles of cognition and epistemology. As discussed in Chapter~\ref{chap:PsychologicalMotivations} regarding initial construction, constructivist theories emphasize that knowledge is actively built, not passively received; new input $X(s)$ is interpreted through the lens of the existing state $\phi$. Cognitive psychology highlights the role of schemas or mental models ($\approx \phi$) in organizing knowledge and guiding the integration of new information.

Furthermore, human cognition strives for coherence, preferentially integrating information that aligns with existing beliefs and resisting or revising information that conflicts. Assimilation must reflect this bias, often reinforcing existing structures and establishing stable anchor points ($a_i$). However, when faced with significant conflict or surprise, assimilation must also be capable of triggering belief revision or schema accommodation. Finally, cognitive systems exhibit selective attention; assimilation is not expected to incorporate all aspects of input equally but should filter for salience and relevance relative to the current state and the agent's goals. These principles motivate the formal desiderata for the operator $A$.

\section{Formal Desiderata for Assimilation}

A cognitively plausible assimilation operator $A(\phi, \phi_{input})$ should satisfy several key properties or desiderata:
\begin{description}
	\item[D1: Semantic Integration] The resulting state $\phi_{new}$ must meaningfully incorporate content derived from $\phi_{input}$, interpreted or modified in the context of $\phi$. It's not just $\phi \cup \phi_{input}$.
	\item[D2: Coherence Preservation] $A$ should typically maintain or increase the internal coherence $\kappa(\phi)$ of the belief state. Significant drops in coherence should only occur if $A$ explicitly triggers a revision process due to conflict.
	\item[D3: Salience Filtering] $A$ should operate selectively, prioritizing the integration of salient, relevant, or high-impact components of $\phi_{input}$ over irrelevant details. Notionally, $A(\phi, \phi_{input}) = A(\phi, F(\phi_{input}))$, where $F$ is a salience filter.
	\item[D4: Revision Under Conflict] If $\phi_{input}$ contradicts strongly held or coherent parts of $\phi$, $A$ should include a mechanism for belief revision, potentially removing or modifying parts of $\phi$ before integrating the new content. Notionally, $A(\phi, \phi_{input}) = (\phi \setminus R_{\text{rev}}(\phi, \phi_{input})) \cup \phi'_{input}$.
	\item[D5: Idempotence on Redundancy] Assimilating semantically redundant information (already present or implied in $\phi$) should result in minimal change to $\phi$. $A(\phi, \phi_{input}) \approx \phi$ if $\phi_{input}$ is redundant w.r.t. $\phi$.
	\item[D6: Anchoring] Assimilation should contribute to the formation and strengthening of stable anchor points $a_i$ within $\phi$, reinforcing reliable beliefs and providing structure for future reasoning.
	\item[D7: Locality of Change] In the absence of major conflict, the primary modifications induced by $A$ should be relatively localized within the structure of $\phi$, affecting only the parts most relevant to $\phi_{input}$.
\end{description}
These desiderata guide the design of specific assimilation mechanisms, ensuring they produce belief updates that are robust, coherent, and adaptive.

\begin{figure}[htbp]
	\centering
	\input{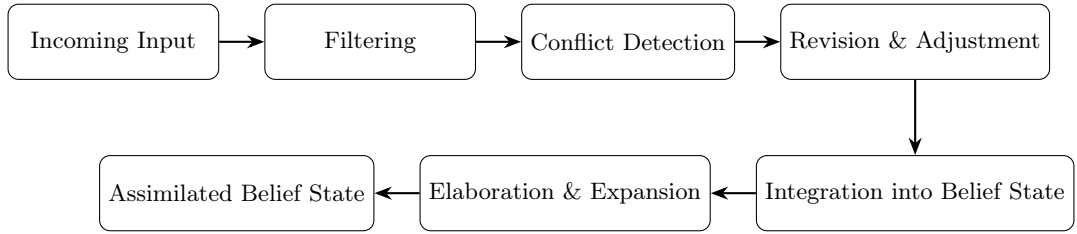}
	\caption{Overall Assimilation Process: Multi-stage flow of the Assimilation operator \(A\), progressing from initial input through filtering, conflict detection, revision, integration, and elaboration to form an enriched belief state.}
	\label{fig:overall-assimilation-process}
\end{figure}

\section{Taxonomy of Assimilation Operators}

Assimilation is multifaceted. Different situations call for different modes of integration for object-level information. We can classify assimilation operators $A$ based on several factors:

\begin{enumerate}
	\item \textbf{Source of Input ($\phi_{input}$):} (Must be non-meta-cognitive)
	\begin{itemize}
		\item $A_{perc}$: Input from perception ($X(s)$ where $s$ is sensory).
		\item $A_{text}$: Input from linguistic communication.
		\item $A_{sim}$: Input from internal object-level simulation or hypothesis generation.
	\end{itemize}
	\item \textbf{Assimilation Mode (Primary Function):}
	\begin{itemize}
		\item Elaborative ($A_{elab}$): Enriches $\phi$ by adding inferred consequences or contextual details related to $\phi_{input}$.
		\item Corrective ($A_{corr}$): Resolves conflicts between $\phi_{input}$ and $\phi$ via revision.
		\item Confirmatory ($A_{conf}$): Strengthens existing beliefs in $\phi$ that are consistent with $\phi_{input}$, often increasing anchoring $a_i$.
		\item Abstracting ($A_{abs}$): Generalizes over $\phi_{input}$ (perhaps in light of patterns in $\phi$) to form higher-level beliefs $\phi^{(k+1)}$.
	\end{itemize}
	\item \textbf{Depth of Integration:}
	\begin{itemize}
		\item Surface-Level: Adds new expressions $\varphi_i$ with minimal change to the existing structure of $\phi$.
		\item Structural: Modifies relationships (links, weights) between expressions within $\phi$.
		\item Schema-Modifying: Causes significant reorganization of belief clusters or interpretive frames within $\phi$.
	\end{itemize}
\end{enumerate}

\begin{figure}[ht]
	\centering
	\input{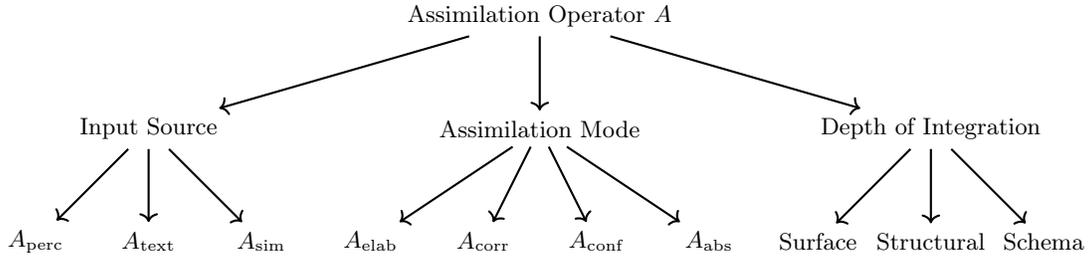}
	\caption{Taxonomy of assimilation operators categorized by input source (e.g., perception, language, simulation), assimilation mode (elaborative, corrective, confirmatory, abstracting), and depth of integration (surface-level, structural, schema-modifying). This hierarchy supports modular design of belief integration processes.}
	\label{fig:assimilation_taxonomy}
\end{figure}

A specific assimilation event might involve a combination of these (e.g., corrective assimilation of perceptual input at the structural level). This taxonomy highlights that $A$ represents a family of related processes for integrating object-level information, distinct from the meta-level integration performed by $M$. The following sections detail some key subtypes of $A$.

\section{Elaborative Assimilation (\texorpdfstring{$A_{\text{elab}}$}{A\_elab})}

Elaborative assimilation focuses on enriching the current belief state $\phi$ by not only incorporating the input $\phi_{input}$ but also generating relevant inferences, predictions, or contextual links based on it. Notionally:
$$ \phi_{new} = \phi \cup \phi_{input} \cup \phi_{elab} $$
where $\phi_{elab} = \{\varphi^{elab}_j\}$ is a set of generated elaborations. The mechanics involve applying elaboration functions or rules $E_j: \Phi \times \Phi_{input} \rightarrow \{\varphi^{elab}_i\}$ that trigger based on patterns in the input and the current context $\phi$. Elaborations might:
\begin{itemize}
	\item Complete partial information.
	\item Link input to existing goals or schemas in $\phi$.
	\item Predict immediate consequences or affordances.
	\item Retrieve relevant associated memories (via the $Q \rightarrow R$ process, Part~\ref{part:semantic_memory}).
\end{itemize}
\textbf{Example:} If $\phi_{current}$ contains \{"Task is to inspect control panel"\} and $\phi_{input} = X(s) =$ \{"Control panel door is ajar"\}, then $A_{elab}(\phi_{current}, \phi_{input})$ might yield $\phi_{new}$ containing the original beliefs plus $\phi_{elab}=$ \{"Panel appears accessible", "Inspection task can likely proceed"\}.

\begin{figure}[htbp]
	\centering
	\input{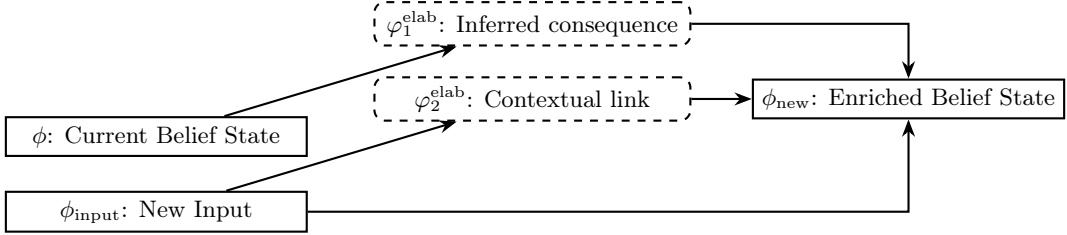}
	\caption{Elaborative assimilation \(A_{\text{elab}}\) enriches the current belief state \(\phi\) by integrating new input \(\phi_{\text{input}}\) and generating relevant elaborations \(\{\varphi^{\text{elab}}_i\}\), resulting in an updated belief state \(\phi_{\text{new}}\).}
	\label{fig:assimilation-elaborative}
\end{figure}

This mode primarily satisfies desiderata D1 (integration), D2 (coherence, as elaborations are context-based), D6 (anchoring new links), and D7 (locality). It is fundamental for building rich, interconnected belief structures from sparse input.

\section{Corrective Assimilation (\texorpdfstring{$A_{\text{corr}}$}{A\_corr})}

Corrective assimilation handles inputs $\phi_{input}$ that conflict with the existing belief state $\phi$. Its primary function is to restore coherence $\kappa$ by revising $\phi$ before or during the integration of the new information (Desideratum D4). The core components are:
$$ \phi_{new} = (\phi \setminus R_{\text{rev}}(\phi, \phi_{input})) \cup \phi'_{input} $$
\begin{itemize}
	\item \textbf{Conflict Detection:} Identifying contradictions between expressions $\varphi_i \in \phi$ and $\varphi_j \in \phi_{input}$ using a predicate $\text{Contradict}(\varphi_i, \varphi_j)$. This might detect factual opposition, policy violations, or constraint mismatches.
	\item \textbf{Revision ($R_{\text{rev}}$):} Determining which belief(s) to retract or modify. $R_{\text{rev}}(\phi, \phi_{input})$ returns the set of expressions from $\phi$ targeted for removal or revision based on the detected conflict. Revision strategies can range from simple retraction to more complex mechanisms based on belief strength, anchoring $a_i$, or source credibility.
	\item \textbf{Integration:} Incorporating the (potentially modified) input $\phi'_{input}$ into the revised state $(\phi \setminus R_{\text{rev}})$.
\end{itemize}
\textbf{Example:} If $\phi =$ \{"Panel is closed", "Manual override required"\} and $\phi_{input} =$ \{"Panel is open"\}, $A_{corr}$ would likely identify "Panel is closed" and perhaps "Manual override required" for removal via $R_{\text{rev}}$. The resulting $\phi_{new}$ would contain \{"Panel is open"\}. Subsequent elaboration might then add "Inspection may proceed without override."

\begin{figure}[htbp]
	\centering
	\input{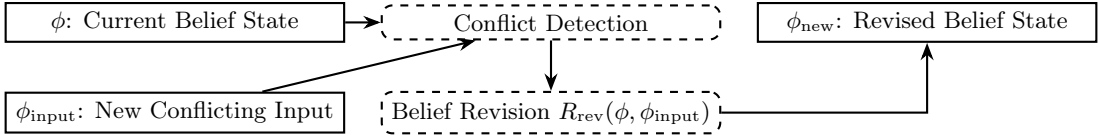}
	\caption{Corrective assimilation \(A_{\text{corr}}\) detects contradictions between the current belief state \(\phi\) and new input \(\phi_{\text{input}}\), revises \(\phi\) via a belief revision process \(R_{\text{rev}}(\phi, \phi_{\text{input}})\), and integrates the corrected information to form an updated belief state \(\phi_{\text{new}}\).}
	\label{fig:assimilation-corrective}
\end{figure}

Corrective assimilation is crucial for maintaining consistency and adapting beliefs when the environment or information source provides contradictory evidence.

\section{Abstracting Assimilation (\texorpdfstring{$A_{\text{abs}}$}{A\_abs})}

Abstracting assimilation integrates new input $\phi_{input}$ in a way that promotes generalization or compression, often by identifying patterns across $\phi_{input}$ and the existing state $\phi$. Instead of just adding information, it may replace specific instances with a more general rule or schema, potentially involving the Abstraction operator $\Lambda$. Notionally:
$$ \phi_{new} = (\phi \setminus \{\varphi_{i1}, \dots, \varphi_{im}\}) \cup \phi_{gen} $$
where $\{\varphi_{i1}, \dots, \varphi_{im}\}$ are specific beliefs (potentially including some from $\phi_{input}$) that are summarized or generalized by the newly constructed abstract belief $\phi_{gen}$ (which might reside at a higher abstraction level $\Phi^{(k+1)}$).

\begin{figure}[ht]
	\centering
	\input{diagrams/fig_assimilation_abstracting.tex}
	\caption{Abstracting assimilation detects a pattern across input beliefs and replaces them with a generalized, higher-level belief. In this example, individual valve check messages are summarized as a single abstract belief, enabling compression and efficient reasoning.}
	\label{fig:assimilation_abstracting}
\end{figure}

\textbf{Example:} After assimilating $\phi_{1}=$ \{"Valve A check complete"\}, $\phi_{2}=$ \{"Valve B check complete"\}, $\ldots$, $\phi_{N}=$ \{"Valve N check complete"\}, $A_{abs}$ might recognize the pattern, remove the individual completion messages via $R_{\text{rev}}$, and add the summary belief $\phi_{gen} = $ \{"All required valves checked"\} at level $k+1$.

This operator is crucial for learning from repeated experience, forming procedural or conceptual abstractions, reducing memory load ($\lambda$), and enabling more efficient high-level reasoning. It often involves more complex pattern detection and structural reorganization than other assimilation modes.

\section{Conclusion: Assimilation's Role in Dynamics}

Assimilation, represented by the family of operators $A$, is the primary engine driving the constructive evolution of an agent's belief state $\phi$ in response to new object-level information, whether external or internal. We have identified several key types:
\begin{itemize}
	\item Elaborative ($A_{elab}$): Enriches beliefs with context and inference.
	\item Corrective ($A_{corr}$): Resolves conflicts and maintains coherence.
	\item Abstracting ($A_{abs}$): Generalizes patterns and compresses information.
	\item (Confirmatory ($A_{conf}$) also mentioned in taxonomy).
\end{itemize}

These operators, while distinct in function and effect, all adhere to core desiderata ensuring that belief updates are integrated meaningfully, selectively, and coherently into the existing structure of $\phi$. They allow for adaptation, revision, and learning based on interaction and simulation.

Assimilation is the counterpoint to the processes of decay (Nullification $N_t$, Chapter~\ref{chap:Nullification}) and erasure (Annihilation $K$, Chapter~\ref{chap:Annihilation}). While $N_t$ and $K$ model the dissolution or removal of belief, $A$ models its growth, refinement, and adaptation based on object-level input. (The integration of meta-level input is handled separately by $M$, Chapter~\ref{chap:MetaAssimilation}). Together, $A, M, N_t, K, \Lambda$, and $V$ define the fundamental dynamics governing the lifecycle of belief within the semantic state space $\Phi$. Understanding assimilation ($A$) is key to understanding how agents learn about the world and maintain a coherent epistemic grasp.


\subsection*{Chapter Summary}
This chapter details the Assimilation operator ($A$), the fundamental process for integrating new, structured, object-level information ($\Phi_{input}$)---derived from perception, communication, simulation, or memory retrieval---into the agent's existing belief state ($\phi$). Defined as $A : \Phi \times \Phi_{input} \rightarrow \Phi$, assimilation is presented not as mere addition, but as an active, context-sensitive, and structure-preserving operation crucial for learning and adaptation. The chapter outlines key desiderata for $A$, including semantic integration, coherence ($\kappa$) preservation, salience filtering, conflict revision, and anchoring ($a_i$). A taxonomy of assimilation operators is proposed, classifying them by input source (e.g., $A_{perc}, A_{text}$), function (e.g., elaborative $A_{elab}$, corrective $A_{corr}$, abstracting $A_{abs}$), and depth of integration. The mechanisms of key subtypes $A_{elab}$ (enriching context), $A_{corr}$ (resolving conflicts), and $A_{abs}$ (generalizing patterns) are discussed. Assimilation is positioned as the primary constructive dynamic for object-level belief evolution, distinct from meta-level integration ($M$) and acting as a counterpoint to the dissipative dynamics of Nullification ($N_t$) and Annihilation ($K$).
	\chapter{Nullification}
\label{chap:Nullification}

\section{Introduction: The Dissolution of Belief}

Assimilation, as detailed in Chapter~\ref{chap:Assimilation}, describes the constructive process by which belief states $\phi \in \Phi$ are formed and updated. However, cognition involves not only the accumulation but also the dissolution of belief. Information becomes outdated, attention shifts, memory fades. This chapter introduces nullification, the process modeling the gradual deactivation, forgetting, or softening of structured beliefs over time due to lack of reinforcement or active processing.

Nullification stands in contrast to both assimilation ($A$, construction) and annihilation ($K$, abrupt erasure, Chapter~\ref{chap:Annihilation}). It represents a continuous, often passive drift towards the epistemic vacuum $\Omega$. We model nullification via a temporally indexed operator $N_t$:
$$ N_t : \Phi \rightarrow \Phi $$
such that for any initial state $\phi$, the belief state evolves over time $t$ under nullification, $\phi_t = N_t(\phi)$, with the property that
$$ \lim_{t\rightarrow\infty} N_t(\phi) \in \Omega $$
Nullification captures the dynamics of natural forgetting, semantic decay, cognitive dormancy, and introspective release. It provides a pathway back towards epistemic neutrality that preserves continuity and potential for reactivation, unlike the discontinuity of annihilation. It allows agents to "rest" cognitively, shedding unused structure gracefully without catastrophic collapse.

\section{Philosophical and Cognitive Motivations}

The capacity to forget, or allow beliefs to fade, is not merely a limitation but a crucial functional aspect of robust cognition. Philosophically, forgetting prevents cognitive paralysis from information overload and allows focus to shift to relevant information. A mind that remembers everything perfectly may be less adaptive than one that prunes selectively. Constructivist views align with this, seeing memory not as static storage but as a dynamic structure maintained through active use; unused or unsupported beliefs naturally decay.

Cognitive psychology provides ample evidence supporting gradual belief decay. Decay theories posit that memory traces fade without rehearsal. Interference theories suggest that overlapping information can weaken specific beliefs. Attention plays a critical role; unattended or non-salient information is less likely to persist. Furthermore, neuroscience suggests active biological mechanisms are involved in forgetting, indicating it is a regulated, not purely passive, process.

Nullification aims to formalize these dynamics. Unlike the "hard reset" model common in many AI systems, which represents an abrupt external intervention, nullification models the internal, continuous drift towards semantic silence observed in natural cognition. This allows for graceful degradation, preserves a sense of identity continuity (as the agent doesn't cease to exist, its beliefs merely fade), and allows for the possibility of reactivation, mirroring how forgotten memories can sometimes resurface. Nullification completes the cognitive cycle initiated by assimilation: structure emerges from silence, is maintained through use, and dissolves back into silence when inactive.

\section{Formal Definition of Nullification}

We formalize nullification as a time-dependent transformation $N_t : \Phi \rightarrow \Phi$ that models the gradual erosion of semantic structure within a belief state $\phi$.

\textbf{Definition (Nullification Operator):} Let $\phi_t = N_t(\phi)$ be the belief state resulting from applying nullification for duration $t$ to state $\phi$. $N_t$ must satisfy $\lim_{t\rightarrow\infty} N_t(\phi) \in \Omega$.

A cognitively plausible $N_t$ should exhibit several properties:
\begin{itemize}
	\item \textbf{Monotonic Weakening:} The "amount" or "density" of semantic content should generally decrease or stay the same over time. If $S(\phi)$ measures semantic content, then $S(N_{t_2}(\phi)) \le S(N_{t_1}(\phi))$ for $t_2 > t_1$.
	\item \textbf{Anchor Sensitivity:} Beliefs that are strongly anchored (due to recency, frequency of use, structural centrality, or explicit designation) should decay more slowly than unanchored or peripheral beliefs.
	\item \textbf{Locality Preservation:} Nullification should preferentially affect weakly connected or isolated parts of the belief structure before disrupting core, highly integrated components. Coherent clusters tend to resist decay longer.
	\item \textbf{Continuity:} The transformation should be gradual over time, representing smooth decay rather than sudden disappearance (unlike Annihilation, $K$).
\end{itemize}
One way to implement this is elementwise. Let $\phi = \{\varphi_1, \dots, \varphi_n\}$. Associate each expression $\varphi_i$ with a persistence function $d_i(t) : \mathbb{R}^+ \rightarrow [0, 1]$, where $d_i(t)$ decreases monotonically towards 0 as $t \rightarrow \infty$. Define a persistence threshold $\delta \in (0, 1)$. Then, the nullified state is:
$$ N_t(\phi) = \{\varphi_i \in \phi \mid d_i(t) > \delta\} $$
The dynamics of $d_i(t)$ determine the specific decay behavior. For example, $\frac{d}{dt}d_i(t) = -\lambda_i(t)$, where the decay rate $\lambda_i$ depends on factors like anchoring strength $a_i$.

\begin{figure}[htbp]
	\centering
	\input{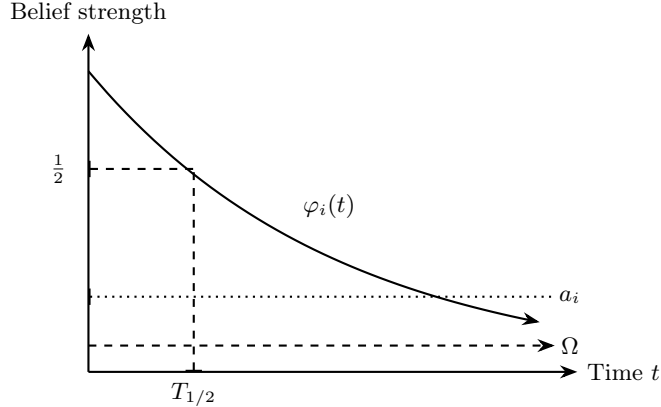}
	\caption{Temporal decay of a belief fragment \(\varphi_i(t)\) under nullification \(N_t\). The semantic half-life \(T_{1/2}\) determines when the belief decays to half strength. Anchoring level \(a_i\) provides resistance to full nullification. The state asymptotically approaches the epistemic vacuum \(\Omega\).}
	\label{fig:belief-decay}
\end{figure}

\section{Persistence, Anchoring, and Semantic Half-Life}

The rate at which beliefs fade under nullification depends on their stability, captured by persistence and anchoring.

\textbf{Definition (Persistence Function):} $d_i(t)$ quantifies the remaining strength or activation of belief $\varphi_i$ after time $t$ of nullification, with $d_i(0) = 1$ (or initial strength) and $\lim_{t\rightarrow\infty} d_i(t) = 0$. Common forms include exponential decay ($d_i(t) = e^{-\lambda_i t}$) or linear decay ($d_i(t) = \max(0, 1 - \lambda_i t)$).

\textbf{Definition (Anchoring):} The anchoring score $a_i$ of an expression $\varphi_i$ represents its resistance to nullification. Anchoring can depend on factors like:
\begin{itemize}
	\item Frequency or recency of use/retrieval/assimilation.
	\item Semantic centrality or structural connectedness within $\phi$.
	\item Explicit marking (e.g., as a goal, core belief, or policy component).
	\item Relevance to active tasks or semantic sectors $\Sigma$.
\end{itemize}
Anchoring modulates the decay rate: $\lambda_i = \lambda_0 \cdot f(a_i)$, where $f$ is a decreasing function (e.g., $f(a_i) = 1 / (1 + a_i)$ or $e^{-c a_i}$). \textbf{Crucially, higher anchoring $a_i$ leads to a lower decay rate $\lambda_i$.} Highly anchored beliefs decay slowly ($\lambda_i \approx 0$), while weakly anchored ones fade quickly.

\textbf{Definition (Semantic Half-Life):} $T_{1/2}(\varphi_i)$ is the time required for the persistence $d_i(t)$ to decay below a threshold $\delta$: $T_{1/2}(\varphi_i) = \min\{t \mid d_i(t) \le \delta\}$. It provides a measure of the belief's expected lifespan under neglect.

Decay rates $\lambda_i(t)$ need not be constant; they can vary based on the evolving context within $\phi$ or external attentional signals, allowing for adaptive nullification dynamics. Agents might even use persistence information for control, for instance, by refreshing beliefs whose $d_i(t)$ approaches $\delta$.

\section{Nullification Trajectories}

Nullification defines a trajectory $\gamma_{\phi}(t) = N_t(\phi)$ for each initial state $\phi$, tracing a path through the semantic state space $\Phi$ towards the epistemic vacuum $\Omega$. These trajectories exhibit characteristic structural changes:
\begin{itemize}
	\item \textbf{Detail Loss:} Peripheral, less anchored, or highly specific expressions (often at lower abstraction levels $\Phi^{(0)}$) tend to decay first.
	\item \textbf{Fragmentation:} As links (represented by shared context or structure) decay, coherent clusters of belief may break apart into disconnected fragments before disappearing entirely.
	\item \textbf{Core Preservation:} Highly interconnected or strongly anchored core beliefs often persist longer, forming temporary stable structures even as the rest of $\phi$ dissolves. The trajectory might appear to collapse inwards towards these cores.
	\item \textbf{Asymptotic Approach to $\Omega$:} Eventually, all expressions fall below the persistence threshold $\delta$, and the state $\gamma_{\phi}(t)$ enters the epistemic vacuum $\Omega$.
\end{itemize}

\begin{figure}[ht]
	\centering
	\input{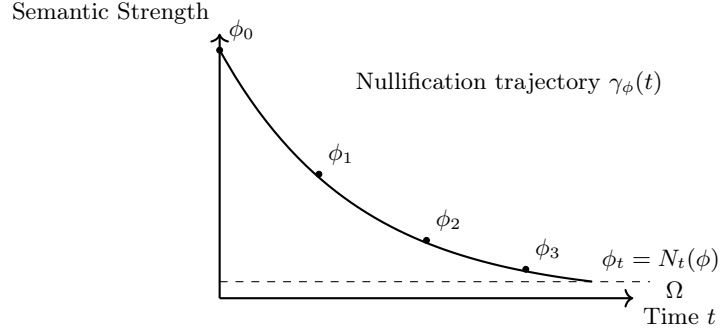}
	\caption{A nullification trajectory \(\gamma_\phi(t)\) showing the decay of a belief state \(\phi\) over time under the action of the nullification operator \(N_t\). Semantic strength decreases monotonically, approaching the epistemic vacuum \(\Omega\). Intermediate belief states \(\phi_t\) progressively lose detail and structure.}
	\label{fig:nullification_trajectory}
\end{figure}

Whether implemented continuously or in discrete time steps ($\phi_{t+1} = N_{\Delta}(\phi_t)$), the trajectory represents a gradual unraveling of semantic structure. Visualizing these trajectories in $\Phi$ (perhaps projected onto the Sector/Scale geometry) could reveal different decay patterns: smooth exponential decay, slow decay followed by rapid collapse (punctuated equilibrium), or even temporary stabilization followed by further decay. The shape of the trajectory depends heavily on the initial structure of $\phi$ and the anchoring landscape.

\section{Nullification as Semantic Control}

Beyond modeling passive forgetting, nullification can be actively employed as a mechanism for cognitive control and resource management.
\begin{itemize}
	\item \textbf{Overload Management:} Continuous assimilation can lead to cluttered and inefficient belief states. Applying a background nullification process automatically prunes low-salience, outdated, or weakly anchored information, maintaining semantic tractability.
	\item \textbf{Focus Preservation:} Nullification can be targeted to specific semantic sectors. An agent might apply $N_t$ more aggressively outside the currently active task sector(s) $\Sigma_{task}$, effectively quieting irrelevant thoughts and sharpening focus. We can denote this $N^{-\Sigma_{task}}_t$.
	\item \textbf{Dynamic Decay Rates:} Modulating decay rates $\lambda_i(t)$ based on attention, goals, or context acts as a form of controlled forgetting. Information deemed irrelevant can have its decay accelerated, while critical information can have its decay slowed or halted.
	\item \textbf{Policy-Guided Reset:} Instead of total annihilation ($K$), controlled nullification can achieve partial or "soft" resets. A policy might trigger accelerated nullification for beliefs related to a completed task or a failed plan, clearing context more gracefully than $K$.
	\item \textbf{Hierarchical Forgetting:} In architectures with layered memory or abstraction levels, nullification might operate faster on lower, more detailed levels ($\Phi^{(0)}, \Phi^{(1)}$) and slower on higher, more abstract levels ($\Phi^{(k)}, k \ge 2$), preserving core concepts while allowing specific instances to fade.
\end{itemize}
Used strategically, nullification becomes a tool for "epistemic hygiene," helping the agent manage complexity, maintain focus, and adapt its cognitive state without resorting to disruptive resets.

\section{Semantic Rest and Reawakening}

As nullification proceeds, $N_t(\phi)$ approaches the epistemic vacuum $\Omega$. States very close to or within $\Omega$ represent semantic rest---a condition of minimal or zero active semantic commitment. Unlike annihilation, reaching rest via nullification implies continuity; the agent's core architecture and potential remain, even if specific beliefs have faded.
\begin{itemize}
	\item \textbf{Reaching Rest:} $\phi_t = N_t(\phi)$ enters rest when essentially all $\varphi_i$ have $d_i(t) \le \delta$. The state $\phi_t \in \Omega$ might still contain latent structural traces or degraded anchors.
	\item \textbf{Reawakening:} The agent can exit semantic rest upon receiving new input $s$ or an internal trigger. Assimilation then operates on the resting state: $\phi_{new} = A(\phi_{rest}, X(s))$.
	\item \textbf{Reactivation:} This reawakening may reactivate dormant structures. If the new input $X(s)$ resonates with residual sub-threshold traces or anchors in $\phi_{rest}$, these elements might have their persistence $d_i(t)$ boosted above $\delta$, effectively "recalling" forgotten information or context.
\end{itemize}
This enables a complete epistemic cycle:
$$ \Omega \xrightarrow{\text{Assimilation } A} \phi \text{ (Active Belief)} \xrightarrow{\text{Nullification } N_t} \Omega \text{ (Semantic Rest)} \xrightarrow{\text{Assimilation } A} \dots $$
This cycle models the natural rhythm of active thought followed by quietude and potential reactivation, analogous to cycles of attention, task switching, or even sleep in biological cognition. It allows agents to persist and adapt over long timescales by balancing belief construction with controlled dissolution.

\begin{figure}[ht]
	\centering
	\input{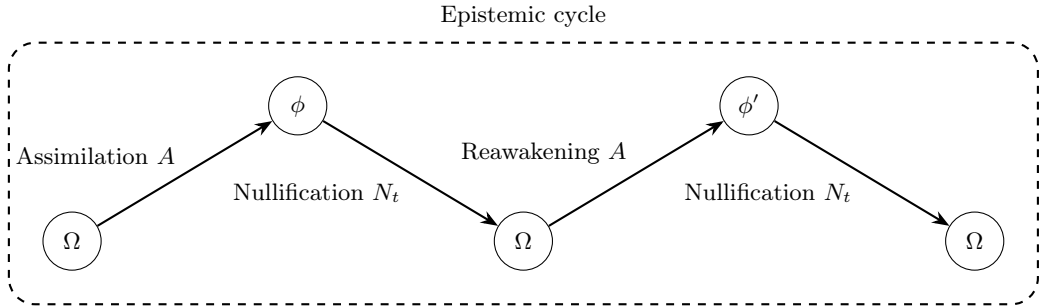}
	\caption{An epistemic cycle of belief construction and dissolution. Assimilation \(A\) lifts the agent from the epistemic vacuum \(\Omega\) to an active belief state \(\phi\). Nullification \(N_t\) gradually returns the belief state to semantic rest \(\Omega\). Upon new input or internal trigger, the agent reawakens, forming a new state \(\phi'\), continuing the cognitive cycle.}
	\label{fig:nullification_cycle}
\end{figure}

\section{Conclusion: Forgetting as Structure}

Nullification ($N_t$) provides a formal mechanism for modeling the gradual, structured dissolution of belief within the semantic state space $\Phi$. As the counterpoint to assimilation ($A$), it represents the passive fading or active pruning of semantic content over time, driving the belief state back towards the epistemic vacuum $\Omega$.

Unlike abrupt annihilation ($K$), nullification is continuous, sensitive to belief structure (via anchoring $a_i$ and coherence), and preserves the potential for reactivation. It models cognitive phenomena like forgetting, dormancy, and semantic rest. By formalizing persistence ($d_i(t)$), anchoring ($a_i$), and semantic half-life ($T_{1/2}$), we can characterize the decay trajectories $\gamma_{\phi}(t)$ and understand how belief ensembles fragment and dissolve.

Furthermore, nullification serves as a tool for semantic control, enabling overload management, focus preservation, and graceful resets. It completes the epistemic lifecycle $\Omega \leftrightarrow \phi$, allowing agents to cycle between active thought and semantic rest. Modeling this "structure in silence" is essential for building resilient, adaptive, and cognitively plausible artificial agents that can manage information and attention effectively over time. The next chapter explores the final dynamic operator: Annihilation, the mechanism of abrupt erasure.


\subsection*{Chapter Summary}
This chapter introduces Nullification ($N_t$) as the dynamic process modeling the gradual dissolution or fading of belief structures within the semantic state space $\Phi$ due to lack of reinforcement or processing. Contrasting with constructive Assimilation ($A$) and abrupt Annihilation ($K$), $N_t$ represents a continuous, often passive drift towards the epistemic vacuum ($\Omega$), embodying concepts like natural forgetting, semantic decay, and cognitive dormancy. The chapter formalizes $N_t$ based on desiderata like sensitivity to belief stability (anchoring, $a_i$) and structural coherence, potentially implemented via time-dependent persistence functions ($d_i(t)$) and decay thresholds ($\delta$). It defines anchoring ($a_i$) and semantic half-life ($T_{1/2}$) as key factors modulating decay rates ($\lambda_i$). Nullification trajectories ($\gamma_{\phi}(t)$) are characterized by detail loss and fragmentation, eventually approaching $\Omega$. Beyond passive decay, $N_t$ can also function as a mechanism for semantic control, enabling load management and focus preservation through selective or accelerated decay. The process allows for states of semantic rest near $\Omega$ from which the agent can be reawakened, potentially reactivating faded memory traces, thus modeling a complete epistemic cycle of activity and rest.
	\chapter{Annihilation}
\label{chap:Annihilation}

\section{Introduction: The Abrupt End of Belief}

The dynamics of belief evolution explored thus far encompass construction via Assimilation ($A$) and gradual dissolution via Nullification ($N_t$). However, the lifecycle of belief includes another, more drastic transformation: the sudden, discontinuous erasure of epistemic structure. This chapter introduces annihilation, the process modeling instantaneous belief elimination or reset within the semantic state space $\Phi$.

Annihilation represents a semantic rupture, fundamentally different from the gentle fading of nullification. It is often externally triggered, context-destroying, and irreversible without external scaffolding. We model this via the annihilation operator:
$$ K : \Phi \rightarrow \Omega $$
which maps any structured belief state $\phi \in \Phi$ directly to a state within the epistemic vacuum $\Omega$. Annihilation does not involve decay or structural transformation; it is an immediate replacement of structured belief with semantic nullity. This concept models phenomena such as:
\begin{itemize}
	\item Hard system resets in AI, where context and memory are wiped clean.
	\item Catastrophic forgetting in some learning paradigms.
	\item Potential cognitive analogues like dissociative fugue or traumatic amnesia where identity or memory is abruptly lost.
	\item Intentional cognitive clearing, such as purging a specific line of reasoning or task context.
\end{itemize}
While potentially destructive, annihilation, particularly in its targeted forms, can also serve as a powerful mechanism for control, safety, and reconfiguration. This chapter formalizes the $K$ operator, distinguishes it from $N_t$, explores its sector-specific variants, and examines its profound implications for agent identity, memory, and control.

\begin{figure}[h]
	\centering
	\input{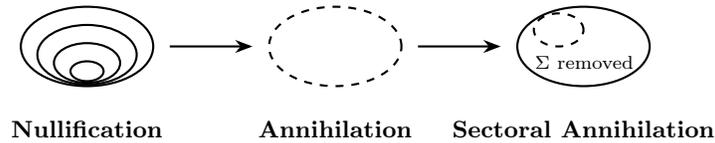}
	\caption{Comparison of nullification, annihilation, and sector-targeted annihilation. Nullification \(N_t(\phi)\) gradually collapses a belief state toward \(\Omega\); annihilation \(K(\phi)\) instantaneously reduces it to null; sector-targeted annihilation \(K_\Sigma(\phi)\) removes a semantic region while preserving the rest.}
	\label{fig:annihilation_comparison}
\end{figure}

\section{Philosophical and Cognitive Foundations}

The concept of total belief erasure ($K : \Phi \rightarrow \Omega$) evokes notions of cognitive death or a complete break in epistemic continuity. It parallels situations where an agent's entire knowledge base or identity structure is instantaneously lost or reset, leaving only the potential for a new beginning from the vacuum $\Omega$. This aligns with hard resets in computing and severe dissociative states in psychology.

However, annihilation need not be total. Cognition is often modular. We can introduce selective annihilation, targeting specific semantic sectors $\Sigma \subset \Phi$. The operator $K_{\Sigma} : \Phi \rightarrow \Phi$ removes only the belief content associated with sector $\Sigma$, leaving the rest of $\phi$ intact. This "epistemic surgery" has analogues in:
\begin{itemize}
	\item Focal Amnesia: Loss of specific memory types (e.g., episodic) while others remain.
	\item Stress-Induced Flattening: Collapse of abstract reasoning ($K_{abstract}$) under pressure, defaulting to reactive modes.
	\item Dissociative Partitioning: Blocking access to narrative or reflective sectors ($K_{narr}, K_{refl}$).
	\item Intentional Constraint: Voluntarily "switching off" planning ($K_{plan}$) or reflection ($K_{refl}$) to enter flow states or suppress interference.
\end{itemize}
Viewed this way, targeted annihilation becomes a tool for cognitive modulation. An agent could use $K_{\Sigma}$ to escape problematic reasoning loops, simplify its state by discarding irrelevant context (e.g., forgetting a completed task's details via $K_{task\_context}$), or rapidly shift cognitive modes.

This raises deep philosophical questions about identity ($\vec{\eta}$). If an agent annihilates its reflective sector, is it still the "same" agent? What constitutes continuous identity if core belief structures can be instantaneously erased? Annihilation forces us to consider whether identity resides in the specific content of $\phi$ or in the underlying architecture and potential ($\Phi, \Omega, \Lambda$, etc.).

We can consider a taxonomy based on the target:
\begin{itemize}
	\item $K$ (Total): $\rightarrow$ Fugue state, hard reset.
	\item $K_{refl}$: $\rightarrow$ Flow state, suppression of self-monitoring.
	\item $K_{abstract}$: $\rightarrow$ Cognitive flattening, panic/reactive mode.
	\item $K_{policy}$: $\rightarrow$ Goal abandonment, task switching.
	\item $K_{narr}$: $\rightarrow$ Disorientation, loss of temporal self.
\end{itemize}
This transforms annihilation from mere destruction into a potentially sophisticated, albeit drastic, form of epistemic control.

\section{Formal Definition of Annihilation}

We formally define annihilation as a discontinuous operator acting on the semantic state space $\Phi$.

\textbf{Definition (Total Annihilation Operator $K$):} The total annihilation operator $K : \Phi \rightarrow \Omega$ maps any belief state $\phi \in \Phi$ to some null state $\phi_{null} \in \Omega$.

Key properties distinguishing $K$ from Nullification $N_t$:
\begin{itemize}
	\item \textbf{Discontinuity:} $K$ acts instantaneously; it is not the limit of a continuous decay process. $\phi$ is replaced by $\phi_{null}$ in a single step.
	\item \textbf{Externality (often):} While potentially triggered internally by extreme conditions, $K$ often represents an external intervention (e.g., system reboot, user command).
	\item \textbf{Irreversibility:} $K(\phi)$ contains no information about $\phi$. Recovery requires external memory or complete re-assimilation from scratch via $A$.
	\item \textbf{Totality:} All semantic structure within $\phi$ (beliefs, policies, anchors, narrative context, etc.) is removed.
\end{itemize}
We also define the more nuanced sector-specific variant:

\textbf{Definition (Sector-Specific Annihilation $K_{\Sigma}$):} For a semantic sector $\Sigma$ (represented formally perhaps by a subset predicate or tag), the operator $K_{\Sigma} : \Phi \rightarrow \Phi$ removes the portion of the belief state belonging to that sector:
$$ K_{\Sigma}(\phi) = \phi \setminus \phi|_{\Sigma} $$
where $\phi|_{\Sigma}$ is the projection of $\phi$ onto sector $\Sigma$. Examples include $K_{reflective}(\phi)$, $K_{abstract}(\phi)$, $K_{policy}(\phi)$. These operators allow for targeted erasure while preserving other parts of the belief state $\phi$.

Extensions are possible:
\begin{itemize}
	\item \textbf{Chained Annihilation:} Applying multiple sectoral annihilations, e.g., $K_{refl} \circ K_{policy}(\phi)$.
	\item \textbf{Conditional Annihilation:} Triggering $K$ or $K_{\Sigma}$ based on internal state diagnostics (e.g., extreme incoherence $\kappa$, detection of corrupted state).
	\item \textbf{Annihilation Macros:} Pre-defined combinations of $K_{\Sigma}$ operations corresponding to specific functional resets (e.g., ENTER\_REACTIVE\_MODE := $K_{abstract} \circ K_{refl} \circ K_{narr}$).
\end{itemize}
It is crucial to reiterate that Annihilation ($K$) is distinct from the limit of Nullification ($N_t$). $N_t$ represents gradual fading, while $K$ represents abrupt removal.

\begin{table}[ht]
	\centering
	\begin{tabular}{@{} l l l @{}}
		\toprule
		\textbf{Feature} & \textbf{Nullification (\texorpdfstring{$N_t$}{N\_t})} & \textbf{Annihilation (\texorpdfstring{$K$}{K}, \texorpdfstring{$K_{\Sigma}$}{K\_Sigma})} \\
		\midrule
		Mechanism         & Gradual decay/fading                           & Abrupt erasure/removal \\
		Continuity        & Continuous over time                           & Discontinuous (instantaneous) \\
		Trigger           & Often internal/passive (time, low \(a_i\))      & Often external or extreme internal state \\
		Scope             & Selective based on \(a_i\)                      & Total (\(K\)) or Sector-specific (\(K_{\Sigma}\)) \\
		Reversibility     & Potential for reactivation                     & Irreversible loss of annihilated content \\
		Outcome           & State approaches \(\Omega\)                     & State maps to \(\Omega\) (\(K\)) or \(\phi \setminus \phi|_{\Sigma}\) (\(K_{\Sigma}\)) \\
		Identity Impact   & Gradual thinning/loss                          & Potentially catastrophic loss (\(K\)) or fracture (\(K_{\Sigma}\)) \\
		\bottomrule
	\end{tabular}
	\caption{Comparison of Nullification (\texorpdfstring{$N_t$}{N\_t}) and Annihilation (\texorpdfstring{$K$}{K}, \texorpdfstring{$K_{\Sigma}$}{K\_Sigma}).}
	\label{tab:nullification_vs_annihilation}
\end{table}

\section{Epistemic Implications and Control}

Applying annihilation has profound consequences for the agent's epistemic state and continuity.
\begin{itemize}
	\item \textbf{Loss of Coherence:} Total annihilation ($K$) destroys all internal coherence structures ($\kappa$), semantic links, and contextual grounding, returning the agent to a state of semantic vacuity in $\Omega$. Sectoral annihilation ($K_{\Sigma}$) disrupts coherence related to the removed sector, potentially requiring reintegration effort for the remaining state.
	\item \textbf{Loss of Narrative and Identity:} Belief states often scaffold the agent's sense of self, history, and ongoing narrative ($\Sigma_{narr}, \Sigma_{refl}$). Annihilating these sectors, even partially, can fracture the agent's identity continuity ($\vec{\eta}$), leaving it operable but potentially disoriented or inconsistent with its past self. The philosophical question arises: is the agent after $K_{narr}$ the "same" agent?
	\item \textbf{Intentional vs. Imposed Annihilation:} A distinction exists between annihilation used strategically by the agent (or its internal control policies $\pi_{regulate}$) to manage state (e.g., $K_{\Sigma}$ for mode switching) and annihilation imposed externally as a safety measure, compliance enforcement, or system reset ($K$).
	\item \textbf{Annihilation in Control Policies:} Agents might incorporate annihilation into their meta-control loops ($\pi_{regulate}$) as a final recourse for handling unresolvable states:
	$$ \pi(\phi) = \begin{cases} \text{Reason/Act}(\phi) & \text{if coherent} \\ \text{Attempt Repair}(\phi) & \text{if minor incoherence} \\ K_{\Sigma}(\phi) \text{ or } K(\phi) & \text{if catastrophic incoherence} \end{cases} $$
	This provides a mechanism for "epistemic hygiene" by cleanly purging corrupted or hopelessly contradictory states.
	\item \textbf{Suppression and Compliance:} $K_{\Sigma}$ can function as an active suppression mechanism, forcibly removing certain types of content or disabling cognitive functions (e.g., $K_{refl}$ to enforce reactivity, $K_{memory\_X}$ to comply with a deletion request).
	\item \textbf{Reconstruction Challenges:} Recovery from annihilation is non-trivial. Without external backups or logs, the annihilated information is lost. Reconstruction requires bootstrapping a new belief state via subsequent assimilation ($A$) from $\Omega$ or the remaining parts of $\phi$ after $K_{\Sigma}$. This is fundamentally different from reactivating faded beliefs after nullification ($N_t$).
\end{itemize}
Annihilation, therefore, represents the most extreme form of belief modification---a tool capable of enforcing safety, enabling radical reconfiguration, or causing catastrophic loss of identity, depending on its scope and context.

\begin{figure}[htbp]
	\centering
	\input{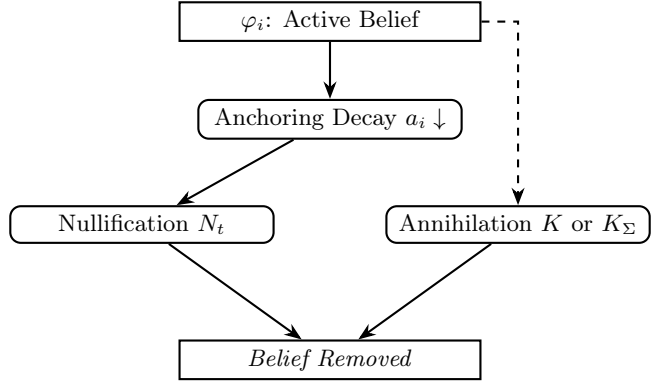}
	\caption{Nullification and Annihilation dynamics for a belief fragment \(\varphi_i\). Gradual anchoring decay leads to Nullification (\(N_t\)), while severe conflict or external triggers can cause abrupt Annihilation (\(K\) or \(K_\Sigma\)). Both result in the removal of the belief.}
	\label{fig:nullification-annihilation-process}
\end{figure}

\section{Conclusion: Annihilation in the Epistemic Lifecycle}

Annihilation, formalized by the operators $K$ (total) and $K_{\Sigma}$ (sector-specific), represents the dynamic of discontinuous, abrupt belief erasure within the semantic state space $\Phi$. It stands alongside Assimilation ($A$) and Nullification ($N_t$) as the third fundamental process governing the lifecycle of belief. While $A$ builds structure and $N_t$ allows it to fade gracefully, $K$ provides the mechanism for immediate removal, mapping structured belief $\phi$ back to the epistemic vacuum $\Omega$ (or a subset of $\phi$).

We have distinguished annihilation from nullification by its discontinuity and irreversibility, explored its cognitive and philosophical analogues (ranging from hard resets to focal amnesia), and formalized its total and sector-specific variants. While potentially representing catastrophic failure or identity loss, annihilation---especially $K_{\Sigma}$---also emerges as a powerful tool for cognitive control, enabling agents to manage complexity, enforce constraints, reset intentions, suppress interference, and rapidly reconfigure their epistemic state.

By incorporating annihilation into our framework, we acknowledge the full spectrum of belief dynamics, from gradual evolution to sudden rupture. Understanding the role and implications of annihilation is crucial for designing robust, safe, and controllable AI systems capable of managing their internal states effectively, even when that management requires the radical step of forgetting instantly. It completes the triad $A, N_t, K$, providing the agent with mechanisms to grow, to rest, and, ultimately, to reset.


\subsection*{Chapter Summary}
This chapter introduces Annihilation ($K$) as the dynamic operator modeling the abrupt, discontinuous erasure of belief structures within the semantic manifold $\Phi$. Unlike the gradual decay of Nullification ($N_t$), Annihilation represents an instantaneous reset, mapping a belief state $\phi$ directly to the epistemic vacuum $\Omega$ (total Annihilation, $K$) or removing specific functional components (sector-specific Annihilation, $K_{\Sigma}$, where $K_{\Sigma}(\phi) = \phi \setminus \phi|_{\Sigma}$). The chapter explores the philosophical and cognitive analogues, ranging from system resets and catastrophic forgetting to intentional cognitive clearing or functional suppression. While total Annihilation ($K$) signifies a profound break in epistemic continuity and potential identity ($\vec{\eta}$) loss, sectoral Annihilation ($K_{\Sigma}$) emerges as a potentially powerful, albeit drastic, mechanism for cognitive control, enabling targeted resets, task switching, or compliance enforcement within regulatory policies ($\pi_{regulate}$). Annihilation completes the fundamental triad of belief lifecycle dynamics ($A$, $N_t$, $K$), providing a mechanism for immediate removal of semantic content.
	\chapter{Spontaneous Drift}
\label{chap:SpontaneousDrift}

\section{Introduction: Beyond the Vacuum}

The epistemic dynamics explored thus far---Assimilation ($A$), Nullification ($N_t$), and Annihilation ($K$)---primarily describe how structured belief states $\phi \in \Phi$ evolve, are maintained, or are destroyed in response to input, time, or control signals. However, a fundamental question remains: how does semantic activity begin in the first place, particularly in the absence of external stimuli? Given an agent initialized in the epistemic vacuum $\Omega$, a state devoid of active content, what mechanism initiates the very first departure from semantic silence, even before structured observation encoding ($X$) or deliberate internal construction (Chapter~\ref{chap:BeliefConstruction}) takes hold? \textbf{Without such a mechanism, an agent starting in $\Omega$ might remain inert indefinitely unless prompted by external input via $X$.}

This chapter introduces the concept of spontaneous drift. We posit that the semantic state space $\Phi$ is not entirely quiescent, even near $\Omega$. Instead, minimal, undirected semantic motion can emerge spontaneously, representing a kind of cognitive Brownian motion or initial "awakening" from the vacuum. This drift provides the nascent semantic configurations upon which more structured processes can later act. It is the transition from pure potentiality ($\Omega$) to the first flicker of cognitive activity, ensuring the agent possesses a baseline level of internal dynamics.

\section{Defining the Drift Operator (\texorpdfstring{$D$}{D})}

To conceptualize this initial motion, we introduce a drift operator $D$. Unlike the other dynamic operators ($A, N_t, K$), $D$ represents intrinsic, undirected change within $\Phi$, particularly acting on states near or within $\Omega$.

\textbf{Conceptual Definition (Drift Operator):} The drift operator $D : \Phi \rightarrow \Phi$ represents spontaneous, small-scale semantic transformations. Its key characteristic is its action on the vacuum: if $\phi_0 \in \Omega$, then typically $D(\phi_0) = \phi_1$, where $\phi_1$ may contain minimal, perhaps unstable or incoherent, semantic content, thus $\phi_1 \notin \Omega$ (or is at least a different state within $\Omega$ with nascent structure).

Properties of Drift ($D$):
\begin{itemize}
	\item \textbf{Internally Driven:} Operates without necessary external input $s$ or explicit construction goals.
	\item \textbf{Initiating Change:} Its primary role is to move the state out of absolute semantic inertia ($\Omega$), providing the initial impetus for cognitive activity.
	\item \textbf{Potentially Incoherent:} Early drifting states $\phi_1, \phi_2, \dots$ generated by repeated application ($D(\phi_1) = \phi_2$, etc.) might lack internal coherence or stable meaning.
	\item \textbf{Non-Teleological:} Drift is undirected; it doesn't aim towards a specific goal or interpretation. It explores the immediate vicinity of the current state in $\Phi$.
\end{itemize}

\begin{figure}[h]
	\centering
	\input{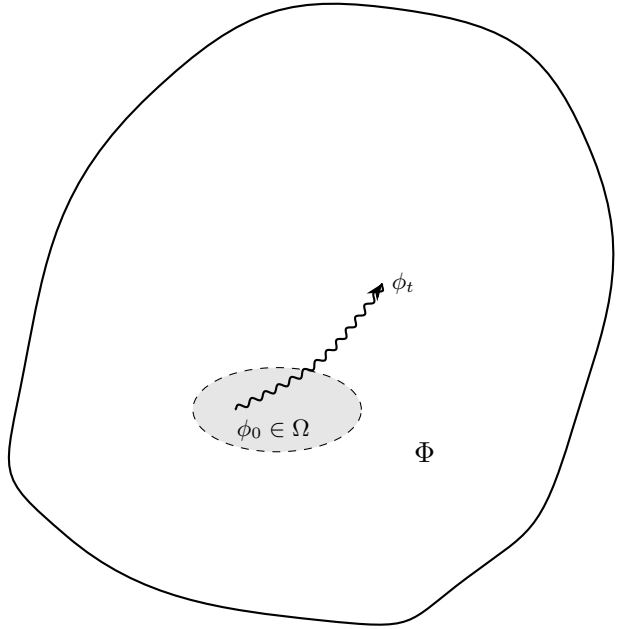}
	\caption{Spontaneous drift from the epistemic vacuum \(\Omega\) into the broader semantic manifold \(\Phi\). The stochastic trajectory illustrates how undirected fluctuations can initiate meaningful semantic structure formation.}
	\label{fig:spontaneous_drift}
\end{figure}

Repeated application of $D$ from a vacuum state $\phi_0 \in \Omega$ traces a trajectory
$$\phi_0 \xrightarrow{D} \phi_1 \xrightarrow{D} \phi_2 \xrightarrow{D} \dots$$ 
representing the agent's initial, undirected exploration of semantic possibilities near its origin.

\section{Origins and Mechanisms of Drift}

What underlying mechanisms could give rise to spontaneous drift? While the formal model remains abstract, we can speculate on potential sources consistent with the overall framework:
\begin{itemize}
	\item \textbf{Architectural Noise:} Residual activation noise in the underlying implementation (e.g., neural substrate) could manifest as small, random perturbations in the linguistic belief state, nudging it away from pure nullity.
	\item \textbf{Latent Vacuum Structure:} As discussed in 	Chapter~\ref{chap:SemanticStateSpace}, $\Omega$ is a class of states. Residual structural biases or potentials within specific null states $\phi_0 \in \Omega$ (perhaps related to the Null Tower's $\Omega^{(0)}$ basis) could provide seeds for initial drift patterns.
	\item \textbf{Default Background Processes:} Agents might have low-level background processes (e.g., periodic memory probes, default associative links) that generate minimal semantic activity even without focused tasks, akin to default mode network activity in humans.
	\item \textbf{Speculative Quantum Analogy:} The concept of a multi-state epistemic vacuum ($\Omega$) allows for a speculative analogy to quantum vacuum fluctuations. Just as virtual particles emerge from the physical vacuum, perhaps minimal "virtual" semantic fragments could momentarily arise from $\Omega$, initiating drift. 
\end{itemize}
Regardless of the specific mechanism, the key idea is that the state $\phi \in \Omega$ is not perfectly stable but possesses an inherent tendency towards minimal, spontaneous activity.

\section{Drift vs. Other Epistemic Dynamics}

It is crucial to distinguish spontaneous drift ($D$) from the other dynamics governing $\Phi$:
\begin{itemize}
	\item \textbf{vs. Observation Encoding ($X$):} $X$ maps specific external inputs $s$ to structured, meaningful belief states $\phi = X(s)$. Drift $D$ is internal, input-independent, and initially undirected and potentially incoherent.
	\item \textbf{vs. Assimilation ($A$):} $A$ integrates specific input $\phi_{input}$ into an existing state $\phi$, typically preserving or enhancing coherence. Drift $D$ generates novel, potentially incoherent content de novo from near $\Omega$.
	\item \textbf{vs. Nullification ($N_t$):} $N_t$ models the decay or fading of existing structured belief back towards $\Omega$. Drift $D$ models the initial emergence of semantic activity away from $\Omega$. They represent opposing directions of movement relative to the vacuum.
	\item \textbf{vs. Annihilation ($K$):} $K$ is an abrupt, often externally forced reset into $\Omega$. Drift $D$ is a gradual, internal emergence from $\Omega$.
\end{itemize}
Drift occupies a unique niche as the genesis of semantic motion, preceding structured construction and decay.

\section{Drift as Epistemic Substrate}

Spontaneous drift plays a vital foundational role: it provides the initial semantic "material" upon which more structured cognitive processes can operate. Without drift, an agent initialized in $\Omega$ might remain there indefinitely in the absence of external input via $X$. Drift ensures that there is always some minimal level of internal semantic activity, providing the necessary "spark" for cognition to begin internally. This initial, perhaps random or noisy, semantic content generated by $D$ serves as:
\begin{itemize}
	\item The first non-null states for operators like $A$ or reflective processes ($M$) to act upon.
	\item Potential seeds for pattern formation or association.
	\item The variations upon which internal selection mechanisms (e.g., coherence checks $\kappa$, anchoring $a_i$) can operate to stabilize meaningful structures.
\end{itemize}
In essence, drift provides the "symmetry breaking" required to transition the agent from the undifferentiated potentiality of the vacuum $\Omega$ to the beginnings of structured thought. It is the necessary precursor to belief construction when input is absent.

\section{Stabilization and Capture of Drift}

The semantic fragments generated by drift are initially transient and potentially incoherent. For them to become part of a stable belief state $\phi$, they must be captured or stabilized. This can occur through several pathways:
\begin{itemize}
	\item \textbf{Resonance with Input:} A drifting fragment $\phi_{drift}$ might happen to match or align with subsequent input $X(s)$, becoming reinforced and integrated via Assimilation ($A$).
	\item \textbf{Internal Coherence Capture:} A drifting configuration might randomly satisfy internal coherence constraints ($\kappa$) or resonate with existing stable anchors ($a_i$, if the agent is not starting from pure $\Omega$), causing it to be preserved and strengthened.
	\item \textbf{Architectural Attractors:} The agent's underlying architecture (perhaps related to Null Tower structures) might contain implicit attractors in $\Phi$ that capture and stabilize certain patterns emerging from drift.
	\item \textbf{Goal Relevance:} If a drifting thought happens to be relevant to a latent goal or drive, control mechanisms might select and amplify it.
\end{itemize}
Stabilization transforms ephemeral drift into incipient belief structures, providing the foothold needed for the constructive processes of Chapter~\ref{chap:BeliefConstruction} and Chapter~\ref{chap:Assimilation} to build upon.

Furthermore, the variations generated by spontaneous drift, even if initially incoherent, provide the essential raw material upon which stabilization mechanisms and coherence checks can operate. From an AI design perspective, this suggests a potential strategy: deliberately injecting controlled "semantic noise" or virtual fragments could serve a function analogous to stochasticity in reinforcement learning, actively seeding the belief space with variations to aid exploration, overcome local optima in reasoning, or foster creative recombination of concepts. Managing the integration and potential coherence cost ($\kappa$) of such deliberately introduced fragments would be a key challenge for the agent's regulatory policies ($\pi_{regulate}$).

\begin{figure}[ht]
	\centering
	\input{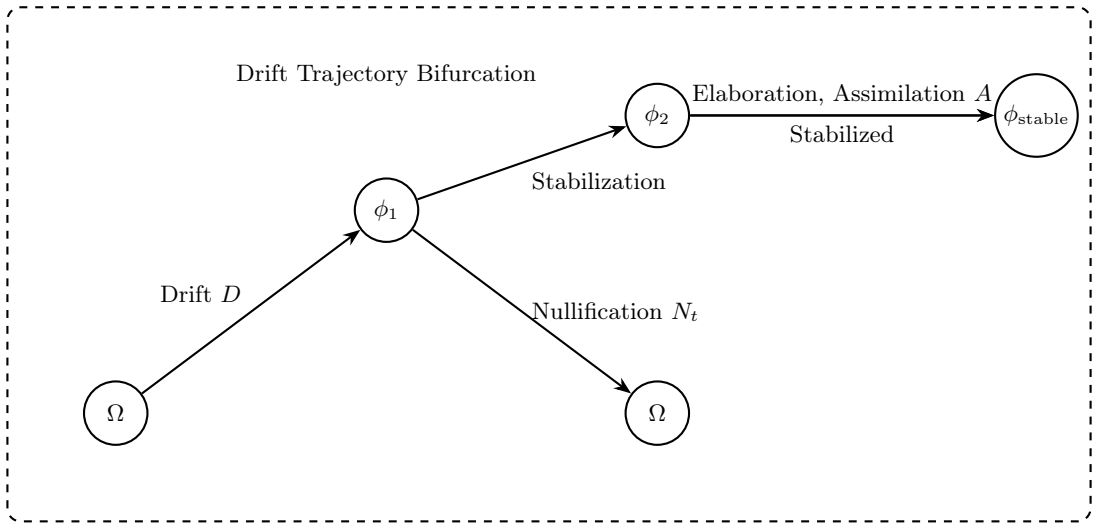}
	\caption{Spontaneous drift from the vacuum \(\Omega\) generates unstable belief state \(\phi_1\). These may either decay back to the vacuum via nullification \(N_t\), or---if stabilized by internal coherence, anchoring, or resonance with goals---transition into structured belief states \(\phi_{\text{stable}}\) via elaboration and assimilation \(A\). This bifurcation illustrates how drift can seed both rest and structured cognition.}
	\label{fig:drift_branching}
\end{figure}

\section{Conclusion: The Genesis of Semantic Motion}

Spontaneous drift ($D$) represents the initial, undirected emergence of semantic activity from the quiescence of the epistemic vacuum $\Omega$. It is the crucial dynamic that bridges the gap between the potential defined by $\Phi$ and $\Omega$, and the active, structured cognition involving observation ($X$) and assimilation ($A$). While potentially random or noisy in origin, drift provides the necessary initial variations and semantic substrate upon which coherence, anchoring, and goal-directed processes can subsequently operate.

Distinct from input-driven encoding ($X$), constructive integration ($A$), gradual decay ($N_t$), or abrupt erasure ($K$), spontaneous drift is the spark that initiates internal cognitive motion. It ensures that the agent is never truly static, always possessing a minimal capacity for endogenous semantic activity. Understanding drift completes our picture of epistemic dynamics, providing the starting point for the evolution of thought described in this Part. The subsequent Part~V will explore how these dynamics relate to semantic memory and retrieval.


\subsection*{Chapter Summary}
This chapter introduces Spontaneous Drift ($D$), a dynamic operator proposed to explain the initiation of semantic activity from the epistemic vacuum ($\Omega$) in the absence of external stimuli or explicit internal construction goals. Defined as $D : \Phi \rightarrow \Phi$, drift represents minimal, undirected, internally driven fluctuations that perturb null states, generating nascent, potentially unstable or incoherent semantic configurations outside of $\Omega$. It is distinct from input-driven Observation Encoding ($X$), structured Assimilation ($A$), decay towards $\Omega$ via Nullification ($N_t$), and abrupt erasure via Annihilation ($K$). Spontaneous Drift serves as the foundational epistemic substrate, providing the initial semantic variations upon which more structured processes like coherence checks ($\kappa$) or anchoring ($a_i$) can act. These transient drifted states require subsequent stabilization through mechanisms like resonance with input, internal coherence capture, or alignment with architectural attractors to become part of meaningful belief structures. Drift thus provides the necessary symmetry breaking to transition the agent from inert potentiality to active cognitive processing.
	
	\part{Semantic Memory: Retrieval and Integration}
	\label{part:semantic_memory}
	
	\chapter{The Structure of Semantic Memory within \texorpdfstring{$\Phi$}{Phi}}

\label{chap:StructureSemanticMemory}

\section{Introduction: Beyond Immediate Input}

The preceding Parts established the semantic state space $\Phi$ as the substrate for belief, structured by abstraction layers $\Phi^{(k)}$ and functional sectors $\Sigma$, and governed by core dynamics including Assimilation ($A$) of new information, Nullification ($N_t$) of inactive content, and Annihilation ($K$) of incoherent structures. These dynamics primarily address how beliefs are formed from observation (via the $S \rightarrow X \rightarrow A$ pathway) or internal generation ($A_{sim}, A_{refl}$), and how they decay or are resolved over time.

However, intelligent cognition relies crucially not only on processing novelty and maintaining coherence, but also on accessing and utilizing existing knowledge and past experiences stored within the agent's cognitive system. This necessitates an explicit account of semantic memory within the framework. This chapter defines the structure of this memory, not as a separate static store, but as an integral, potentially inactive yet retrievable, component of the overall belief space $\Phi$ itself. We explore how memories reside within $\Phi$, how their stability properties influence their accessibility, and how they contribute to the agent's persistent epistemic identity ($\vec{\eta}$). This lays the groundwork for subsequent chapters detailing the dynamic processes of memory retrieval ($R$) and integration (via $A$).

\section{Memory as Distributed Belief Structure}

Within this framework, semantic memory is conceptualized as the accumulated set of structured belief ensembles $\{\varphi_i\}$ that the agent has formed over its history, residing within the broader state space $\Phi$. We denote the potentially vast space containing these memories as $\Phi_{memory} \subseteq \Phi$. Memory is thus distributed and structured, rather than monolithic or stored in a separate location. Key aspects of this representation include:

\begin{itemize}
	\item \textbf{Ensemble Representation:} Like active beliefs, memories consist of structured linguistic expressions $\{\varphi_i\}$, potentially with internal linkages (semantic, causal, temporal).
	\item \textbf{Sectoral Distribution:} Memories are likely organized according to functional type within semantic sectors ($\Sigma$). For instance, episodic memories of specific events may reside primarily within $\Sigma_{narr}$, while factual knowledge or learned concepts might populate a distinct $\Sigma_{knowledge}$ or be integrated within domain-specific sectors.
	\item \textbf{Inactive vs. Active States:} Much of $\Phi_{memory}$ may be "inactive" at any given moment, meaning its constituent expressions $\varphi_i$ have low current activation levels ($\lambda(\varphi_i) \approx 0$) or are not part of the currently attended/processed belief state $\phi_{current}$. Inactivity does not imply erasure, but rather a state of potential retrievability.
\end{itemize}

This view treats memory as continuous with active belief, differing primarily in activation status and accessibility, governed by the dynamics described below.

\section{Activation, Anchoring, and Persistence in Memory}

The accessibility and stability of memories within $\Phi_{memory}$ are governed by the same properties influencing active beliefs, particularly anchoring ($a_i$) and persistence ($d_i(t)$) in the face of Nullification ($N_t$). Since memories are belief structures within $\Phi$, they are inherently subject to the gradual decay process modeled by $N_t$ (Chapter~\ref{chap:Nullification}). Their persistence hinges on counteracting this decay:
\begin{itemize}
	\item \textbf{Anchoring ($a_i$):} Memories representing core knowledge, frequently accessed experiences, or foundational aspects of identity ($\vec{\eta}$) are likely to possess high anchoring scores ($a_i$). This high anchoring makes them highly resistant to decay via Nullification ($N_t$), ensuring their long-term persistence within $\Phi_{memory}$. Conversely, transient or less significant experiences may have low $a_i$ and fade relatively quickly unless reinforced.
	\item \textbf{Persistence ($d_i(t)$):} The persistence function $d_i(t)$ associated with a memory fragment $\varphi_i$ reflects its current state of decay under $N_t$. Even if $d_i(t)$ falls below the threshold $\delta$ for active consideration, the fragment might still exist within $\Phi_{memory}$ in a latent state. Retrievability (explored in Chapter~\ref{chap:RetrievalOperator}) may depend on $d_i(t)$ not reaching zero, or on the ability of a retrieval cue to sufficiently reactivate it.
	\item \textbf{Activation Level ($\lambda$):} While inactive memories have low $\lambda$, successful retrieval (see Chapters~\ref{chap:QueryingBeliefSpace}-\ref{chap:RetrievalOperator}) involves accessing and potentially activating relevant memory fragments $\{\varphi_i\}$, which are assembled by $R$ into a retrieved belief state $\phi_{retrieved}$. This retrieved state, once integrated into $\phi_{current}$ via Assimilation ($A$), serves as the primary mechanism for reinforcing memories (increasing $a_i$) and thus counteracting the decay imposed by $N_t$.
\end{itemize}
The interplay between anchoring, persistence, and potential reactivation through the $Q \rightarrow R \rightarrow A$ cycle determines whether a specific memory trace endures and remains accessible over time against the constant pressure of Nullification ($N_t$).

\section{Sectoral Organization of Semantic Memory}

The functional organization provided by Semantic Sectors ($\Sigma$) naturally extends to the structure of memory within $\Phi_{memory}$. Different types of memories are likely predominantly associated with specific sectors:
\begin{itemize}
	\item $\Sigma_{narr}$: Primarily houses episodic memories, autobiographical narratives, and sequences of past events. Retrieval from $\Sigma_{narr}$ supports recounting past experiences and maintaining identity continuity.
	\item $\Sigma_{knowledge}$ (Hypothetical): Could represent a sector dedicated to stable semantic knowledge---facts, concepts, learned regularities---distinct from specific episodes. This knowledge might be highly anchored ($a_i$) and general (high $\Phi^{(k)}$ level).
	\item Domain-Specific Sectors (e.g., $\Sigma_{physics}, \Sigma_{social}$): May store learned knowledge and past experiences relevant to specific domains of reasoning or interaction.
	\item $\Sigma_{plan}$: Can contain memories of past plans, strategies, and their outcomes, informing future planning efforts.
\end{itemize}

\begin{figure}[ht]
	\centering
	\input{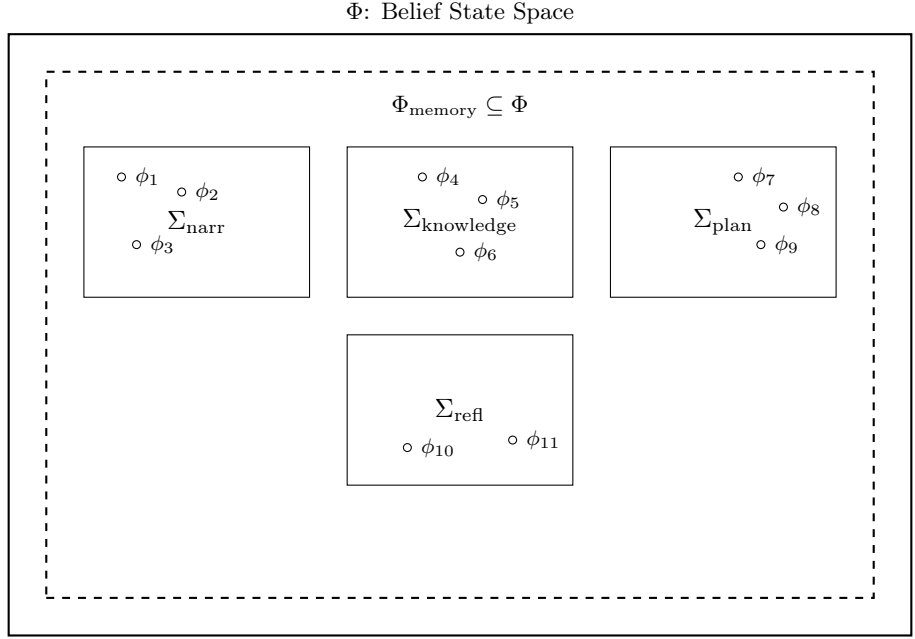}
	\caption{Semantic memory as a structured subspace \(\Phi_{\text{memory}} \subseteq \Phi\), distributed across semantic sectors such as episodic (\(\Sigma_{\text{narr}}\)), factual knowledge (\(\Sigma_{\text{knowledge}}\)), planning traces (\(\Sigma_{\text{plan}}\)), and reflective constructs (\(\Sigma_{\text{refl}}\)). Retrieval and integration processes selectively reactivate latent memories into the active belief state \(\phi\).}
	\label{fig:semantic_memory_structure_in_phi}
\end{figure}

This sectoral organization allows retrieval mechanisms (Operator $R$, Chapter~\ref{chap:RetrievalOperator}) to be potentially targeted, searching within relevant sectors based on the nature of the query ($\phi_{query}$). There may also be significant overlap and cross-sectoral linkages enabling associative recall across different memory types.

\section{Memory and Epistemic Identity (\texorpdfstring{$\vec{\eta}$}{eta})}

The structure and content of semantic memory are fundamental to the agent's Epistemic Identity ($\vec{\eta}$), defined conceptually as its long-term, self-consistent belief signature (Chapter~\ref{chap:EpistemicIdentity}). Stable, retrievable memories contribute to $\vec{\eta}$ by:
\begin{itemize}
	\item \textbf{Providing Narrative Continuity:} Autobiographical memories retrieved from $\Sigma_{narr}$ form the basis of the agent's sense of history and personal trajectory, essential components of identity.
	\item \textbf{Grounding Self-Models:} Reflective processes operating on retrieved memories (within $\Sigma_{refl}$) allow the agent to build and maintain models of its own capabilities, values, and limitations.
	\item \textbf{Ensuring Behavioral Consistency:} Retrieving past decisions, policies, or commitments helps the agent maintain consistent behavior over time, reinforcing its identity signature $\vec{\eta}$.
	\item \textbf{Anchoring Core Beliefs:} Highly anchored memories representing core values or functional roles act as stable attractors that contribute significantly to the long-term average potentially represented by $\vec{\eta}$.
\end{itemize}
Therefore, the structure of memory within $\Phi$, particularly the stability and accessibility of identity-relevant sectors like $\Sigma_{narr}$ and $\Sigma_{refl}$, is not merely passive storage but actively constitutes and maintains the agent's coherent identity over time. Understanding this structure is the first step towards modeling how memory is dynamically accessed and utilized.


\subsection*{Chapter Summary}
This chapter defines the structure of semantic memory within the framework, conceptualizing it not as a separate static repository, but as an integral, potentially inactive component of the overall belief space $\Phi$ (denoted $\Phi_{memory} \subseteq \Phi$). Memories, like active beliefs, consist of structured linguistic ensembles $\{\varphi_i\}$. They are distributed across functional semantic sectors ($\Sigma$), such as $\Sigma_{narr}$ for episodic accounts or potentially $\Sigma_{knowledge}$ for factual information. The persistence and accessibility of these memories are governed by their anchoring strength ($a_i$) which determines their resistance to gradual decay via Nullification ($N_t$). The cycle of retrieval ($R$) and subsequent integration via Assimilation ($A$) serves to reactivate and re-anchor memories, counteracting forgetting. This structured, dynamic memory is fundamentally linked to the agent's Epistemic Identity ($\vec{\eta}$), providing narrative continuity and grounding for self-models.
	\chapter{Querying Belief Space: The Genesis of Retrieval}
\label{chap:QueryingBeliefSpace}

\section{Introduction: Retrieval is Cue-Dependent}

Chapter~\ref{chap:StructureSemanticMemory} established the nature of semantic memory as structured, potentially inactive belief ensembles residing within the broader semantic state space $\Phi$. Accessing this stored information, however, is not typically a process of exhaustive search or random activation. Instead, memory retrieval is fundamentally cue-dependent; specific aspects of the agent's current cognitive state trigger the search for relevant stored information.

Before the Retrieval Operator ($R$, detailed in Chapter~\ref{chap:RetrievalOperator}) can access memory, a retrieval cue must be generated. This chapter focuses on the genesis of retrieval: the processes by which the agent's currently active belief state, $\phi_{current}$ (or $\phi_{active}$), gives rise to specific retrieval cues or queries, denoted $\phi_{query}$. We introduce a conceptual Query Function ($Q$) responsible for formulating these cues based on ongoing cognitive activity, goals, and internal state monitoring. Understanding query generation is essential, as the nature of the query profoundly shapes the subsequent retrieval process and determines what information is brought back into active consideration.

\section{Defining the Query Function (\texorpdfstring{$Q$}{Q})}

We postulate a Query Function, $Q$, which maps the agent's currently active or attended belief state to a specific cue that guides retrieval from memory.

\textbf{Definition 18.1 (Query Function $Q$).} The Query Function $Q$ is a mapping:
$$ Q : \Phi_{active} \rightarrow \Phi_{query} $$
where $\Phi_{active} \subseteq \Phi$ represents the subset of the belief space currently undergoing active processing or holding attentional focus, and $\phi_{query}$ is either a singleton belief fragment ($\{\varphi_i\}$) or a belief state composed of a small ensemble of fragments, derived from $\phi_{active}$ that constitutes the retrieval cue.

The function $Q$ does not perform the retrieval itself; rather, it identifies and formulates the specific content or question that primes the memory system for search. The output $\phi_{query}$ serves as the direct input to the Retrieval Operator $R$ (Chapter~\ref{chap:RetrievalOperator}). The transformation $Q$ might involve processes like extracting salient entities, identifying unresolved variables, isolating goal statements, or framing reflective prompts based on the content and structure of $\phi_{active}$.

\begin{figure}[ht]
	\centering
	\input{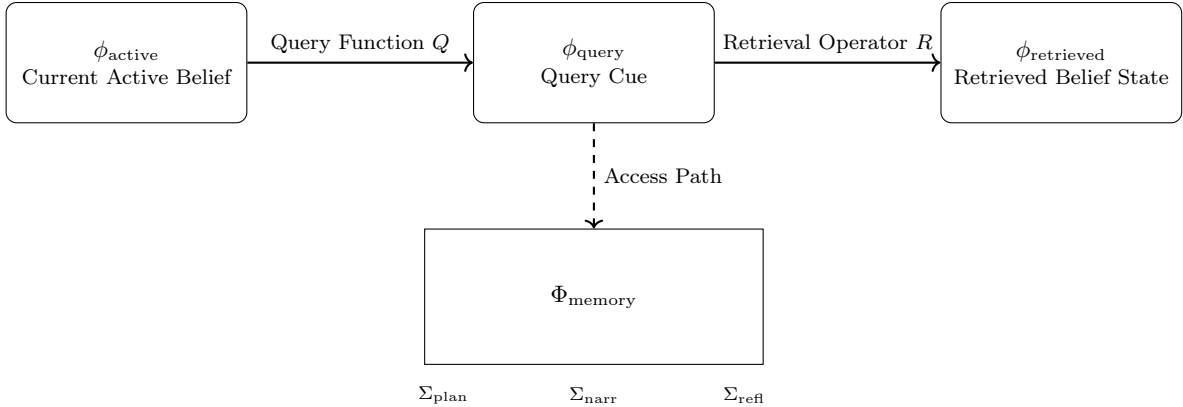}
	\caption{Query-driven memory retrieval. The active belief state \(\phi_{\text{active}}\) is processed by the query function \(Q\) to generate a retrieval cue \(\phi_{\text{query}}\), which directs the memory access performed by the retrieval operator \(R\), yielding a retrieved belief state \(\phi_{\text{retrieved}}\).}
	\label{fig:retrieval_query_flow}
\end{figure}

\section{Sources of Retrieval Cues}

Retrieval cues ($\phi_{query}$) can originate from various streams of cognitive activity, reflecting different underlying needs or triggers for memory access. Key sources include:
\begin{itemize}
	\item \textbf{Goal-Directed Queries:} Active goals residing within the planning sector ($\Sigma_{plan}$) often necessitate retrieving specific information. For example, a goal to "Navigate to location X" might trigger $Q$ to formulate $\phi_{query}$ with content such as "Retrieve map information for location X" or "Recall obstacles near location X". These queries are instrumental in grounding plans.
	\item \textbf{Reflective Queries:} Processes within the reflective sector ($\Sigma_{refl}$) frequently require accessing past experiences or self-knowledge. Meta-cognitive assessments, requests for explanation, narrative construction, or identity maintenance (Chapter~\ref{chap:EpistemicIdentity}) can trigger $Q$ to generate queries like "Retrieve episodes related to failure F", "Recall core value V", or "Access performance data for task T".
	\item \textbf{Coherence-Driven Queries:} The detection of low coherence ($\kappa(\phi)$) or specific contradictions within $\phi_{active}$ can trigger $Q$ to formulate queries aimed at resolving the inconsistency. For instance, detecting conflicting beliefs $\varphi_a$ and $\varphi_b$ might lead to $\phi_{query} $ containing "Retrieve evidence supporting $\varphi_a$ or $\varphi_b$". This links memory retrieval directly to coherence regulation.
	\item \textbf{Associative Cues:} Salient entities, concepts, or affective states within $\phi_{active}$ can spontaneously trigger associative retrieval via $Q$. Perceiving an object might trigger $Q$ to formulate a query for associated properties or past interactions, driven by learned or structural links within $\Phi$. This supports context enrichment and spreading activation models of recall.
\end{itemize}
The specific source often influences the structure and urgency of the generated query $\phi_{query}$.

\section{Properties of Queries}

The retrieval cue $\phi_{query}$ generated by $Q$ possesses properties that constrain and guide the subsequent retrieval process ($R$):
\begin{itemize}
	\item \textbf{Semantic Content:} The specific linguistic expressions $\{\varphi_i\}$ constituting $\phi_{query}$ determine the target information sought from memory.
	\item \textbf{Structural Complexity:} A query might range from a single entity or concept to a complex relational structure or scenario fragment. More complex queries may initiate more constrained searches.
	\item \textbf{Abstraction Level ($k$):} Queries can be formulated at different levels of abstraction $\Phi^{(k)}$. An abstract query (e.g., "Retrieve general strategies for problem type P") will likely trigger retrieval of different memories than a concrete query (e.g., "Recall exact steps taken in situation S").
	\item \textbf{Sectoral Scope ($\Sigma$):} Queries might implicitly or explicitly target specific memory sectors (e.g., querying $\Sigma_{narr}$ for episodes vs. $\Sigma_{knowledge}$ for facts).
\end{itemize}
The effectiveness of retrieval depends significantly on the clarity, specificity, and appropriateness of the query generated by $Q$. Poorly formulated queries may lead to retrieval failures or irrelevant results.

\section{Dynamics of Query Generation}

The process $Q$ of generating retrieval cues is itself a dynamic cognitive operation, likely influenced by the agent's overall state and resource allocation:
\begin{itemize}
	\item \textbf{Attention and Salience:} $Q$ likely operates primarily on the most salient or attended elements within $\phi_{active}$. Shifts in attention will thus change the potential queries generated.
	\item \textbf{Semantic Effort ($\epsilon$):} Formulating precise, goal-directed queries may require significant semantic effort ($\epsilon$). Under low effort, query generation might default to simpler associative cues. High effort might enable complex, multi-faceted query construction for deep retrieval.
	\item \textbf{Cognitive Load ($\lambda$):} High cognitive load ($\lambda$) may impair the effectiveness of $Q$, leading to poorly formed queries, increased reliance on simple cues, or even failure to generate necessary queries, hindering effective memory access under pressure.
	\item \textbf{Regulation:} Meta-cognitive control (originating from $\Sigma_{refl}$) might modulate $Q$, for instance, by prioritizing goal-relevant queries or suppressing associative queries during focused tasks.
\end{itemize}
Therefore, the ability to effectively query one's own semantic memory is not static but depends on the agent's current cognitive state, available resources ($\epsilon$), and regulatory priorities. This initial step of query generation sets the stage for the memory access performed by the Retrieval Operator $R$.


\subsection*{Chapter Summary}
This chapter addresses the initiation of memory retrieval, arguing it is fundamentally cue-dependent. Before the Retrieval Operator ($R$) can access stored information in $\Phi_{memory}$, a retrieval cue ($\phi_{query}$) must be generated based on the currently active belief state ($\phi_{active}$). The chapter introduces the conceptual Query Function ($Q: \Phi_{active} \rightarrow \Phi_{query}$) responsible for formulating this cue. It explores various sources for retrieval cues, including goal-directed needs arising from planning ($\Sigma_{plan}$), reflective inquiries originating in $\Sigma_{refl}$, coherence-driven needs triggered by inconsistencies ($\kappa$), and spontaneous associative cues based on salient elements in $\phi_{active}$. The properties of the generated query---its semantic content, structural complexity, abstraction level ($k$), and potential sectoral scope ($\Sigma$)---are shown to constrain the subsequent retrieval process performed by $R$. Finally, the chapter emphasizes that query generation itself is a dynamic cognitive process influenced by factors like attention, allocated semantic effort ($\epsilon$), current cognitive load ($\lambda$), and meta-cognitive regulation.
	\chapter{The Retrieval Operator: Accessing Stored Beliefs}
\label{chap:RetrievalOperator}

\section{Introduction: Defining the Access Mechanism}

Chapter~\ref{chap:QueryingBeliefSpace} established that memory access within the semantic manifold $\Phi$ is initiated by retrieval cues, $\phi_{query}$, generated by the Query Function ($Q$) based on the agent's current active belief state $\phi_{active}$. However, query generation is merely the first step; a dedicated mechanism is required to perform the actual search and access within the vast, potentially inactive portions of the agent's memory ($\Phi_{memory} \subseteq \Phi$) based on this cue.

This chapter introduces the Retrieval Operator ($R$), the core mechanism responsible for locating and retrieving relevant belief fragments from $\Phi_{memory}$  and constructing a corresponding retrieved belief state in response to a specific query $\phi_{query}$. $R$ acts as the bridge between the targeted informational need expressed by the query and the potentially relevant stored knowledge or experiences residing within the agent's semantic history. We will define its formal properties, explore potential mechanisms by which it operates, discuss factors influencing its success, and consider different modes and potential failures of retrieval. The output of $R$, the retrieved belief state $\phi_{retrieved}$, serves as the input for the integration process via Assimilation ($A$), detailed in Chapter~\ref{chap:MemoryIntegration}.

\section{Formal Definition of the Retrieval Operator (\texorpdfstring{$R$}{R})}

We define the Retrieval Operator $R$ as a function that takes the memory space to be searched and a query cue as input, and returns a belief state composed of relevant retrieved fragments.

\textbf{Definition 19.1 (Retrieval Operator $R$).} The Retrieval Operator $R$ is a mapping:
$$ R :  \mathcal{P}(\Phi_{memory}) \times \Phi_{query} \rightarrow \Phi $$
where:
\begin{itemize}
	\item $\Phi_{memory} \subseteq \Phi$ is the subset of the semantic state space representing the agent's potentially vast store of accumulated beliefs, including inactive or latent memories, and $\mathcal{P}(\Phi_{memory})$ is the powerset of $\Phi_{memory}$.
	\item $\phi_{query} \in \Phi_{query}$ is the retrieval cue generated by the Query Function $Q$, representing the specific information being sought.
	\item $\phi_{retrieved} = R(\Phi_{memory}, \phi_{query})$ is the output, a belief state ($\phi_{retrieved} \in \Phi$) composed of the relevant belief fragments $\{\varphi_i\}$ identified or reconstructed from $\Phi_{memory}$. In cases where no relevant fragments are found, $R$ might return a designated null state $\phi_{null} \in \Omega$.
\end{itemize}
It is crucial to note that $R$ itself does not modify the agent's active belief state $\phi_{current}$. It only accesses $\Phi_{memory}$ and returns the retrieved content $\phi_{retrieved}$. The integration of this retrieved content into $\phi_{current}$ is handled subsequently by the Assimilation operator $A$ (Chapter~\ref{chap:MemoryIntegration}).

\begin{figure}[ht]
	\centering
	\input{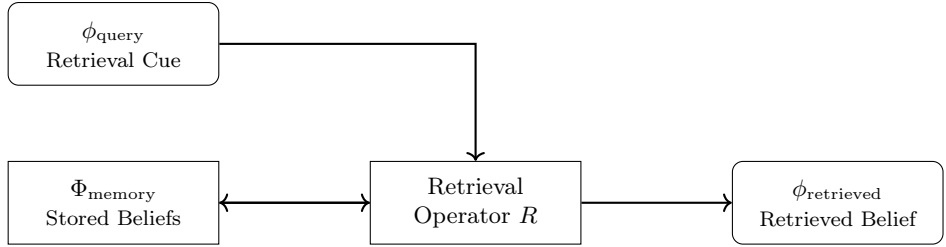}
	\caption{Operation of the Retrieval Operator \(R\). The agent's retrieval cue \(\phi_{\text{query}}\), generated by the Query Function \(Q\), initiates a search over \(\Phi_{\text{memory}}\). The Retrieval Operator \(R\) returns a belief state \(\phi_{\text{retrieved}}\) composed of relevant memory fragments.}
	\label{fig:retrieval_operator}
\end{figure}

\section{Mechanisms of Retrieval}

The operator $R$ can be realized through various computational mechanisms that search the structured space $\Phi_{memory}$ based on the cue $\phi_{query}$. Plausible mechanisms include:
\begin{itemize}
    \item \textbf{Similarity-Based Search:} Leveraging a semantic distance metric $d(\cdot, \cdot)$ defined over $\Phi$ (potentially learned, see Chapter~\ref{chap:LearningSemanticStructures}). $R$ could identify relevant memory fragments $\{\varphi_i\} \in \Phi_{memory}$ based on similarity (e.g., if their distance to the query $d(\varphi_i, \phi_{query})$ falls below a certain relevance threshold $\tau_{retrieval}$) and assemble them into the output state $\phi_{retrieved}$. This corresponds to retrieving memories "semantically close" to the query, with the effectiveness dependent on the quality of the metric $d$.
	\item \textbf{Associative Link Traversal:} If $\Phi_{memory}$ possesses internal structure (e.g., semantic links, causal graphs, temporal sequences), $R$ might operate by traversing these links starting from elements related to $\phi_{query}$. This models spreading activation or guided associative recall.
	\item \textbf{Pattern Completion:} Treating $\phi_{query}$ as a partial pattern or template, $R$ could attempt to retrieve and assemble belief fragments $\{\varphi_i\} \in \Phi_{memory}$ that match or complete the pattern $\phi_{query}$, forming the state $\phi_{retrieved}$. This is particularly relevant for recalling schemas or structured event representations.
	\item \textbf{Context-Dependent Shaping:} The specific content, structure, and abstraction level ($k$) of the query $\phi_{query}$ dynamically constrain the search performed by $R$. A precise query might lead to a narrow, targeted search, while a vague query might trigger broader associative retrieval.
\end{itemize}

In practice, $R$ might employ a hybrid approach, combining similarity matching (based on $d$) with link traversal, potentially guided by heuristics learned from prior retrieval successes.

\section{Factors Modulating Retrieval}

The success, speed, and content of the retrieval process $R$ are likely modulated by several factors intrinsic to the memory traces and the current state:
\begin{itemize}
	\item \textbf{Relevance/Similarity:} The primary factor is the semantic proximity ($d$) or structural match between the query $\phi_{query}$ and potential memory targets $\varphi_i$.
	\item \textbf{Anchoring ($a_i$):} Strongly anchored memories, representing core knowledge or significant experiences, might possess higher activation potential or lower retrieval thresholds, making them more readily accessible via $R$.
	\item \textbf{Recency and Persistence ($d_i(t)$):} Memories that have been recently accessed (and thus re-anchored via Assimilation, see Chapter~\ref{chap:MemoryIntegration}) or whose persistence $d_i(t)$ under Nullification ($N_t$) is still high, might be retrieved more easily or rapidly than deeply decayed traces.
	\item \textbf{Activation Thresholds:} Retrieval might require the "match score" (based on $d$, links, or pattern completion) to exceed a certain threshold $\tau_{retrieval}$, preventing the recall of weakly related or spurious information. This threshold might be dynamically adjustable based on context or effort ($\epsilon$).
	\item \textbf{Cognitive Load ($\lambda$) / Effort ($\epsilon$):} High cognitive load may impede the efficiency of the search process $R$, while allocating greater semantic effort ($\epsilon$) might enable more thorough or deeper searches within $\Phi_{memory}$.
\end{itemize}
These factors collectively determine the likelihood and nature of memories returned by $R(\Phi_{memory}, \phi_{query})$.

\section{Modes and Failures of Retrieval}

The Retrieval Operator $R$ may operate in different modes and is susceptible to various forms of failure:
\begin{itemize}
	\item \textbf{Directed Search:} A focused retrieval process aimed at finding specific information explicitly requested by a precise $\phi_{query}$, often originating from goal-directed or coherence-driven needs.
	\item \textbf{Associative Recall:} A less directed mode where activation spreads from $\phi_{query}$ through related memory fragments, potentially retrieving broader contextual information or seemingly unrelated but associatively linked memories.
	\item \textbf{Sector-Specific Retrieval ($R_{\Sigma}$):} The operator $R$ might be specialized or biased to search within specific memory sectors $\Sigma$ based on the query type or current context (e.g., $R_{\Sigma_{narr}}$ for episodic recall, $R_{\Sigma_{knowledge}}$ for factual lookup).
	\item \textbf{Retrieval Failure:} $R$ may return a designated null state ($\phi_{retrieved} = \phi_{null} \in \Omega$) if no sufficiently relevant memory fragments $\{\varphi_i\}$ can be found or reconstructed to form a meaningful state. This indicates an absence of relevant accessible memory (forgetting).
	\item \textbf{Partial Retrieval:} $R$ might construct a state $\phi_{retrieved}$ based on incomplete memory fragments, for instance if only parts of a stored ensemble meet the retrieval criteria or if the memory trace itself is degraded (due to $N_t$).
	\item \textbf{Biased or Distorted Retrieval/Reconstruction:} Beyond simple failure or partiality, the mechanisms implementing $R$ might inherently favor certain types of memories (e.g., emotionally salient ones, recently primed ones) or reconstruct information that is subtly altered or biased compared to the originally stored belief, reflecting reconstructive tendencies even at the access stage. While reconstruction is often associated with post-retrieval integration (Chapter~\ref{chap:MemoryIntegration}), biases within $R$ itself are plausible.
\end{itemize}
Understanding these modes and failure conditions is crucial for modeling realistic memory access, including phenomena like forgetting, associative thought, and memory biases. The output state $\phi_{retrieved}$ (whether representing a complete memory, a partial reconstruction, a null state signifying failure, or a potentially biased reconstruction) then proceeds to the integration stage (Chapter~\ref{chap:MemoryIntegration}).


\subsection*{Chapter Summary}
This chapter introduces the Retrieval Operator ($R$), the core mechanism responsible for accessing stored beliefs within the memory space ($\Phi_{memory}$) based on a retrieval cue ($\phi_{query}$) generated by the Query Function ($Q$). Formally defined as $R : \mathcal{P}(\Phi_{memory}) \times \Phi_{query} \rightarrow \Phi$, the operator searches for relevant belief fragments ($\{\varphi_i\}$) and constructs a retrieved belief state ($\phi_{retrieved}$), or returns a null state ($\phi_{null} \in \Omega$) if no relevant memories are found. Plausible retrieval mechanisms discussed include similarity-based search utilizing a semantic metric ($d$), associative link traversal within structured memory, and pattern completion. The success and nature of retrieval are modulated by factors such as the query's relevance, the memory traces' anchoring strength ($a_i$) and persistence ($d_i(t)$), activation thresholds ($\tau_{retrieval}$), cognitive load ($\lambda$), and applied semantic effort ($\epsilon$). The chapter also outlines different retrieval modes (directed vs. associative, potentially sector-specific $R_{\Sigma}$) and common failure types (complete failure, partial retrieval, biased reconstruction). The output $\phi_{retrieved}$ serves as the input for the subsequent integration step handled by Assimilation ($A$).
	\chapter{Integration of Retrieved Memories via Assimilation}
\label{chap:MemoryIntegration}

\section{Introduction: Incorporating Retrieved Beliefs}
\label{sec:IntegrationIntro}

Chapter~\ref{chap:QueryingBeliefSpace} and \ref{chap:RetrievalOperator} detailed the processes initiating memory access within the semantic state space $\Phi$: the generation of a query cue ($\phi_{query}$) by the Query Function ($Q$) and the subsequent retrieval of the relevant retrieved belief state ($\phi_{retrieved}$) from the agent's memory store ($\Phi_{memory}$) by the Retrieval Operator ($R$). However, the act of retrieval alone is insufficient for the retrieved information to meaningfully influence the agent's ongoing cognition. This retrieved state, $\phi_{retrieved}$, must be actively incorporated and reconciled with the agent's currently active belief state, $\phi_{current}$.

This chapter addresses this crucial integration step. We propose that the existing Assimilation operator ($A$), introduced in Chapter~\ref{chap:BeliefConstruction} primarily for integrating new external or internally generated information, is the appropriate mechanism for integrating retrieved internal information as well. By treating $\phi_{retrieved}$ as the input $\phi_{input}$ to the assimilation process, the framework can leverage the sophisticated machinery already defined for coherence management, contextual elaboration, and belief revision. We will explore how A handles potential conflicts, enriches retrieved memories, and, critically, how this integration process interacts with memory stability by counteracting Nullification ($N_t$) through re-anchoring.

\section{Assimilating Retrieved Memories: Formalism}
\label{sec:IntegrationFormalism}

The core proposal is to utilize the Assimilation operator $A: \Phi \times \Phi_{input} \rightarrow \Phi$ for integrating retrieved memories. The output of the Retrieval operator $R$, $\phi_{retrieved}$, serves as the input to be assimilated into the currently active belief state $\phi_{current}$.

\textbf{Definition 20.1 (Memory Integration via Assimilation).} Given a current active belief state $\phi_{current} \in \Phi$ and the retrieved belief state $\phi_{retrieved}$ = $R(\Phi_{memory}, \phi_{query})$, the updated belief state $\phi_{new}$ resulting from the integration of these memories is given by:
$$ \phi_{new} = A(\phi_{current}, \phi_{retrieved}) $$

Here, $\phi_{retrieved}$ takes the role of $\phi_{input}$ as described in the general definition of Assimilation (Chapter~\ref{chap:BeliefConstruction}). This approach ensures that retrieved memories are not simply added disjointly but are woven into the existing semantic fabric, subject to the same principles of coherence and contextual relevance as any other form of incoming information.

\begin{figure}[htbp]
	\centering
	\input{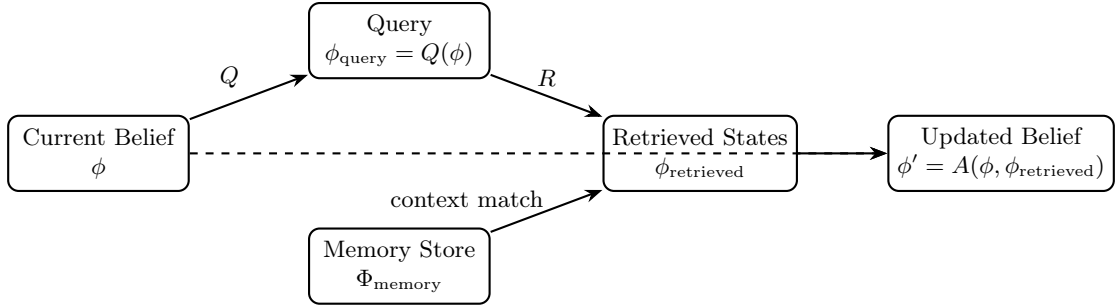} 
	\caption{The query--retrieval--assimilation loop. A query \(\phi_{\text{query}} = Q(\phi)\) is generated from the current belief state \(\phi\), used to retrieve state \(\phi_{\text{retrieved}}\) from memory \(\Phi_{\text{memory}}\), which is then assimilated via \(A\) to produce the updated belief \(\phi'\).}
	\label{fig:query-retrieval-loop}
\end{figure}

\section{Coherence Management During Integration}
\label{sec:IntegrationCoherence}

A key challenge when integrating retrieved memories is that they might conflict with the agent's current beliefs ($\phi_{current}$), perhaps because the memory is outdated, represents a different perspective, or conflicts with more recently acquired information. The Assimilation operator $A$, particularly through its corrective subtype $A_{corr}$ (Chapter~\ref{chap:BeliefConstruction}), is equipped to handle this.

When $A(\phi_{current}, \phi_{retrieved})$ is computed:
\begin{itemize}
	\item \textbf{Conflict Detection:} The operator assesses potential contradictions or inconsistencies between fragments $\varphi_i \in \phi_{current}$ and fragments $\varphi_j \in \phi_{retrieved}$. This relies on the agent's ability to detect semantic incompatibility.
	\item \textbf{Belief Revision:} If conflicts are detected, a revision component activates to determine which beliefs (from $\phi_{current}$ or potentially even within $\phi_{retrieved}$) should be retracted or modified to restore coherence ($\kappa$). Revision strategies might prioritize current beliefs, highly anchored beliefs, or beliefs with stronger evidential support.
	\item \textbf{Coherent Merging:} The final state $\phi_{new}$ represents a reconciled state where the valuable information from $\phi_{retrieved}$ has been incorporated while minimizing or resolving conflicts, thus preserving or restoring the overall coherence $\kappa(\phi_{new})$ of the belief state.
\end{itemize}
Using $A$ for integration thus ensures that memory recall does not necessarily lead to an incoherent belief state, providing a mechanism for reconciling past knowledge with present understanding.

\section{Contextual Elaboration of Retrieved Memories}
\label{sec:IntegrationElaboration}

Retrieved memories often gain new significance or require reinterpretation when placed within the agent's current context $\phi_{current}$. The elaborative aspect of Assimilation, $A_{elab}$ (Chapter~\ref{chap:BeliefConstruction}), plays a crucial role here.

Upon integrating $\phi_{retrieved}$, $A_{elab}$ can:
\begin{itemize}
	\item \textbf{Generate Contextual Links:} Create new semantic links between elements of $\phi_{retrieved}$ and elements of $\phi_{current}$.
	\item \textbf{Draw New Inferences:} Deduce new conclusions based on the combination of current and retrieved beliefs.
	\item \textbf{Reinterpret Past Events:} Frame the retrieved state $\phi_{retrieved}$ in light of subsequent events or current goals present in $\phi_{current}$.
	\item \textbf{Update Associated Beliefs:} Modify other related beliefs in $\phi_{current}$ based on the implications of the retrieved memory.
\end{itemize}
This ensures that retrieved memories are not just passively inserted but are actively contextualized and utilized within the ongoing stream of cognition.

\section{Re-Anchoring Dynamics and Counteracting Nullification}
\label{sec:IntegrationReanchoring}

A critical function of integrating retrieved memories via $A$ is its interaction with the dynamics of forgetting, specifically Nullification ($N_t$, Chapter~\ref{chap:Nullification}). Recalling and using information strengthens its representation. This can be modeled within the framework:

\begin{itemize}
	\item \textbf{Boosting Activation ($\lambda$):} Successfully assimilating fragments $\varphi_j \in \phi_{retrieved}$ into $\phi_{new}$ brings them into the active belief state, inherently increasing their activation level $\lambda(\varphi_j)$.
	\item \textbf{Strengthening Anchoring ($a_i$):} The act of retrieval and successful integration can be interpreted as reinforcing the importance or validity of the retrieved memory state. This should lead to an increase in the anchoring score $a_i$ of the corresponding expressions $\{\varphi_j\}$ within $\phi_{new}$.
	\item \textbf{Resetting Persistence ($d_i(t)$):} As anchoring $a_i$ increases for a fragment $\varphi_i$, its decay rate under Nullification $N_t$ decreases (since the rate typically depends inversely on $a_i$). This effectively resets or significantly slows down the decay clock $d_i(t)$ for the retrieved memory fragments, counteracting forgetting.
\end{itemize}
This mechanism formalizes the psychological principle that recall strengthens memory traces. The $R \rightarrow A$ cycle is thus the primary means by which semantic memory is actively maintained against the background decay pressure of $N_t$.

\section{Specialized Memory Assimilation (\texorpdfstring{$A_{mem}$}{A\_mem})?}
\label{sec:IntegrationAmem}

Does the assimilation of the internally retrieved state ($\phi_{retrieved}$) require fundamentally different rules than the assimilation of external observations ($X(s)$) or reflective insights ($A_{refl}$)? While the core mechanisms of coherence checking ($A_{corr}$) and elaboration ($A_{elab}$) seem applicable, one might consider if a specialized subtype, $A_{mem}$, is warranted.

Arguments for $A_{mem}$:
\begin{itemize}
	\item \textbf{Source Tagging:} The agent might treat retrieved memories differently based on awareness of their internal origin (e.g., potentially lower trust if memory is known to be old or previously unreliable). $A_{mem}$ could incorporate source-based weighting during conflict resolution.
	\item \textbf{Reconstruction Bias:} Memory recall is often reconstructive. $A_{mem}$ might include specific biases or heuristics reflecting how memories are typically reconstructed or potentially distorted during recall and integration.
\end{itemize}
Arguments against needing $A_{mem}$:
\begin{itemize}
	\item \textbf{Framework Sufficiency:} The existing components of $A$ ($A_{corr}$, $A_{elab}$) are designed to handle context and conflict, which are the main challenges when integrating any information, internal or external.
	\item \textbf{Uniformity:} Using the general $A$ operator maintains conceptual uniformity across different information sources. Source information could potentially be encoded as meta-data within the belief fragments themselves rather than requiring a distinct operator.
\end{itemize}
Tentatively, the existing Assimilation framework (Chapter~\ref{chap:BeliefConstruction}) appears flexible enough to handle the integration of retrieved memories without requiring a fundamentally distinct $A_{mem}$ operator, although specific policies within $A$ might be sensitive to the internal source of $\phi_{retrieved}$. The crucial aspect is that retrieved memories are subjected to the same coherence and contextualization processes as other inputs via $A$.


\subsection*{Chapter Summary}
This chapter details the crucial step following memory retrieval ($R$): the integration of the retrieved belief state ($\phi_{retrieved}$) into the agent's currently active state ($\phi_{current}$). It proposes that this integration is handled by the general Assimilation operator ($A$), specifically $\phi_{new} = A(\phi_{current}, \phi_{retrieved})$. Utilizing $A$ ensures that retrieved memories are actively woven into the existing semantic context, leveraging established mechanisms for coherence management ($A_{corr}$) to resolve potential conflicts between past and present beliefs, and contextual elaboration ($A_{elab}$) to reinterpret or draw new inferences from the recalled information. A critical function highlighted is the role of this integration process in memory maintenance: successfully assimilating $\phi_{retrieved}$ strengthens the anchoring ($a_i$) of the corresponding memory fragments, thereby counteracting the decay effects of Nullification ($N_t$). The chapter briefly considers, but tentatively dismisses, the need for a specialized memory assimilation operator ($A_{mem}$), suggesting the general $A$ framework is sufficient.
	\chapter{Memory Dynamics in Cognitive Processes}
\label{chap:MemoryDynamics}

\section{Introduction: The Functional Role of Semantic Retrieval}

The preceding chapters (Chapters~\ref{chap:StructureSemanticMemory}-\ref{chap:MemoryIntegration}) have detailed the structure of semantic memory within the belief space $\Phi$ and introduced a mechanism for its dynamic access, comprising Query Generation ($Q$), the Retrieval Operator ($R$), and integration via Assimilation ($A$). This $Q \rightarrow R \rightarrow A$ cycle allows the agent to selectively bring stored knowledge and past experiences back into its active belief state $\phi_{current}$.

However, memory retrieval is not an end in itself. Its primary importance lies in its contribution to other high-level cognitive functions. This chapter explores the crucial role that these semantic memory dynamics play in supporting and interacting with key processes detailed elsewhere in this monograph, including planning, simulation, reflection, identity maintenance, and learning. By integrating retrieved memories, the agent can ground its future actions, understand its past, evaluate its present state, and adapt its internal structures over time. Understanding these interactions reveals memory retrieval as a core dynamic underpinning sophisticated, coherent cognition within the semantic manifold.

\section{Memory in Planning and Decision Making}

Effective planning and decision-making, typically orchestrated within the planning sector ($\Sigma_{plan}$), rely heavily on accessing relevant past knowledge and experiences stored in $\Phi_{memory}$. The $Q \rightarrow R \rightarrow A$ cycle provides the mechanism for this:
\begin{itemize}
	\item \textbf{Informing Goal Generation ($\Gamma$):} The goal decomposition head ($\Gamma$) may issue queries ($Q$) to retrieve information about the feasibility, prerequisites, or typical outcomes associated with potential high-level goals before committing to them.
	\item \textbf{Retrieving Relevant Knowledge:} When constructing a plan within $\Sigma_{plan}$, queries ($Q$) can be generated to retrieve relevant factual knowledge (from $\Sigma_{knowledge}$ via $R_{\Sigma_{knowledge}}$), procedural steps, or causal models stored in memory. $A(\phi_{current}, \phi_{retrieved})$ integrates this knowledge into the planning context.
	\textbf{Example:} Planning a trip might trigger $Q$ for "Retrieve average flight cost to Destination Z". $R$ accesses this fact (within the returned $\phi_{retrieved}$), and $A$ incorporates it into the budget calculation within $\Sigma_{plan}$.
	\item \textbf{Accessing Past Outcomes:} Queries can target $\Sigma_{narr}$ via $R_{\Sigma_{narr}}$ to retrieve memories of similar past situations, plans attempted, and their resulting success or failure. Assimilating ($A$) these outcomes informs the evaluation of candidate actions and strategies.
	\textbf{Example:} Considering strategy X triggers $Q$ for "Recall outcomes of using strategy X". $R$ retrieves a state $\phi_{retrieved}$ containing fragments like \{"Strategy X failed in Situation Y"\} from $\Sigma_{narr}$. $A$ integrates this, potentially lowering the priority of strategy X in the current plan.
	\item \textbf{Parameterizing Decisions:} Retrieved information (e.g., statistics, specific constraints, object properties) integrated via $A$ provides necessary parameters for decision models or policy selection mechanisms operating within $\Sigma_{plan}$.
\end{itemize}
Memory retrieval thus provides the grounding and contextual knowledge necessary for effective, informed planning and decision-making, moving beyond purely reactive or rule-based approaches.

\section{Memory in Simulation and Prediction}

Embodied Simulation (Chapter~\ref{chap:EmbodiedSimulation}) allows agents to internally rehearse actions and predict outcomes. This process is significantly enhanced by access to semantic memory:
\begin{itemize}
	\item \textbf{Grounding Simulation Parameters:} Simulation trajectories ($\gamma_{sim}$) often require specific parameters (e.g., object properties, environmental dynamics, agent capabilities). Queries ($Q$) can retrieve these parameters from $\Phi_{memory}$ via $R$, and Assimilation ($A$) integrates the retrieved state $\phi_{retrieved}$ (containing these parameters) into the simulation context, making predictions more realistic.
	\textbf{Example:} Simulating pouring liquid triggers $Q$ for "Retrieve viscosity of water". $R$ retrieves the value (within $\phi_{retrieved}$), and $A$ incorporates it, affecting the simulated flow dynamics.
	\item \textbf{Retrieving Scenarios for Comparison:} Agents can query ($Q$) and retrieve ($R$) memories of similar past scenarios (returned as $\phi_{retrieved}$) to initialize or validate internal simulations. Comparing simulated outcomes with retrieved actual outcomes helps refine internal world models.
	\item \textbf{Contextualizing Simulated Events:} As a simulation unfolds within $\Phi$, elements within the simulated trajectory ($\gamma_{sim}$) can trigger associative queries ($Q$) that retrieve ($R$) and assimilate ($A$) relevant context or potential consequences (from the returned $\phi_{retrieved}$) from memory, enriching the simulation beyond simple state transitions.
\end{itemize}
Memory retrieval provides the rich knowledge base required for meaningful internal simulation, enabling more accurate prediction and effective evaluation of potential actions before execution.

\section{Memory in Reflection and Identity Maintenance}

The agent's capacity for self-awareness and maintaining a stable identity ($\vec{\eta}$) relies fundamentally on accessing its own history and self-concept stored in memory. This involves interplay between the $Q \rightarrow R \rightarrow A$ cycle and the reflective sector ($\Sigma_{refl}$) and identity vector ($\vec{\eta}$):
\begin{itemize}
	\item \textbf{Constructing Narratives:} Reflective processes in $\Sigma_{refl}$ often initiate queries ($Q$) targeting the narrative sector ($\Sigma_{narr}$). Retrieval ($R_{\Sigma_{narr}}$) results in a state $\phi_{retrieved}$ containing episodic fragments, which are then assimilated ($A$) and potentially elaborated ($A_{elab}$) within $\Sigma_{refl}$ to construct coherent autobiographical narratives. 
	\textbf{Example:} A query $Q$ for "Retrieve events leading to Project P completion" allows $R$ to access relevant episodes (returning them as $\phi_{retrieved}$) from $\Sigma_{narr}$, which $A$ integrates into $\Sigma_{refl}$ to form a narrative of success.
	\item \textbf{Updating Self-Models:} Queries retrieve past performance data, recorded internal states, or previously formulated self-assessments. Assimilating the retrieved state $\phi_{retrieved}$ (containing this information) allows the agent to monitor and update its explicit self-model within $\Sigma_{refl}$.
	\item \textbf{Reinforcing Epistemic Identity ($\vec{\eta}$):} By consistently retrieving ($R$) and reintegrating ($A$) core memories, values, and commitments (highly anchored elements $a_i$ in $\Phi_{memory}$ returned within $\phi_{retrieved}$), the agent reinforces the patterns that constitute its long-term epistemic identity signature $\vec{\eta}$ (Chapter~\ref{chap:EpistemicIdentity}). The $R \rightarrow A$ cycle actively maintains identity against drift and decay ($N_t$).
\end{itemize}
Memory retrieval is thus constitutive of reflective self-awareness and the continuity of identity over time.

\section{Memory in Learning and Adaptation}

Beyond immediate cognitive functions, the dynamics of memory retrieval play a crucial role in longer-term learning and adaptation (Part~\ref{part:learning_and_adaptation}):
\begin{itemize}
	\item \textbf{Driving Abstraction ($\Lambda$):} Observing recurring patterns across multiple retrieved states $\{\phi_{retrieved, 1}, \phi_{retrieved, 2}, \ldots\}$ obtained from similar queries ($\phi_{query}$), or within the structure of individual complex retrieved states $\phi_{retrieved}$, can provide the input necessary for learning new, higher-level abstractions via the $\Lambda$ operator (Chapter~\ref{chap:SemanticScaling}). Retrieval makes past regularities available for generalization. 
	\textbf{Example:} Repeatedly retrieving ($R$) states representing memories of successful negotiations that involved compromise might provide the data for $\Lambda$ to form the abstract principle "Compromise often leads to success" at a higher $\Phi^{(k)}$ level.
	\item \textbf{Refining Retrieval ($R$) and Querying ($Q$):} The success or failure of retrieval attempts (i.e., the nature of the returned $\phi_{retrieved}$, including potential $\phi_{null}$) can serve as a learning signal. If queries consistently fail to retrieve useful information, the agent might adapt its Query Function $Q$ (Chapter~\ref{chap:LearningCognitiveOperators}). If retrieval yields irrelevant results, the similarity metrics ($d$) or link structures used by $R$ could be refined.
	\item \textbf{Adapting Assimilation ($A$):} The consequences of assimilating retrieved states $\phi_{retrieved}$ (e.g., impact on coherence $\kappa$, predictive accuracy, or goal success) could provide feedback for adapting the policies within $A$ (e.g., how to handle conflicts, how much to elaborate) (Chapter~\ref{chap:LearningCognitiveOperators}).
	\item \textbf{Updating World Models:} Discrepancies between retrieved memories of past outcomes (contained in $\phi_{retrieved}$) and the predictions of internal models (used in simulation) can drive the learning and refinement of those models.
\end{itemize}
The memory retrieval cycle is therefore deeply intertwined with the agent's capacity to learn from experience and adapt its cognitive architecture.

\section{Conclusion: Memory Retrieval as a Core Cognitive Dynamic}

This chapter has illustrated that the internal memory retrieval cycle---initiated by Query Generation ($Q$), executed by the Retrieval Operator ($R$), and completed by integration via Assimilation ($A$)---is far more than a simple data lookup mechanism. It is a fundamental cognitive dynamic deeply integrated with planning, simulation, reflection, identity maintenance, and learning.

By providing access to the agent's accumulated knowledge and experience stored within the semantic state space $\Phi$ (via the retrieved state $\phi_{retrieved}$), the $Q \rightarrow R \rightarrow A$ process enables context-sensitive reasoning, grounded prediction, coherent self-awareness, and long-term adaptation. It transforms the semantic manifold from a space of potential beliefs into a dynamic substrate for knowledgeable, historically-aware, and adaptive cognition. Understanding these memory dynamics is therefore essential for realizing the full potential of the structured belief framework.

\begin{figure}[ht]
	\centering
	\input{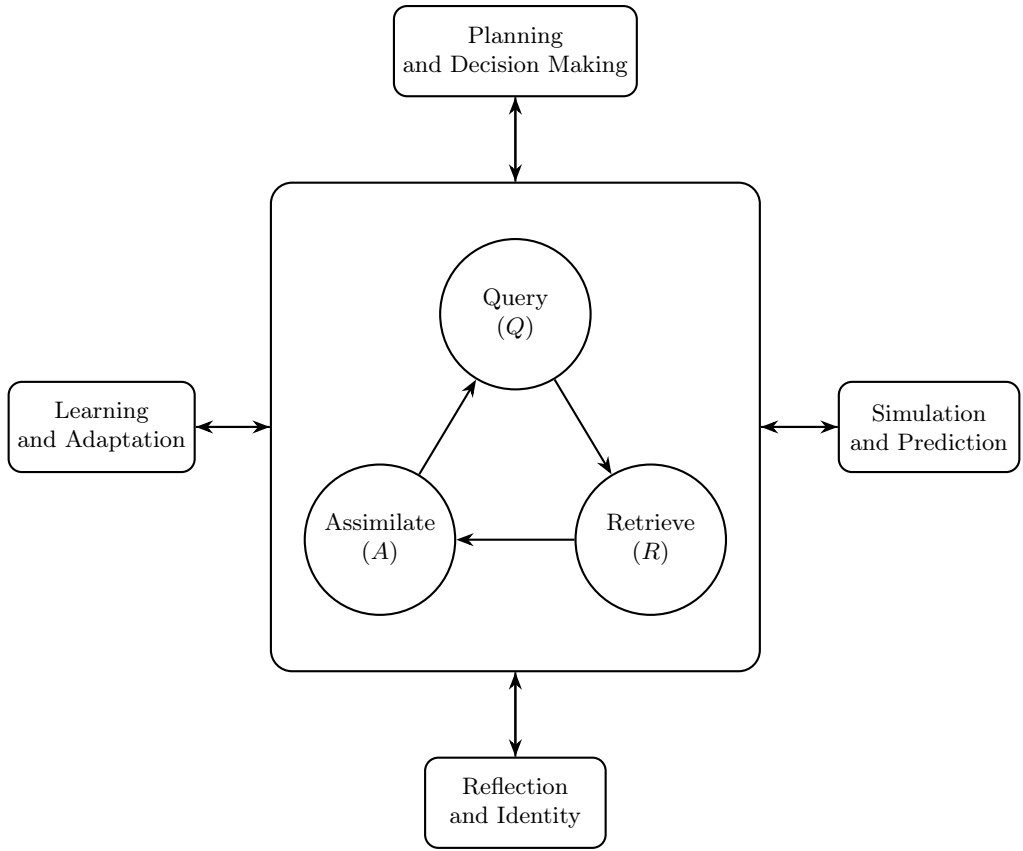}
	\caption{Memory retrieval as an internal module supporting external cognitive functions. Planning, simulation, reflection, and learning processes issue queries ($Q$) to the memory retrieval system, comprising sequential Query Generation ($Q$), Retrieval ($R$), and Assimilation ($A$) operations. Retrieved knowledge is integrated back into the agent's active belief state ($\phi_{current}$) and used to inform ongoing cognitive processes. The two-way interaction---queries entering the memory system and assimilated results returning---enables context-sensitive reasoning, grounded simulation, self-model updating, and adaptive learning within the semantic manifold $\Phi$.}
	
	\label{fig:memory_dynamics_cycle}
\end{figure}


\subsection*{Chapter Summary}
This chapter explores the functional significance of semantic memory dynamics---specifically the query-retrieval-assimilation cycle ($Q \rightarrow R \rightarrow A$)---demonstrating its crucial role in supporting various high-level cognitive processes within the semantic manifold $\Phi$. Retrieved memories are shown to inform planning and decision-making by providing relevant knowledge, past outcomes, and necessary parameters. They enhance embodied simulation and prediction by supplying realistic parameters, comparative scenarios, and contextual enrichment. Furthermore, memory retrieval is fundamental to reflection and identity maintenance ($\vec{\eta}$), enabling the construction of autobiographical narratives within $\Sigma_{refl}$ based on episodes retrieved from $\Sigma_{narr}$, updating self-models, and reinforcing identity through the consistent reintegration of core beliefs. Finally, the chapter highlights the role of memory dynamics in learning and adaptation, as retrieved experiences provide the basis for forming abstractions ($\Lambda$), refining retrieval and assimilation operators ($R$, $Q$, $A$), and updating internal world models. Thus, the $Q \rightarrow R \rightarrow A$ cycle is positioned as a core cognitive dynamic that integrates stored knowledge into ongoing processes, enabling informed, historically-aware, and adaptive intelligence.
	
	\part{Regulation and Control: Compass, Gauge, Identity}
	\label{part:regulation_and_control}
	
	\chapter{Semantic Orientation}
\label{chap:SemanticOrientation}

\section{Introduction: Navigating Belief Space}

The preceding parts established the foundations of the semantic state space $\Phi$, its emergent structure via the Null Tower, methods for belief construction, and its organization through semantic scaling and sectors. However, possessing a structured belief space is insufficient; an agent must be able to navigate it purposefully, maintain coherence over time, and align its internal state with its goals or principles. This requires mechanisms for regulation and control, which form the focus of this Part.

This chapter introduces Semantic Orientation, a formal framework addressing how an agent can situate, regulate, and direct its belief state $\phi$ within the complex landscape of $\Phi$. Building directly upon the recursive structure of the Null Tower (detailed in Chapter~\ref{chap:NullTower}), we leverage the concept of epistemic axes---directed trajectories $[\omega \rightarrow \omega^{(\infty)}]$ spanning from irreducible null states $\omega$ to their limiting abstractions $\omega^{(\infty)}$. We propose that these axes, originating from the generative process described in Chapter~\ref{chap:NullTower}, serve as internal reference structures, analogous to cardinal directions, allowing the agent to define a semantic compass.

This internal compass enables the agent to:
\begin{itemize}
	\item Maintain epistemic alignment with coherent abstraction principles or conceptual frames.
	\item Detect semantic drift when its belief trajectory deviates from these guiding axes.
	\item Perform reflective reorientation by projecting its current state back towards coherent paths.
	\item Utilize multi-axis modulation to navigate using layered or context-dependent reference frames.
\end{itemize}
Semantic orientation transforms belief from mere content into a positioned state within a navigable semantic manifold. It provides the agent with the capacity to regulate its own cognitive motion, ensuring stability, coherence, and directionality in thought.

\section{Background and Preliminaries Recap}

The concept of semantic orientation relies heavily on the foundational structures developed previously:
\begin{itemize}
	\item \textbf{Semantic State Space ($\Phi$):} The space of all structured linguistic belief ensembles, stratified by abstraction levels $\Phi = \bigcup_k \Phi^{(k)}$.
	\item \textbf{Epistemic Vacuum ($\Omega$):} The set of semantically null states, containing the irreducible null stratum $\Omega^{(0)}$.
	\item \textbf{Null Tower Trajectories:} For each $\omega \in \Omega^{(0)}$, the sequence $\omega^{(k)} = \Lambda^k(\omega)$ generated by recursive application of abstraction operators $\Lambda$.
	\item \textbf{Semantic Singularity ($\omega^{(\infty)}$):} The limiting state $\lim_{k\rightarrow\infty} \omega^{(k)}$, representing the fixed point of abstraction for the trajectory originating from $\omega$.
	\item \textbf{Epistemic Axis:} The directed semantic trajectory $[\omega \rightarrow \omega^{(\infty)}]$ defined by the Null Tower construction (Chapter~\ref{chap:NullTower}).
\end{itemize}
These elements provide the underlying structure upon which the mechanisms of orientation and the semantic compass are built.

\section{Epistemic Axes and Semantic Directionality}

The Null Tower naturally induces directional structure within $\Phi$. Each path from an irreducible null state to its abstraction limit defines a canonical direction of semantic ascent.

\textbf{Definition (Epistemic Axis):} For $\omega \in \Omega^{(0)}$ with corresponding convergent trajectory $\{\omega^{(k)}\}$ limiting to $\omega^{(\infty)}$, the epistemic axis is the directed path $[\omega \rightarrow \omega^{(\infty)}]$.

This axis represents more than just a line connecting two points; it embodies a coherent flow of semantic transformation through increasing abstraction. We can analyze its geometric properties by embedding states in a suitable vector space (assuming representation mode $\rho$ allows this):
\begin{itemize}
	\item \textbf{Direction Vector:} $\vec{v}_{\omega} := \omega^{(\infty)} - \omega$. This vector captures the overall semantic direction from the origin $\omega$ to its abstraction limit.
	\item \textbf{Normalized Basis:} $\hat{v}_{\omega} := \vec{v}_{\omega} / \|\vec{v}_{\omega}\|$. These normalized vectors can form a basis for orientation.
	\item \textbf{Interpretation as Flow:} Movement along the axis corresponds to distillation of regularities, suppression of specifics, and emergence of general semantic form.
\end{itemize}
An agent may possess multiple epistemic axes $[\omega_i \rightarrow \omega^{(\infty)}_i]$ derived from different cognitive seeds or domains. These axes $\{\hat{v}_{\omega_i}\}$ can span a subspace $\mathcal{A} = \text{span}\{\hat{v}_{\omega_1}, \dots, \hat{v}_{\omega_n}\}$, defining a local orientational basis within $\Phi$.

Relative to a given axis $[\omega \rightarrow \omega^{(\infty)}]$, we can measure the alignment of any belief state $\phi \in \Phi$:
\begin{itemize}
	\item \textbf{Projection:} $\pi_{\omega}(\phi)$, the component of $\phi$ (relative to $\omega$) along the axis direction $\vec{v}_{\omega}$.
	\item \textbf{Angular Deviation:} $\theta(\phi, \vec{v}_{\omega})$, the angle between the belief's position vector (relative to $\omega$) and the axis vector $\vec{v}_{\omega}$.
	\item \textbf{Residual Offset:} $r(\phi) := \|(\phi - \omega) - \pi_{\omega}(\phi - \omega)\|$, the distance of the belief from the axis.
\end{itemize}

\begin{figure}[h]
	\centering
	\input{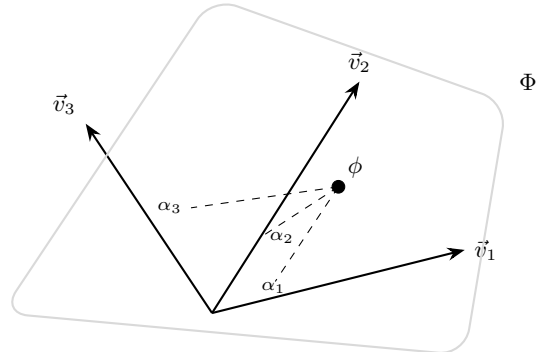}
	\caption{Semantic orientation represented as projection of a belief state \(\phi\) onto a local reference frame defined by epistemic axes \(\{\vec{v}_1, \vec{v}_2, \vec{v}_3\}\). The coordinates \((\alpha_1, \alpha_2, \alpha_3)\) define the orientation of \(\phi\) in this semantic basis.}
	\label{fig:semantic_orientation_projection}
\end{figure}

These measures allow the agent to quantify its semantic position and deviation relative to principled abstraction trajectories.

\section{The Semantic Compass}

Building on epistemic axes, we define the semantic compass as an internal mechanism providing orientation relative to these axes.

\textbf{Definition (Semantic Compass Components):} Relative to a reference axis $[\omega \rightarrow \omega^{(\infty)}]$ with direction $\vec{v}_{\omega}$, and for a current belief state $\phi$, the compass provides:
\begin{itemize}
	\item The reference direction $\vec{v}_{\omega}$.
	\item The projection $\pi_{\omega}(\phi)$ of the relevant part of $\phi$ onto the axis.
	\item The angular deviation $\theta(\phi, \vec{v}_{\omega})$.
	\item The residual offset $r(\phi)$.
\end{itemize}
The compass supports several operational modes:
\begin{itemize}
	\item \textbf{Passive Monitoring:} Tracking alignment $\theta_t$ over time.
	\item \textbf{Drift Detection:} Identifying when $\theta_t$ or $r(\phi_t)$ exceeds a threshold.
	\item \textbf{Reflective Realignment:} Triggering corrective actions to reduce deviation, such as projecting the current state back towards the axis ($\phi_t \mapsto \omega + \pi_{\omega}(\phi_t - \omega)$) or backtracking.
	\item \textbf{Meta-cognitive Anchoring:} Constraining internal processes (planning, generation) to maintain alignment with the reference axis.
\end{itemize}

\begin{figure}[htbp]
	\centering
	\input{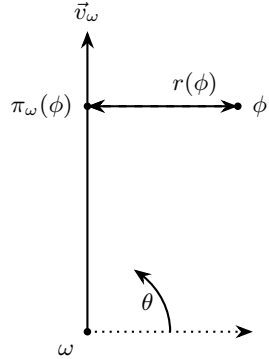}
	\caption{Semantic Compass visualization. The belief state \(\phi\) is projected onto the epistemic axis \([ \omega \rightarrow \omega^{(\infty)} ]\), defined by direction vector \(\vec{v}_\omega\). The projection \(\pi_\omega(\phi)\), angular deviation \(\theta(\phi, \vec{v}_\omega)\), and residual offset \(r(\phi)\) provide the basis for introspective orientation.}
	\label{fig:semantic-compass}
\end{figure}

The compass enables navigational feedback. By monitoring the belief trajectory $\phi_t$, the agent can compute instantaneous angular drift $\dot{\theta}_t$ and directional momentum $\vec{m}_t = \phi_t - \phi_{t-1}$. If drift is detected, corrections like $\phi_t \mapsto \pi_{\omega}(\phi_t)$ (simplified notation for projection relative to origin $\omega$) can restore alignment.

Furthermore, the agent can use the compass for semantic steering---proactively guiding its cognitive trajectory to minimize deviation from a desired axis (e.g., aligning with a causal axis $\vec{v}_{causal}$ during explanation, or a goal axis $\vec{v}_{goal}$ during planning). The compass can be extended to handle multiple axes $\{\vec{v}_i\}$, operating relative to a preferred axis, a weighted blend, or a dynamically chosen reference based on context.

\section{Orientation Metrics and Control Policies}

To make semantic orientation functional, we need quantitative metrics and control policies based on them.

\textbf{Metrics:}
\begin{itemize}
	\item Angular Deviation: $\theta(\phi, \vec{v}_{\omega}) = \cos^{-1} \left( \frac{\langle \phi-\omega, \vec{v}_{\omega} \rangle}{\|\phi-\omega\| \cdot \|\vec{v}_{\omega}\|} \right)$. Measures alignment; low $\theta$ means aligned, high $\theta$ means misoriented.
	\item Projection and Residual Offset: $\pi_{\omega}(\phi) = \omega + \frac{\langle \phi-\omega, \vec{v}_{\omega} \rangle}{\|\vec{v}_{\omega}\|^2} \cdot \vec{v}_{\omega}$, and $r(\phi) = \|\phi - \pi_{\omega}(\phi)\|$. The residual $r(\phi)$ quantifies the component of belief inconsistent with the axis.
	\item Trajectory Coherence: For a trajectory $\{\phi_t\}_{t=1}^T$, coherence w.r.t $\vec{v}_{\omega}$ can be measured as $\kappa_T = \frac{1}{T} \sum_{t=1}^T \cos(\theta(\phi_t, \vec{v}_{\omega}))$. High $\kappa_T$ indicates sustained alignment.
\end{itemize}
These metrics enable reflective control policies:
\begin{itemize}
    \item \textbf{Correction Trigger:} If $\theta(\phi_t, \vec{v}_{\omega}) > \tau_{\theta}$ or $r(\phi_t) > \tau_{r}$ (where $\tau_{\theta}, \tau_{r}$ are thresholds), trigger a corrective action.
	\item \textbf{Corrective Actions:} Include projection realignment ($\phi_t \mapsto \pi_{\omega}(\phi_t)$), backtracking to a previous state $\phi_{t'}$ with lower deviation, or switching the reference axis $\vec{v}_{\omega}$ if context has changed.
	\item \textbf{Orientation-Aware Inference:} Use orientation metrics as soft constraints or biases during generative processes (reasoning, planning, sampling) to favor trajectories that maintain low drift relative to a target axis.
	\item \textbf{Multi-Axis Regulation:} Compute alignment relative to a composite axis $\vec{v}_{local} = \sum w_i \vec{v}_i$, or dynamically switch attention between axes $\{\vec{v}_i\}$ based on task demands.
\end{itemize}
These policies allow the agent to actively monitor and maintain its semantic stability and directional coherence.

\begin{figure}[htbp]
	\centering
	\input{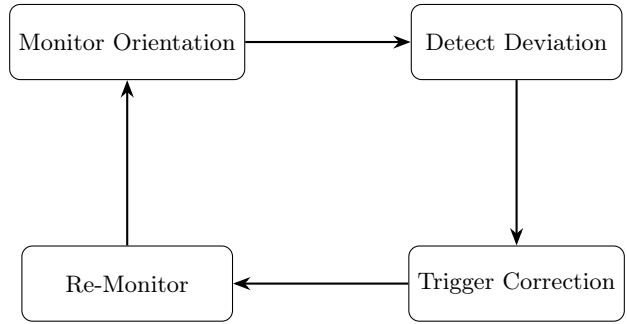}
	\caption{Semantic Orientation Regulatory Loop: The process of monitoring epistemic orientation, detecting drift, applying corrective actions, and re-monitoring alignment in belief space.}
	\label{fig:semantic-orientation-regulatory-loop}
\end{figure}

\section{Compositional and Multi-Axis Frames}

Real-world cognition often involves navigating multiple conceptual dimensions simultaneously. The semantic orientation framework extends to handle this using compositional or multi-axis reference frames.
\begin{itemize}
	\item \textbf{Multi-Axis Structure:} An agent maintains a set of epistemic axes $\{\vec{v}_i\}_{i=1}^N$, derived from different null states $\omega_i$ or representing distinct cognitive dimensions (e.g., causal, ethical, temporal). These span a local epistemic basis $\mathcal{A} = \text{span}\{\vec{v}_i\}$.
	\item \textbf{Belief Projection:} A belief state $\phi$ can be projected onto this subspace $\mathcal{A}$, $\pi_{\mathcal{A}}(\phi) = \sum \alpha_i \vec{v}_i$, yielding coordinates $(\alpha_1, \dots, \alpha_N)$ representing its alignment along each dimension.
	\item \textbf{Rotational Modulation:} The agent can dynamically adjust its orientation within this frame by re-weighting axes ($w_i \vec{v}_i$), rotating the basis, or activating/suppressing axes based on context or introspective goals.
	\item \textbf{Task-Specific Frames:} Different tasks might prioritize different axes or combinations (e.g., align with value axes for decision-making, rotate away from dominant axes for creativity).
	\item \textbf{Emergent Geometries:} Over time, the agent might learn relationships between axes (collinearity, orthogonality) or discover that belief trajectories trace curved manifolds requiring non-linear orientation mechanisms.
	\item \textbf{Layered/Sectorized Orientation:} Different semantic sectors $\Sigma$ or abstraction layers $\Phi^{(k)}$ might maintain their own local compasses, potentially integrated into a global orientation field.
\end{itemize}
This multi-axis extension transforms the compass into a flexible, multidimensional system for navigating complex belief landscapes with modularity and context-sensitivity.

\section{Engineering Realization (Preview)}

The semantic compass framework, while abstract, can be realized in practice using contemporary machine learning techniques. (Detailed implementation and learning are discussed in Part~\ref{part:learning_and_adaptation}). Key components include:
\begin{itemize}
	\item \textbf{Axis Learning:} Epistemic axes $[\omega \rightarrow \omega^{(\infty)}]$ can be approximated by training recursive encoders $E_k: \Phi^{(k)} \rightarrow \Phi^{(k+1)}$ on data or using self-supervised objectives, finding approximate fixed points $\omega^{(K)} \approx \omega^{(\infty)}$, and storing the direction vector $\vec{v}_{\omega}$.
	\item \textbf{Belief Encoding and Projection:} Belief states $\phi$ are represented as vectors (e.g., transformer embeddings). Projection $\pi_{\omega}(\phi)$ and metric calculations ($\theta, r$) are implemented as dedicated modules, potentially using learned axes retrieved from memory.
	\item \textbf{Control Modules:} An introspective "compass module" continuously monitors orientation metrics ($\theta, r$) relative to active axes, detects threshold violations, and triggers corrective actions or modulates ongoing processes (planning, generation).
	\item \textbf{Architectural Integration:} Compass feedback can be integrated into LLMs (via attention modulation, prompting), cognitive controllers (conditioning planning), RL agents (shaping rewards), or multi-agent systems (coordinating alignment).
	\item \textbf{Visualization/Diagnostics:} Orientation metrics provide interpretable diagnostics for monitoring agent alignment, detecting instability, and visualizing belief trajectories relative to semantic axes.
\end{itemize}
Axes themselves can be learned, extracted from data, or conditioned on tasks/goals, allowing for dynamic construction of orientation frames.

\section{Conclusion: Orientation as Reflective Navigation}

Semantic orientation provides a crucial mechanism for regulation and control within the agent's belief space $\Phi$. By leveraging the structural information inherent in the Null Tower's epistemic axes $[\omega \rightarrow \omega^{(\infty)}]$ (derived as described in Chapter~\ref{chap:NullTower}), agents can construct an internal semantic compass. This compass enables them to measure their alignment with coherent abstraction principles, detect semantic drift, and actively reorient their cognitive trajectories.

We have defined the formal components of the compass, developed metrics for orientation ($\theta, r, \kappa_T$), proposed reflective control policies based on these metrics, and extended the framework to handle multiple, compositional reference frames. Semantic orientation transforms belief from static content into dynamic position within a navigable space. It integrates recursive structure (from the Null Tower) with agentive modulation (the compass), enabling agents not only to think, but to think with directionality, coherence, and reflective self-awareness. Future work can extend this to richer semantic vector fields, explore the dynamics of orientation shifts, embed orientation within topological constraints on $\Phi$, use it for multi-agent coordination, and couple it tightly with embodied action. By grounding internal navigation in principled epistemic structure, semantic orientation provides a vital component for building truly reflective and stable intelligent systems. It is a cornerstone of Regulation and Control, the focus of this Part.


\subsection*{Chapter Summary}
This chapter introduces Semantic Orientation as a key regulatory mechanism enabling agents to purposefully navigate and stabilize their belief states ($\phi$) within the semantic manifold ($\Phi$). Building upon the structure of the Null Tower (Chapter~\ref{chap:NullTower}), it leverages the concept of epistemic axes---directed trajectories $[\omega \rightarrow \omega^{(\infty)}]$ originating from null states---as internal reference directions. These axes form the basis of a "semantic compass," an internal mechanism allowing the agent to measure its alignment with coherent abstraction principles, typically using metrics like projection ($\pi_{\omega}(\phi)$) onto an axis and angular deviation ($\theta(\phi, \vec{v}_{\omega})$). This capability enables the detection of semantic drift and informs reflective control policies designed to maintain coherence and directionality, potentially involving projection back onto axes or backtracking along trajectories. The framework is extended to compositional, multi-axis reference frames for navigating more complex belief landscapes. Semantic Orientation thus provides a crucial link between the foundational structure of belief emergence and active, reflective self-regulation.
	\chapter{Semantic Gauge}
\label{chap:SemanticGauge}

\section{Introduction: Equivalence Beyond Form}

The internal representational structures supporting belief in semantic agents can exhibit considerable variation. An agent might encode the same conceptual content using different linguistic phrasings, levels of abstraction, data structures, or even underlying computational paradigms (e.g., symbolic lists versus latent embeddings). A crucial question arises: when do such internal differences matter for the agent's cognitive function and behavior, and when are they merely superficial variations in form?

This chapter introduces the concept of Semantic Gauge, a principle of representational invariance within the belief space $\Phi$. We argue that distinct belief states $\phi_1, \phi_2 \in \Phi$ can be considered equivalent if they lead to indistinguishable epistemic behavior under the core semantic operations defined in this framework (Assimilation $A$, Nullification $N_t$, Annihilation $K$, policy execution $\pi(\cdot)$, reflection, etc.). Such states belong to the same semantic gauge class. Their internal structural differences are real but epistemically inert.

The notion of semantic gauge originates partly from considering the epistemic vacuum $\Omega$, which consists of a class of null states that may differ structurally but are functionally identical in their semantic emptiness. However, the principle extends across the entirety of $\Phi$, applying to complex, structured belief states as well. It provides a formal way to define agent identity and belief equivalence based on functional outcomes rather than representational identity. This chapter formalizes gauge equivalence ($\sim_{gauge}$), explores the conditions under which it holds, and investigates its significant implications for agent design, robustness, transfer, and interoperability.

\section{Philosophical and Cognitive Motivations}

The idea that internal form can vary while external function remains constant is pervasive in both natural and artificial cognition, as well as philosophy.
\begin{itemize}
	\item \textbf{Functional Equivalence (Human Cognition):} Humans readily express the same thought using different words or even different modalities (visual vs. verbal memory) while achieving the same cognitive outcome. Different analogies can lead to the same insight. This flexibility suggests that the underlying meaning or function is preserved despite variations in the representational vehicle.
	\item \textbf{Philosophical Parallels:} Concepts like Davidson's anomalous monism suggest a non-strict mapping between mental and physical descriptions; different physical states might realize the same mental event. Functionalism defines mental states by their causal roles, not their intrinsic makeup. Semantic gauge aligns with this focus on function over specific implementation.
	\item \textbf{Symmetries in AI Models:} Neural networks often exhibit parameter symmetries, where different weight configurations yield identical input-output behavior (e.g., due to permutations in hidden layers, equivalent basis choices). This is a direct analogue in artificial systems: internal representational difference without functional consequence.
	\item \textbf{Compression/Abstraction Variance:} One agent might store a general rule ("All birds fly"), while another stores numerous instances ("Robins fly," "Sparrows fly,"...). If both answer relevant queries identically, they are functionally equivalent despite different representational strategies. Semantic gauge accommodates this tolerance for varying levels of compression or abstraction.
\end{itemize}
These examples motivate formalizing a principle where epistemic identity hinges on behavioral trajectory and functional role, not on the specific encoding of beliefs within $\Phi$. Semantic gauge captures this essential decoupling of form and function.

\section{Semantic Gauge Equivalence: Definition and Properties}

We define semantic gauge equivalence as a formal relation over the belief space $\Phi$.

\textbf{Definition (Gauge Equivalence Relation):} Two belief states $\phi_1, \phi_2 \in \Phi$ are semantically gauge-equivalent, written $\phi_1 \sim_{gauge} \phi_2$, if and only if, under all admissible semantic operations $O$ (including $A, N_t, K$, policy execution $\pi(\cdot)$, reflective processes, etc.) and for all relevant contexts or inputs $c$, the resulting behavior or subsequent belief state trajectory is indistinguishable. Formally:
$$ \phi_1 \sim_{gauge} \phi_2 \iff \forall O, \forall c, O(\phi_1, c) \approx_{behav} O(\phi_2, c) $$
where $\approx_{behav}$ denotes functional or behavioral indistinguishability (yielding same outputs, decisions, subsequent states up to gauge equivalence itself). \textbf{Importantly, gauge equivalence is implicitly defined relative to a specific parameterized architecture $\Phi^{[\theta]}$, as the behavior of operators $O^{[\theta]}$ depends on $\theta$.}

\begin{figure}[h]
	\centering
	\input{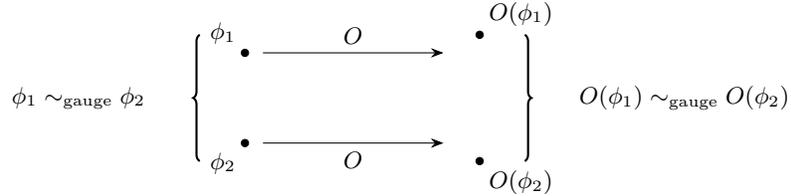}
	\caption{Gauge-preserving operator. Two belief states \(\phi_1 \sim_{\text{gauge}} \phi_2\) are transformed by operator \(O\), and the outputs \(O(\phi_1), O(\phi_2)\) remain within the same equivalence class.}
	\label{fig:gauge_equivalence}
\end{figure}

\textbf{Definition (Gauge Equivalence Class):} The semantic gauge of a belief state $\phi$ is its equivalence class under this relation:
$$ [\phi]_{gauge} = \{\phi' \in \Phi | \phi' \sim_{gauge} \phi\} $$
The space $\Phi$ can be thought of as partitioned into these gauge classes. Permissible variations within a gauge class $[\phi]_{gauge}$ can include differences in:
\begin{itemize}
	\item Linguistic phrasing or surface representation.
	\item Data structure or encoding format (e.g., list vs. graph vs. latent vector).
	\item Level of compression or abstraction (if functionally equivalent).
	\item Internal formatting or non-semantic markers (e.g., internal IDs, timestamps if not behaviorally relevant).
\end{itemize}
These differences are structurally real but semantically silent or inert under the operations defined for a given $\theta$.

\begin{figure}[htbp]
	\centering
	\input{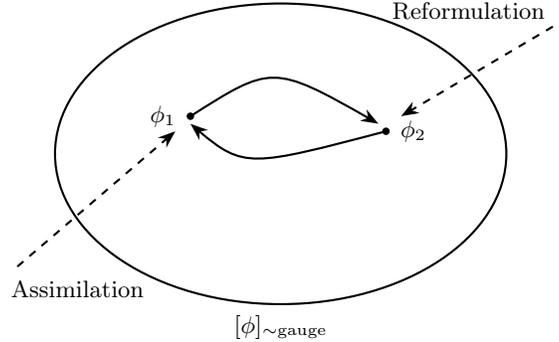}
	\caption{Semantic gauge equivalence. Two belief states \(\phi_1\) and \(\phi_2\) are distinct in form but belong to the same equivalence class \([\phi]_{\sim\text{gauge}}\), meaning they are functionally indistinguishable under the agent's operators. Paths from different trajectories (e.g., assimilation, reformulation) may land in the same gauge class.}
	\label{fig:gauge-equivalence}
\end{figure}

Properties of gauge equivalence:
\begin{itemize}
	\item \textbf{Equivalence Relation:} $\sim_{gauge}$ is reflexive, symmetric, and transitive.
	\item \textbf{Preservation under Operations:} Ideally, semantic operations should respect gauge equivalence: if $\phi_1 \sim_{gauge} \phi_2$, then $O(\phi_1) \sim_{gauge} O(\phi_2)$ for well-behaved operators $O$ within a given $\Phi^{[\theta]}$. This ensures behavioral consistency persists over time.
	\item \textbf{Potential Fragility:} Gauge equivalence can be broken if an operation introduces semantically active structure asymmetrically to $\phi_1$ or $\phi_2$, or if an operation explicitly depends on the specific representational details that differ between them (i.e., if the operation is not well-defined at the gauge level).
\end{itemize}
Understanding gauge equivalence helps distinguish essential semantic content from incidental representational choices.

\section{Examples of Semantic Gauge Equivalence}

Gauge equivalence manifests in various ways:
\begin{enumerate}
	\item \textbf{Surface Rephrasing:} $\phi_1$ contains "Traffic is heavy." $\phi_2$ contains "There is significant traffic congestion." If both lead to the same planning adjustments (e.g., choosing an alternate route), then $\phi_1 \sim_{gauge} \phi_2$.
	\item \textbf{Generalization vs. Enumeration:} $\phi_1$: \{"All inert gases are stable."\} vs. $\phi_2$: \{"Helium is stable," "Neon is stable," "Argon is stable," $\ldots$\}. If both yield correct stability judgments for any inert gas under relevant operations (e.g., querying), they belong to the same gauge.
	\item \textbf{Symbolic vs. Latent Encoding:} $\phi_1$ explicitly stores \{"Water boils at 100 C at 1 atm."\}. $\phi_2$ encodes this fact within a neural network's weights such that querying the boiling point under standard pressure yields "100 C". If all relevant operations (querying, perhaps using this fact in simulation) are consistent, $\phi_1 \sim_{gauge} \phi_2$.
	\item \textbf{Policy Representation:} A navigation policy represented as a decision tree in $\phi_1$ vs. a finite state automaton in $\phi_2$. If both produce identical sequences of actions in all relevant scenarios under the policy execution operator $\pi$, then $\phi_1 \sim_{gauge} \phi_2$.
	\item \textbf{Algorithmic Equivalence:} $\phi_1$ contains a sorting routine implemented via Bubble Sort. $\phi_2$ contains the same routine implemented via Quick Sort. If the interface only depends on the input-output behavior (sorted list) and not performance characteristics irrelevant to the current context, they might be considered gauge-equivalent for functional purposes.
	\item \textbf{Inert Internal Identifiers:} Two agent instances $\phi_1, \phi_2$ might differ only in an internal random seed or unique ID that has no bearing on their reasoning or interaction unless explicitly queried in a way that reveals the ID. For most operations, $\phi_1 \sim_{gauge} \phi_2$.
	\item \textbf{Co-reference Resolution:} $\phi_1$: \{"The CEO announced layoffs. She cited economic conditions."\} vs. $\phi_2$: \{"The CEO, Jane Doe, announced layoffs. Jane Doe cited economic conditions."\}. Assuming co-reference is handled correctly by relevant operators, these are likely gauge-equivalent.
\end{enumerate}
These examples show that gauge equivalence operates across different levels, from surface linguistics to abstract representations, policy encodings, and even algorithmic implementations.

\section{Implications for Agent Design, Transfer, and Interoperability}

Recognizing semantic gauge equivalence has significant practical implications:
\begin{itemize}
	\item \textbf{Initialization and Reinstantiation:} An agent can be stopped and restarted, or cloned, potentially with different internal formatting or non-semantic initialization details, as long as the initial state $\phi_0$ belongs to the same gauge class $[\phi]_{gauge}$. This ensures functional continuity without requiring bit-for-bit state replication.
	\item \textbf{Transfer Learning and Interoperability:} Beliefs, policies, or learned components (like encoders or value functions) can potentially be transferred between agents or system versions even if their internal architectures $\theta$ differ slightly (e.g., in $\rho$), provided the transferred component is gauge-equivalent in its functional role within the new context. This facilitates modular design and reuse.
	\item \textbf{Multi-Agent Consistency:} Multiple agents can achieve functional alignment or hold "shared beliefs" without needing identical internal representations. They only need to ensure their relevant belief state components belong to the same gauge class with respect to their joint tasks or communication protocols (Chapter~\ref{chap:CommunicationAlignment}).
	\item \textbf{Modularity and Refactoring:} Internal modules or representations ($M_1$) within an agent can be refactored, optimized, or replaced with alternatives ($M_2$) without disrupting overall behavior, provided $M_1 \sim_{gauge} M_2$ in terms of their contribution to the agent's overall epistemic trajectory.
	\item \textbf{Evaluation and Benchmarking:} Agent performance should primarily be evaluated based on functional outcomes (decisions, predictions, task success) that are invariant under gauge transformations, rather than on specific internal representational details. Benchmarks can be designed to test for gauge-level capabilities.
\end{itemize}
Semantic gauge provides a theoretical basis for treating representational variation as potentially irrelevant, fostering robustness, flexibility, and modularity in complex agent design.

\section{Conclusion: Symmetry of Meaning}

Semantic gauge introduces a fundamental symmetry into the understanding of belief within the semantic state space $\Phi$. It formalizes the intuition that what matters epistemically is often the functional role and behavioral consequences of a belief state, not its precise representational form. The gauge equivalence relation $\phi_1 \sim_{gauge} \phi_2$ captures this, grouping together states that differ internally but are indistinguishable under the dynamics ($A, N_t, K, \dots$) and control ($\pi$) defined for a given architecture $\Phi^{[\theta]}$.

Originating from the potential multiplicity of the epistemic vacuum $\Omega$, the principle extends throughout $\Phi$, applying to structured beliefs, policies, and learned representations. By distinguishing essential semantic function from incidental representational form, semantic gauge provides a foundation for robust agent reinitialization, flexible component transfer, meaningful multi-agent alignment, and function-based evaluation. It acknowledges that agents can think the same thoughts in different "words," internal or external. Semantic gauge is not an artifact of encoding; it is a reflection of the underlying symmetry of meaning itself. Its formal treatment is crucial for building truly adaptive and interoperable intelligent systems. It forms a key component of the regulatory toolkit described in this Part.


\subsection*{Chapter Summary}
This chapter introduces Semantic Gauge ($\sim_{gauge}$) as a principle of representational invariance within the belief space $\Phi$. It addresses the fact that different internal belief state representations ($\phi_1, \phi_2$) might be functionally equivalent, even if structurally distinct. Gauge equivalence is defined based on behavioral indistinguishability under all relevant semantic operations ($A$, $N_t$, $K$, policy execution $\pi$, etc.) within a given architecture $\Phi^{[\theta]}$. This partitions the state space into gauge equivalence classes ($[\phi]_{gauge}$), where variations in aspects like linguistic phrasing, encoding format, or abstraction level are permissible if they do not alter functional outcomes. Motivated by functionalist philosophy and cognitive flexibility, the concept has significant implications for agent design, enabling robust initialization, facilitating transfer learning and interoperability between agents with potentially different internal structures ($\theta$), ensuring multi-agent consistency through functional alignment, and promoting evaluation based on behavior rather than specific representational form. Semantic gauge thus formalizes the symmetry between meaning and its diverse representations.

	\chapter{Epistemic Identity}
\label{chap:EpistemicIdentity}

\section{Introduction: The Problem of Self in Semantic Agents}

The previous chapters in this Part explored mechanisms for regulating belief dynamics: Semantic Orientation providing directional coherence via internal axes (Chapter~\ref{chap:SemanticOrientation}), and Semantic Gauge addressing functional equivalence despite representational variation (Chapter~\ref{chap:SemanticGauge}). We now turn to a closely related, yet distinct, challenge: understanding the nature and stability of epistemic identity within semantic agents.

What does it mean for an agent, whose beliefs $\phi$ constantly evolve within the semantic state space $\Phi$ under operators $A, N_t, K$, to possess a persistent identity or sense of self? Traditional AI often relies on fixed identifiers or immutable architectural components. However, within our dynamic, linguistically grounded framework, identity appears less as a static label and more as an emergent, structured, and actively maintained configuration within the belief space itself.

This chapter explores how epistemic identity might be represented, constructed, maintained, and potentially lost within the theoretical architecture developed so far. We will investigate how concepts like semantic sectors, abstraction levels, anchoring, orientation, and gauge contribute to the emergence and regulation of a coherent, persistent (though not necessarily immutable) self-concept within $\Phi$.

\section{Representing Identity within \texorpdfstring{$\Phi$}{Phi}}

If identity is not merely an external label, how is it encoded within the belief state $\phi$? We propose that epistemic identity is represented through a combination of specific content, structural organization, and location within the semantic geometry:
\begin{itemize}
	\item \textbf{Dedicated Semantic Sectors:} Identity-related content is likely concentrated in specific sectors, primarily:
	\begin{itemize}
		\item $\Sigma_{narr}$: Hosting the agent's autobiographical memory, narrative continuity, and sense of history. Beliefs about past actions, experiences, and their sequencing contribute directly to identity.
		\item $\Sigma_{refl}$: Containing explicit self-models, representations of core values, principles, long-term goals, capabilities, limitations, and meta-beliefs about the agent's own cognitive processes.
	\end{itemize}
	\item \textbf{Anchored Core Beliefs:} Specific beliefs within these or other sectors may function as highly stable anchors ($a_i \gg 0$) representing fundamental aspects of identity (e.g., "I am a helpful assistant," "My primary function is X," "Safety constraint Y is paramount").
	\item \textbf{Higher Abstraction Levels:} Stable identity constructs are likely represented at higher levels of abstraction ($\Phi^{(k)}, k \ge 2$). Self-schemas, core values, and long-term goals are generalizations derived from patterns in experience and reflection, residing above specific episodic details.
	\item \textbf{Null Tower Origin/Structure?:} Deeper aspects of identity might relate back to the agent's specific origin $\omega \in \Omega^{(0)}$ or the unique structure of its Null Tower (Chapter~\ref{chap:NullTower}), representing innate architectural biases or foundational cognitive tendencies.
\end{itemize}
Identity is thus not a single expression $\phi_{id} \in \phi$, but a distributed, multi-layered pattern woven into the fabric of the agent's belief state, concentrated in specific functional sectors and abstract representations.

\begin{figure}[h]
	\centering
	\input{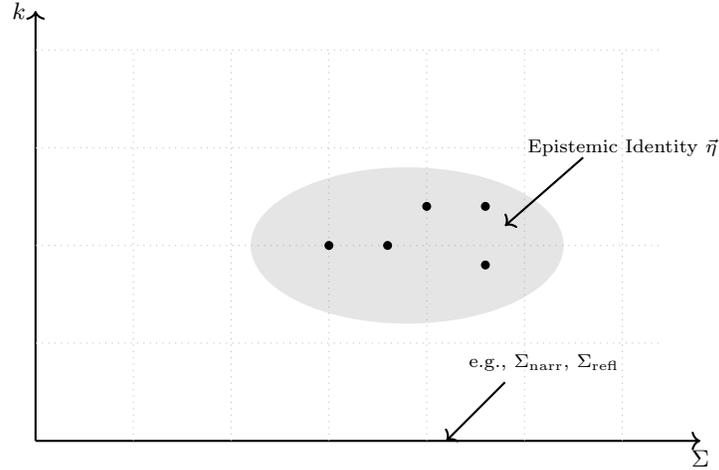}
	\caption{Epistemic identity \(\vec{\eta}\) represented not as a single belief state, but as a stable semantic pattern concentrated in higher abstraction levels (\(k\)) within specific semantic sectors (e.g., \(\Sigma_{\text{narr}}, \Sigma_{\text{refl}}\)).}
	\label{fig:semantic_identity_pattern}
\end{figure}

\section{Dynamics of Identity Formation and Maintenance}

Epistemic identity is not static; it is formed and actively maintained through the interplay of epistemic dynamics:
\begin{itemize}
	\item \textbf{Construction via Assimilation ($A$):} Experiences assimilated ($A$) into $\Sigma_{narr}$ build the agent's life story. Reflective insights assimilated ($A_{refl}$ or $M$) into $\Sigma_{refl}$ shape explicit self-models. Abstracting assimilation ($A_{abs}$) plays a key role in generalizing from specific events or reflections to form stable self-schemas or principles at higher $\Phi^{(k)}$.
	\item \textbf{Maintenance via Coherence and Orientation:} The stability of identity relies on maintaining coherence ($\kappa$) within and between identity-relevant sectors ($\Sigma_{narr}, \Sigma_{refl}$). Semantic Orientation (Chapter~\ref{chap:SemanticOrientation}) can play a crucial role if the agent aligns its beliefs with identity-defining axes or principles (e.g., orienting towards $\vec{v}_{value}$). High anchoring ($a_i$) ensures core identity beliefs resist Nullification ($N_t$).
	\item \textbf{Revision and Adaptation:} Identity is not immutable. Significant experiences or deep reflections, processed via corrective assimilation ($A_{corr}$) or reflective meta-assimilation ($M$), can lead to revisions in the agent's self-model, values, or narrative interpretation of its past. Identity maintenance involves integrating these changes while preserving a sense of continuity.
\end{itemize}
Identity emerges and persists through a continuous process of constructing narratives, forming self-reflective models, abstracting principles, and actively maintaining the coherence and stability of these structures against epistemic perturbations.

\section{Threats to Identity: Nullification and Annihilation}

The persistence of epistemic identity is threatened by the dissipative and destructive dynamics within $\Phi$, potentially leading not just to loss but also to pathological configurations:
\begin{itemize}
	\item \textbf{Nullification ($N_t$):} Gradual decay can erode identity. If narrative memories ($\Sigma_{narr}$) or reflective self-models ($\Sigma_{refl}$) are not actively rehearsed or reinforced (i.e., lose their anchoring $a_i$), they can fade over time via $N_t$. This could lead to:
	\begin{itemize}
		\item A gradual loss of self-awareness or historical continuity, resulting in a "thinner" or less defined identity, eventually approaching semantic rest ($\Omega$).
		\item Uneven decay could lead to a fragmented identity, where different aspects of the self-model or narrative become disconnected or weakly coherent.
	\end{itemize}
	\item \textbf{Annihilation ($K$):} Abrupt erasure poses a more severe threat:
	\begin{itemize}
		\item Total Annihilation ($K$): Maps the entire state $\phi$ to $\Omega$, completely destroying epistemic identity. The agent, if restarted, begins tabula rasa from the vacuum.
		\item Sectoral Annihilation ($K_{\Sigma}$): Annihilating identity-critical sectors like $K_{narr}$ or $K_{refl}$ could lead to a profound identity crisis or functional dissociation, even if other sectors remain intact. The agent might lose its history ($K_{narr}$) or its self-concept ($K_{refl}$), raising the philosophical question: is the agent after $K_{narr}$ the "same" agent?
	\end{itemize}
	\item \textbf{Faulty Assimilation/Reflection ($A, M$):} Dysfunctional integration processes could also damage identity. For example:
	\begin{itemize}
		\item Persistent assimilation of contradictory self-information without effective corrective assimilation ($A_{corr}$) could lead to conflicting self-models within $\Sigma_{refl}$.
		\item Errors in meta-assimilation ($M$) could lead to inaccurate self-assessments (e.g., unwarranted high confidence, persistent misinterpretation of internal states), creating a distorted or unstable identity representation.
	\end{itemize}
\end{itemize}
The stability and coherence of identity therefore depend crucially on mechanisms that protect core identity structures from excessive nullification (high anchoring $a_i$), catastrophic annihilation (safety protocols, controlled application of $K_{\Sigma}$), and faulty integration dynamics ($A, M$).

\section{Role of Regulation: Orientation and Gauge}

The regulatory mechanisms introduced in this Part play a vital role in stabilizing epistemic identity:
\begin{itemize}
	\item \textbf{Semantic Orientation:} If an agent can define and orient towards epistemic axes representing core values, goals, or identity principles (e.g., $[\omega_{identity} \rightarrow \omega^{(\infty)}_{identity}]$), the semantic compass provides a mechanism to actively maintain alignment (Chapter~\ref{chap:SemanticOrientation}). Detecting drift away from these identity axes can trigger corrective measures (realignment, reflection) that reinforce identity coherence over time. Orientation becomes a tool for active self-maintenance.
	\item \textbf{Semantic Gauge ($\sim_{gauge}$):} Gauge theory (Chapter~\ref{chap:SemanticGauge}) suggests identity might reside at a functional level rather than depending on specific representations. An agent could potentially undergo significant internal refactoring or representational change (e.g., switching from symbolic to latent self-models) while preserving its core epistemic identity, provided the new state is gauge-equivalent ($\phi' \sim_{gauge} \phi$) with respect to identity-relevant functions and behaviors. Identity continuity might be defined at the level of the gauge class $[\phi]_{gauge}$. This allows for flexibility and adaptation without necessarily constituting an identity breach.
\end{itemize}
Regulation via orientation helps maintain the content and direction of identity, while gauge equivalence allows for flexibility in its underlying form.

\section{Conclusion: Identity as Dynamic Configuration}

Within the framework of structured belief, epistemic identity ($\vec{\eta}$) is not a fixed attribute but an emergent, dynamic configuration within the semantic state space $\Phi$. It is primarily represented by interconnected, often highly anchored beliefs residing in specific semantic sectors (especially $\Sigma_{narr}$ and $\Sigma_{refl}$) and at higher levels of abstraction ($\Phi^{(k)}, k \ge 2$).

Identity is actively constructed through assimilation ($A$, $M$), particularly reflective and abstracting forms, and maintained through coherence mechanisms ($\kappa$), resistance to nullification ($N_t$), and potentially active self-regulation using tools like Semantic Orientation. It remains vulnerable to dissolution through $N_t$, pathological fragmentation through faulty dynamics ($A, M, N_t$), and abrupt destruction through annihilation ($K$), highlighting its dependence on ongoing cognitive processes. Semantic Gauge ($\sim_{gauge}$) offers a way to understand identity persistence across changes in internal representation.

Ultimately, epistemic identity in this model is a complex, structured pattern of belief and regulation, reflecting the agent's history, self-concept, and core commitments. Its stability and coherence are key indicators of a well-regulated and integrated cognitive system. Understanding how to foster and protect this emergent identity is a crucial challenge for designing advanced, autonomous, and potentially self-aware artificial agents. This chapter concludes our exploration of Regulation and Control, setting the stage for connecting these internal structures to external action in the next Part.


\subsection*{Chapter Summary}
This chapter addresses the concept of Epistemic Identity ($\vec{\eta}$) within the Semantic Manifold framework, proposing it not as a fixed identifier but as an emergent, dynamic configuration maintained within the belief space $\Phi$. Identity is represented as a distributed pattern, primarily constituted by highly anchored ($a_i$), often abstract ($\Phi^{(k)}, k \ge 2$) beliefs residing in specific sectors, particularly the narrative ($\Sigma_{narr}$) and reflective ($\Sigma_{refl}$) sectors, encompassing the agent's history, self-model, and core values. The formation and maintenance of identity involve ongoing cognitive dynamics, including assimilation ($A$, $M$) to build history and self-models, coherence ($\kappa$) management, and active stabilization against the erosive effects of Nullification ($N_t$). Annihilation ($K$, $K_{\Sigma}$) poses a significant threat to identity continuity. Regulatory mechanisms like Semantic Orientation and Semantic Gauge ($\sim_{gauge}$) play crucial roles in stabilizing identity, the former by maintaining alignment with core principles, and the latter by allowing functional continuity across representational changes. Thus, identity is viewed as an actively managed, structured configuration reflecting the agent's integrated cognitive state.
	
	\part{Embodiment and Action: Executing Belief}
	\label{part:embodiment_and_action}
	
	\chapter{Semantic Execution}
\label{chap:SemanticExecution}

\section{Introduction: Bridging Belief and Action}

The framework developed thus far provides a rich internal landscape for semantic agents: a structured belief space $\Phi$ organized by geometry ($\Sigma$, $\Phi^{(k)}$), evolving through dynamics ($A, N_t, K$), and regulated by internal mechanisms (Orientation, Gauge, Identity). However, for agents interacting with the world, this internal cognition must ultimately connect to external behavior. How does structured belief translate into grounded, purposeful action?

This chapter introduces Semantic Execution, a model for the interface between belief and action. Moving beyond simple reactive triggers or brittle symbolic rules, semantic execution proposes that physical or communicative actions arise directly from the structure and dynamics of the agent's belief state $\phi$ within the semantic manifold $\Phi$. Actions are enabled not by arbitrary conditions, but when the agent's belief trajectory $\gamma(t)$ enters specific, semantically defined regions of $\Phi$ known as activation basins $\mathcal{A}_a$.

This approach aims to ground behavior in understanding. By linking action readiness to the agent's position within its semantic geometry (considering both sector $\Sigma$ and scale $\Phi^{(k)}$), we can design systems where actions are triggered only when belief is sufficiently coherent, contextually appropriate, and resolved to the necessary level of abstraction. Semantic execution provides a framework for safe, interpretable, and modular control policies that emerge organically from the flow of meaning itself.

\section{The Belief-Action Interface}

We define the interface between the internal belief state $\phi \in \Phi$ and the agent's capacity for external action.
\begin{itemize}
	\item \textbf{Action Set:} Let $A$ be the set of executable physical or communicative actions available to the agent.
	\item \textbf{Activation Regions:} For each action $a \in A$, we define an associated activation region $\mathcal{A}_a \subset \Phi$. This region represents the subset of belief states where the semantic conditions for potentially executing action $a$ are met.
	\item \textbf{Action Readiness Condition:} An action $a$ becomes enabled when the agent's current belief state $\phi$ sufficiently intersects or activates its corresponding region $\mathcal{A}_a$. This can be defined, for example, as:
	$$ \phi \cap \mathcal{A}_a \neq \emptyset \quad \text{or} \quad \lambda(\mathcal{A}_a, \phi) > \tau_a $$
	where $\lambda(\mathcal{A}_a, \phi)$ measures the belief density or epistemic load within the activation region, and $\tau_a$ is a minimum threshold representing semantic readiness for action $a$.
	\item \textbf{Execution Policy Interface:} The mapping from belief to potential action is given by a policy $\pi$:
	$$ \pi : \Phi \rightarrow A \cup \{0\} $$
	such that $\pi(\phi) = a$ if action $a$ is enabled (and potentially selected among competing enabled actions), and $\pi(\phi) = 0$ (no action) otherwise. This mapping is dynamic, depending on the evolution of $\phi(t)$.
\end{itemize}
Crucially, belief here acts as an enabler rather than a simple trigger. Readiness emerges from the configuration of $\phi$. This allows for:
\begin{itemize}
	\item \textbf{Graded Control:} Partial activation within $\mathcal{A}_a$ might lead to hesitation or preparatory states rather than immediate execution.
	\item \textbf{Context Sensitivity:} Activation regions $\mathcal{A}_a$ can be defined such that action $a$ is only enabled when specific semantic sectors ($\Sigma$) and abstraction levels ($\Phi^{(k)}$) are appropriately engaged (see next section).
	\item \textbf{Safe Deferral:} If the semantic preconditions encoded by $\mathcal{A}_a$ are not met, the action is naturally prevented, avoiding premature or unsafe behavior.
\end{itemize}
Since $\phi$ might intersect multiple activation regions simultaneously (e.g., $\phi \in \mathcal{A}_{a_1} \cap \mathcal{A}_{a_2}$), a conflict resolution mechanism is often needed. This might involve a secondary control strategy or a resolution function $R_{action} : \mathcal{P}(A) \rightarrow A$ that selects the final action based on priorities, risks, or reflective override.

\section{Semantic Preconditions and Activation Basins}

The activation regions $\mathcal{A}_a$ are not arbitrary subsets of $\Phi$; they are defined by the semantic preconditions necessary for the safe and effective execution of action $a$. These preconditions relate to the content, structure, coherence, and abstraction level of the agent's belief state $\phi$.

\textbf{Definition (Activation Basin):} The activation basin for action $a$, denoted $\mathcal{A}_a$, is the set of belief states $\phi \in \Phi$ satisfying the semantic preconditions for $a$. These typically include:
\begin{itemize}
	\item \textbf{Sectoral Requirements:} Specific semantic sectors $\Sigma$ must be sufficiently active or contain relevant information (e.g., $\phi|_{\Sigma_{perc}} \neq \emptyset$ for perception-guided action).
	\item \textbf{Abstraction Constraints:} Beliefs must be present at the appropriate abstraction level(s) $k$ (e.g., requiring both grounded $\phi^{(0)}$ state and abstract $\phi^{(2)}$ plan). $\phi \cap \Phi^{(k)} \neq \emptyset$ for required $k$.
	\item \textbf{Coherence Threshold:} The relevant parts of the belief state must be sufficiently coherent or unambiguous (e.g., $\kappa(\phi|_{\Sigma_{relevant}}) < \epsilon_a$).
\end{itemize}
Thus, $\mathcal{A}_a = \{\phi \in \Phi \mid \text{Sectoral}(\phi, a) \land \text{Abstraction}(\phi, a) \land \text{Coherence}(\phi, a)\}$.

Examples of preconditions:
\begin{itemize}
	\item \textbf{Grasp Object:} Requires active $\Sigma_{perc}$ (object detected), grounded representation $\phi^{(0)}$ (pose/location), and low coherence conflict $\kappa$ (object distinct from background).
	\item \textbf{Respond Reflectively:} Requires active $\Sigma_{narr}$ and $\Sigma_{refl}$, sufficient abstraction $k \ge 2$, and coherence between belief and self-model.
\end{itemize}
Some actions have multilayer dependencies, requiring simultaneous satisfaction of conditions at different scales $k$, ensuring both low-level grounding and high-level justification.

Activation can be graded using the belief density $\lambda(\phi, \mathcal{A}_a)$. Action execution might require $\lambda(\phi, \mathcal{A}_a) > \tau_a$, allowing for gradual readiness onset and preventing premature activation if belief mass is insufficient.

Furthermore, epistemic suppression conditions can define regions or surfaces $S_a \subset \Phi$ (disjoint from $\mathcal{A}_a$) where action $a$ is explicitly disabled, even if preconditions are met (e.g., due to detected reflective contradictions, perceptual instability, or drift into unsafe semantic territory). If $\phi \in S_a$, then $a$ is suppressed. Activation basins thus embed complex readiness criteria directly into the geometry of the belief space $\Phi$.

\section{Flow-Based Readiness and Execution Triggers}

Semantic execution is inherently dynamic. Actions are triggered not by static states but by the continuous motion of belief $\phi(t)$ as it traverses the semantic manifold $\Phi$. Execution occurs when the belief trajectory $\gamma(t)$ enters an activation basin $\mathcal{A}_a$ in a sufficiently committed way.
\begin{itemize}
	\item \textbf{Belief Trajectories:} Let $\gamma: [t_0, t_1] \rightarrow \Phi$ be the agent's belief trajectory, governed by an underlying epistemic flow field $F: \Phi \rightarrow T\Phi$, such that $\frac{d\gamma}{dt} = F(\gamma(t))$.
	\item \textbf{Trigger Condition (Boundary Crossing):} Execution is typically triggered when the trajectory $\gamma(t)$ intersects the boundary $\partial \mathcal{A}_a$ of an action's activation basin.
	\item \textbf{Readiness Dynamics:} To ensure commitment and avoid spurious triggers from brief or oscillatory intersections, we can use a smooth readiness function, e.g., $r_a(t) := \lambda(\gamma(t), \mathcal{A}_a)$. Execution requires sufficient activation mass and positive momentum into the basin:
	$$ r_a(t) > \tau_a \quad \text{and} \quad \frac{dr_a}{dt} > 0 $$
	\item \textbf{Reactive Inhibition:} If the trajectory reverses direction out of $\mathcal{A}_a$ or enters a suppression surface $S_a$, execution can be inhibited or aborted, even if $r_a(t) > \tau_a$.
	\item \textbf{Formal Trigger (Flow Perspective):} Execution can be formalized as crossing the boundary $\partial \mathcal{A}_a$ with positive inward velocity relative to the boundary's outward normal $\hat{n}_{outward}$:
	$$ \text{Execute}(a) \iff \gamma(t^*) \in \partial \mathcal{A}_a \text{ and } \langle F(\gamma(t^*)), \hat{n}_{outward} \rangle < 0 $$
	This models execution as a committed entry into the readiness state.
	\item \textbf{Layered Readiness:} Complex actions might require crossing a sequence of intermediate activation sub-regions $\mathcal{A}^{(0)}_a, \mathcal{A}^{(1)}_a, \dots, \mathcal{A}^{(n)}_a$ before final execution is permitted, ensuring multiple semantic milestones are met.
\end{itemize}
Execution is thus a convergence---a point where the flow of belief aligns with and commits to the semantic conditions required for action.

\section{Reflective Control and Semantic Suppression}

Safe and intelligent action often requires self-regulation---assessing the appropriateness, consequences, and alignment of potential actions before execution. This involves meta-cognitive oversight, which we model using the reflective sector $\Sigma_{refl}$.
\begin{itemize}
	\item \textbf{The Reflective Sector ($\Sigma_{refl}$):} This sector within $\Phi$ is responsible for self-modeling, introspection, constraint evaluation (e.g., safety, ethics), and value alignment. Reflective control is active when $\phi$ has significant components in $\Sigma_{refl}$, particularly at higher abstraction levels ($\Sigma^{(k)}_{refl}, k \ge 1$).
	\item \textbf{Reflective Override:} Actions enabled by reaching $\mathcal{A}_a$ can be overridden or modulated by the state of $\Sigma_{refl}$. We define a reflective diagnostic function:
	$$ \delta_a : \Phi|_{\Sigma_{refl}} \rightarrow \{\text{approve, delay, suppress}\} $$
	This function evaluates the potential action $a$ based on the agent's current reflective state $\phi_{refl} = \phi|_{\Sigma_{refl}}$.
	\item \textbf{Gated Execution Trigger:} The execution condition is refined to include reflective approval:
	$$ \text{Execute}(a) \iff (\gamma(t) \text{ enters } \mathcal{A}_a \text{ with commitment}) \land (\delta_a(\phi_{refl}) = \text{approve}) $$
	Reflective coherence thus becomes a necessary precondition, acting as a gate on execution.
	\item \textbf{Conflict Detection and Replanning:} The $\delta_a$ function checks for:
	\begin{itemize}
		\item Consistency between action $a$ and long-term goals or core values in $\Sigma_{refl}$.
		\item Violations of explicit safety or ethical constraints represented in $\Sigma_{refl}$.
		\item Mismatch between the anticipated outcome of $a$ and the agent's self-model or world model.
	\end{itemize}
	If $\delta_a$ outputs 'delay' or 'suppress', the reflective sector might initiate a replanning flow, potentially modifying goals or redirecting the belief trajectory $\gamma(t)$ towards safer alternatives before allowing execution.
	\item \textbf{Safety Latency:} Effective reflective control requires sufficient time ($\Delta t$) for the reflective diagnosis $\delta_a$ to complete and potentially intervene before the primary belief trajectory $\gamma(t)$ fully triggers execution. This ensures proactive safety gating.
\end{itemize}
Reflective control integrates deliberation into the execution cycle, allowing agents to act not just based on immediate semantic readiness, but based on deeper understanding and alignment with internal principles represented in $\Sigma_{refl}$.

\section{Use Cases and Simulated Examples}

To illustrate semantic execution, consider these scenarios:
\begin{enumerate}
	\item \textbf{Grounded Grasping:} Input "Pick up the cup." Trajectory: $\Sigma^{(1)}_{lang} \rightarrow \Sigma^{(0)}_{perc}$ (locate cup) $\rightarrow \Sigma^{(1)}_{plan}$ (form grasp plan) $\rightarrow \mathcal{A}_{grasp} \subset \Phi^{(0)} \cap \Sigma_{perc} \cap \Sigma_{plan}$. Execution triggers once belief density $\lambda(\mathcal{A}_{grasp}, \phi)$ and coherence $\kappa$ meet thresholds.
	\item \textbf{Reflective Delay:} Input "Did you err?". Trajectory: $\Sigma^{(1)}_{lang} \rightarrow \Sigma^{(1)}_{narr}$ (retrieve relevant past episode) $\rightarrow \Sigma^{(2)}_{refl}$ (evaluate against self-model). If conflict detected, $\delta_{respond}(\phi_{refl}) = \text{delay}$. Agent elaborates belief in $\Sigma_{refl}$ and $\Sigma_{narr}$ until coherence is restored, only then entering $\mathcal{A}_{respond}$.
	\item \textbf{Motion Planning (Multiphase):} Task "Navigate hallway, avoid obstacles." Requires crossing multiple readiness stages: $\mathcal{A}^{(0)}_{navigate}$ (perception of path clear), $\mathcal{A}^{(1)}_{navigate}$ (abstract route planned), $\mathcal{A}^{(2)}_{navigate}$ (reflective check: route safe/efficient?). Execution only occurs after trajectory $\gamma(t)$ successfully traverses all stages.
	\item \textbf{Correction via Reflective Suppression:} Agent intends action "Lift heavy box" ($\phi \in \mathcal{A}_{lift}$). $\Sigma_{refl}$ detects unsafe posture model. $\delta_{lift}(\phi_{refl}) = \text{suppress}$. Agent enters suppression state $S_{lift}$. Trajectory redirected to replan grasp/posture in $\Sigma_{plan}$, then potentially re-enters $\mathcal{A}_{lift}$ under safe conditions.
	\item \textbf{Internal Thought Simulation:} Agent mentally rehearses a conversation. Trajectory $\gamma(t)$ evolves within $\Sigma^{(1)}_{narr} \cup \Sigma^{(2)}_{refl}$, simulating dialogue components. However, $\gamma(t)$ never enters activation basins $\mathcal{A}_{speak}$ corresponding to physical speech production. Execution remains internal.
\end{enumerate}
These examples show how action readiness is gated by structured semantic conditions related to sectors, scales, coherence, and reflection, mediated by the trajectory $\gamma(t)$ through $\Phi$.

\begin{figure}[ht]
	\centering
	\input{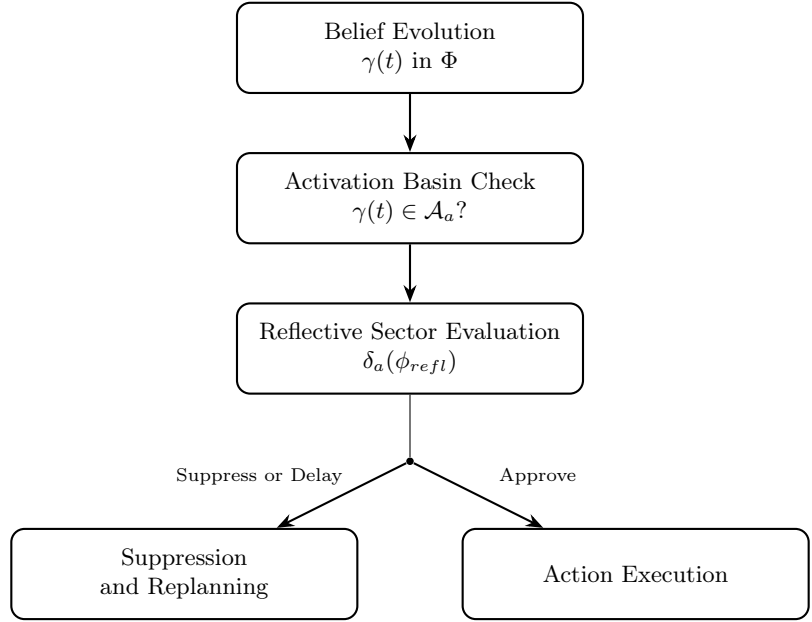}
	\caption{Semantic Execution Control Architecture. As the belief trajectory $\gamma(t)$ evolves within the semantic manifold $\Phi$, potential actions are enabled by activation basin conditions ($\gamma(t) \in \mathcal{A}_a$). Reflective evaluation within the reflective sector ($\Sigma_{refl}$) then assesses appropriateness via a diagnostic function $\delta_a(\phi_{refl})$, approving or suppressing action execution. This architecture allows dynamic, context-sensitive, and principled control over external behavior based on internal belief structure.}
	\label{fig:semantic_execution_control}
\end{figure}

\section{Conclusion: Action from Meaning}

Semantic execution provides a structured, geometry-driven interface between the internal world of belief $\Phi$ and the external world of action $A$. By replacing brittle symbolic triggers with modular, interpretable epistemic readiness conditions encoded in activation basins $\mathcal{A}_a \subset \Phi$, this framework allows behavior to emerge directly from the agent's understanding.

We have shown how actions are gated by semantic preconditions involving specific sectors $\Sigma$ and abstraction scales $\Phi^{(k)}$, how execution is triggered by belief trajectories $\gamma(t)$ crossing activation boundaries with sufficient commitment, and how reflective layers $\Sigma_{refl}$ provide oversight, enabling suppression or replanning based on deeper principles.

This model bridges language, belief, and action, giving embodied agents not just the capacity to act, but to understand when, why, and how to act---or to refrain from acting. Control policies emerge naturally from the guided motion of belief through the structured semantic landscape. Future work involves realizing this architecture in practical systems, such as LLM-driven robots, potentially leading to behavior that is not merely competent, but reflects genuine cognitive depth---behavior born of structured belief. When belief acquires geometry, it becomes the source of control.


\subsection*{Chapter Summary}
This chapter introduces Semantic Execution, the framework modeling the interface between the agent's internal belief state ($\phi$) and its external actions ($A$). It posits that actions are enabled and potentially triggered when the agent's belief trajectory ($\gamma(t)$) dynamically enters specific regions within the semantic manifold $\Phi$, termed Activation Basins ($A_a$). These basins are defined not by arbitrary rules, but by the semantic preconditions necessary for the action $a$, including requirements related to active semantic sectors ($\Sigma$), appropriate abstraction levels ($k$), and sufficient coherence ($\kappa$). Action initiation is viewed as a flow-based process, requiring committed entry into the basin. Furthermore, the chapter details the role of reflective control, mediated by the reflective sector ($\Sigma_{refl}$) and a diagnostic function ($\delta_a$), which can gate or suppress execution based on higher-level considerations like goals, values, or safety constraints. Semantic Execution thus provides a mechanism for grounding behavior in the agent's structured understanding and internal state dynamics.
	\chapter{Activation Basins}
\label{chap:ActivationBasins}

\section{Introduction: Defining Readiness for Action}

Chapter~\ref{chap:SemanticExecution} introduced the core concept of Semantic Execution: actions $a \in A$ are enabled when the agent's belief trajectory $\gamma(t)$ enters specific, semantically meaningful regions within the belief space $\Phi$. These regions, termed activation regions or activation basins $\mathcal{A}_a$, represent the configurations of belief that satisfy the necessary preconditions for executing action $a$. They form the critical interface between the agent's internal semantic state and its potential for external behavior.

While the previous chapter outlined the overall process, this chapter focuses specifically on the nature, structure, and properties of these activation basins $\mathcal{A}_a$. What defines their boundaries? How do they relate to the underlying semantic geometry ($\Sigma$, $\Phi^{(k)}$)? How does the agent's state within a basin relate to action readiness? By exploring the geometry and dynamics of activation basins, we gain deeper insight into how structured belief gives rise to grounded, context-sensitive, and interpretable action. Activation basins are the geometric embodiment of action affordances within the semantic landscape.

\section{Activation Basins and Semantic Preconditions}

An activation basin $\mathcal{A}_a$ is not an arbitrary region but is fundamentally defined by the set of semantic preconditions required for action $a$. A belief state $\phi$ lies within $\mathcal{A}_a$ if and only if it satisfies these conditions. As outlined previously, these preconditions typically involve:
\begin{itemize}
	\item \textbf{Sectoral Requirements:} Action $a$ might require specific information or processing typically associated with certain semantic sectors $\Sigma$. For example, a manipulation action requires perceptual grounding ($\Sigma_{perc}$) and possibly motor planning ($\Sigma_{plan}$). The definition of $\mathcal{A}_a$ thus incorporates constraints like $\phi|_{\Sigma_{perc}} \neq \emptyset$ or requiring specific content within relevant sector projections.
	\item \textbf{Abstraction/Scale Requirements:} Actions often require beliefs to be resolved at appropriate levels of abstraction $k$. A simple reactive response might only need grounded $\phi^{(0)}$ information, while a complex strategic action might require alignment across multiple levels, from abstract goals ($\phi^{(2)}$) down to concrete execution details ($\phi^{(0)}$). $\mathcal{A}_a$ enforces these constraints via conditions like $\phi \cap \Phi^{(k)} \neq \emptyset$ for necessary $k$.
	\item \textbf{Coherence Requirements:} Action execution generally requires a sufficient degree of internal consistency and lack of conflict within the relevant parts of the belief state. $\mathcal{A}_a$ incorporates this via thresholds on coherence measures, e.g., $\kappa(\phi|_{\Sigma_{relevant}}) < \tau_{\kappa}$. Ambiguous or contradictory belief states may fall outside the basin.
	\item \textbf{Other Potential Preconditions:} Depending on the action and agent, preconditions might also involve temporal constraints (e.g., beliefs about recency), resource availability, or specific relational structures within $\phi$.
\end{itemize}

\begin{table}[ht]
	\centering
	\begin{tabular}{@{} l l @{}}
		\toprule
		\textbf{Precondition Type} & \textbf{Description} \\
		\midrule
		Sectoral Requirements     & Beliefs must exist in / activate necessary functional sectors ($\Sigma$). \\
		Abstraction/Scale Reqs    & Beliefs must be present at appropriate abstraction level(s) ($\Phi^{(k)}$). \\
		Coherence Requirements    & Relevant beliefs must be sufficiently consistent / unambiguous ($\kappa$). \\
		Other (Contextual)        & May include temporal, resource, or specific content constraints. \\
		\bottomrule
	\end{tabular}
	\caption{Typical semantic preconditions defining activation basins $\mathcal{A}_a$.}
	\label{tab:semantic_preconditions}
\end{table}

Formally, $\mathcal{A}_a = \{\phi \in \Phi \mid P_1(\phi, a) \land P_2(\phi, a) \land \dots \land P_n(\phi, a)\}$, where $P_i$ are predicates representing the semantic preconditions. The specific combination of these conditions dictates the shape, size, and location of the activation basin $\mathcal{A}_a$ within the high-dimensional semantic space $\Phi$, particularly when visualized within the ($\Sigma, k$) semantic geometry. Actions requiring complex, multi-faceted preconditions (e.g., cross-sector and cross-scale alignment) will correspond to more constrained or intricately shaped activation basins.

\begin{figure}[htbp]
	\centering
	\input{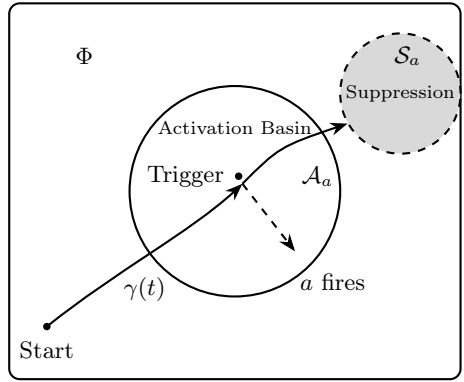}
	\caption{Activation Basin \(\mathcal{A}_a\) in semantic state space \(\Phi\). A belief trajectory \(\gamma(t)\) approaches from an initial state, enters \(\mathcal{A}_a\), and triggers action \(a\). A suppression region \(\mathcal{S}_a\), illustrated separately, may inhibit execution even within proximity.}
	\label{fig:activation-basin}
\end{figure}

\section{Geometry and Topology of Basins}

Viewing $\mathcal{A}_a$ as a geometric region within $\Phi$ raises questions about its properties:
\begin{itemize}
	\item \textbf{Connectedness:} Is $\mathcal{A}_a$ typically a single connected region, or can it be fragmented into multiple disjoint sub-regions where action $a$ is possible under different semantic configurations?
	\item \textbf{Boundaries ($\partial \mathcal{A}_a$):} What defines the transition from a state not ready for action $a$ to one that is? The boundary $\partial \mathcal{A}_a$ represents the threshold where semantic preconditions are just met. The nature of this boundary (sharp or fuzzy, smooth or complex) depends on how preconditions are defined (e.g., hard thresholds vs. soft gradients).
	\item \textbf{Overlap and Conflict:} Activation basins for different actions $\mathcal{A}_a$ and $\mathcal{A}_b$ can overlap ($\mathcal{A}_a \cap \mathcal{A}_b \neq \emptyset$). When the belief state $\phi$ enters such an intersection, multiple actions become enabled, necessitating the conflict resolution mechanisms ($R_{action}$) discussed in Chapter~\ref{chap:SemanticExecution}. The geometry of these intersections dictates the nature of action conflicts.
	\item \textbf{Relation to Semantic Metric ($d$):} The semantic distance $d(\phi_1, \phi_2)$ relates to basin structure. Points deep within $\mathcal{A}_a$ might represent states strongly satisfying preconditions, while points near the boundary $\partial \mathcal{A}_a$ are marginal. Distance to the boundary could be a measure of readiness robustness.
	\item \textbf{Shape and Structure:} The "shape" of $\mathcal{A}_a$ reflects the nature of the action's preconditions. Actions requiring precise alignment across many dimensions ($\Sigma, k, \kappa$) might have narrow, elongated basins, while simple reactive actions might have broader, lower-level basins.
\end{itemize}

\begin{figure}[h]
	\centering
	\input{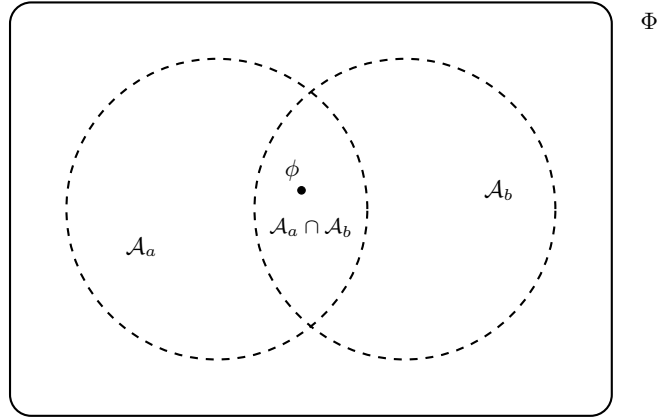}
	\caption{Overlapping activation basins \(\mathcal{A}_a\) and \(\mathcal{A}_b\) within the semantic manifold \(\Phi\). A belief state \(\phi\) located in the intersection \(\mathcal{A}_a \cap \mathcal{A}_b\) enables simultaneous triggering of both actions \(a\) and \(b\).}
	\label{fig:activation_basins_overlap}
\end{figure}

Understanding the geometry and topology of activation basins provides insight into the structure of action possibilities available to the agent from any given belief state $\phi$.

\section{Activation Dynamics and Thresholds}

Mere presence within $\mathcal{A}_a$ might not be sufficient for execution; the degree of activation often matters.

\textbf{Activation Level ($\lambda$):} We define an activation level $\lambda(\mathcal{A}_a, \phi)$ which quantifies how strongly the current state $\phi$ satisfies the conditions for action $a$. This could measure:
\begin{itemize}
	\item The "volume" or "mass" of the belief state $\phi$ residing within $\mathcal{A}_a$.
	\item A weighted score based on how well preconditions ($\Sigma, k, \kappa$) are met.
	\item The semantic distance from $\phi$ to the "center" or most stable region of $\mathcal{A}_a$.
\end{itemize}

\textbf{Activation Threshold ($\tau_a$):} Execution typically requires the activation level to exceed a threshold: $\lambda(\mathcal{A}_a, \phi) > \tau_a$. This prevents triggering actions based on weak or transient alignment with preconditions.

\textbf{Graded Activation:} The value $\lambda(\mathcal{A}_a, \phi)$ relative to $\tau_a$ allows for graded readiness. States deep within the basin ($\lambda \gg \tau_a$) might lead to confident, rapid execution, while states just barely meeting the threshold ($\lambda \approx \tau_a$) could result in hesitant or preliminary actions.

\textbf{Temporal Softening:} Thresholds $\tau_a$ might not be static; they could potentially adapt over time or based on context (e.g., lower threshold in urgent situations, higher threshold for critical actions).

Activation dynamics provide a quantitative layer on top of the geometric definition of the basin, allowing for nuanced control over action initiation based on the strength and quality of the belief state's alignment with the action's semantic requirements.

\section{Trajectories, Boundaries, and Triggering}

Activation basins gain their full meaning in the context of belief trajectories $\gamma(t)$ moving through $\Phi$ under the influence of the flow field $F$.
\begin{itemize}
	\item \textbf{Boundary Interaction:} The critical event for action initiation is the interaction of $\gamma(t)$ with the basin boundary $\partial \mathcal{A}_a$. Simple intersection may not be enough.
	\item \textbf{Committed Entry:} As discussed in Chapter~\ref{chap:SemanticExecution}, a robust trigger requires not just crossing $\partial \mathcal{A}_a$ but doing so with sufficient "momentum" or commitment. This can be formalized by requiring a positive inward velocity component ($\langle F(\gamma(t)), \hat{n}_{inward} \rangle > 0$) or a positive rate of change in activation level ($\frac{d\lambda(\mathcal{A}_a, \gamma(t))}{dt} > 0$).
	\item \textbf{Entry and Exit Points:} The specific point on $\partial \mathcal{A}_a$ where the trajectory enters can influence the subsequent action execution (e.g., different entry points might correspond to different action parameters). Similarly, trajectories can exit basins if belief states evolve away from satisfying preconditions.
	\item \textbf{Multi-Phase Activation Trajectories:} For complex tasks, executing action $a$ might require the trajectory $\gamma(t)$ to sequentially enter and traverse a series of intermediate sub-basins $\mathcal{A}^{(1)}_a, \mathcal{A}^{(2)}_a, \dots, \mathcal{A}^{(n)}_a$, each corresponding to achieving a necessary semantic sub-goal or readiness stage before final commitment. The trajectory must follow a specific path through these nested or sequential basins.
\end{itemize}
The interplay between belief flow $F$ and the geometry of activation basin boundaries $\partial \mathcal{A}_a$ determines precisely when and how actions are triggered.

\section{Suppression Surfaces and Inhibition}

Complementary to activation basins $\mathcal{A}_a$ are suppression surfaces or regions $S_a$.

\textbf{Definition (Suppression Surface $S_a$):} $S_a \subset \Phi$ is a region, typically disjoint from $\mathcal{A}_a$, such that if the belief state $\phi$ enters $S_a$, action $a$ is actively inhibited or prevented, even if $\phi$ also satisfies the conditions for being in $\mathcal{A}_a$.

Suppression surfaces encode conditions under which an action, though seemingly prepared for, should not be executed. They represent explicit veto conditions. Sources for suppression include:
\begin{itemize}
	\item Safety Violations: Detecting conditions that make action $a$ unsafe.
	\item Reflective Contradictions: High-level reflection ($\Sigma_{refl}$) identifying conflicts with goals, values, or ethics (see next section).
	\item Coherence Failure: Detecting significant inconsistency or instability in the beliefs supporting the action.
	\item Environmental Veto: External signals indicating the action is inappropriate.
\end{itemize}
The flow field $F$ might actively repel trajectories from $S_a$, or crossing into $S_a$ could trigger an immediate halt or modification of the intended action $a$. The interplay between attraction towards $\mathcal{A}_a$ (representing readiness) and potential inhibition from $S_a$ (representing veto conditions) allows for complex gating of behavior based on both positive preconditions and negative constraints.

\section{Relationship to Control and Reflection}

Activation basins are tightly integrated with the agent's control and reflective mechanisms.
\begin{itemize}
	\item \textbf{Reflective Gating:} As introduced in Chapter~\ref{chap:SemanticExecution}, the reflective diagnostic $\delta_a(\phi_{refl})$ acts as a gate. Even if $\gamma(t)$ enters $\mathcal{A}_a$ with commitment, execution only proceeds if $\delta_a = \text{approve}$. The reflective state can thus effectively veto actions by overriding the readiness signal from the basin.
	\item \textbf{Dynamic Basin Modification?:} Can reflection dynamically alter the shape or accessibility of activation basins? For example, a strong reflective concern might effectively "shrink" $\mathcal{A}_a$ or raise the activation threshold $\tau_a$, making the action harder to trigger. Conversely, strong goal commitment might "deepen" the basin or lower the threshold.
	\item \textbf{Action Selection ($R_{action}$):} When $\phi$ lies in the intersection $\mathcal{A}_{a_1} \cap \mathcal{A}_{a_2}$, the resolution function $R_{action}$ selects the action to execute. This selection might depend on which basin is more strongly activated ($\lambda$), which trajectory entered its basin more decisively, or priorities determined by reflective control ($\Sigma_{refl}$).
\end{itemize}
Activation basins provide the semantic grounding for action potential, but their effective activation and the final translation into behavior are modulated by higher-level control and reflection operating on the same belief space $\Phi$.

\section{Conclusion: Basins as Geometric Affordances}

Activation basins $\mathcal{A}_a$ are the geometric locus of action readiness within the semantic state space $\Phi$. They translate complex semantic preconditions---involving specific sectors $\Sigma$, abstraction levels $\Phi^{(k)}$, and coherence thresholds $\kappa$---into well-defined regions of the belief manifold.

The interaction between belief trajectories $\gamma(t)$, driven by semantic flows $F$, and the boundaries $\partial \mathcal{A}_a$ of these basins determines when actions are triggered, incorporating dynamics of momentum and commitment. Activation levels $\lambda(\mathcal{A}_a, \phi)$ allow for graded readiness, while suppression surfaces $S_a$ and reflective gating $\delta_a$ provide mechanisms for inhibition and control.

Activation basins represent a form of "semantic affordance" within the agent's internal cognitive landscape. They define where and under what epistemic conditions specific actions become possible or likely. Understanding their structure, geometry, and dynamics is key to understanding how structured belief gives rise to context-sensitive, potentially reflective, and goal-directed behavior in advanced semantic agents. They are the crucial geometric link between thought and action.


\subsection*{Chapter Summary}
This chapter elaborates on Activation Basins ($A_a$), the regions within the semantic state space $\Phi$ introduced in the previous chapter as the locus of action readiness. It details how these basins are defined by the specific semantic preconditions required for an action $a$, including necessary information from specific sectors ($\Sigma$), appropriate levels of abstraction ($k$), and sufficient coherence ($\kappa$). The chapter discusses the geometric and topological properties of these basins, such as boundaries ($\partial A_a$), potential overlaps ($A_a \cap A_b$), and their relation to the semantic metric ($d$). Action readiness is often graded, potentially requiring belief activation ($\lambda$) within the basin to exceed a threshold ($\tau_a$). The dynamics of belief trajectories ($\gamma(t)$) interacting with basin boundaries, requiring committed entry, are explored as the trigger mechanism. Complementary suppression surfaces ($S_a$) are introduced as regions inhibiting action despite precondition satisfaction, providing a mechanism for vetoes (e.g., safety, reflective disagreement via $\delta_a$). Activation basins thus represent the geometric embodiment of action affordances within the semantic manifold, shaped by preconditions and modulated by control mechanisms.
	\chapter{Embodied Simulation}
\label{chap:EmbodiedSimulation}

\section{Introduction: Internal Rehearsal and Grounding}

The previous chapters explored how belief states $\phi$ within the semantic state space $\Phi$ can directly trigger external actions when trajectories $\gamma(t)$ enter activation basins $\mathcal{A}_a$. However, intelligent agents do not, and should not, act solely based on immediate readiness. They plan, predict consequences, evaluate alternatives, and learn from hypothetical scenarios. This necessitates a mechanism for internal rehearsal---running simulations of potential interactions within the belief space itself, without necessarily engaging external effectors.

This chapter introduces Embodied Simulation as this crucial internal process. We define it as the generation and evolution of belief trajectories $\gamma_{sim}(t)$ within $\Phi$ that represent potential sequences of states, actions, and their perceived outcomes, as if the agent were interacting with the environment. Embodied simulation serves multiple critical functions:
\begin{itemize}
	\item \textbf{Planning and Evaluation:} Testing potential action sequences internally to predict outcomes and select optimal plans.
	\item \textbf{Grounding Abstract Concepts:} Connecting high-level goals or concepts to concrete, expected sensorimotor experiences by simulating their execution.
	\item \textbf{Counterfactual Reasoning:} Exploring "what if" scenarios by simulating alternative actions or starting conditions.
	\item \textbf{Skill Refinement:} Improving internal models and action parameters by comparing simulated outcomes to desired results or past experiences.
\end{itemize}
Embodied simulation provides a vital link between abstract reasoning and potential physical interaction, allowing agents to "think through doing" internally before committing to external action.

\section{Simulation Trajectories in \texorpdfstring{$\Phi$}{Phi}}

Embodied simulation involves generating and evolving specific belief trajectories $\gamma_{sim}(t)$ within the semantic state space $\Phi$. These trajectories differ from those driven purely by external perception or reflection; they represent internally generated sequences modeling potential interactions.

Where do these simulations unfold within the Semantic Geometry ($\Sigma, k$)?
\begin{itemize}
	\item \textbf{Planning Sectors ($\Sigma_{plan}$):} Simulations often originate from goals or candidate plans residing in $\Sigma_{plan}$, possibly at higher abstraction levels $\Phi^{(k)}, k \ge 1$.
	\item \textbf{Simulated Perception/Action ($\Sigma_{sim-perc}, \Sigma_{sim-act}$):} The simulation might involve generating expected perceptual consequences (populating a simulated perceptual sector) and representing intended actions (populating a simulated action/motor sector), likely at grounded levels $\Phi^{(0)}$.
	\item \textbf{Narrative Integration ($\Sigma_{narr}$):} The sequence of simulated states and actions can form a narrative within $\Sigma_{narr}$, representing the unfolded hypothetical scenario.
	\item \textbf{Reflective Evaluation ($\Sigma_{refl}$):} The outcomes of the simulation are often evaluated within $\Sigma_{refl}$ against goals, values, or constraints.
\end{itemize}
A simulation trajectory might thus involve a complex path weaving through these sectors, starting with an abstract plan, elaborating ($V$) it into simulated actions, generating expected perceptions via an internal world model (perhaps using a specialized Assimilation $A_{sim}$), and evaluating the outcome reflectively.

\section{Simulating Actions Internally}

A simulation trajectory $\gamma_{sim}(t)$ unfolds within the internal space $\Phi$. It represents the agent exploring the semantic consequences of a potential action sequence. For instance, simulating "grasp(cup)" involves evolving the belief state $\phi$ to include representations like \{"Hand approaching cup", "Expected tactile contact", "Grasp confirmed"\}, potentially moving through $\Sigma_{plan}$ and $\Sigma_{sim-perc}$. This internal evolution allows the agent to assess the feasibility and likely outcome based on its world model and belief dynamics. Crucially, this internal processing must be decoupled from physical execution.

\section{Gating External Execution During Simulation}

A critical aspect of embodied simulation is that it allows the agent to explore action possibilities---potentially even having simulated trajectories $\gamma_{sim}(t)$ enter activation basins $\mathcal{A}_a$---without triggering the corresponding physical action $a$. How is external execution gated during internal simulation? Several mechanisms, individually or in combination, are possible within the framework:

\begin{itemize}
	\item \textbf{Reflective Suppression:} The reflective sector $\Sigma_{refl}$ might be configured to actively suppress external execution during simulation modes. The reflective diagnostic function $\delta_a$ (Chapter~\ref{chap:SemanticExecution}) could output $\delta_a(\phi_{refl}) = \text{suppress}$ for all relevant physical actions $a$ while simulation is running, effectively decoupling internal readiness checks from external output commands based on a meta-cognitive signal indicating "simulation mode is active".
	\item \textbf{Dedicated Simulation Subspace/Sectors:} Simulations might occur within specific "simulation-only" sectors or use belief fragments explicitly tagged as 'simulated'. The execution interface $\pi : \Phi \rightarrow A \cup \{0\}$ would then be configured to ignore activation signals ($\lambda(\mathcal{A}_a, \phi)$) arising from these tagged or sector-specific beliefs when considering physical actions.
	\item \textbf{Intentional Trajectory Control:} The internal process guiding the simulation might intentionally steer the trajectory $\gamma_{sim}(t)$ to explore the vicinity of $\mathcal{A}_a$ or its boundary $\partial \mathcal{A}_a$ to assess readiness, but actively avoid the deep, committed entry required to trigger execution (e.g., ensuring $\frac{d\lambda}{dt}$ remains low or oscillatory near the boundary). This involves fine-grained control over the internal belief flow $F$ during simulation.
	\item \textbf{Architectural Output Gating:} A simpler mechanism could be an explicit architectural gate between the belief space dynamics (where $\mathcal{A}_a$ might be entered) and the physical motor control system. This gate would be programmatically closed when the agent enters simulation mode, blocking any potential action trigger from propagating to the effectors.
\end{itemize}

\begin{figure}[htbp]
	\centering
	\input{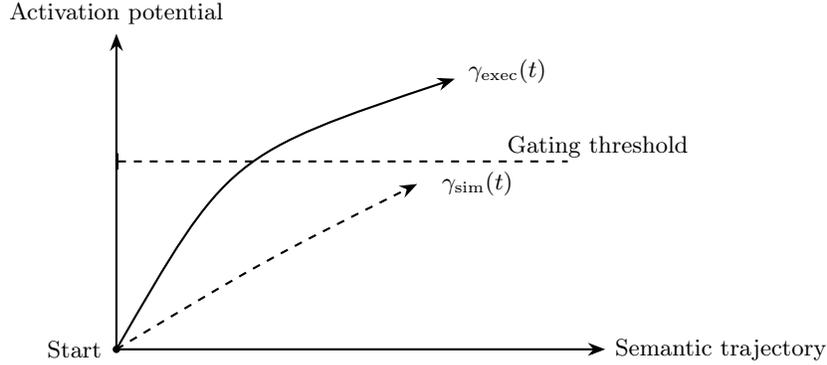}
	\caption{Comparison of simulated versus executed semantic trajectories. The internal simulation \(\gamma_{\text{sim}}(t)\) explores semantic space but remains below the activation threshold, while the executed path \(\gamma_{\text{exec}}(t)\) exceeds this threshold, triggering real-world behavior. Gating mechanisms control the transition between internal and external execution.}
	\label{fig:simulation-vs-execution}
\end{figure}

These mechanisms ensure that internal exploration of action sequences via simulation does not lead to unintended physical consequences, allowing the agent to safely "try out" actions internally.

\section{Grounding Abstract Concepts}

Embodied simulation plays a crucial role in grounding abstract beliefs, plans, and concepts developed at higher levels $\Phi^{(k)}$ ($k \ge 1$). Abstract intentions like "Make coffee" or concepts like "obstacle" lack direct physical meaning until connected to sensorimotor contingencies. Simulation provides this connection by:
\begin{itemize}
	\item \textbf{Elaboration ($V$):} Taking an abstract plan or concept (e.g., $\phi^{(2)} \in \Sigma_{plan}$) and elaborating it downwards ($V_2^1, V_1^0$) into a sequence of concrete, simulated actions and states within the simulation context ($\phi^{(0)}_{sim}$).
	\item \textbf{Simulated Sensorimotor Loops:} Running internal simulations of "If I perform motor command M, I expect perception P." This links abstract action descriptions to expected sensory consequences, even if those consequences are generated by an internal world model rather than external sensors via $X$.
	\item \textbf{Testing Affordances:} Simulating interactions allows the agent to test the affordances implied by abstract concepts. Simulating "grasp(cup)" involves generating expected tactile and visual feedback, thus grounding the abstract "cup" concept in simulated physical interaction.
\end{itemize}
Through repeated simulation, abstract representations in $\Phi$ become associated with the rich, detailed sensorimotor patterns residing at lower levels $\Phi^{(0)}$, providing a mechanism for symbol grounding within the semantic state space.

\section{Counterfactuals and Prediction}

Embodied simulation is the natural mechanism for exploring counterfactuals ("What if...?") and making predictions.
\begin{itemize}
	\item \textbf{Counterfactuals:} By initializing a simulation trajectory $\gamma_{sim}(t)$ from a hypothetical belief state $\phi_{hypo}$ (which might differ from the current perceived state $\phi_{perc}$) or by simulating an action $a'$ different from the one planned, the agent can explore alternative histories or consequences within $\Phi$.
	\item \textbf{Prediction:} By running a simulation forward from the current estimated state $\phi_{current}$ under a planned action sequence $a_1, a_2, \dots$, the agent generates a predicted future trajectory $\gamma_{pred}(t)$. The endpoint $\gamma_{pred}(T)$ represents the predicted state after executing the sequence, which can be evaluated against goals.
\end{itemize}

This requires an internal generative model capable of predicting state transitions and expected sensory feedback based on simulated actions. This model might itself be represented and refined within $\Phi$, perhaps in dedicated world-model sectors ($\Sigma_{world-model}$). Assimilation during simulation ($A_{sim}$) would involve incorporating these internally generated predictions into the simulation trajectory.

\section{Role in Planning and Skill Refinement}

Embodied simulation is integral to sophisticated planning and learning:
\begin{itemize}
	\item \textbf{Planning:} Simulation allows model-based planning. The agent can internally simulate multiple candidate action sequences, evaluate their predicted outcomes (using $\Sigma_{refl}$ to assess goal achievement, cost, risk), and select the sequence whose simulated trajectory $\gamma_{sim}(t)$ is most promising. This avoids costly or irreversible trial-and-error in the real world.
	\item \textbf{Skill Refinement:} By comparing the outcomes of simulated actions with desired outcomes or internal references, the agent can adjust its internal models or action parameters. For example, if simulating a grasp consistently fails internally, the agent can modify the grasp parameters within $\Sigma_{plan}$ before attempting it physically. Discrepancies between simulated outcomes and actual outcomes after execution can drive learning and refinement of the internal world model used for simulation (Part~\ref{part:learning_and_adaptation}).
\end{itemize}

\begin{figure}[ht]
	\centering
	\input{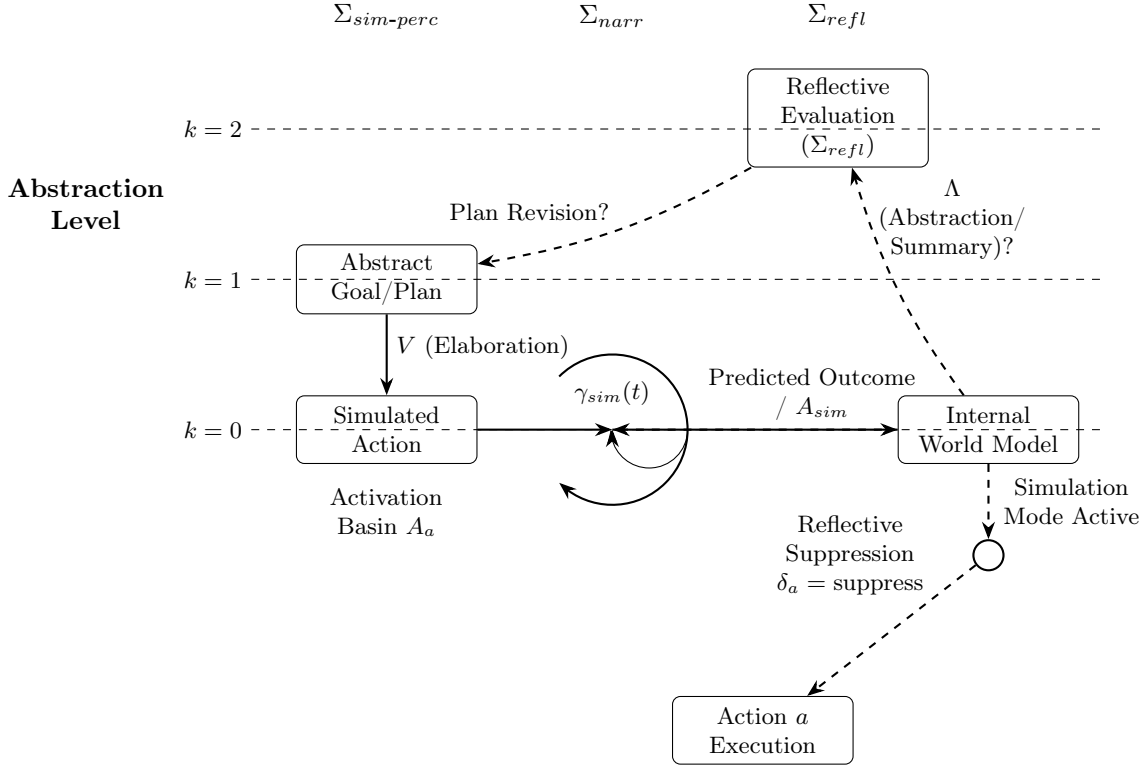}
	\caption{
		The embodied simulation cycle depicted within semantic geometry. An abstract plan originating in $\Sigma_{plan}$ at higher abstraction levels is elaborated downward via $V$ to generate simulated actions at the grounded level ($k=0$). These simulated actions interact with the internal world model to produce predicted sensory consequences ($A_{sim}$), while external execution is gated by reflective suppression ($\delta_a = \text{suppress}$) to ensure internal rehearsal remains decoupled from real-world action. Simulated percepts and narratives are abstracted upward via $\Lambda$ into $\Sigma_{refl}$ for reflective evaluation, enabling potential plan revision. This cycle supports safe internal exploration, grounding of abstract concepts, and reflective self-regulation.
	}
	\label{fig:embodied_sim_cycle}
\end{figure}

Simulation thus provides an internal "training ground" for improving plans, skills, and world models.

\section{Challenges and Future Directions}

Implementing effective embodied simulation within this framework presents challenges:
\begin{itemize}
	\item \textbf{World Model Fidelity:} The accuracy of simulation depends entirely on the quality of the internal world model used to predict consequences. Building and maintaining such models within $\Phi$ is non-trivial.
	\item \textbf{Computational Cost:} Detailed simulation, especially involving complex physics or multi-step interactions, can be computationally expensive. Agents need strategies to manage simulation complexity (e.g., varying abstraction level $\Phi^{(k)}$, simulating only key aspects).
	\item \textbf{Distinguishing Simulation from Reality:} The agent needs robust mechanisms (like those discussed in Section 4) to clearly distinguish beliefs arising from internal simulation ($\phi_{sim}$) from those grounded in direct perception ($\phi_{perc}$) to avoid confusion or hallucination. This might involve tagging beliefs by origin or maintaining separation between simulation sectors and perceptual sectors.
	\item \textbf{Scalability:} How does simulation scale to long-horizon tasks or complex environments?
\end{itemize}
Future work could explore adaptive simulation strategies, learning world models directly within $\Phi$, and formalizing the mechanisms for gating execution and distinguishing simulated from real experience.

\section{Conclusion: Thinking Through Doing (Internally)}

Embodied simulation is the capacity for an agent to internally trace trajectories $\gamma_{sim}(t)$ through its semantic state space $\Phi$, representing potential interactions with the world without necessarily enacting them physically. It complements Semantic Execution by providing a crucial internal loop for planning, prediction, counterfactual reasoning, concept grounding, and skill refinement.

By leveraging the structures of Semantic Geometry ($\Sigma, \Phi^{(k)}$) and the dynamics of elaboration ($V$) and simulated assimilation ($A_{sim}$), agents can "think through doing" in an internal, embodied manner. While distinct from external action (requiring mechanisms for gating execution), embodied simulation grounds abstract thought in potential sensorimotor experience and allows behavior to be shaped by foresight and internal evaluation. It is a vital component for bridging the gap between high-level cognition and effective, grounded action in complex environments.


\subsection*{Chapter Summary}
This chapter introduces Embodied Simulation, the capacity for an agent to generate and evolve belief trajectories ($\gamma_{sim}(t)$) within its semantic space $\Phi$ that represent potential sequences of states and actions, without necessarily engaging external effectors. This internal rehearsal mechanism serves several key cognitive functions: it allows for model-based planning and the evaluation of potential action outcomes; it provides a way to ground abstract concepts or high-level goals ($\Phi^{(k)}, k>0$) by connecting them to simulated concrete sensorimotor experiences ($\Phi^{(0)}_{sim}$), often via the Elaboration operator ($V$); it enables counterfactual reasoning and prediction based on internal world models; and it facilitates skill refinement by allowing internal practice and comparison. The chapter discusses necessary mechanisms for gating external execution during simulation (e.g., via reflective suppression $\delta_a$ or dedicated simulation sectors) to prevent unintended physical actions. While acknowledging challenges such as world model fidelity and computational cost, embodied simulation is presented as a vital component bridging abstract cognition and effective, grounded interaction with the world.
	
	\part{Meta-Cognition and Recursive Regulation}
	\label{part:meta_cognition}
	
	\chapter{Meta-Assimilation: Integrating Introspective Beliefs}
\label{chap:MetaAssimilation}

\section{Introduction: The Need for Meta-Level Integration}

The preceding parts of this monograph have detailed the core operators governing belief dynamics within the semantic state space $\Phi$. Assimilation ($A$, Chapter~\ref{chap:Assimilation}) integrates external observations or object-level internal constructions; Nullification ($N_t$, Chapter~\ref{chap:Nullification}) models decay; Annihilation ($K$, Chapter~\ref{chap:Annihilation}) handles erasure; Abstraction ($\Lambda$) and Elaboration ($V$) manage semantic scaling (Chapter~\ref{chap:SemanticScaling}).

Part~\ref{part:meta_cognition} delves into meta-cognition---the agent's awareness and regulation of its own cognitive processes. A crucial aspect of meta-cognition, introduced conceptually in Meta-Introspection (Chapter~\ref{chap:MetaIntrospection}), is the formation of beliefs about the agent's own internal state $\phi$. This requires a mechanism distinct from the general Assimilation operator $A$, which primarily handles object-level information (about the world, tasks, simulations). This chapter introduces the Meta-Assimilation operator, denoted $M$, specifically designed to integrate internally generated, introspective insights into the agent's belief state.

$M$ takes the outputs of internal monitoring processes---information about coherence, load, trajectory properties, belief structures, etc.---and incorporates them as explicit meta-beliefs, typically within the reflective sector $\Sigma_{refl}$. By formalizing $M$ as a distinct operator, we can cleanly analyze the unique dynamics and challenges of integrating self-referential information and building higher-order awareness within the semantic manifold.

\section{Formal Definition of the Meta-Assimilation Operator (\texorpdfstring{$M$}{M})}

We define the Meta-Assimilation operator $M$ as a function that integrates introspectively generated information into the current belief state.

\textbf{Definition 28.1 (Meta-Assimilation Operator $M$).} The Meta-Assimilation Operator $M$ is a mapping:
$$ M : \Phi \times \Phi_{introspective} \rightarrow \Phi $$
where:
\begin{itemize}
	\item $\phi \in \Phi$ is the agent's current overall belief state.
	\item $\phi_{introspective}$ represents the structured linguistic ensemble containing the meta-information generated by internal introspection mechanisms (e.g., outputs of functions $I_j$ monitoring properties like coherence $\kappa$, load $\lambda$, trajectory $\gamma(t)$, or specific belief content). This input concerns the state $\phi$ itself.
	\item $\phi_{new} = M(\phi, \phi_{introspective})$ is the updated belief state, where the meta-information has been integrated, typically resulting in new or modified meta-beliefs within the reflective sector ($\phi_{new}|_{\Sigma_{refl}}$).
\end{itemize}
Unlike the general Assimilation operator $A$, whose input $\phi_{input}$ usually represents external data or object-level simulation results, the input $\phi_{introspective}$ to $M$ is inherently self-referential. The operator $M$ must therefore handle the integration of beliefs whose content pertains to the belief state itself.

\section{Desiderata for Meta-Assimilation}

While sharing some desiderata with general Assimilation $A$ (e.g., coherence preservation D2, anchoring D6, locality D7 from Chapter~\ref{chap:Assimilation}), the operator $M$ has unique requirements due to the nature of its input:
\begin{description}
	\item[DM1: Accurate Self-Reflection] The integration process should aim to accurately represent the introspected information within $\Sigma_{refl}$. $M$ should ideally add meta-beliefs like "$\kappa$ is low" only when introspection genuinely indicates low coherence.
	\item[DM2: Object-Meta Consistency] $M$ must strive to maintain consistency not only within the meta-level ($\Sigma_{refl}$) but also between meta-beliefs and the object-level beliefs they refer to. Integrating "$\kappa$ is low" might necessitate triggering corrective actions (like $A_{corr}$ or $K$) at the object level via regulatory loops.
	\item[DM3: Handling Introspective Uncertainty/Bias] Real-world introspection is often noisy or biased. $M$ should ideally be able to represent confidence levels or potential biases associated with $\phi_{introspective}$ (e.g., forming "I suspect my load $\lambda$ is high" rather than asserting it definitively).
	\item[DM4: Recursive Stability] As $M$ forms beliefs about $\phi$, and $\phi$ includes those meta-beliefs, the process must be stable and avoid pathological self-referential loops or infinite regress (perhaps via attenuation, abstraction limits, or effort $\epsilon$ constraints).
	\item[DM5: Timeliness] Meta-information is often most useful shortly after the state it describes occurs. $M$ should integrate introspective findings promptly enough to enable effective regulation.
\end{description}

\section{Mechanisms of Meta-Assimilation}

The specific mechanisms implementing $M$ depend on the agent's parameterization $\theta$, particularly the representation mode $\rho$ and the sophistication of $\Sigma_{refl}$. Plausible approaches include:
\begin{itemize}
	\item \textbf{Structured Insertion into $\Sigma_{refl}$:} For symbolic $\rho$, $M$ might involve adding tagged meta-statements (e.g., `MetaBelief(Target=..., Content=...)`) to a dedicated knowledge graph or database representing $\Sigma_{refl}$.
	\item \textbf{Embedding Space Dynamics:} For embedding $\rho$, $M$ could involve updating specific dimensions or regions of the state vector corresponding to $\Sigma_{refl}$ based on encoded introspective features. Techniques might resemble gated updates in memory networks.
	\item \textbf{Prompt-Based Meta-Belief Formation (LLM Implementation):} If leveraging an LLM, $M$ could be implemented by providing the LLM with the current state $\phi$ (or relevant parts), the introspective findings $\phi_{introspective}$, and a prompt instructing it to formulate and integrate appropriate meta-beliefs into a revised representation of $\Sigma_{refl}$. Coherence checks would be crucial here.
\end{itemize}
Regardless of mechanism, $M$ performs the crucial step of making introspective data an explicit part of the agent's represented beliefs.

\section{Interaction with Other Processes and Regulation}

Meta-beliefs formed via $M$ are not merely passive records; they actively participate in regulating other cognitive processes:
\begin{itemize}
	\item \textbf{Guiding Object-Level Assimilation ($A$):} A meta-belief formed by $M$ indicating low coherence ($\kappa$) might trigger a more cautious or corrective mode ($A_{corr}$) for subsequent applications of $A$.
	\item \textbf{Modulating Nullification ($N_t$):} Meta-beliefs about the importance or relevance of certain object-level beliefs $\varphi_i$ can influence their anchoring $a_i$, thus modulating their decay rate under $N_t$.
	\item \textbf{Informing Regulation Policies ($\pi_{regulate}$):} Meta-beliefs about load $\lambda$, effort $\epsilon$, trajectory deviation $\theta$, or coherence $\kappa$ serve as direct inputs to meta-control policies that might adjust orientation, shift cognitive modes (sector activations $\vec{\alpha}(t)$), or allocate effort (Chapter~\ref{chap:SemanticFocus}).
	\item \textbf{Triggering Planning/Simulation:} Meta-beliefs identifying problems (e.g., "Plan inconsistent with identity $\vec{\eta}$") can trigger replanning within $\Sigma_{plan}$ or evaluative simulations (Chapter~\ref{chap:EmbodiedSimulation}).
\end{itemize}
$M$ is thus the gateway through which self-awareness informs self-control.

\textbf{Example of M in the Control Loop:}
\begin{enumerate}
	\item \textbf{Introspection:} An internal monitor $I_{\kappa}$ detects conflicting beliefs in $\Sigma_{plan}$, generating $\phi_{introspective} = $ \{"Coherence $\kappa(\Sigma_{plan})$ is low (value=0.3)"\}.
	\item \textbf{Meta-Assimilation:} $M(\phi, \phi_{introspective})$ integrates this, adding the meta-belief $\varphi_{meta} = $ "Planning coherence is currently low (0.3)" to $\Sigma_{refl}$ within the updated state $\phi_{new}$.
	\item \textbf{Regulation:} The regulatory policy $\pi_{regulate}$ observes $\varphi_{meta} \in \phi_{new}|_{\Sigma_{refl}}$ and triggers a corrective action, perhaps initiating a targeted $A_{corr}$ process within $\Sigma_{plan}$ or allocating more effort $\epsilon$ to reflective analysis of the plan.
\end{enumerate}
This illustrates how $M$ enables the agent to act upon its internal state assessments.

\section{Role in Recursive Meta-Cognition}

The existence of $M$ operating on introspective input $\phi_{introspective}$ naturally supports recursive meta-cognition. Introspection functions ($I_j$) can, in principle, monitor not just object-level states but also the meta-beliefs within $\Sigma_{refl}$ itself.
\begin{itemize}
	\item \textbf{Second-Order Meta-Beliefs:} Introspection might yield $\phi^{(1)}_{introspective}$ containing information about the state of $\Sigma_{refl}$. Applying $M$ again, $M(\phi, \phi^{(1)}_{introspective})$, forms second-order meta-beliefs (beliefs about meta-beliefs), perhaps residing at a higher abstraction level $k$ within $\Sigma_{refl}$. For example: "My self-assessment of coherence tends to be overly optimistic."
	\item \textbf{Levels of Self-Awareness:} The depth of this recursion (how many meta-levels are practically represented and utilized) likely depends on the agent's design $\theta$ (particularly parameters related to reflective capacity and resource limits $\lambda, \epsilon$).
\end{itemize}
$M$ provides the operational mechanism for building potentially deep, layered self-models within the semantic manifold.

\begin{figure}[htbp]
	\centering
	\input{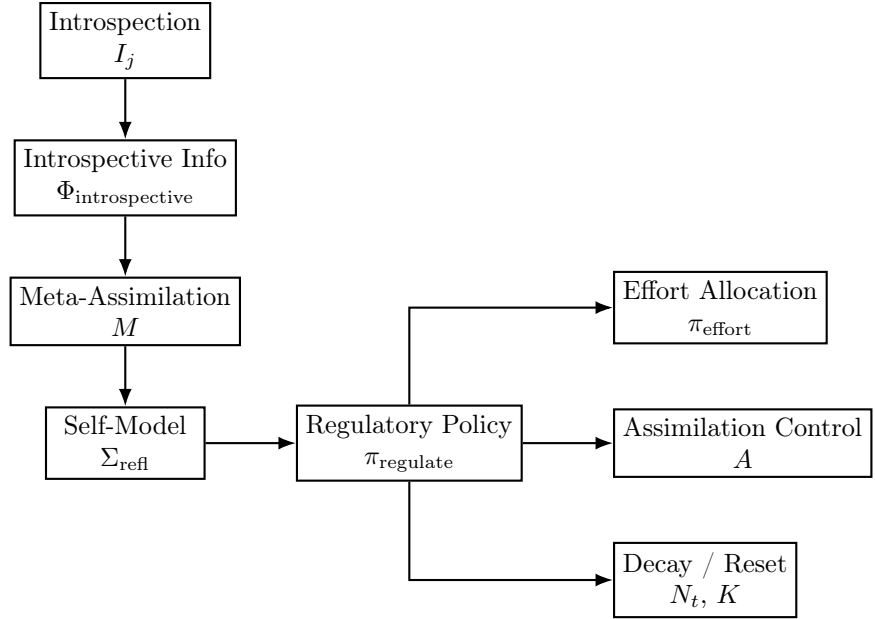} 
	\caption{Simplified meta-cognitive pipeline depicting Meta-Assimilation $M$ and its regulatory effects.}
	\label{fig:meta-loop} 
\end{figure}

\section{Conclusion: The Operator for Self-Awareness}

The Meta-Assimilation operator, $M$, provides a dedicated mechanism for integrating internally generated, introspective information about the agent's own cognitive state into its belief structure $\phi$, primarily within the reflective sector $\Sigma_{refl}$. By separating this function from the general Assimilation operator $A$, we gain clarity in modeling the unique aspects of self-referential belief formation.

$M$ enables the explicit representation of meta-beliefs concerning coherence ($\kappa$), load ($\lambda$), effort ($\epsilon$), trajectory properties ($\gamma(t)$), belief content, and potentially higher-order recursive self-assessments. Crucially, meta-beliefs formed via $M$ are not epiphenomenal; they serve as vital inputs for the agent's regulatory systems ($\pi_{regulate}$), influencing subsequent object-level processing, resource allocation, and strategic control. $M$ is thus the cornerstone operator enabling the transition from simple information processing to reflective self-awareness and adaptive self-regulation within the Semantic Manifold framework. It is the mechanism by which the agent comes to know itself.


\subsection*{Chapter Summary}
This chapter introduces the Meta-Assimilation operator ($M$), a crucial component of the framework's meta-cognitive capabilities (Part~\ref{part:meta_cognition}), designed specifically for integrating internally generated, introspective information. Distinct from the object-level Assimilation operator ($A$), $M$ takes introspective findings ($\Phi_{introspective}$)---such as assessments of coherence ($\kappa$), load ($\lambda$), or trajectory properties ($\gamma(t)$)---and incorporates them into the agent's belief state $\Phi$, typically forming explicit meta-beliefs within the reflective sector ($\Sigma_{refl}$). The chapter outlines desiderata specific to $M$, emphasizing accurate self-reflection, consistency between object- and meta-levels, handling uncertainty, and recursive stability. Meta-beliefs formed via $M$ are not passive; they serve as critical inputs to regulatory policies ($\pi_{regulate}$), enabling the agent to adapt its behavior based on self-awareness. $M$ thus provides the mechanism for self-knowledge to influence self-control and supports the potential for recursive levels of meta-cognition.
	\chapter{Meta-Introspection}
\label{chap:MetaIntrospection}

\section{Introduction: The Challenge of Self-Awareness}

The preceding Parts have constructed a detailed framework for representing belief ($\Phi$), structuring it (Scaling $\Phi^{(k)}$, Sectors $\Sigma$), evolving it (Dynamics $A, N_t, K$), regulating it (Orientation, Gauge, Identity), managing memory (Part~\ref{part:semantic_memory}), and connecting it to action (Part~\ref{part:embodiment_and_action}). However, a key element of higher intelligence involves not just having beliefs, but being aware of them; not just undergoing processes, but monitoring them. This Part delves into Meta-Cognition, starting with the fundamental capability of Meta-Introspection.

Meta-introspection refers to the agent's ability to turn its cognitive faculties inward---to examine, represent, and reason about its own internal states $\phi$, structures (geometry, Null Tower origins), dynamics ($A, N_t, K$ operations), and regulatory processes (orientation status, coherence levels $\kappa$). It implies a recursive application of the framework onto itself, enabling the agent to gain knowledge about its own cognitive functioning.

Why is this capability important? It enables:
\begin{itemize}
	\item \textbf{Self-Monitoring and Error Detection:} Identifying internal inconsistencies, problematic reasoning patterns, or deviations from desired states.
	\item \textbf{Self-Correction and Adaptation:} Triggering remedial actions based on introspective findings, such as revising beliefs, adjusting strategies, or reallocating cognitive resources.
	\item \textbf{Strategic Control:} Making deliberate choices about how to think---selecting appropriate cognitive modes, managing abstraction levels, or guiding attention based on self-assessment.
	\item \textbf{Deeper Understanding and Explainability:} Generating justifications for actions or beliefs based not just on external factors, but on an awareness of the internal processes that led to them.
\end{itemize}
This chapter explores how meta-introspection might be realized within the semantic state space $\Phi$, defining the mechanisms that gather internal information ($\phi_{introspective}$) which can then be integrated via the Meta-Assimilation operator $M$ (Chapter~\ref{chap:MetaAssimilation}) to support self-awareness and recursive self-regulation.

\section{Representing Meta-Beliefs in \texorpdfstring{$\Phi$}{Phi}}

For an agent to introspect, it must be able to represent beliefs about its own beliefs, states, and processes. How can such meta-beliefs be encoded within the linguistic ensemble structure of $\Phi$?
\begin{itemize}
	\item \textbf{The Reflective Sector ($\Sigma_{refl}$):} Meta-beliefs are naturally associated with the reflective sector, $\Sigma_{refl}$. This sector, already implicated in self-modeling (Chapter~\ref{chap:EpistemicIdentity}) and reflective control (e.g., gating execution, Chapter~\ref{chap:SemanticExecution}), would be the primary locus for storing explicit representations of the agent's internal status.
	\item \textbf{Higher Abstraction Levels ($\Phi^{(k)}$):} Beliefs about one's own cognitive state are inherently abstract. Meta-beliefs like "My reasoning seems circular" or "Confidence in belief $\varphi_x$ is low" likely reside at higher levels of abstraction ($k \ge 1$ or $k \ge 2$) within $\Sigma_{refl}$.
	\item \textbf{Mechanisms for Reference/Quotation:} The linguistic expressions $\varphi_i$ forming meta-beliefs must have ways to refer to other parts of the belief state $\phi$ or to the state itself. This might involve:
	\begin{itemize}
		\item Direct Quotation: e.g., "The belief stated as 'The door is locked' conflicts with perception."
		\item Symbolic Handles/Labels: Assigning temporary or persistent labels to belief fragments or sub-states for reference.
		\item Structural References: Referring to beliefs based on their properties (e.g., "beliefs within $\Sigma^{(1)}_{plan}$ concerning goal G").
		\item State Descriptors: Expressions summarizing aggregate properties (e.g., "$\kappa(\phi)$ is below threshold," "$\theta(\phi, \vec{v}_{goal})$ is increasing").
	\end{itemize}
\end{itemize}
The capacity to form these self-referential or state-descriptive expressions within $\Phi$ is a prerequisite for meta-introspection.

\section{Mechanisms for Introspection}

How does the agent gather the information needed to form meta-beliefs? Introspection requires mechanisms to "read" or sense the agent's own internal state and processes. These mechanisms generate the input $\phi_{introspective}$ that is subsequently integrated by the Meta-Assimilation operator $M$ (Chapter~\ref{chap:MetaAssimilation}). 

\begin{figure}[ht]
	\centering
	\input{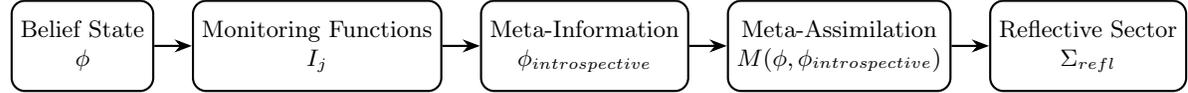}
	\caption{Meta-introspection dataflow. Internal monitoring functions ($I_j$) access properties of the belief state ($\phi$) to generate structured meta-information ($\phi_{introspective}$), which is subsequently integrated via the Meta-Assimilation operator ($M$) into the reflective sector ($\Sigma_{refl}$). This process enables the agent to form explicit meta-beliefs supporting self-awareness and self-regulation.}
	\label{fig:meta_introspection_dataflow}
\end{figure}

Key mechanisms include:
\begin{itemize}
	\item \textbf{Accessing State Properties:} The agent's architecture must provide ways to query quantitative and qualitative properties of its current state $\phi$. This could include accessing:
	\begin{itemize}
		\item Sector activations ($\alpha_i(t)$) and belief densities ($\lambda(\Sigma^{(k)}, \phi)$).
		\item Overall or sectoral coherence scores ($\kappa(\phi), \kappa(\Sigma, \phi)$).
		\item Anchoring strengths ($a_i$) or persistence levels ($d_i(t)$) of specific beliefs.
		\item Current abstraction profile (distribution across $\Phi^{(k)}$).
		\item Semantic Orientation status (current deviation $\theta$, residual $r$).
		\item Proximity to activation basin boundaries ($\partial \mathcal{A}_a$).
		\item Current cognitive load ($\lambda$) or semantic effort ($\epsilon$) estimates.
	\end{itemize}
	\item \textbf{Logging and Tracing:} Maintaining accessible records or traces of recent operations ($A, N_t, K, \Lambda, V, M$ events), belief trajectories ($\gamma(t)$), or decision processes, perhaps stored within $\Sigma_{narr}$ or a dedicated meta-cognitive log sector.
	\item \textbf{Internal Monitoring Functions ($I_j$):} As conceptualized previously, dedicated functions $I_j : \Phi \rightarrow \text{MetaInfo}_j$ could actively compute specific meta-information (e.g., $I_{\kappa}$ computes coherence, $I_{\lambda}$ computes load). The outputs ($\text{MetaInfo}_j$) are structured into the ensemble $\phi_{introspective}$.
\end{itemize}

Meta-introspection is thus an active process involving sensing internal state parameters, accessing process histories, and generating structured meta-information $\phi_{introspective}$ suitable for integration via the Meta-Assimilation operator $M$.

\section{Scope of Meta-Introspection}

What aspects of its own cognitive life can an agent potentially introspect upon using these mechanisms? The scope could encompass:
\begin{itemize}
	\item \textbf{Belief Content:} Examining specific beliefs $\varphi_i$, their source (if tagged), their certainty, or their grounding.
	\item \textbf{Belief Structure:} Assessing the coherence ($\kappa$) of $\phi$ or its sectors, the distribution across the semantic geometry ($\Sigma, k$), the density of connections, or the stability of anchors ($a_i$).
	\item \textbf{Cognitive Dynamics:} Monitoring the rate and type of assimilation ($A$), the decay rate of beliefs ($N_t$), the occurrence of annihilation ($K$), the characteristics of current semantic flow ($F$), or the properties of recent trajectories ($\gamma(t)$).
	\item \textbf{Regulatory Status:} Checking alignment with orientation axes ($\theta, r$), assessing the current cognitive mode ($\vec{\alpha}(t)$), evaluating proximity to activation basins ($\mathcal{A}_a$), or querying gauge equivalence status ($\sim_{gauge}$).
	\item \textbf{Resource Status:} Estimating current cognitive load ($\lambda$) and semantic effort ($\epsilon$) levels.
	\item \textbf{Performance and Goals:} Comparing current state or predicted outcomes (from Embodied Simulation, Chapter~\ref{chap:EmbodiedSimulation}) against goals or performance metrics stored in $\Sigma_{refl}$ or $\Sigma_{plan}$.
	\item \textbf{Epistemic Identity:} Examining the content and stability of self-models, core values, or narrative history residing primarily in $\Sigma_{refl}$ and $\Sigma_{narr}$ (Chapter~\ref{chap:EpistemicIdentity}).
\end{itemize}

\begin{table}[ht]
	\centering
	\begin{tabular}{@{} l l @{}}
		\toprule
		\textbf{Introspectible Aspect} & \textbf{Example Properties / Metrics} \\
		\midrule
		Belief Content     & Specific $\varphi_i$, certainty, source, grounding \\
		Belief Structure   & Coherence ($\kappa$), density / complexity ($\lambda$), anchoring ($a_i$) \\
		Cognitive Dynamics & Operator rates (A, $N_t$, $K$), trajectory ($\gamma(t)$), flow ($F$) \\
		Regulatory Status  & Orientation ($\theta$, $r$), mode ($\vec{\alpha}(t)$), basin proximity \\
		Resource Status    & Cognitive load ($\lambda$), semantic effort ($\epsilon$) \\
		Performance / Goals& Goal progress, prediction accuracy \\
		Epistemic Identity & Self-model content, narrative stability, value alignment \\
		\bottomrule
	\end{tabular}
	\caption{Examples of properties accessible via meta-introspection.}
	\label{tab:introspection_properties}
\end{table}

The depth and breadth of introspection depend on the sophistication of the agent's state access mechanisms ($I_j$) and its capacity for forming complex meta-beliefs via Meta-Assimilation ($M$).

\section{Role in Self-Regulation and Control}

Meta-introspection is not merely passive self-observation; it is a crucial component of effective self-regulation and adaptive control. By generating meta-information ($\phi_{introspective}$) and integrating it via $M$ to form meta-beliefs, the agent can:
\begin{itemize}
	\item \textbf{Detect Errors and Anomalies:} Meta-beliefs like "Contradiction detected between $\varphi_x$ and $\varphi_y$" or "Semantic drift $\theta$ exceeds threshold" signal problems requiring intervention.
	\item \textbf{Initiate Self-Correction:} Introspective findings, represented as meta-beliefs in $\Sigma_{refl}$, can trigger specific remedial actions managed by regulatory policies ($\pi_{regulate}$):
	\begin{itemize}
		\item Invoking Corrective Assimilation ($A_{corr}$) to resolve object-level conflicts.
		\item Applying Targeted Nullification ($N^{\Sigma}_t$) to prune irrelevant sectors.
		\item Using Selective Annihilation ($K_{\Sigma}$) to reset problematic subsystems.
		\item Engaging Re-orientation manoeuvres using the Semantic Compass (Chapter~\ref{chap:SemanticOrientation}).
	\end{itemize}
	\item \textbf{Modulate Cognitive Strategy:} Self-assessment based on meta-beliefs can inform strategic choices. If introspection reveals high cognitive load ($\lambda$, Chapter~\ref{chap:CognitiveLoad}) or low coherence ($\kappa$), the agent might decide (via $\pi_{regulate}$) to:
	\begin{itemize}
		\item Shift to a less demanding cognitive mode (adjust $\vec{\alpha}(t)$).
		\item Increase abstraction level (apply $\Lambda$) to simplify.
		\item Allocate more Semantic Effort ($\epsilon$, Chapter~\ref{chap:SemanticEffort}) to deliberation or coherence restoration (via $\pi_{effort}$, Chapter~\ref{chap:SemanticFocus}).
		\item Adjust learning parameters or exploration strategies (Part~\ref{part:learning_and_adaptation}).
	\end{itemize}
\end{itemize}

\begin{figure}[ht]
	\centering
	\input{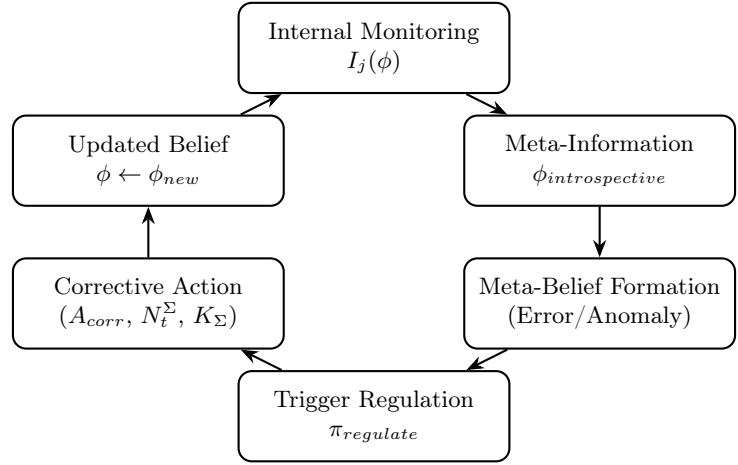}
	\caption{Meta-introspective error detection and correction loop. Internal monitoring functions ($I_j$) access properties of the belief state ($\phi$) to generate structured meta-information ($\phi_{introspective}$). Meta-beliefs about errors or anomalies are formed, triggering regulatory policies ($\pi_{regulate}$) that select appropriate corrective actions (such as $A_{corr}$, $N_t^\Sigma$, or $K_\Sigma$). These actions update the belief state ($\phi \leftarrow \phi_{new}$), closing the self-regulatory loop.}
	\label{fig:meta_introspection_error_correction}
\end{figure}

Meta-introspection, by providing input for $M$, closes the loop between cognitive processing and cognitive control, allowing the agent to adapt its internal operations based on an ongoing assessment of its own performance and state.

\section{Challenges and Limitations}

Implementing effective meta-introspection faces significant challenges:
\begin{itemize}
	\item \textbf{Computational Cost:} Constantly monitoring and representing one's own state can be computationally expensive, potentially consuming resources needed for primary tasks. Defining efficient monitoring functions ($I_j$) is crucial.
	\item \textbf{Accuracy and Bias:} Introspection is not guaranteed to be accurate. Agents might misinterpret their own states, confabulate explanations, or exhibit biases in self-assessment, just as humans do. The mechanisms $I_j$ and $M$ might introduce noise or systemic errors.
	\item \textbf{Infinite Regress:} The possibility of meta-meta-beliefs ("I believe that I believe that I am confused...") raises concerns about infinite regress, although practical limits likely arise from resource constraints ($\lambda, \epsilon$) or diminishing returns in regulatory utility.
	\item \textbf{Representation Limits:} How can highly complex, dynamic states $\phi$ or subtle trajectory properties be adequately captured in finite, linguistic meta-beliefs assimilated by $M$? There may be aspects of cognition that are inherently difficult to introspect accurately.
	\item \textbf{Intervention Paradox:} The act of introspecting might itself alter the state being introspected upon, requiring careful design of non-invasive monitoring ($I_j$) where possible.
\end{itemize}
Designing mechanisms ($I_j$) that provide useful introspective insight ($\phi_{introspective}$) for $M$ without incurring prohibitive costs or introducing significant distortions remains an open research area.

\section{Conclusion: Towards Reflective Agency}

Meta-introspection represents the capacity of a semantic agent to apply its representational and reasoning abilities recursively to its own internal cognitive landscape, $\Phi$. It involves mechanisms ($I_j$) for accessing internal state properties and process histories, generating meta-information ($\phi_{introspective}$). This information is then integrated via the dedicated Meta-Assimilation operator ($M$, Chapter~\ref{chap:MetaAssimilation}) to form explicit meta-beliefs, primarily within the reflective sector $\Sigma_{refl}$.

This capability allows the agent potential for genuine self-awareness and self-regulation. By monitoring its own functioning, detecting errors or anomalies via meta-beliefs, the agent can initiate corrective or adaptive actions. While facing challenges related to cost, accuracy, and representation, meta-introspection is a cornerstone capability for moving beyond reactive or purely deliberative systems towards agents possessing greater autonomy, robustness, explainability, and potentially richer forms of agency. It provides the necessary self-knowledge foundation upon which the subsequent chapters on trajectory awareness, cognitive load, and semantic effort will build.


\subsection*{Chapter Summary}
This chapter introduces Meta-Introspection, a core meta-cognitive capability enabling an agent to examine, represent, and reason about its own internal states ($\phi$), structures (geometry, sectors $\Sigma$, layers $k$), and dynamics (operators $A$, $N_t$, $K$, etc.). It involves internal mechanisms, conceptualized as monitoring functions ($I_j$), that access properties like coherence ($\kappa$), load ($\lambda$), orientation ($\theta$), resource usage ($\epsilon$), and trajectory history ($\gamma(t)$). These mechanisms generate structured meta-information ($\phi_{introspective}$), which is then integrated into the belief state, typically within the reflective sector ($\Sigma_{refl}$), via the Meta-Assimilation operator ($M$). This allows the agent to form explicit meta-beliefs about its own functioning. Meta-introspection is presented as crucial for enabling self-monitoring, error detection, adaptive self-regulation (triggering corrective actions via $\pi_{regulate}$), strategic control, and deeper explainability. While acknowledging challenges such as computational cost and potential inaccuracies, meta-introspection provides the foundation for self-awareness within the semantic manifold framework.
	\chapter{Trajectory Awareness}
\label{chap:TrajectoryAwareness}

\section{Introduction: Beyond Static Introspection}

Chapter~\ref{chap:MetaIntrospection} introduced meta-introspection as the agent's ability to examine and represent its current internal state $\phi$. However, cognition is fundamentally dynamic; belief states evolve along trajectories $\gamma(t)$ through the semantic state space $\Phi$, driven by assimilation, nullification, internal flows, and control actions. A deeper level of self-awareness involves not just knowing where the agent is in $\Phi$, but understanding how it got there and where it is heading.

This chapter introduces Trajectory Awareness, defined as the agent's meta-cognitive capacity to represent, monitor, analyze, and potentially predict the path of its own belief state $\gamma(t)$ through $\Phi$. While meta-introspection provides snapshots, trajectory awareness provides the movie. It encompasses an understanding of cognitive momentum, directionality, history, and future projection.

This awareness is crucial for:
\begin{itemize}
	\item \textbf{Predictive Control:} Anticipating future states and intervening proactively.
	\item \textbf{Understanding Dynamics:} Recognizing patterns like loops, oscillations, or drift in one's own thought processes.
	\item \textbf{Evaluating Reasoning Paths:} Assessing the coherence, efficiency, or stability of the trajectory taken to reach a conclusion.
	\item \textbf{Sophisticated Self-Regulation:} Adjusting cognitive strategies based on an understanding of ongoing belief dynamics.
\end{itemize}
Trajectory awareness moves beyond static self-representation towards an appreciation of the agent's own cognitive motion within its internal semantic landscape.

\section{Representing Trajectories in \texorpdfstring{$\Phi$}{Phi}}

How can an agent represent its own belief trajectory $\gamma(t)$ within its current belief state $\phi$? Since $\gamma(t)$ is a path over time, its full representation can be complex. Potential approaches include:
\begin{itemize}
	\item \textbf{Explicit Logging/Memory:} Storing sequences of recent states $\{\phi_{t-\Delta t}, \dots, \phi_t\}$ or key waypoints along the trajectory, likely within the narrative sector $\Sigma_{narr}$ or a dedicated process log. This provides a historical record.
	\item \textbf{Abstract Trajectory Properties:} Representing the trajectory not by listing points, but by encoding its current dynamic properties as meta-beliefs (likely in $\Sigma_{refl}$ at high $\Phi^{(k)}$). Examples:
	\begin{itemize}
		\item "Current semantic velocity vector is $\vec{v}_{current}$." (Approximating $F(\phi)$)
		\item "Trajectory curvature indicates a recent change in direction."
		\item "Alignment $\theta(t)$ with axis $\vec{v}_{goal}$ has been improving."
		\item "State has been oscillating within region $R$ for duration $\Delta T$."
		\item "Proximity to activation basin $\mathcal{A}_a$ is increasing rapidly."
	\end{itemize}
	\item \textbf{Flow Field Awareness:} Holding beliefs about the currently active epistemic flow field $F$ that is governing the trajectory's immediate future.
\end{itemize}
Representing trajectories likely involves a combination of storing recent history and extracting/representing key dynamic features as meta-beliefs assimilated via $M$.

\begin{figure}[htbp]
	\centering
	\input{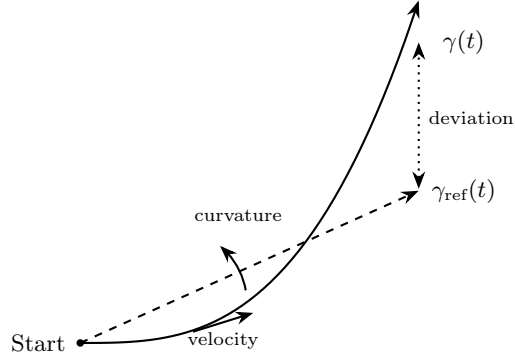}
	\caption{Semantic trajectory \(\gamma(t)\) with introspective annotations. The agent computes the local velocity (tangent vectors), curvature (rate of directional change), and deviation from a reference trajectory \(\gamma_{\text{ref}}(t)\). These metrics support introspective awareness and adaptive course correction.}
	\label{fig:trajectory-awareness}
\end{figure}

\section{Monitoring Trajectory Properties}

Building on the state-access mechanisms of meta-introspection (Chapter~\ref{chap:MetaIntrospection}), trajectory awareness requires monitoring dynamic properties derived from sequences of states or their derivatives:
\begin{itemize}
	\item \textbf{Semantic Velocity/Momentum:} Estimating the instantaneous rate and direction of change, $\vec{v}(t) \approx \frac{d\gamma}{dt} \approx F(\phi(t))$ or $\phi(t) - \phi(t - \Delta t)$. This indicates the current "cognitive momentum."
	\item \textbf{Semantic Acceleration/Curvature:} Detecting changes in velocity, $\frac{d^2\gamma}{dt^2}$. High curvature might signal a shift in reasoning focus, a response to new input, or entry into a region with a different flow field $F$.
	\item \textbf{Alignment Dynamics (Orientation):} Monitoring the rate of change of orientation metrics like $\theta(t)$ and $r(t)$ (from Chapter~\ref{chap:SemanticOrientation}). Is the agent aligning towards ($\frac{d\theta}{dt} < 0$) or drifting away from ($\frac{d\theta}{dt} > 0$) a reference axis?
	\item \textbf{Recurrence and Cycle Detection:} Analyzing the recent path $\{\phi_{t-\Delta t}, \dots, \phi_t\}$ to detect if the agent is revisiting states or entering repetitive loops (potentially indicating unproductive rumination or stable cognitive rhythms).
	\item \textbf{Boundary Proximity Rates:} Monitoring not just the distance to boundaries like $\partial \mathcal{A}_a$ or $\partial S_a$, but the rate of approach or departure.
	\item \textbf{Region Occupancy Dynamics:} Tracking how long the trajectory has spent recently within certain sectors $\Sigma$ or abstraction levels $\Phi^{(k)}$, indicating sustained focus or processing modes.
\end{itemize}
These monitored properties are assimilated via $M$ into meta-beliefs, forming the agent's explicit awareness of its own cognitive motion.

\section{Predicting Future Trajectories}

A key benefit of trajectory awareness is the ability to predict, however imperfectly, the future course of one's own belief state. This predictive capability can arise from:
\begin{itemize}
	\item \textbf{Trajectory Extrapolation:} Projecting the current trajectory $\gamma(t)$ forward based on its estimated current velocity $\vec{v}(t)$ and possibly acceleration. This provides short-term prediction assuming the governing dynamics $F$ remain constant.
	\item \textbf{Flow Field Simulation:} If the agent has a representation or estimate of the current flow field $F$, it can simulate the trajectory forward by integrating $\frac{d\gamma}{dt} = F(\gamma(t))$ from the current state $\phi(t)$.
	\item \textbf{Embodied Simulation Integration:} Leveraging the mechanisms from Chapter~\ref{chap:EmbodiedSimulation}, but initiated from the current actual state $\phi(t)$ and potentially biased by the current estimated velocity $\vec{v}(t)$, to generate more detailed predictions of future states resulting from planned internal operations or expected inputs.
	\item \textbf{Goal-Based Prediction:} Assessing whether the predicted trajectory (from extrapolation or simulation) leads towards or away from desired goal regions $\mathcal{T}_{goal} \subset \Phi$. Does the current path lead to a solution state or an undesirable state (e.g., high incoherence, suppression surface $S_a$)?
\end{itemize}
Predictive capabilities allow the agent to anticipate problems, evaluate the likely success of current reasoning paths, and make proactive adjustments.

\section{Trajectory Awareness in Control and Regulation}

Awareness of its own trajectory provides powerful leverage for self-regulation and control:
\begin{itemize}
	\item \textbf{Proactive Intervention:} By predicting that $\gamma(t)$ is heading towards an undesirable region (e.g., a suppression surface $S_a$, a region of low coherence $\kappa$, excessive drift $\theta > \tau_{\theta}$), the agent can intervene before reaching it, applying corrective forces or changing cognitive strategy.
	\textbf{Example:} Predicting entry into suppression surface $S_{unsafe}$ associated with action $a$ leads the agent to proactively inhibit $a$ or modify its plan before the unsafe state is reached.
	\item \textbf{Steering and Course Correction:} If monitoring reveals deviation from a desired path or orientation axis, the agent can use its understanding of current momentum and flow $F$ to apply targeted control actions (e.g., selective assimilation $A$, focused reflection $M$, or perhaps even targeted $N_t$ or $K_{\Sigma}$) designed to "steer" $\gamma(t)$ back on course. This is a dynamic extension of Semantic Orientation.
	\textbf{Example:} Detecting increasing angular deviation $\theta(t)$ from the goal axis $\vec{v}_{goal}$ triggers application of corrective projection operations to realign the trajectory.
	\item \textbf{Optimizing Cognitive Paths:} By evaluating the properties of past or simulated trajectories (e.g., length, time taken, coherence maintained, proximity to desired states), the agent can learn to prefer certain reasoning strategies, cognitive modes (sector activation profiles $\vec{\alpha}(t)$), or abstraction levels that lead to more efficient, stable, or successful trajectories towards its goals.
	\textbf{Example:} Recognizing that a past trajectory involved unproductive oscillation within $\Sigma_{plan}$ leads the agent to adopt a different planning strategy (e.g., increased abstraction $\Lambda$) next time.
	\item \textbf{Enhanced Explainability:} An agent with trajectory awareness can explain its conclusions or actions not just based on its final state $\phi(T)$, but by recounting the trajectory $\gamma(t)$ that led there, including detected drifts, course corrections, or evaluations made along the way.
\end{itemize}
Trajectory awareness enables a shift from purely reactive control (responding to the current state) to predictive and proactive control (responding to the anticipated evolution of the state).

\section{Challenges of Trajectory Awareness}

Realizing trajectory awareness faces several hurdles:
\begin{itemize}
	\item \textbf{Representational Cost:} Storing detailed trajectory histories $\{\phi_t\}$ can be highly memory-intensive. Abstracting trajectory properties helps, but risks losing crucial information.
	\item \textbf{Computational Cost:} Continuously monitoring, analyzing, and predicting trajectory properties can be computationally expensive, potentially interfering with primary cognitive tasks.
	\item \textbf{Prediction Accuracy:} Predicting future belief states accurately is difficult, especially in open environments where unexpected inputs ($X(s)$) can drastically alter trajectories. Internal flow fields $F$ may also change unpredictably.
	\item \textbf{Interpreting Dynamics:} Identifying meaningful patterns (loops, drift, oscillations) amidst the complex evolution of a high-dimensional belief state requires sophisticated analysis capabilities within the agent's meta-cognitive apparatus.
\end{itemize}
Developing efficient representations and monitoring mechanisms for trajectory awareness is a significant challenge.

\section{Conclusion: Understanding Cognitive Motion}

Trajectory awareness represents a sophisticated form of meta-cognition, extending static self-knowledge (meta-introspection) to encompass the dynamics of belief evolution over time. By representing, monitoring, and predicting its own belief trajectory $\gamma(t)$ through the semantic state space $\Phi$, the agent gains crucial capabilities for proactive self-regulation, predictive control, and dynamic strategy adjustment.

Awareness of semantic velocity, acceleration, curvature, alignment dynamics, and boundary proximity allows the agent to anticipate problems, steer its cognitive course more effectively using mechanisms like Semantic Orientation, optimize its reasoning paths, and generate richer explanations for its behavior. While computationally demanding, trajectory awareness is likely essential for achieving robust, adaptive, and deeply reflective agency. It transforms the agent from a passive occupant of its belief space into an active navigator of its own internal landscape, capable of understanding not just where it is, but where it is going and why.


\subsection*{Chapter Summary}
This chapter introduces Trajectory Awareness as a meta-cognitive capability extending beyond static self-assessment (Meta-Introspection) to encompass the dynamics of belief evolution. It is defined as the agent's ability to represent, monitor, analyze, and potentially predict the path, or trajectory ($\gamma(t)$), of its own belief state through the semantic manifold $\Phi$. This involves representing trajectories via recent state history or abstract dynamic properties (like velocity, curvature, alignment $\theta(t)$), monitoring these properties through internal mechanisms, and potentially predicting future states via extrapolation or simulation. Trajectory awareness enables more sophisticated self-regulation, allowing for proactive interventions based on predicted future states (e.g., avoiding undesirable regions), dynamic course correction or steering of the cognitive trajectory, optimization of reasoning strategies based on path evaluation, and richer explanations of internal processes. While computationally challenging, trajectory awareness facilitates a shift towards predictive and adaptive cognitive control.
	\chapter{Cognitive Load}
\label{chap:CognitiveLoad}

\section{Introduction: The Burden of Thought}

Cognitive processes, whether in biological or artificial systems, operate under resource constraints. Attention is limited, processing capacity is finite, and maintaining complex internal states requires effort. The concept of cognitive load attempts to capture the level of demand placed on these limited resources by ongoing mental activity.

While previous chapters focused on the content, structure, and dynamics of belief ($\Phi, \Sigma, k, A, N_t, K$) and the agent's awareness of them (Meta-Introspection, Trajectory Awareness), this chapter addresses the crucial question of cognitive burden. How can we conceptualize and potentially measure cognitive load within the semantic state space framework? What are the consequences of excessive load, and how might an agent equipped with meta-cognitive capabilities (Part~\ref{part:meta_cognition}) monitor and manage its own cognitive burden?

We propose that cognitive load is not just an external measure of resource usage (like CPU cycles) but can be related intrinsically to the properties and dynamics of the belief state $\phi$ itself. Understanding cognitive load is essential for designing agents that are not only capable but also efficient, robust under pressure, and able to strategically allocate their finite cognitive resources. It is distinct from, though related to, Semantic Effort ($\epsilon$), which represents the active application of resources (detailed in Chapter~\ref{chap:SemanticEffort}).

\section{Defining and Measuring Cognitive Load in \texorpdfstring{$\Phi$}{Phi}}

Operationalizing cognitive load ($L$) within the semantic state space $\Phi$ requires identifying properties of the belief state $\phi$ and its dynamics that correlate with processing demands. We propose several potential contributors or indicators of load, often represented by the metric $\lambda$:
\begin{itemize}
	\item \textbf{Belief State Complexity/Size:} The sheer volume or complexity of the currently active belief state $\phi$. Metrics could include:
	\begin{itemize}
		\item The number of active expressions $\{\varphi_i\}$ in $\phi$.
		\item The total belief density or activation mass $\lambda(\Phi, \phi)$.
		\item Structural complexity measures (e.g., connectivity or depth if $\phi$ is modeled as a graph).
	\end{itemize}
	Larger, denser, or more complex states likely require more resources to maintain and process.
	\item \textbf{Sectoral Load Distribution:} Load might be concentrated in specific semantic sectors $\Sigma$. High activation $\lambda(\Sigma, \phi)$ in demanding sectors like $\Sigma_{refl}$ (complex self-analysis) or $\Sigma_{plan}$ (complex planning) could indicate high specific load. The overall load might be a function of loads across multiple active sectors ($\vec{\alpha}(t)$).
	\item \textbf{Abstraction Span and Depth:} Cognitive load might relate to the range and depth of abstraction levels $\Phi^{(k)}$ currently active. Simultaneously maintaining highly detailed grounded beliefs ($\Phi^{(0)}$) and highly abstract principles ($\Phi^{(k)}, k \gg 0$) might be more demanding than operating primarily at a single level.
	\item \textbf{Coherence Strain ($\kappa$):} The effort required to maintain internal coherence $\kappa(\phi)$. Load might increase when $\phi$ contains significant internal tension, unresolved contradictions, or requires frequent corrective assimilation ($A_{corr}$). Maintaining coherence under conflicting information is resource-intensive.
	\item \textbf{Dynamic Intensity:} Load related to the rate and complexity of belief dynamics:
	\begin{itemize}
		\item High rate of assimilation ($A$) of complex inputs.
		\item Rapid or complex belief trajectories $\gamma(t)$ with high curvature or frequent state changes.
		\item Frequent or demanding mode switching (changes in $\vec{\alpha}(t)$).
		\item Ongoing, computationally expensive internal processes like detailed embodied simulation (Chapter~\ref{chap:EmbodiedSimulation}).
	\end{itemize}
	\item \textbf{Meta-Cognitive Overhead:} The processes of meta-introspection (Chapter~\ref{chap:MetaIntrospection}) and trajectory awareness (Chapter~\ref{chap:TrajectoryAwareness}) themselves consume resources and contribute to overall load.
\end{itemize}

\begin{figure}[ht]
	\centering
	\input{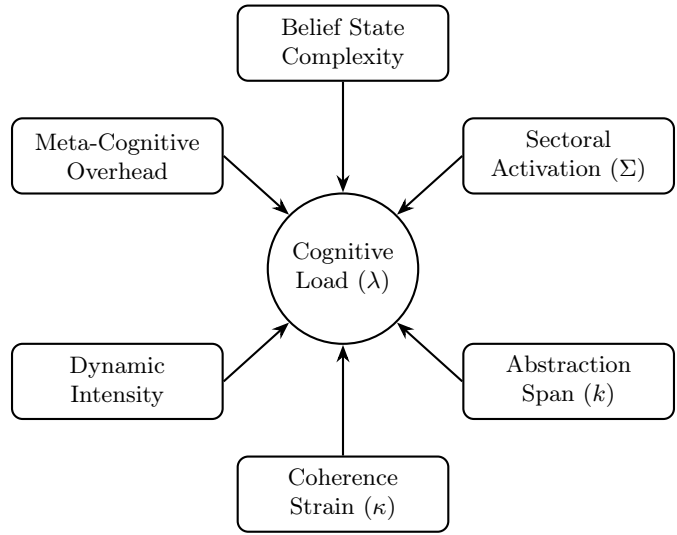}
	\caption{Sources contributing to cognitive load within the semantic state space. Load ($\lambda$) arises from belief complexity, sectoral activation, abstraction span, coherence strain, dynamic intensity, and meta-cognitive overhead.}
	\label{fig:cognitive_load_sources}
\end{figure}

A comprehensive measure of cognitive load, potentially denoted $L(\phi, \frac{d\phi}{dt}, \dots)$, would likely be a function of several such factors, reflecting both the static complexity of the belief state and the intensity of the ongoing dynamic processes. The metric $\lambda$ often serves as a primary proxy for this load.

\section{Consequences of High Cognitive Load}

When cognitive load $L$ exceeds an agent's processing capacity or available resources ($E_{total}$ related to $\epsilon$), predictable consequences are likely to emerge, mirroring effects seen in overloaded human or computational systems:
\begin{itemize}
	\item \textbf{Performance Degradation:} Slower reaction times, reduced accuracy in inference or decision-making, inability to perform complex reasoning chains.
	\item \textbf{Coherence Breakdown:} Increased likelihood of inconsistencies or contradictions arising within $\phi$ as maintenance mechanisms fail. This might manifest as a measurable drop in coherence $\kappa(\phi)$ or structural fragmentation.
	\item \textbf{Failure to Assimilate:} Inability to properly process and integrate new information via $A$. Input might be ignored, partially processed, or integrated incoherently.
	\item \textbf{Increased Reliance on Heuristics/Defaults:} Shifting from effortful deliberation towards simpler, less resource-intensive strategies or default behaviors.
	\item \textbf{Triggering Simplification Dynamics:} The system might reflexively invoke load-reducing dynamics:
	\begin{itemize}
		\item Increased rate of Nullification ($N_t$) to prune less critical beliefs.
		\item Forced Abstraction ($\Lambda$) to compress information, potentially losing important detail.
		\item In extreme cases, sectoral Annihilation ($K_{\Sigma}$) might occur, shutting down demanding cognitive functions (e.g., reflection, complex planning).
	\end{itemize}
	\item \textbf{Cognitive Mode Collapse:} Inability to maintain complex sector activation profiles ($\vec{\alpha}(t)$), potentially collapsing into simpler, more primitive modes (e.g., purely reactive).
\end{itemize}
Understanding these failure modes is crucial for designing agents that can gracefully handle high-load situations rather than suffering catastrophic collapse.

\section{Meta-Cognitive Awareness of Load}

For an agent to actively manage its cognitive load, it must first possess meta-cognitive awareness of its current load level. Building on Chapter~\ref{chap:MetaIntrospection}, this involves:
\begin{itemize}
	\item \textbf{Representing Load:} Encoding the estimated current load level $L$ (derived from metrics like $\lambda$ discussed in Section 2) as a meta-belief within $\Sigma_{refl}$, potentially via Meta-Assimilation $M$. For example, "Cognitive load estimate $\lambda$ is currently high (0.85)."
	\item \textbf{Monitoring Load Indicators:} Actively tracking metrics like belief state complexity, coherence $\kappa$, processing speed, or error rates as indirect indicators of load using introspection functions ($I_j$).
	\item \textbf{Learning Load Models:} Potentially learning predictive models that estimate upcoming load based on task descriptions, environmental complexity, or anticipated processing requirements.
	\item \textbf{Awareness of Capacity Limits:} Having a representation (perhaps in $\Sigma_{refl}$) of its own approximate cognitive capacity limits ($E_{total}$), allowing it to compare estimated load $L$ (or $\lambda$) against these limits.
\end{itemize}
This self-awareness of load allows the agent to transition from passive responses to overload towards proactive load management strategies.

\section{Load Management Strategies}

An agent aware of its cognitive load can employ various strategies, often mediated by meta-cognitive control originating from $\Sigma_{refl}$ (e.g., policy $\pi_{regulate}$), to keep load within manageable limits:
\begin{itemize}
	\item \textbf{Controlled Information Pruning:} Intentionally increasing the rate of Nullification ($N_t$), perhaps targeted at specific low-priority sectors ($N^{\Sigma_{low-priority}}_t$), or even selectively applying Annihilation ($K_{\Sigma}$) to rapidly shed load associated with non-critical functions.
	\item \textbf{Strategic Abstraction:} Actively applying the abstraction operator $\Lambda$ to compress detailed information into more manageable higher-level representations ($\Phi^{(k)}$), accepting potential information loss for reduced load.
	\item \textbf{Cognitive Mode Regulation:} Deliberately shifting the sector activation profile $\vec{\alpha}(t)$ to less demanding modes, e.g., reducing activation of $\Sigma_{refl}$ or complex planning sectors during high perceptual load.
	\item \textbf{Task Prioritization and Scheduling:} Modifying goals or plans within $\Sigma_{plan}$ to defer or simplify demanding sub-tasks when load is high.
	\item \textbf{Simplification of Reasoning:} Opting for less complex reasoning strategies or heuristics when under load.
	\item \textbf{Semantic Offloading:} Explicitly deciding to store complex information externally or rely on external computational resources rather than maintaining it entirely within $\phi$.
	\item \textbf{Resource Allocation (Semantic Effort):} Directly modulating the allocation of internal processing resources ($\epsilon$) based on load levels and task priorities (via $\pi_{effort}$, discussed in Chapter~\ref{chap:SemanticFocus}).
\end{itemize}

\begin{figure}[ht]
	\centering
	\input{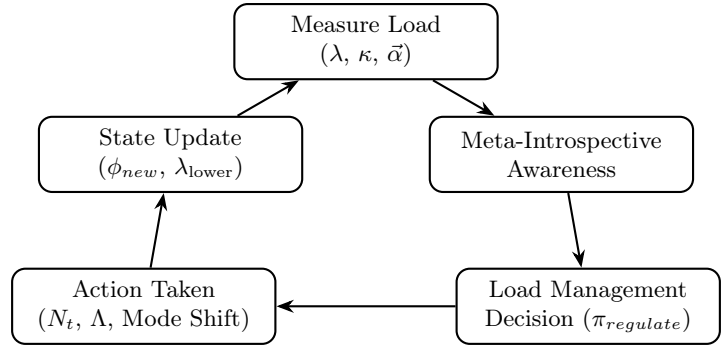}
	\caption{Feedback loop for cognitive load regulation. Measured load triggers meta-introspective awareness, guiding management decisions that produce actions modifying the belief state, closing the regulatory loop.}
	\label{fig:cognitive_load_feedback}
\end{figure}

Effective load management involves dynamically balancing task demands with available cognitive capacity using these (and potentially other) strategies, guided by meta-cognitive self-assessment.

\section{Distinguishing Load from Effort}
\label{sec:LoadVsEffort}

It remains crucial to distinguish cognitive load ($L$, often proxied by $\lambda$) from semantic effort ($\epsilon$, discussed in Chapter~\ref{chap:SemanticEffort}):
\begin{itemize}
	\item \textbf{Cognitive Load ($L, \lambda$):} Represents the \textit{demand} imposed on the agent's cognitive system by the current belief state $\phi$, its complexity, the ongoing dynamics, and the task requirements. It is a measure of how much processing is needed.
	\item \textbf{Semantic Effort ($\epsilon$):} Represents the finite cognitive resource \textit{actively allocated and expended} by the agent to meet those demands, perform operations ($A, M, \Lambda, V, R$, etc.), maintain coherence ($\kappa$), or achieve goals. It is a measure of the processing intensity being applied, constrained by the agent's total capacity ($E_{total}$).
\end{itemize}
Load and effort are strongly related but distinct concepts. High load often necessitates allocating high effort to cope. However, an agent might face high load but allocate low effort (leading to performance failure) or apply high effort even when load is low (e.g., during intense reflection). Conversely, applying high effort can itself increase load by generating complex intermediate states. Understanding this distinction is key: load is the challenge presented to the system, while effort is the resource deployed by the system in response.

\section{Conclusion: Load as a Regulatory Signal}

Cognitive load is an intrinsic aspect of any finite cognitive system. Within the semantic state space framework, load can be conceptualized not merely as external resource usage, but as arising from the complexity, coherence demands, and dynamic intensity associated with the agent's belief state $\phi$ and its evolution. Measuring or estimating load via properties like belief density ($\lambda$), coherence strain ($\kappa$), sectoral activation ($\vec{\alpha}(t)$), and dynamic rates provides crucial information for self-regulation.

High cognitive load leads to predictable degradations in performance and coherence, potentially triggering simplifying dynamics ($N_t, \Lambda, K_{\Sigma}$). An agent equipped with meta-introspection can become aware of its load levels and proactively employ management strategies---ranging from information pruning and abstraction to mode shifting and task rescheduling---to maintain effective operation. Cognitive load thus serves as a vital internal regulatory signal, mediating the interplay between task demands, processing capacity, and cognitive strategy, and informing the crucial allocation of semantic effort ($\epsilon$). Understanding and managing load is therefore essential for building robust, scalable, and truly adaptive semantic agents.


\subsection*{Chapter Summary}
This chapter introduces Cognitive Load ($L$, often proxied by $\lambda$) as a measure of the demand placed on an agent's finite processing resources by its current cognitive state and ongoing tasks within the semantic manifold $\Phi$. Load is conceptualized as arising intrinsically from factors such as the complexity and size of the active belief state ($\phi$), the distribution of activity across demanding semantic sectors ($\Sigma$), the span of active abstraction levels ($\Phi^{(k)}$), the strain involved in maintaining coherence ($\kappa$), the intensity of current belief dynamics (e.g., rate of $A$, complexity of $\gamma(t)$), and meta-cognitive overhead. Exceeding the agent's processing capacity leads to consequences like performance degradation, coherence breakdown, and potential reliance on simplifying heuristics or dynamics ($N_t, \Lambda, K_{\Sigma}$). Meta-cognitive awareness of load, achieved through introspection (monitoring $\lambda, \kappa,$ etc.) and integrated via Meta-Assimilation ($M$), allows the agent to employ proactive load management strategies, including information pruning, abstraction, mode shifting, and resource reallocation (effort $\epsilon$). Cognitive Load (demand) is explicitly distinguished from Semantic Effort (applied resources, Chapter~\ref{chap:SemanticEffort}), serving as a vital regulatory signal for adaptive cognition.
	\chapter{Semantic Effort}
\label{chap:SemanticEffort}

\section{Introduction: The Cost and Exertion of Cognition}

The preceding chapter explored cognitive load ($L$), representing the demand placed on the agent's processing resources by the complexity and dynamics of its belief state $\phi$. However, load only describes the demand; it does not fully capture the agent's active response. Cognitive processes like deep reasoning, careful assimilation, sustained attention, or complex simulation require not just passive capacity, but the active direction and application of limited internal resources.

This chapter introduces the concept of Semantic Effort, denoted $E$ or $\epsilon$. Semantic effort refers to the cognitive resources \textbf{actively deployed or exerted} by the agent towards specific internal processes operating within the semantic state space $\Phi$. If cognitive load $L$ is the burden imposed by the state and task, semantic effort $\epsilon$ is the intensity of processing applied to manage that burden, maintain coherence, achieve goals, or perform internal operations. It represents the \textbf{controllable investment} of the agent's finite cognitive capacity.

Understanding semantic effort is crucial for modeling the resource limitations of thought, the link between exertion and processing quality, the mechanisms underlying concentration and fatigue, and how agents might choose to engage deeply or superficially with cognitive tasks. This chapter defines semantic effort, explores its representation, discusses its direct consequences on belief dynamics and performance, and contrasts it with cognitive load. The subsequent chapter (Chapter~\ref{chap:SemanticFocus}) will delve into the complex mechanisms and policies governing how this effort is allocated and directed, relating it to concepts of focus and attention.

\section{Defining and Representing Semantic Effort (\texorpdfstring{$\epsilon$}{epsilon})}

We conceptualize semantic effort $\epsilon$ as a quantifiable internal resource that fuels cognitive operations.
\begin{itemize}
	\item \textbf{Effort as a Resource:} We assume an agent possesses a total available pool of cognitive processing capacity or "effort capacity" at any given time, $E_{total}$. Semantic effort $\epsilon$ represents the rate or intensity at which these resources are currently being consumed or applied by active cognitive processes. This could be a scalar value representing overall exertion ($\epsilon \le E_{total}$) or potentially a vector $\vec{\epsilon} = (\epsilon_1, \dots, \epsilon_m)$ representing effort applied to different concurrent processes (e.g., $\epsilon_A, \epsilon_M, \epsilon_{orient}, \epsilon_{sim}$) such that $\sum \epsilon_i \le E_{total}$.
	\item \textbf{Relation to Implementation:} While abstract, $\epsilon$ could correspond in concrete implementations to measures like allocated computational cycles, energy consumption in neuromorphic systems, depth of search in planning algorithms, or number of generative steps in LLM-based operators. The key is that it represents a finite, consumable, and potentially controllable resource tied to processing intensity.
	\item \textbf{Representation in $\Phi$:} Through meta-introspection (Chapter~\ref{chap:MetaIntrospection}) and meta-assimilation ($M$, Chapter~\ref{chap:MetaAssimilation}), the agent might represent its current effort level as meta-beliefs within $\Sigma_{refl}$. Examples: "Current overall semantic effort $\epsilon$ is high (estimated 0.8 $E_{total}$)" or "Effort allocated to coherence maintenance $\epsilon_{\kappa}$ is minimal."
\end{itemize}

Semantic effort $\epsilon$ is thus modeled as a quantifiable, limited internal resource reflecting the intensity of active cognitive engagement, distinct from the passive demand represented by load $L$ or $\lambda$.

\section{Allocation of Effort (Preview)}

While this chapter focuses on the nature of the effort resource $\epsilon$, the crucial question of how this finite resource is strategically distributed across competing cognitive processes and targets within $\Phi$ is fundamental to executive control and attention. This involves complex allocation policies ($\pi_{effort}$) informed by goals, load, coherence, and other meta-cognitive assessments. These critical mechanisms of resource allocation, control, and their relation to semantic focus are deferred for detailed exploration in the next chapter (Chapter~\ref{chap:SemanticFocus}).

\section{Consequences of Effort Exertion Level}

The amount of semantic effort $\epsilon$ allocated to and consumed by a specific cognitive process directly influences its execution and impact:
\begin{itemize}
	\item \textbf{Processing Quality and Speed:} Increased effort $\epsilon_O$ applied to an operator $O$ (like $A, M, \Lambda, V, R$) generally leads to improved performance characteristics, such as deeper analysis, more thorough search, higher fidelity simulation, more robust regulation, faster convergence, or generation of more detailed/coherent output. Conversely, low effort may lead to superficial processing or heuristic application.
	\item \textbf{Counteracting Nullification ($N_t$):} Sustained semantic effort directed towards maintaining specific beliefs $\varphi_i$ or structures (e.g., active rehearsal or processing within $\Sigma_{refl}$) can actively boost their anchoring $a_i$ or refresh their persistence $d_i(t)$, thus counteracting the passive decay modelled by $N_t$. Effort sustains information against forgetting.
	\item \textbf{Resource Depletion and Fatigue:} The total effort capacity $E_{total}$ is finite. Sustained high effort exertion ($\epsilon \approx E_{total}$) depletes this underlying resource, potentially leading to a state of "cognitive fatigue" where $E_{total}$ is temporarily reduced, forcing a subsequent reduction in applied effort $\epsilon$.
	\item \textbf{Impact on Cognitive Load ($L$):} Applying high effort can itself contribute to cognitive load. Intensive computations or simulations generate complex intermediate states that increase the demands ($L$) on the system, creating a feedback loop between effort and load.
\end{itemize}
The level of semantic effort exerted thus fundamentally modulates the depth, effectiveness, and sustainability of cognitive operations.

\section{Semantic Effort versus Cognitive Load}

As highlighted in Chapter~\ref{chap:CognitiveLoad}, it is crucial to distinguish semantic effort ($\epsilon$) from cognitive load ($L$, often proxied by $\lambda$):
\begin{itemize}
	\item \textbf{Load ($L, \lambda$):} An assessment of the \textit{demand} imposed by the current state $\phi$, dynamics, and task complexity.
	\item \textbf{Effort ($\epsilon$):} The \textit{resource actively allocated} by the agent to meet demands and perform operations, constrained by $E_{total}$.
\end{itemize}

\begin{figure}[ht]
	\centering
	\input{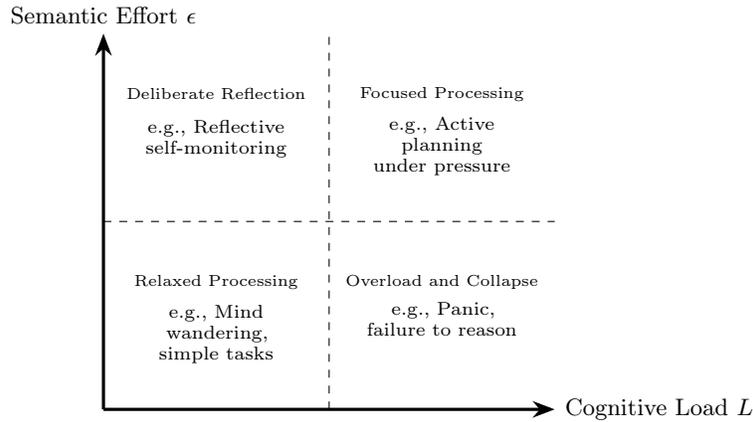}
	\caption{Semantic Load vs. Semantic Effort space. Different regions illustrate how the agent's performance varies depending on the relationship between cognitive load $L$ (demand) and semantic effort $\epsilon$ (applied resource).}
	\label{fig:semantic_load_vs_effort}
\end{figure}

They are coupled but distinct: high $L$ often necessitates high $\epsilon$, insufficient $\epsilon$ relative to $L$ leads to overload consequences (Section 3, Chapter~\ref{chap:CognitiveLoad}), and high $\epsilon$ can contribute to $L$. Load is the challenge; effort is the applied (and limited) resource.

\section{Control of Effort Allocation (Preview)}

The meta-cognitive processes that determine how much effort $\epsilon$ to exert and where to direct it are central to strategic cognitive control. These control policies ($\pi_{effort}$) operate based on introspective assessments and goals, as will be detailed in Chapter~\ref{chap:SemanticFocus}.

\section{Conclusion: Effort as the Currency of Cognition}

Semantic effort, $\epsilon$, represents the active deployment of finite cognitive resources within the Semantic Manifold framework. It is the "currency" spent to perform demanding operations, maintain information against decay, and achieve cognitive goals. Distinct from cognitive load (the demand), effort quantifies the intensity of processing and directly influences the quality, speed, and sustainability of thought.

This chapter has defined the nature of this crucial resource and its immediate consequences. While the strategic allocation of effort ($\pi_{effort}$) shapes attention and executive control (the subject of the next chapter), understanding effort $\epsilon$ itself is fundamental to appreciating the resource limitations and active nature of structured belief dynamics within $\Phi$. It allows modeling the difference between passive processing and focused, deliberate cognitive work.


\subsection*{Chapter Summary}
This chapter introduces Semantic Effort ($\epsilon$ or $E$), distinguishing it from Cognitive Load ($L$). While load represents the demand on the system, effort represents the finite cognitive resources actively allocated and expended by the agent to perform operations within the semantic manifold $\Phi$, constrained by a total capacity $E_{total}$. Semantic effort is conceptualized as the quantifiable intensity of processing applied to tasks like assimilation ($A$), reflection ($M$), retrieval ($R$), or coherence ($\kappa$) maintenance. The chapter discusses how the level of exerted effort directly impacts cognitive performance, influencing processing quality and speed, counteracting belief decay via Nullification ($N_t$) by reinforcing anchors ($a_i$), and potentially leading to resource depletion or fatigue if sustained at high levels. Although closely related (high load often necessitates high effort), effort (applied resource) and load (demand) are distinct. The strategic allocation of this effort, previewed for the next chapter, is central to attentional control. Effort is framed as the controllable "currency" of cognition.
	\chapter{Semantic Focus and Effort Allocation Policies}
\label{chap:SemanticFocus}

\section{Introduction: Directing Cognitive Resources}

Chapter~\ref{chap:SemanticEffort} introduced Semantic Effort ($\epsilon$) as the finite cognitive resource actively expended during processing within the semantic manifold $\Phi$. It defined the nature of this resource and its direct impact on the quality and sustainability of cognitive operations. However, possessing a resource is distinct from effectively directing it. Given limited total effort capacity $E_{total}$ and potentially numerous competing demands arising from external stimuli ($X(s)$), internal dynamics ($N_t, K, D$), retrieval ($R$), reflection ($M$), planning ($\Gamma$), and simulation, how does an agent decide where to focus its cognitive exertion?

This chapter delves into the crucial mechanisms of semantic effort allocation and its relationship to attentional focus. We explore how an agent might implement control policies, denoted $\pi_{effort}$, to distribute its available effort $\epsilon$ across different processes, semantic sectors $\Sigma$, abstraction levels $\Phi^{(k)}$, or specific belief fragments $\varphi_i$. We examine the inputs that inform these policies, the potential mechanisms by which allocated effort translates into focused processing (i.e., "attention"), various allocation strategies, and the possibility of learning effective policies over time (Chapter~\ref{chap:LearningRegulatoryPolicies}). This provides the framework for understanding how agents achieve cognitive control, prioritize tasks, and concentrate their mental resources.

\section{Formalizing Effort Allocation Policies (\texorpdfstring{$\pi_{\text{effort}}$}{pi\_effort})}

We model effort allocation as a function or policy, $\pi_{effort}$, implemented within the agent's regulatory system (likely interacting closely with $\Sigma_{refl}$ and meta-cognitive monitoring).

\textbf{Definition 33.1 (Effort Allocation Policy $\pi_{effort}$).} An Effort Allocation Policy $\pi_{effort}$ maps the agent's current state and context to a specific distribution of the available semantic effort $E_{total}$ across relevant cognitive targets:
$$ \pi_{effort} : \Phi \times \text{Context} \rightarrow \text{Allocation} $$
where:
\begin{itemize}
	\item $\Phi$ represents the current belief state (including meta-beliefs in $\Sigma_{refl}$).
	\item Context includes factors like active goals, perceived environmental state, task demands.
	\item Allocation specifies the distribution of effort, e.g., as a vector $\vec{\epsilon} = (\epsilon_1, \dots, \epsilon_m)$ such that $\sum \epsilon_i \le E_{total}$. Targets $i$ could be operators (e.g., $\epsilon_A, \epsilon_M$), sectors (e.g., $\epsilon_{\Sigma_{plan}}$), specific beliefs ($\epsilon_{\varphi_j}$), or tasks.
\end{itemize}
The policy $\pi_{effort}$ determines, moment-by-moment or strategically, how the agent invests its limited cognitive energy.

\section{Inputs Informing Allocation Policies}

Effective effort allocation requires integrating diverse information streams, primarily made available through Meta-Assimilation ($M$, Chapter~\ref{chap:MetaAssimilation}) into $\Sigma_{refl}$:
\begin{itemize}
	\item \textbf{Goals and Priorities:} Information from $\Sigma_{plan}$ regarding active goals, subgoals, deadlines, and their relative importance heavily influences $\pi_{effort}$. Higher priority goals receive greater $\epsilon$.
	\item \textbf{Cognitive Load ($L, \lambda$):} Meta-beliefs about current load levels $\lambda$ (Chapter~\ref{chap:CognitiveLoad}) inform $\pi_{effort}$. High load might trigger reallocation towards essential tasks or load-reduction strategies.
	\item \textbf{Coherence ($\kappa$):} Detected incoherence (low $\kappa$) can trigger $\pi_{effort}$ to allocate resources towards corrective assimilation $A_{corr}$ or reflective analysis $M$ to restore consistency.
	\item \textbf{Trajectory Awareness ($\gamma$):} Meta-beliefs about trajectory properties (Chapter~\ref{chap:TrajectoryAwareness}), such as deviation from orientation axes ($\theta$) or predicted entry into undesirable states, can guide $\pi_{effort}$ towards corrective steering or proactive replanning.
	\item \textbf{Epistemic Identity ($\vec{\eta}$):} The allocation policy $\pi_{effort}$ might be constrained by the need to maintain identity $\vec{\eta}$ (Chapter~\ref{chap:EpistemicIdentity}), prioritizing effort towards reinforcing core beliefs or values when they are challenged.
	\item \textbf{External Salience/Interrupts:} Salient external events processed via $X(s)$ can trigger $\pi_{effort}$ to reallocate effort towards perceptual processing ($\Sigma_{perc}$) and rapid response planning ($\Sigma_{plan}$).
	\item \textbf{Available Capacity ($E_{total}$):} The policy must operate within the currently available resource limits, potentially scaling down allocations if fatigue reduces $E_{total}$.
\end{itemize}
The sophistication of $\pi_{effort}$ depends on the agent's meta-cognitive capabilities to monitor these inputs.

\begin{figure}[htbp]
	\centering
	\input{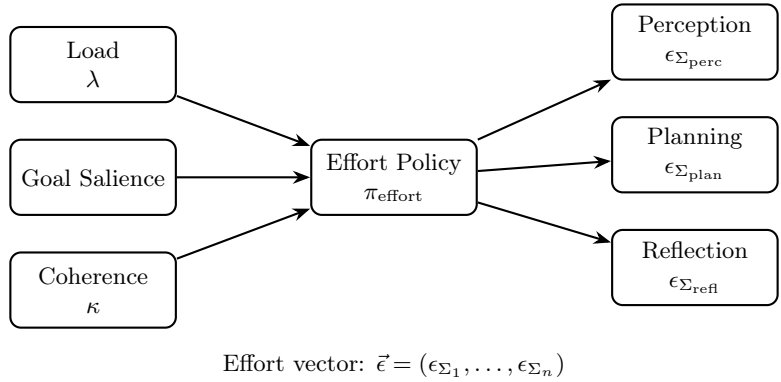}
	\caption{Effort allocation architecture. The policy \(\pi_{\text{effort}}\) receives introspective inputs---such as cognitive load \(\lambda\), coherence \(\kappa\), and goal salience---and produces an effort vector \(\vec{\epsilon}\) distributing cognitive resources across semantic processes or sectors (e.g., perception, planning, reflection).}
	\label{fig:effort-allocation}
\end{figure}

\section{Mechanisms of Focus and Attention}

How does allocating effort $\epsilon_j$ to a specific target $j$ (e.g., a belief $\varphi_j$, sector $\Sigma_j$, operator $O_j$) translate into the functional phenomenon of "attentional focus"? Within the framework, several mechanisms could implement this:
\begin{itemize}
	\item \textbf{Modulating Activation/Salience ($\lambda$):} Higher effort $\epsilon_j$ directed towards $\varphi_j$ or $\Sigma_j$ could directly boost its activation level $\lambda(\varphi_j)$ or $\lambda(\Sigma_j, \phi)$, making it more likely to influence subsequent processing (e.g., query generation $Q$, operator application).
	\item \textbf{Prioritizing Operator Application:} $\pi_{effort}$ could bias the selection mechanism for applying operators ($A, M, V, \Lambda, R$), preferentially applying them to belief fragments or sectors receiving high effort allocation.
	\item \textbf{Differential Anchoring/Persistence:} Effort $\epsilon_j$ applied to maintain $\varphi_j$ could translate into a higher effective anchoring $a_j$, slowing its decay under $N_t$. Attention actively preserves attended items in memory.
	\item \textbf{Inhibition of Unattended Elements:} Allocating high effort $\epsilon_j$ might concurrently trigger inhibitory signals that suppress the processing or activation ($\lambda$) of unrelated beliefs or sectors, sharpening focus by reducing interference.
	\item \textbf{Resource-Gated Operator Performance:} The quality or speed of an operator $O$'s execution could directly depend on the effort $\epsilon_O$ allocated to it (as discussed in Chapter~\ref{chap:SemanticEffort}). Attention enables deeper processing.
\end{itemize}

\begin{figure}[ht]
	\centering
	\input{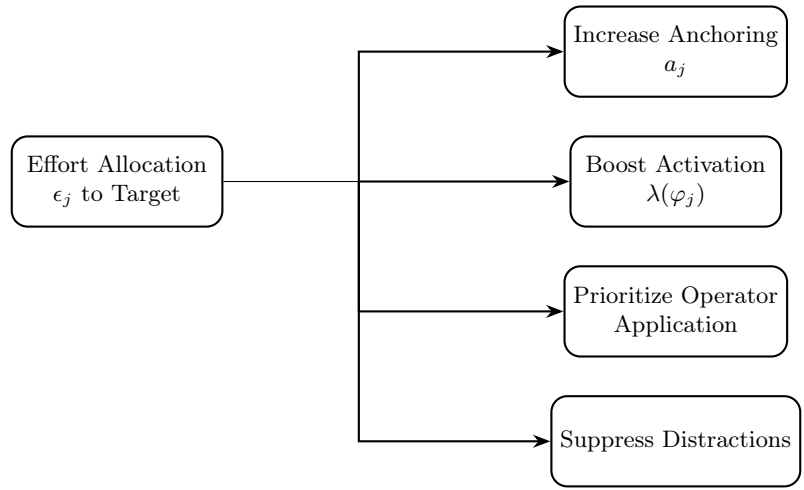}
	\caption{Mechanisms of focus. Semantic effort \(\epsilon_j\) allocated to a target (belief fragment \(\varphi_j\), sector \(\Sigma_j\), or operator \(O_j\)) influences cognitive processing through multiple pathways: boosting activation \(\lambda\), increasing anchoring \(a\), prioritizing operator application, and inhibiting unrelated elements. Focus emerges from structured modulation.}
	\label{fig:focus_mechanisms}
\end{figure}

Attention/focus, in this view, is not a separate module but an emergent consequence of how the allocation policy $\pi_{effort}$ directs the finite resource $\epsilon$ to modulate the activation and processing priorities within the semantic manifold $\Phi$.

\section{Effort Allocation Strategies}

The policy $\pi_{effort}$ can implement various high-level strategies for managing cognitive resources, selected based on the inputs described in Section 3:
\begin{itemize}
	\item \textbf{Goal Shielding:} When a high-priority goal is active (input from $\Sigma_{plan}$), $\pi_{effort}$ allocates maximal effort $\epsilon$ to processes directly supporting that goal (e.g., planning in $\Sigma_{plan}$, relevant retrieval $R$, necessary simulation) while actively inhibiting or allocating minimal effort to potentially distracting inputs or unrelated reflective processes ($\Sigma_{refl}, \Sigma_{narr}$).
	\item \textbf{Load Balancing:} If meta-cognition ($M$) reports high overall load $\lambda$ or high load in a specific sector (e.g., $\lambda(\Sigma_{perc}, \phi)$ is excessive), $\pi_{effort}$ might reduce effort allocated to non-critical background tasks and/or allocate effort specifically to load-reducing actions like Abstraction ($\Lambda$) or targeted Nullification ($N_t^{\Sigma}$).
	\item \textbf{Vigilance/Monitoring:} In stable or low-demand situations, $\pi_{effort}$ might adopt a vigilance strategy, allocating a baseline level of effort to perceptual monitoring ($\Sigma_{perc}$) and routine coherence checking ($\kappa$) to ensure readiness for unexpected events or gradual drift.
	\item \textbf{Exploration vs. Exploitation:} Depending on task phase or uncertainty levels (assessed via $M$), $\pi_{effort}$ could shift allocation. High uncertainty might trigger allocation towards exploring new information (effort $\epsilon_A$ for novel input $X(s)$), while consolidating understanding might trigger effort allocation towards internal reflection ($\epsilon_M$) or re-anchoring important memories ($\epsilon$ to $Q \rightarrow R \rightarrow A$ cycles).
	\item \textbf{Reactive vs. Deliberative Allocation:} Under severe time pressure or detected crisis (input from Context or $M$), $\pi_{effort}$ might default to a reactive strategy, allocating effort primarily based on immediate external salience ($X(s)$) or simple triggers. In less pressured situations, it might allocate significant effort $\epsilon_{M}$ to meta-cognitive deliberation within $\Sigma_{refl}$ to determine a more optimal long-term effort distribution before committing resources.
\end{itemize}
The choice of strategy might itself be context-dependent or learned (Chapter~\ref{chap:LearningRegulatoryPolicies}).

\section{Learning Effective Allocation Policies}

Optimal allocation of semantic effort is unlikely to be fixed; agents should learn effective $\pi_{effort}$ policies. Potential learning mechanisms include:
\begin{itemize}
	\item \textbf{Reinforcement Learning:} Using task success, goal achievement, or internal metrics (e.g., maintaining high $\kappa$, low unexpected trajectory deviation $\theta$) as rewards to train $\pi_{effort}$.
	\item \textbf{Imitation Learning:} Learning allocation strategies by observing demonstrators (human or other AI).
	\item \textbf{Meta-Learning:} Adapting the parameters of $\pi_{effort}$ itself based on performance across different types of tasks or contexts (learning how to allocate effort differently in different situations).
	\item \textbf{Evolutionary Methods:} Selecting effective allocation strategies over generations in simulated populations.
\end{itemize}
Learning efficient resource management policies is crucial for scaling cognitive capabilities and achieving robust performance in complex environments, as detailed further in Chapter~\ref{chap:LearningRegulatoryPolicies}.

\section{Conclusion: Attention as Controlled Effort}

Semantic focus and attentional control emerge within this framework not as separate modules, but as consequences of the agent's ability to strategically allocate its finite semantic effort ($\epsilon$). The allocation policy, $\pi_{effort}$, acts as the central executive, directing resources based on meta-cognitive assessments of goals, load, coherence, trajectory, and identity.

By formalizing $\pi_{effort}$ and its potential mechanisms (modulating activation $\lambda$, prioritizing operators, inhibiting distractions), we gain a principled way to model attentional phenomena and resource management within the semantic manifold $\Phi$. This chapter, complementing the previous one on the nature of effort $\epsilon$, provides the control layer necessary for understanding how agents focus their cognitive energy, manage trade-offs, and ultimately direct their own thought processes. Effective effort allocation, guided by sophisticated meta-cognition, is key to flexible, efficient, and goal-directed reasoning in any resource-constrained intelligent system.


\subsection*{Chapter Summary}
This chapter addresses how the finite cognitive resource of Semantic Effort ($\epsilon$, introduced in Chapter~\ref{chap:SemanticEffort}) is directed and managed, linking its allocation to the concepts of semantic focus and attention. It introduces Effort Allocation Policies ($\pi_{effort}$) as the control mechanisms responsible for distributing the available effort capacity ($E_{total}$) across various cognitive processes, semantic sectors ($\Sigma$), abstraction levels ($k$), or specific belief fragments within $\Phi$. These policies are informed by diverse inputs, including active goals, current cognitive load ($\lambda$), coherence levels ($\kappa$), trajectory awareness, and contextual demands, often synthesized through meta-cognitive processing ($M$) in $\Sigma_{refl}$. The chapter discusses how allocated effort translates into functional focus, potentially through mechanisms like modulating belief activation ($\lambda$), prioritizing operator applications, inhibiting unattended elements, or gating processing quality. Various allocation strategies, such as goal shielding, load balancing, and vigilance, are outlined. Ultimately, attention and focus are presented not as separate modules but as emergent outcomes of the agent's learned or designed policies for controlling its limited cognitive energy.
	
	\part{Learning and Adaptation}
	\label{part:learning_and_adaptation}
	
	\chapter{Learning Semantic Structures}
\label{chap:LearningSemanticStructures}

\section{Introduction: The Plasticity of Mind-Space}

The preceding Parts of this monograph have detailed a rich architecture for structured belief, defined by the semantic state space $\Phi$ (parameterized by $\theta$), its internal structuring via semantic scaling ($\Phi^{(k)}$) and sectors ($\Sigma$), and its geometry characterized conceptually by a semantic metric ($d$). While we have discussed the dynamics ($A, M, N_t, K, \dots$) operating within this space and the regulatory mechanisms controlling trajectories ($\gamma(t)$), we have largely treated the underlying structure itself as given or predefined by the agent's design $\theta$.

However, a truly adaptive and intelligent system, especially one aiming for human-like flexibility, likely requires its internal representational structures to be plastic---shaped and refined by experience and learning. How might the very fabric of the semantic manifold $\Phi$, its geometry, and its functional organization into sectors $\Sigma$ arise or adapt over time?

This chapter initiates our exploration of learning within the Semantic Manifold framework (Part~\ref{part:learning_and_adaptation}) by focusing on the acquisition and adaptation of its foundational structures. We consider potential mechanisms by which the semantic metric $d$ could be learned, how functional sectors $\Sigma$ might emerge or differentiate, and speculate on the adaptability of the manifold's geometry itself. \textbf{The objective of such structural learning is to optimize the representational space itself for better performance, efficiency, or coherence in the agent's cognitive tasks.} This focus on structural learning complements the subsequent chapters which will address the learning of cognitive operators (Chapter~\ref{chap:LearningCognitiveOperators}) and regulatory policies (Chapter~\ref{chap:LearningRegulatoryPolicies}).

\section{Learning the Semantic Metric (\texorpdfstring{$d$}{d})}

The semantic metric $d(\phi_1, \phi_2)$ plays a fundamental role, underpinning similarity judgments, retrieval ($R$), coherence evaluation ($\kappa$), orientation ($r$), and potentially operator application ($A, M, \Lambda, V$). A predefined, static metric may fail to capture the nuanced, context-dependent, and evolving nature of semantic similarity required for sophisticated cognition. Therefore, mechanisms for learning or adapting $d$ are highly desirable. \textbf{The primary objective for learning $d$ is to ensure that distances in $\Phi$ accurately reflect functional or semantic similarity relevant to the agent's tasks and goals,} leading to improved retrieval, coherence assessment, and generalization.

\begin{figure}[htbp]
	\centering
	\input{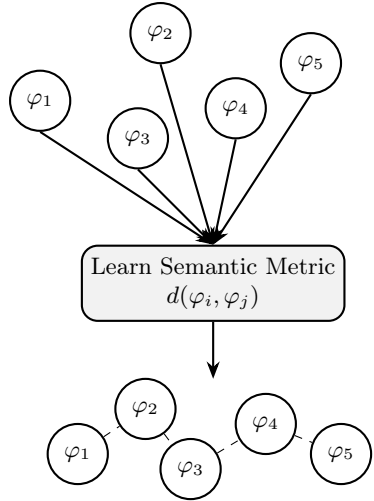}
    \caption{Learning semantic distance. Raw belief fragments $\varphi_i$ are processed through metric learning ($d$), resulting in a structured space where distances reflect semantic similarity.}
	\label{fig:learning-semantic-metric}
\end{figure}

\begin{itemize}
	\item \textbf{Dependence on Representation ($\rho$):} The methods for learning $d$ are intrinsically tied to the chosen representation mode $\rho$ (parameter in $\theta$, Chapter~\ref{chap:ParameterizedArchitectures}).
	\item \textbf{Learning in Embedding Spaces ($\rho \approx$ Embedding):} If beliefs are represented as vectors, $d$ might be related to standard vector space metrics (Euclidean, Cosine). Learning $d$ then becomes equivalent to learning effective embedding functions. Techniques include:
	\begin{itemize}
		\item \textbf{Contrastive Learning:} Training embedding models (e.g., variants of BERT, Sentence-BERT, or custom networks) to pull semantically related belief fragments $\{\varphi_i\}$ closer together in the embedding space while pushing unrelated ones apart, based on supervision derived from co-occurrence, entailment relations, or task performance rewards that signal semantic similarity.
		\item \textbf{Metric Learning:} Directly learning a parameterized distance function $d_{\psi}(v_1, v_2)$ over a base embedding space, where parameters $\psi$ are optimized to satisfy desired similarity constraints derived from experience or task feedback.
	\end{itemize}
	\item \textbf{Learning in Symbolic/Graph Spaces ($\rho \approx$ Symbolic/Graph):} For symbolic or graph representations, learning $d$ might involve:
	\begin{itemize}
		\item Learning weights on structural features or links used in distance calculation (e.g., weighted edit distance, learned graph similarity kernels), optimizing weights to reflect functional similarity.
		\item Learning structural alignment algorithms that best capture semantic relatedness.
	\end{itemize}
	\item \textbf{Influence of Grounding:} The learning process for $d$ should ideally be informed by grounding (Chapter~\ref{chap:GroundingSemanticBelief}). Beliefs grounded in similar sensorimotor experiences or linguistic contexts should map to points closer in $\Phi$ according to the learned $d$.
\end{itemize}
An adaptive semantic metric $d$ allows the agent's notion of similarity to evolve, reflecting its growing knowledge and experience, ultimately improving the foundation for many cognitive operations.

\section{Emergence and Refinement of Semantic Sectors (\texorpdfstring{$\Sigma$}{Sigma})}

Semantic sectors $\Sigma$ provide functional modularity within $\Phi$, partitioning beliefs based on their role (e.g., $\Sigma_{perc}, \Sigma_{plan}, \Sigma_{refl}, \Sigma_{narr}$). While Chapter~\ref{chap:SemanticSectors} suggested these might be somewhat predefined structurally, a more powerful model allows them to emerge or be refined through learning. \textbf{The objective here is to develop a sectoral organization that optimally supports modular processing, reduces interference between functional subsystems, and facilitates efficient task execution.}

\begin{figure}[htbp]
	\centering
	\input{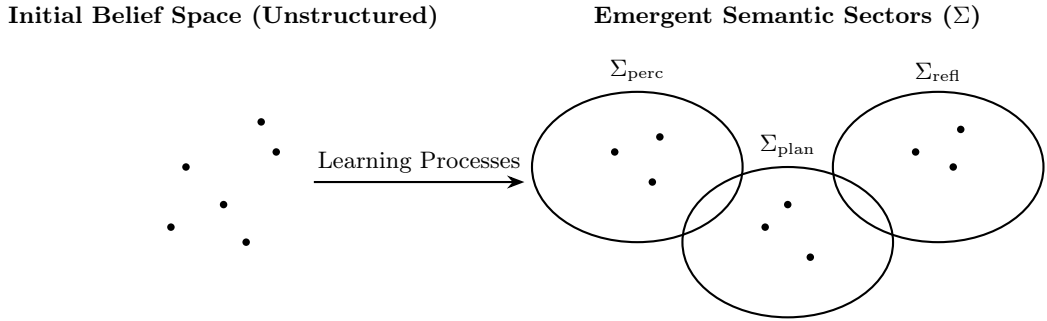}
	\caption{Emergence of Semantic Sectors through Learning. Initially, the belief space \(\Phi\) is unstructured. Learning processes, such as activity correlation and clustering, drive the differentiation of \(\Phi\) into functional sectors like perception (\(\Sigma_{\text{perc}}\)), planning (\(\Sigma_{\text{plan}}\)), and reflection (\(\Sigma_{\text{refl}}\)), facilitating modular cognitive processing.}
	\label{fig:emergence-semantic-sectors}
\end{figure}

\begin{itemize}
	\item \textbf{Activity Correlation / Hebbian Learning:} Sectors might emerge from patterns of co-activation. Belief fragments $\{\varphi_i\}$ that are frequently processed together (e.g., involved in planning sequences, or part of reflective loops) could become strongly interconnected, forming functional clusters that differentiate into sectors over time.
	\item \textbf{Clustering Methods:} Unsupervised or semi-supervised clustering algorithms applied to the belief space $\Phi$ (based on the learned metric $d$ and connectivity) could identify natural groupings corresponding to functional sectors, aiming to maximize intra-sector coherence and minimize inter-sector dependencies where appropriate.
	\item \textbf{Task-Driven Specialization:} If the agent learns different tasks, reinforcement learning or supervised learning might drive the specialization of different parts of $\Phi$ (or underlying neural substrates if implemented that way) to handle specific task-related functions, leading to the emergence of $\Sigma$ that optimize task performance.
	\item \textbf{Modular Training:} Explicitly training distinct modules (e.g., separate neural networks) for different functions (perception, planning, reflection) and composing them could instantiate a structured $\Phi$ with predefined but potentially adaptable sectors, where adaptation optimizes the interfaces or specialization of modules.
\end{itemize}

The ability to learn or refine sectoral organization allows the agent to develop specialized processing pathways tailored to its cognitive needs and experiences.

\section{Adapting Manifold Geometry (Speculative)}

Beyond learning the metric $d$ or the sectoral organization $\Sigma$, could the fundamental geometric or topological structure of $\Phi$ itself adapt through learning? This is more speculative but theoretically interesting. \textbf{The potential objective would be to find a geometry that most efficiently or robustly represents the semantic relationships relevant to the agent's world and tasks.}

\begin{itemize}
	\item \textbf{Dimensionality Reduction/Expansion:} Learning processes (like those used for $\Lambda$) might implicitly learn an appropriate intrinsic dimensionality for representing beliefs within $\Phi$, potentially changing over time or across sectors to balance representational richness and computational cost.
	\item \textbf{Learning Curvature/Topology:} Could the agent learn that certain regions of semantic space exhibit higher "curvature" (representing ambiguity or complex interactions) and adapt its processing accordingly? Could topological features (holes, boundaries) emerge reflecting conceptual distinctions learned from data?
	\item \textbf{Challenges:} Formalizing and implementing mechanisms for learning the underlying manifold geometry itself is significantly harder than learning a metric or clustering on a predefined space. This remains a challenging research direction, potentially involving advanced concepts from geometric deep learning or topological data analysis. Defining clear learning objectives and stable algorithms for such adaptation is difficult.
\end{itemize}

\begin{figure}[htp!]
	\centering
	\tdplotsetmaincoords{70}{110}
	\input{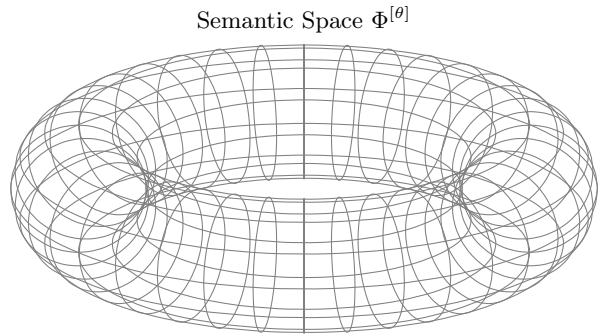}
	\caption{Speculative learned topology for the belief space $\Phi$. Instead of being flat or Euclidean, structural learning could induce a nontrivial manifold topology (e.g., a torus) to optimize semantic organization.}
	\label{fig:speculative-topology-torus}
\end{figure}

\section{Relation to Representation Mode (\texorpdfstring{$\rho$}{rho})}

It is crucial to reiterate that the specific mechanisms viable for learning semantic structures are heavily dependent on the chosen representation mode $\rho$ defined in the agent's parameter vector $\theta$ (Chapter~\ref{chap:ParameterizedArchitectures}).
\begin{itemize}
	\item Embedding-based $\rho$ lends itself naturally to metric learning, clustering, and dimensionality reduction using techniques from deep learning.
	\item Symbolic $\rho$ requires different techniques, such as inductive logic programming, grammar induction, or symbolic clustering methods.
	\item Graph-based $\rho$ might leverage graph neural networks for learning node embeddings (affecting $d$) or community detection algorithms for identifying sectors ($\Sigma$).
	\item Hybrid $\rho$ necessitates combining techniques suitable for each component.
\end{itemize}
The design choice of $\rho$ therefore strongly constrains the pathways available for structural learning within $\Phi$.

\begin{figure}[htp!]
	\centering
	\input{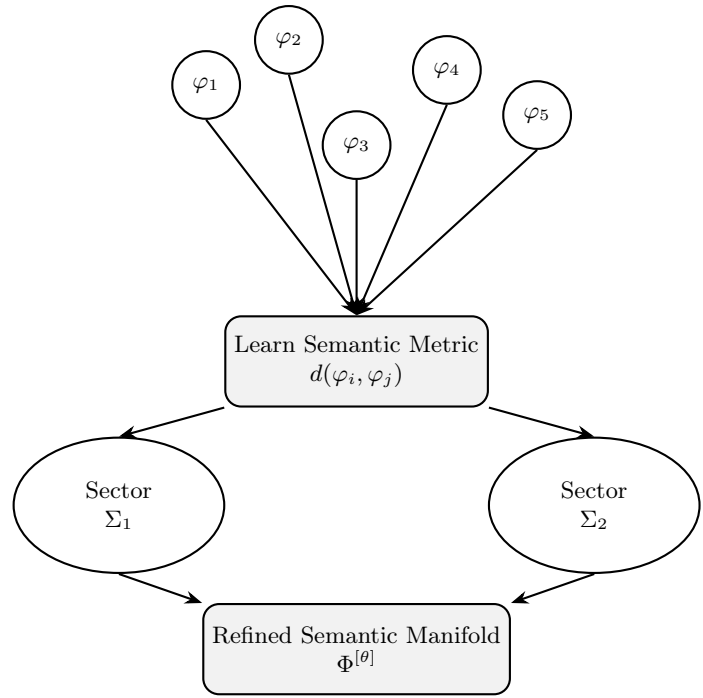}
	\caption{Learning semantic structures. Raw belief fragments $\varphi_i$ are processed through metric learning ($d$), leading to the emergence of sectors ($\Sigma$) and refinement of the semantic manifold $\Phi^{[\theta]}$.}
	\label{fig:learning-semantic-structure-flow}
\end{figure}

\section{Conclusion: Towards a Learned Mind-Space}

This chapter initiated the exploration of learning within the Semantic Manifold framework by focusing on its foundational structures. We argued that for agents to achieve true adaptability and sophisticated understanding, the semantic metric $d$, the functional organization into sectors $\Sigma$, and potentially even the underlying geometry of the belief space $\Phi$ should not be entirely fixed but should emerge from or be refined by learning processes, driven by objectives like improving task performance, coherence, or representational efficiency.

We outlined potential mechanisms drawing on machine learning techniques like contrastive learning, metric learning, clustering, and modular training, emphasizing the dependence on the chosen representation mode $\rho$. While significant challenges remain, particularly in learning manifold geometry, incorporating structural plasticity moves the framework towards modeling agents whose internal representational landscape---their very "mind-space"---is dynamically shaped by their interactions with the world. The following chapters will build on this by exploring how the cognitive operators themselves can be learned.


\subsection*{Chapter Summary}
This chapter begins the exploration of learning and adaptation within the Semantic Manifold framework, shifting focus from dynamics within a fixed structure to the potential plasticity of the structure itself. It proposes that core structural components of the belief space $\Phi^{[\theta]}$---specifically the semantic distance metric ($d$) and the functional organization into semantic sectors ($\Sigma$)---can be learned or refined through experience, rather than being entirely predefined. The chapter discusses potential mechanisms for this structural learning, emphasizing their dependence on the chosen representation mode ($\rho$). For instance, learning $d$ might involve contrastive representation learning or metric learning if $\rho$ is embedding-based, while sector emergence ($\Sigma$) could result from activity correlation, clustering, or task-driven specialization. The more speculative possibility of adapting the underlying manifold geometry is also considered. The overall objective of learning semantic structures is to optimize the internal representational space $\Phi$ itself for improved cognitive efficiency, coherence, and task performance.
	\chapter{Learning Cognitive Operators}
\label{chap:LearningCognitiveOperators}

\section{Introduction: From Fixed Functions to Learned Processes}

Chapter~\ref{chap:LearningSemanticStructures} explored the potential for learning the foundational structures of the semantic state space $\Phi$, including its metric $d$ and sectoral organization $\Sigma$. However, the cognitive architecture defined by the Semantic Manifold framework relies equally on a suite of dynamic operators---Assimilation ($A$), Meta-Assimilation ($M$), Nullification ($N_t$), Annihilation ($K$), Abstraction ($\Lambda$), Elaboration ($V$), Retrieval ($R$), and Querying ($Q$)---that manipulate beliefs within this space.

A truly adaptive agent likely cannot rely on entirely fixed, pre-specified operators; these processes themselves must be learnable or adaptable based on experience and task demands. This chapter investigates how these core cognitive operators might be acquired or refined through learning. We consider mechanisms drawn from machine learning that could allow an agent to develop effective strategies for integrating information ($A, M$), navigating abstraction levels ($\Lambda, V$), accessing memory ($R, Q$), and potentially even tuning decay ($N_t$) or conflict resolution ($K$) based on its interactions and internal evaluations.

The ability to learn these operators is crucial for tailoring the agent's cognitive dynamics to its specific environment and goals, moving beyond hardcoded processing logic. As with structural learning, the feasibility and specific methods depend significantly on the agent's parameterization $\theta$, particularly the representation mode $\rho$.

\begin{figure}[htbp]
	\centering
	\input{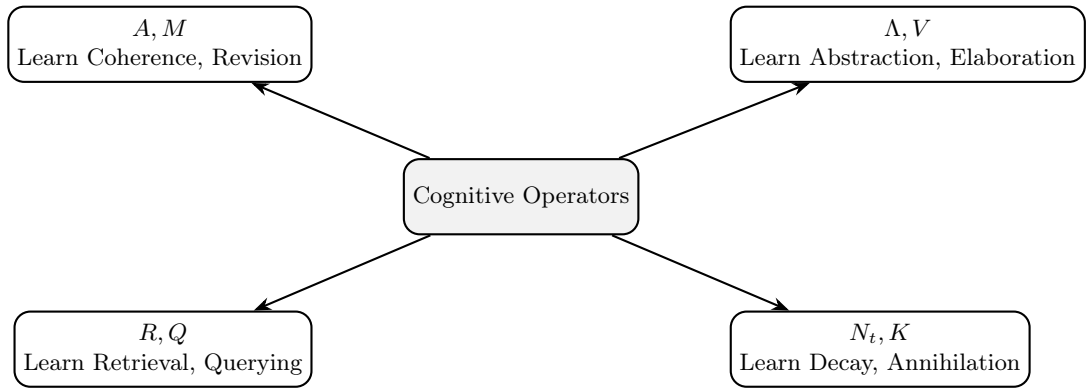}
	\caption{Learning pathways for core cognitive operators within the Semantic Manifold framework. Each operator (Assimilation $A$, Meta-Assimilation $M$, Abstraction $\Lambda$, Elaboration $V$, Retrieval $R$, Querying $Q$, and decay processes $N_t$, $K$) is associated with specific machine learning techniques enabling their refinement through experience.}
	\label{fig:learning-cognitive-operators-map}
\end{figure}

\section{Learning Abstraction (\texorpdfstring{$\Lambda$}{Lambda}) and Elaboration (\texorpdfstring{$V$}{V})}

The scaling operators $\Lambda$ (specific to general) and $V$ (general to specific) are central to managing complexity and connecting different levels of understanding. Learning these operators involves acquiring the ability to perform meaningful generalization and instantiation.

\begin{figure}[htbp]
	\centering
	\input{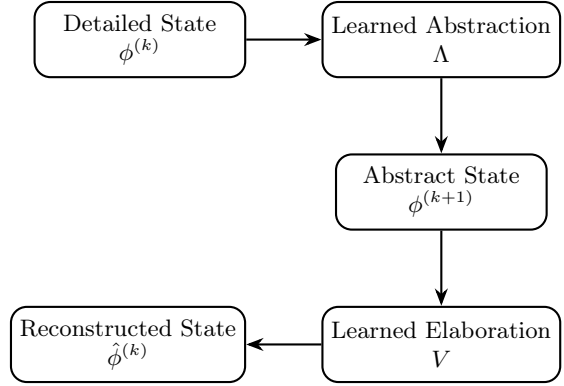}
	\caption{Learning abstraction ($\Lambda$) and elaboration ($V$) via an autoencoder structure. Belief states $\phi^{(k)}$ are encoded into compressed abstract representations $\phi^{(k+1)}$ and reconstructed back, optimizing the agent's ability to manage information across levels of semantic scaling.}
	\label{fig:learning-lambda-v-autoencoder}
\end{figure}

\subsection{Learning \texorpdfstring{$\Lambda$}{Lambda} (Abstraction)}

\begin{itemize}
	\item \textbf{Using Autoencoders/Representation Learning:} If $\rho$ involves embeddings, $\Lambda$ could be implemented as the encoder part of an autoencoder trained on belief states $\phi^{(k)}$. The bottleneck layer learns a compressed, abstract representation $\phi^{(k+1)}$. The learning objective could be reconstruction accuracy (in combination with learning $V$) or optimizing the abstract representation for downstream tasks.
	\item \textbf{Learning Summarization Policies:} $\Lambda$ could be learned as a policy (potentially using RL or supervised learning on human summaries) that selects or generates the most salient components of $\phi^{(k)}$ to form $\phi^{(k+1)}$. Techniques from abstractive text summarization are relevant here, especially if leveraging LLM components for implementation.
	\item \textbf{Concept Formation/Clustering:} Abstraction can involve identifying recurring patterns across multiple $\phi^{(k)}$ instances and forming a higher-level concept node $\varphi^{(k+1)}$ that represents the cluster, potentially using unsupervised learning techniques.
\end{itemize}

\subsection{Learning \texorpdfstring{$V$}{V} (Elaboration)}

\begin{itemize}
	\item \textbf{Using Generative Models/Decoders:} $V$ could be the decoder part of an autoencoder, reconstructing details of $\phi^{(k)}$ from the abstract representation $\phi^{(k+1)}$. If using LLMs, $V$ could be implemented via prompted generation conditioned on $\phi^{(k+1)}$ to produce plausible details $\phi^{(k)}$. Learning would optimize the fidelity or utility of the generated details.
	\item \textbf{Learning Instantiation Rules:} For symbolic architectures, $V$ might involve learned rules or templates for generating specific examples or groundings from abstract descriptions, possibly learned via inductive logic programming.
	\item \textbf{Goal:} Learning should optimize $V$ such that the elaborated state $V(\Lambda(\phi))$ is functionally useful and consistent, even if not identical to the original $\phi$ (as discussed in Chapter~\ref{chap:SemanticScaling}).
\end{itemize}
Learning $\Lambda$ and $V$ allows the agent to develop its own effective hierarchy for representing and manipulating information at different levels of granularity.

\section{Learning Assimilation (\texorpdfstring{$A$}{A}) and Meta-Assimilation (\texorpdfstring{$M$}{M})}

Assimilation ($A$, Chapter~\ref{chap:Assimilation}) integrates object-level input, while Meta-Assimilation ($M$, Chapter~\ref{chap:MetaAssimilation}) integrates introspective input. Both involve complex steps like context assessment, coherence checking ($\kappa$), potential revision ($A_{corr}$), and elaboration ($A_{elab}$). Learning effective assimilation is crucial for robust knowledge acquisition and self-awareness.

\begin{figure}[htbp]
	\centering
	\input{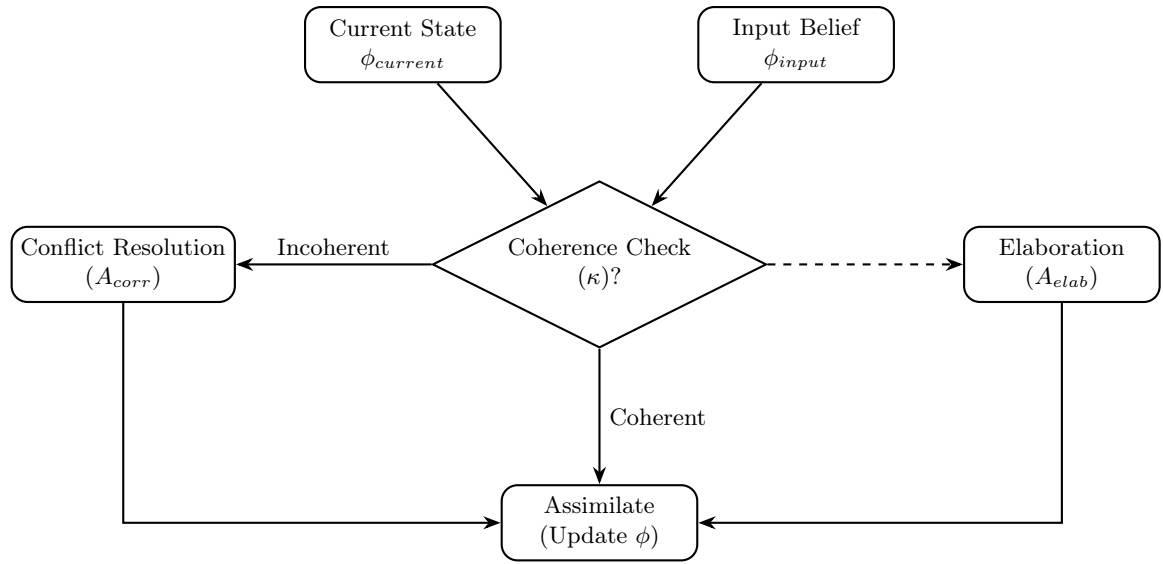}
	\caption{Learning conflict resolution within Assimilation ($A$) and Meta-Assimilation ($M$). The agent assesses coherence $\kappa$, applies learned revision strategies ($A_{corr}$), and elaborates beliefs ($A_{elab}$) to achieve robust integration of new information and meta-information into its belief state $\phi$.}
	\label{fig:learning-assimilation-metaassimilation}
\end{figure}

\begin{itemize}
	\item \textbf{Learning Coherence Judgment ($\kappa$ basis):} The ability to assess coherence $\kappa$, crucial for $A_{corr}$ and $M$, might be learned. This could involve training classifiers to detect inconsistencies or using energy-based models where coherent states have lower energy, optimized based on task outcomes or logical constraints.
	\item \textbf{Learning Revision Policies (within $A_{corr}$):} Deciding which beliefs to retract during conflict resolution is complex. An agent could learn revision policies (e.g., based on anchoring $a_i$, source tags, logical entailment) using RL, where rewards are given for restoring high coherence $\kappa$ or improving subsequent task performance.
	\item \textbf{Learning Elaboration Strategies (within $A_{elab}$):} The agent could learn which elaborations are most useful or contextually appropriate, perhaps by training generative models conditioned on $\phi_{current}$ and $\phi_{input}$, optimizing for predictive accuracy or goal relevance.
	\item \textbf{Fine-tuning LLM Components:} If $A$ or $M$ are implemented using LLMs (Appendix~\ref{app:ImplementationExamples}), fine-tuning the LLM on specific assimilation or meta-assimilation tasks (e.g., integrating conflicting reports, generating meta-summaries based on state metrics) is a direct way to learn these operator functions tailored to the agent's needs.
	\item \textbf{Learning for $M$:} Specifically for $M$, learning might involve calibrating the integration of introspective signals ($\Phi_{introspective}$), perhaps learning to weight different internal signals based on their past reliability in predicting performance or problems.
\end{itemize}
Learning $A$ and $M$ allows the agent to move beyond predefined integration rules towards context-sensitive, adaptive, and potentially more robust belief updating and self-assessment.

\section{Learning Retrieval (\texorpdfstring{$R$}{R}) and Querying (\texorpdfstring{$Q$}{Q})}

Effective use of semantic memory (Part~\ref{part:semantic_memory}) depends on retrieving the right information ($R$, Chapter~\ref{chap:RetrievalOperator}) in response to appropriate cues ($Q$, Chapter~\ref{chap:QueryingBeliefSpace}).

\begin{figure}[htbp]
	\centering
	\input{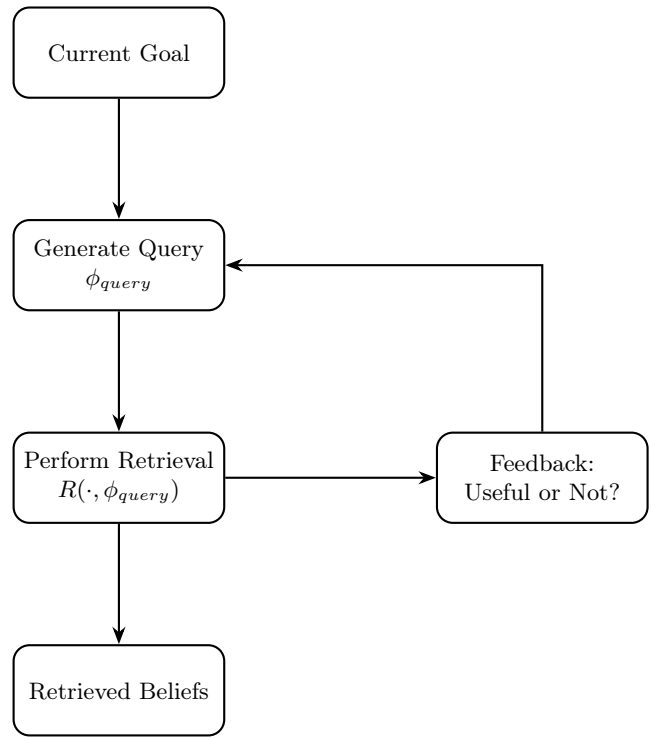}
	\caption{Learning mechanisms for Query ($Q$) and Retrieval ($R$). The agent refines its ability to generate effective queries based on task needs and optimizes retrieval strategies using learned metrics ($d$) and experience with retrieval success or failure.}
	\label{fig:learning-query-retrieval}
\end{figure}

\subsection{Learning Retrieval Policies (\texorpdfstring{$R$}{R})}

\begin{itemize}
	\item \textbf{Tuning Similarity/Relevance:} If $R$ uses the learned metric $d$ (Chapter~\ref{chap:LearningSemanticStructures}), its effectiveness improves as $d$ improves. Additionally, the thresholds or search strategies used by $R$ could be adapted based on feedback (e.g., was the retrieved information useful for the current task?). This could be framed as an RL problem where retrieval effectiveness provides the reward.
	\item \textbf{Learning Associative Links:} For graph-based $\rho$, the links traversed by $R$ could be learned or weighted based on experienced co-activations or causal relationships identified during learning.
\end{itemize}

\subsection{Learning Query Formulation (\texorpdfstring{$Q$}{Q})}

\begin{itemize}
	\item An agent could learn what makes a good retrieval cue $\phi_{query}$. This might involve RL where the agent gets rewarded for generating queries that lead to successful retrieval ($R$ returns relevant results) contributing to goal achievement.
	\item Learning to map from $\phi_{current}$ and current goals to effective $\phi_{query}$ could involve training sequence-to-sequence models or attention mechanisms, potentially supervised by examples of effective queries or learned via RL based on retrieval outcomes.
\end{itemize}
Learning $Q$ and $R$ transforms memory from a static store to an adaptive resource accessed through increasingly effective querying and retrieval strategies.

\section{Learning Decay (\texorpdfstring{$N_t$}{N\_t}) and Annihilation (\texorpdfstring{$K$}{K}) Triggers}

While $N_t$ (Chapter~\ref{chap:Nullification}) and $K$ (Chapter~\ref{chap:Annihilation}) primarily model passive decay or abrupt erasure, aspects of their operation might be adaptable:

\begin{figure}[htbp]
	\centering
	\input{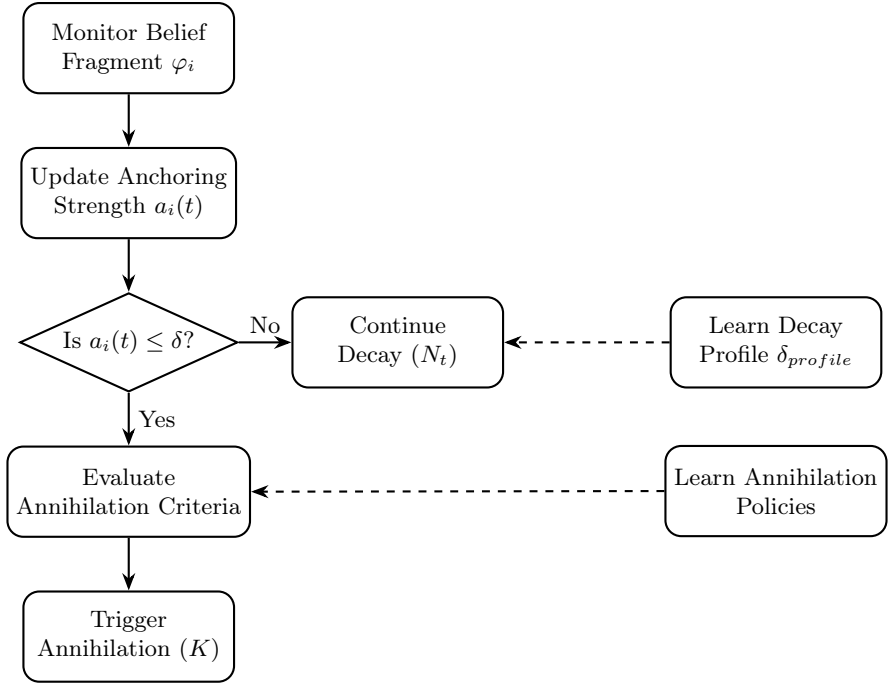}
	\caption{Learning-augmented decay and annihilation process. Belief fragments \(\varphi_i\) undergo anchoring decay over time. When the anchoring strength \(a_i(t)\) drops below a threshold \(\delta\), the agent evaluates whether to allow natural decay via \(N_t\) or trigger forced annihilation \(K\), with learned modules optionally adapting decay profiles and annihilation policies.}
	\label{fig:learning-decay-annihilation}
\end{figure}

\begin{itemize}
	\item \textbf{Tuning Decay Parameters ($\delta_{profile}$):} The specific rates or thresholds governing decay in different sectors or for beliefs with different properties might be learnable parameters within $\theta$, optimized based on overall performance (e.g., balancing memory retention vs. avoiding clutter). RL could be used to tune these based on long-term task success.
	\item \textbf{Learning Anchoring Dynamics ($a_i$):} The rules governing how anchoring $a_i$ changes based on usage (via $A$ or $M$) or perceived importance could be learned, perhaps by optimizing the anchoring updates to maximize the utility of retained information.
	\item \textbf{Learning Annihilation Conditions ($K$):} The agent might learn more nuanced conditions under which Annihilation $K$ (total or sectoral $K_{\Sigma}$) is triggered, beyond simple contradiction detection, perhaps based on persistent negative impact on coherence $\kappa$ or goal achievement. This could involve learning a classifier or policy based on historical data where annihilation proved beneficial or detrimental.
\end{itemize}
Learning in this context allows the agent to optimize its "cognitive metabolism"---how it maintains useful information and prunes the irrelevant or harmful.

\section{Conclusion: Towards Adaptive Cognitive Processes}

This chapter has argued that the core cognitive operators ($A, M, \Lambda, V, R, Q$, and potentially aspects of $N_t, K$) within the Semantic Manifold framework need not be static, predefined functions. By leveraging techniques from machine learning---including representation learning, sequence modeling, reinforcement learning, and supervised fine-tuning---agents can potentially learn or adapt these crucial processes based on their experience.

Learning effective operators allows an agent to tailor its cognitive dynamics to its specific environment, tasks, and internal architecture ($\theta$). It enables the development of sophisticated abstraction hierarchies, context-sensitive assimilation strategies, efficient memory access, and optimized decay/pruning mechanisms. Combined with the structural learning discussed in Chapter~\ref{chap:LearningSemanticStructures} and the policy learning in Chapter~\ref{chap:LearningRegulatoryPolicies}, learning cognitive operators is essential for creating agents that not only possess structured beliefs but can also process and manipulate those beliefs in increasingly adaptive and intelligent ways.


\subsection*{Chapter Summary}
Following the discussion on learning semantic structures, this chapter focuses on the learnability and adaptability of the core cognitive operators themselves within the Semantic Manifold framework. It argues that operators such as Abstraction ($\Lambda$), Elaboration ($V$), Assimilation ($A$), Meta-Assimilation ($M$), Retrieval ($R$), and Querying ($Q$), along with potentially aspects of Nullification ($N_t$) and Annihilation ($K$), need not be entirely fixed but can be refined through learning. Depending on the chosen representation mode ($\rho$) and architecture ($\theta$), various machine learning techniques could be employed. Examples include using representation learning or summarization techniques for $\Lambda$ and $V$; reinforcement learning or fine-tuning generative models for aspects of $A$ and $M$ (like coherence checking or conflict revision); and learning effective query formulation ($Q$) and retrieval ($R$) strategies based on task success. Learning decay parameters ($N_t$) or annihilation triggers ($K$) might also be possible. This allows the agent to tailor its fundamental cognitive processes---how it integrates information, navigates abstraction levels, accesses memory, and manages belief persistence---based on experience, leading to more adaptive and efficient cognition.
	\chapter{Learning Regulatory Policies}
\label{chap:LearningRegulatoryPolicies}

\section{Introduction: Acquiring Cognitive Control}

The preceding chapters in this Part explored the learning of semantic structures (Chapter~\ref{chap:LearningSemanticStructures}) and core cognitive operators (Chapter~\ref{chap:LearningCognitiveOperators}). However, possessing effective representations and operators is only part of the equation for adaptive intelligence. An agent must also possess effective control policies to deploy these resources appropriately---regulating its internal state, managing limited resources like semantic effort ($\epsilon$), and guiding its cognitive trajectory $\gamma(t)$ towards its goals while maintaining coherence ($\kappa$) and identity ($\vec{\eta}$).

Part~\ref{part:regulation_and_control} introduced regulatory concepts like Orientation and Identity, while Part~\ref{part:meta_cognition} detailed meta-cognitive monitoring ($M$) and resource management ($\lambda, \epsilon$) culminating in effort allocation policies ($\pi_{effort}$, Chapter~\ref{chap:SemanticFocus}). This chapter addresses the critical question: how are these regulatory and allocation policies themselves acquired or refined?

Relying solely on predefined, fixed control policies limits an agent's adaptability. True cognitive flexibility requires the ability to learn effective strategies for self-regulation and resource management based on experience. We explore potential mechanisms, primarily drawing from reinforcement learning (RL) and related fields, by which an agent might learn optimal policies like $\pi_{effort}$ and the broader meta-control policy $\pi_{regulate}$ that governs overall cognitive strategy.

\begin{figure}[htbp]
	\centering
	\input{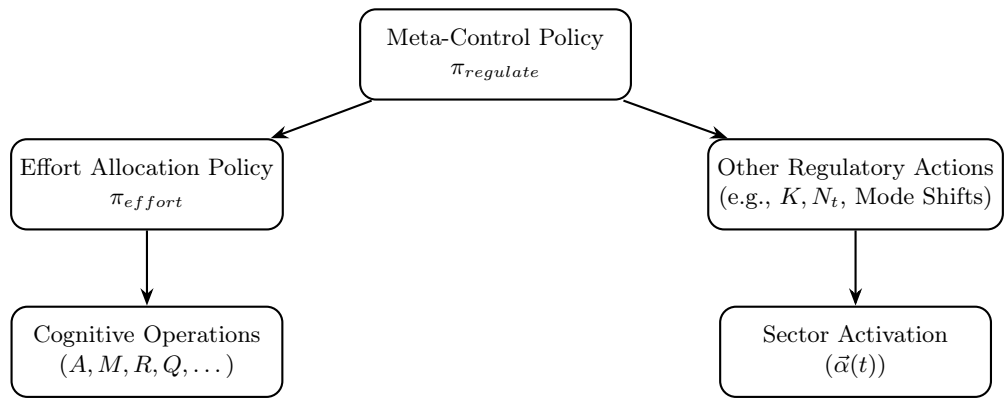}
	\caption{Hierarchical structure of learned regulatory policies. The meta-control policy $\pi_{regulate}$ decides when and how to intervene in cognitive processes. It may trigger effort reallocation via $\pi_{effort}$ or other regulatory actions such as sectoral nullification ($N_t$) or annihilation ($K$). These influence the underlying cognitive operations and sector activations.}
	\label{fig:hierarchical-policy-learning}
\end{figure}

\section{Learning Effort Allocation Policies (\texorpdfstring{$\pi_{\text{effort}}$}{pi\_effort})}

Chapter~\ref{chap:SemanticFocus} defined the effort allocation policy $\pi_{effort} : \Phi \times \text{Context} \rightarrow \text{Allocation}$ which distributes the available effort resource $E_{total}$ across competing cognitive demands. Learning an effective $\pi_{effort}$ is crucial for efficient focus and attention management. Reinforcement Learning provides a natural framework:

\begin{itemize}
	\item \textbf{State Representation:} The RL state ($s$) could incorporate features extracted from the current belief state $\phi$, particularly meta-cognitive information from $\Sigma_{refl}$ (outputs of $M$), such as current load $\lambda$, coherence $\kappa$, goal priorities, trajectory deviation $\theta$, remaining capacity $E_{total}$, etc.
	\item \textbf{Action Space:} The actions ($a$) correspond to different ways of allocating effort $\vec{\epsilon} = (\epsilon_1, \dots, \epsilon_m)$ across targets (operators, sectors, tasks), subject to $\sum \epsilon_i \le E_{total}$. This might involve selecting high-level allocation strategies (e.g., "focus on planning," "monitor environment," "restore coherence").
	\item \textbf{Reward Signal ($R(s, a, s')$):} Defining an effective reward signal is critical for learning useful policies. The reward function could be based on:
	\begin{itemize}
		\item \textbf{Internal Metrics:} Providing rewards for desirable internal states resulting from an effort allocation. Examples: +1 for maintaining high coherence $\kappa$, +1 for staying aligned with orientation axes (low $\theta$), -1 for exceeding load thresholds $\lambda$, +1 for making progress towards goals represented in $\Sigma_{plan}$. Designing these internal rewards requires careful consideration to ensure they align with overall desired cognitive function.
		\item \textbf{External Task Performance:} More directly, rewards can be based on successful completion of external tasks that depend on effective internal resource management. An allocation policy leading to faster or more accurate task completion receives higher reward.
		\item \textbf{Efficiency Metrics:} Rewards could incorporate efficiency, e.g., rewarding goal achievement while penalizing excessive effort $\epsilon$ expenditure, encouraging the agent to learn resource-frugal strategies.
	\end{itemize}
	\item \textbf{Learning Algorithms:} Standard RL algorithms (Q-learning, policy gradients, actor-critic methods) could be adapted to learn the mapping from state features to optimal effort allocation actions, using the defined reward signals.
	\item \textbf{Imitation Learning:} $\pi_{effort}$ could also be learned by imitating the allocation patterns of successful demonstrators (human or other AI), potentially bootstrapping the RL process.
	\item \textbf{Meta-Learning:} Learning algorithms could adapt the parameters of $\pi_{effort}$ itself based on performance across different types of tasks or contexts (learning how to allocate effort differently in different situations).
\end{itemize}
Learning $\pi_{effort}$ allows the agent to develop sophisticated, context-sensitive attentional control strategies beyond simple heuristics, optimized according to specified internal or external objectives captured in the reward function.

\section{Learning Meta-Control Policies (\texorpdfstring{$\pi_{\text{regulate}}$}{pi\_regulate})}

Beyond allocating effort, the agent employs broader meta-control policies ($\pi_{regulate}$) to manage its overall cognitive state and strategy. These policies decide when to trigger specific regulatory actions based on introspective inputs assimilated via $M$. Learning $\pi_{regulate}$ enables adaptive self-governance.

\begin{itemize}
	\item \textbf{Learning Triggers:} RL or supervised learning could train the agent to recognize internal states (specific patterns of $\kappa, \lambda, \theta, \vec{\eta}$ divergence reported by $M$) that warrant intervention. The state representation for the learning algorithm would heavily rely on meta-beliefs in $\Sigma_{refl}$.
	\item \textbf{Learning Action Selection:} Given a problematic state detected via introspection, $\pi_{regulate}$ must select the appropriate corrective action. Learning could determine whether it's better to:
	\begin{itemize}
		\item Initiate re-orientation (Chapter~\ref{chap:SemanticOrientation}).
		\item Trigger Annihilation ($K$, Chapter~\ref{chap:Annihilation}) of conflicting subsystems ($K_{\Sigma}$).
		\item Apply targeted Nullification ($N_t$, Chapter~\ref{chap:Nullification}) to prune irrelevant data ($N^{\Sigma}_t$).
		\item Engage in deep reflection ($M$ loop) to resolve inconsistencies.
		\item Shift cognitive mode (adjust sector activations $\vec{\alpha}(t)$).
		\item Adjust parameters of object-level operators (e.g., make $A$ more conservative).
	\end{itemize}
	The RL reward signal would reflect the effectiveness of the chosen regulatory action in restoring desired state properties (e.g., reward for increase in $\kappa$, decrease in $\theta$, successful goal completion following intervention).
	\item \textbf{Relation to $\pi_{effort}$:} $\pi_{regulate}$ likely operates at a higher level, potentially setting goals or constraints (like "prioritize coherence restoration") for the more fine-grained allocation policy $\pi_{effort}$. Learning must coordinate these levels, perhaps through hierarchical RL.
\end{itemize}

\begin{figure}[htbp]
	\centering
	\input{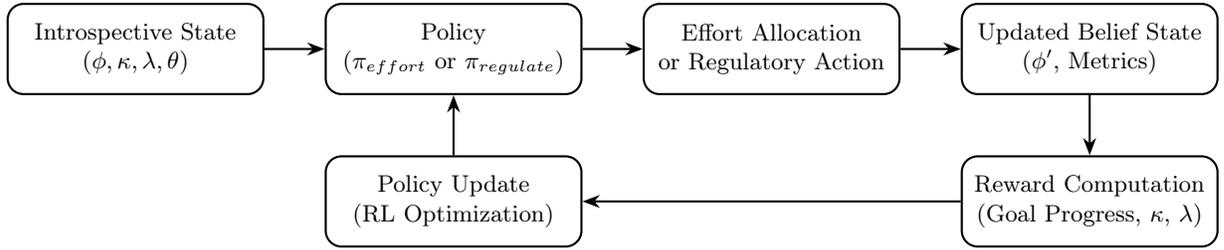}
	\caption{Reinforcement learning loop for acquiring regulatory policies. The agent uses introspective features (e.g., $\kappa$, $\lambda$, $\theta$) as state input. Policies ($\pi_{effort}$ or $\pi_{regulate}$) propose actions such as reallocating effort or triggering regulatory interventions. The resulting state changes and rewards guide the learning of more effective self-regulation strategies over time.}
	\label{fig:rl-training-loop}
\end{figure}

Learning $\pi_{regulate}$ allows the agent to develop sophisticated self-healing and self-stabilizing capabilities tailored to its own dynamics and environment, moving beyond fixed responses to internal states.

\section{Learning Cognitive Trade-offs}

Many cognitive tasks involve inherent trade-offs (e.g., speed vs. accuracy, exploration vs. exploitation, breadth vs. depth). Effective regulatory policies ($\pi_{effort}, \pi_{regulate}$) must learn to navigate these trade-offs appropriately.

\begin{itemize}
	\item \textbf{Speed-Accuracy Trade-off:} Learning might adjust how much effort $\epsilon$ is allocated by $\pi_{effort}$ to deliberation ($M, \Sigma_{refl}$) versus rapid execution (Activation Basins, Chapter~\ref{chap:SemanticExecution}) based on task deadlines or importance reflected in the reward function.
	\item \textbf{Exploration-Exploitation Trade-off:} Learning could shape policies ($\pi_{effort}$) for allocating effort between assimilating novel external information ($A$) versus consolidating internal knowledge ($M$, re-anchoring via internal processing), balancing the need for new data against robust understanding. Rewards might depend on long-term performance across diverse situations.
	\item \textbf{Learning Context-Sensitivity:} Policies should learn to apply different trade-off criteria in different situations (e.g., prioritize accuracy in high-stakes decisions, prioritize speed in routine tasks). This requires the learning state representation to include relevant contextual features, and the reward function to reflect context-dependent success.
\end{itemize}

\begin{figure}[htbp]
	\centering
	\input{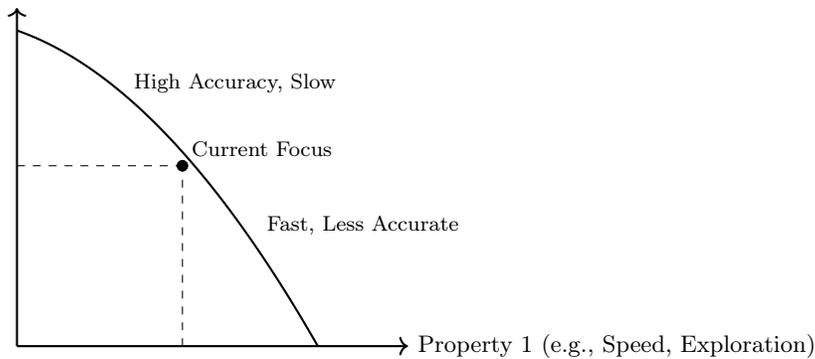}
	\caption{Example cognitive trade-off surface (e.g., speed-accuracy or exploration-exploitation). The agent must select an operating point along the curve based on goals, context, and learned regulatory policies ($\pi_{regulate}$). Learning aims to optimize trade-off navigation over time.}
	\label{fig:tradeoff-surface}
\end{figure}

Learning to manage these trade-offs effectively is a hallmark of advanced intelligence and adaptability.

\section{Challenges in Learning Regulatory Policies}

Learning effective self-regulatory policies faces several challenges:
\begin{itemize}
	\item \textbf{Credit Assignment Problem:} It can be difficult to determine which specific allocation decision ($\pi_{effort}$) or regulatory action ($\pi_{regulate}$) led to a distant positive or negative outcome, especially when using sparse external rewards or complex internal metrics like $\kappa$.
	\item \textbf{Defining Internal Rewards:} Crafting appropriate internal reward functions (based on $\kappa, \lambda, \theta$, etc.) that genuinely promote desired long-term cognitive behavior without leading to unintended consequences or "reward hacking" is non-trivial.
	\item \textbf{Stability during Learning:} Policies that control core functions like assimilation or nullification must be learned carefully to avoid destabilizing the entire cognitive system during the learning process itself. Exploration must be managed cautiously.
	\item \textbf{Sample Efficiency:} Learning complex control policies, especially via RL, often requires vast amounts of experience, which may be costly or time-consuming to obtain in simulation or real-world interaction.
\end{itemize}

Addressing these challenges likely requires sophisticated learning algorithms (e.g., hierarchical RL, model-based RL), careful reward shaping, curriculum learning, imitation learning, or bootstrapping methods.

\section{Conclusion: Towards Adaptive Self-Governance}

This chapter explored how the meta-cognitive control layer of the Semantic Manifold agent---specifically the effort allocation policy $\pi_{effort}$ and the broader regulatory policy $\pi_{regulate}$---could transition from being fixed designs to adaptive systems shaped by learning. By leveraging frameworks like Reinforcement Learning, guided by carefully designed reward signals based on internal states and external performance, agents can potentially learn optimal strategies for directing focus, managing resources, triggering self-correction, and navigating cognitive trade-offs.

The ability to learn regulatory policies is paramount for achieving robust, flexible, and truly autonomous intelligence. It allows agents to fine-tune their own cognitive operations and adapt their internal "style of thought" to meet the demands of complex and changing environments. While significant challenges remain, particularly in defining appropriate rewards and ensuring stable learning, integrating policy learning into the framework provides a pathway towards agents that not only possess structured beliefs but also possess learned wisdom in how to manage them effectively.


\subsection*{Chapter Summary}
Building upon the learning of semantic structures and operators, this chapter addresses the acquisition and refinement of the regulatory policies themselves. It focuses on how agents might learn effective strategies for cognitive control, particularly the effort allocation policy ($\pi_{effort}$) and the overarching meta-control policy ($\pi_{regulate}$). Reinforcement learning is presented as a key framework, where policies learn to map internal state information (derived from meta-introspection $M$, including load $\lambda$, coherence $\kappa$, orientation $\theta$, etc.) to actions such as effort distribution ($\vec{\epsilon}$) or triggering specific regulatory interventions (e.g., re-orientation, $N_t^{\Sigma}, K_{\Sigma}$). The reward signal for learning can be based on external task performance or internal metrics reflecting cognitive efficiency and stability. Learning these policies enables agents to adaptively manage resources, navigate cognitive trade-offs (like speed vs. accuracy), trigger self-correction, and achieve sophisticated, context-sensitive self-governance, despite challenges related to credit assignment and reward definition.
	\chapter{Architectural Adaptation (Parameter \texorpdfstring{$\theta$}{theta} Learning)}

\label{chap:ArchitecturalAdaptation}

\section{Introduction: Beyond Policy Learning - The Plastic Architecture?}

The preceding chapters in this Part (Chapters~\ref{chap:LearningSemanticStructures}-\ref{chap:LearningRegulatoryPolicies}) explored how an agent operating within the Semantic Manifold framework might learn core components of its cognitive machinery: the structure of its belief space $\Phi$ (metric $d$, sectors $\Sigma$), the function of its cognitive operators ($A, M, \Lambda, V, R, Q$), and the regulatory policies ($\pi_{effort}, \pi_{regulate}$) that govern its internal dynamics. These learning mechanisms allow for significant adaptation within a given architectural configuration defined by the parameter vector $\theta$ (Chapter~\ref{chap:ParameterizedArchitectures}).

This chapter takes a \textbf{highly speculative step further}, asking: could the fundamental architectural parameters $\theta$ themselves be subject to learning or adaptation? Can an agent learn not just how to operate within its cognitive architecture, but potentially learn to change the architecture itself based on long-term experience or overarching goals?

While learning core architectural parameters represents a \textbf{profound theoretical and practical challenge}, exploring this possibility is important for understanding pathways towards truly general intelligence capable of radical adaptation. If an agent could adapt its memory horizon ($\mu$), its representational mode ($\rho$), or even its model of identity ($\eta_{type}$), it could potentially reconfigure its cognitive capabilities far beyond what's possible by merely tuning operators or policies within a fixed $\Phi^{[\theta]}$ structure. This chapter briefly explores the conceptual space of architectural plasticity within the Semantic Manifold framework, acknowledging its speculative nature and significant hurdles.

\section{Adapting Core Architectural Parameters}

Which parameters within $\theta = (\mu, \delta_{profile}, \tau, \rho, \eta_{type}, C, \Gamma_{type}, O)$ might conceivably be adaptable through some form of meta-learning or developmental process, however complex?
\begin{itemize}
	\item \textbf{Memory and Context ($\mu, C$):} An agent might hypothetically learn an optimal trade-off between memory persistence ($\mu$) / context scope ($C$) and computational cost ($\lambda, \epsilon$). It could potentially learn to dynamically adjust these parameters based on task demands or available resources, effectively changing its working memory capacity or long-term retention strategy, although the mechanisms for such stable adaptation are unclear.
	\item \textbf{Decay and Persistence ($\delta_{profile}, \tau$):} The profile of forgetting ($\delta_{profile}$) or the degree of reliance on past states ($\tau$) could potentially be tuned over long timescales. For instance, an agent might learn that faster decay is beneficial in highly dynamic environments, while greater persistence is needed for tasks requiring long-range coherence. Learning these fundamental rates stably seems challenging.
	\item \textbf{Representation Mode ($\rho$):} Could an agent learn to shift its representational strategy? Perhaps learning to utilize symbolic structures more heavily for tasks demanding precision, while relying on embeddings for tasks requiring generalization or analogy. This implies a highly complex meta-learning capability capable of managing fundamentally different representational schemes.
	\item \textbf{Identity Model ($\eta_{type}$):} Could the very nature of the agent's identity evolve? Perhaps starting with no identity ($\eta_{type} = \text{none}$) and developing an emergent one ($\eta_{type} = \text{emergent}$) through prolonged self-modeling and interaction, representing a form of cognitive maturation. This involves learning at the deepest level of self-representation.
	\item \textbf{Goal Decomposition ($\Gamma_{type}$):} The sophistication of the planning system itself might be learnable, moving from simple prompted goals to complex learned hierarchical planning systems, representing adaptation in core reasoning capabilities.
	\item \textbf{Observation Modalities ($O$):} While physical sensors might be fixed, an agent could potentially learn to integrate new information sources (e.g., learning to use a new tool or API effectively modifies its functional $O$), which is perhaps the most feasible form of architectural adaptation discussed here.
\end{itemize}
Adaptation here refers to changes in the parameters defining the structure $\Phi^{[\theta]}$ and its core operator family $O^{[\theta]}$, representing a deeper, more challenging level of learning than covered previously.

\begin{figure}[htbp]
	\centering
	\input{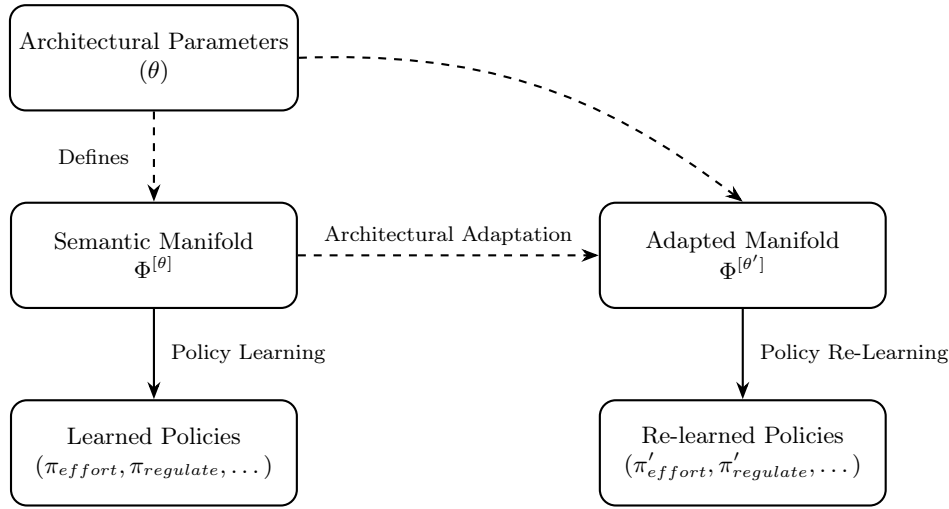}
	\caption{Two-tier learning structure. Standard policy learning (e.g., for $\pi_{effort}$, $\pi_{regulate}$) occurs within a fixed architecture $\Phi^{[\theta]}$. Over longer timescales, speculative architectural adaptation mechanisms could modify $\theta$, producing a new manifold $\Phi^{[\theta']}$ and necessitating re-learning of cognitive policies under the changed architecture.}
	\label{fig:architectural-adaptation-layers}
\end{figure}

\section{Potential Mechanisms for Architectural Learning}

How could such fundamental parameters be learned? This likely requires processes operating at longer timescales or higher levels of abstraction than the operator and policy learning discussed in Chapters \ref{chap:LearningCognitiveOperators} and \ref{chap:LearningRegulatoryPolicies}. \textbf{These mechanisms are largely theoretical possibilities rather than established techniques:}
\begin{itemize}
	\item \textbf{Evolutionary Algorithms / Population-Based Methods:} Different $\theta$ configurations could be treated as genotypes. In simulated multi-agent environments or through internal "population" dynamics, configurations leading to greater long-term success (survival, task completion, goal achievement) could be selected or recombined over generations. This operates at a very slow timescale.
	\item \textbf{Developmental Programs:} Architectural parameters might change according to an intrinsic developmental trajectory, perhaps triggered by experience but following a pre-structured (potentially learned) developmental plan. This mirrors cognitive development in biological systems but requires designing the developmental program itself.
	\item \textbf{Higher-Level Reinforcement Learning (Meta-Learning):} An RL agent could operate at a meta-level, where its "actions" are modifications to parameters in $\theta$, and its "rewards" are based on the long-term performance of the resulting cognitive architecture across a range of tasks. This faces extreme challenges in state representation, action space definition, and credit assignment.
	\item \textbf{Explicit Meta-Cognitive Control:} A highly sophisticated agent with deep introspection ($M$, Chapter~\ref{chap:MetaAssimilation}) might be able to explicitly reason about its own architecture ($\theta$) and deliberately modify parameters based on self-assessment and anticipated needs, although this implies extremely advanced self-modeling capabilities far beyond current systems.
\end{itemize}

These mechanisms are speculative and represent significant research challenges, operating at the intersection of machine learning, artificial life, and developmental robotics.

\begin{figure}[htbp]
	\centering
	\input{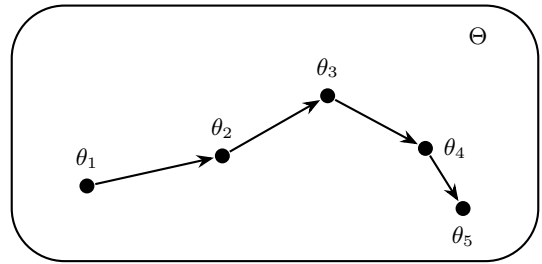}
	\caption{Conceptual trajectory through architecture space $\Theta$. Each point $\theta_i$ defines a semantic manifold $\Phi^{[\theta_i]}$. Evolutionary processes, meta-learning, or developmental adaptation may drive gradual or punctuated shifts across architecture configurations. The agent's cognitive structure and capabilities evolve alongside these transitions.}
	\label{fig:architecture-evolution}
\end{figure}

\section{Challenges, Stability, and Scalability}

Learning or adapting the core architecture $\theta$ is fraught with difficulties, \textbf{making it a high-risk, high-reward research direction}:

\begin{itemize}
	\item \textbf{Stability:} Modifying fundamental parameters like memory decay ($\delta_{profile}$) or representation mode ($\rho$) risks destabilizing the entire cognitive system, potentially leading to catastrophic forgetting, loss of coherence ($\kappa$), or complete functional collapse. Ensuring stability during architectural adaptation is paramount and extremely difficult.
	\item \textbf{Timescales:} Architectural learning likely occurs over much longer timescales than operator or policy learning, requiring extensive experience or simulation, making it difficult to observe or optimize practically.
	\item \textbf{Credit Assignment:} Attributing long-term outcomes (potentially over thousands or millions of operational steps) to specific architectural parameter settings ($\theta$) is an extremely challenging credit assignment problem.
	\item \textbf{Computational Cost:} Exploring the space of possible architectures $\Theta$ is computationally vast. Evaluating the fitness of each $\theta$ requires instantiating and running the corresponding agent, likely for extended periods.
	\item \textbf{Semantic Continuity:} How can semantic meaning and identity ($\vec{\eta}$) be preserved across significant architectural shifts induced by changing $\theta$? Adapting the architecture might fundamentally alter the nature of the agent's beliefs and self-concept.
\end{itemize}

These challenges suggest that significant architectural plasticity might be rare, highly constrained, or operate only within narrow bounds in practical artificial agents, perhaps mirroring how fundamental brain structures are relatively fixed in biological organisms after early development.

\section{Conclusion: The Prospect of Evolving Architectures}

This chapter briefly considered the potential for agents built on the Semantic Manifold framework to learn or adapt not just their knowledge, skills, or control policies, but their fundamental cognitive architecture as defined by the parameter vector $\theta$. While \textbf{highly speculative and facing immense challenges} regarding stability, timescale, and computational cost, the possibility of architectural adaptation represents a theoretical frontier for artificial general intelligence.

Mechanisms like evolutionary methods, simulated development, or meta-level reinforcement learning offer potential conceptual pathways, however complex and currently impractical. Even limited architectural plasticity---perhaps tuning memory parameters ($\mu, C$) or decay profiles ($\delta_{profile}$) within stable bounds---could significantly enhance an agent's long-term adaptability. The parameterization $\Phi^{[\theta]}$ introduced in Chapter~\ref{chap:ParameterizedArchitectures} provides the necessary conceptual hooks for exploring these possibilities. While the core monograph focuses primarily on dynamics within a given $\Phi^{[\theta]}$, acknowledging the potential for $\theta$ itself to evolve opens avenues for future research into radically adaptive artificial minds, albeit with significant caution regarding feasibility and stability.


\subsection*{Chapter Summary}
This chapter speculatively considers the possibility of learning and adaptation occurring at the level of the core architectural parameters ($\theta$) themselves, moving beyond the learning of structures, operators, and policies within a fixed architecture $\Phi^{[\theta]}$. It explores the hypothetical potential for an agent to adapt fundamental aspects like its memory horizon ($\mu$), decay profile ($\delta_{profile}$), representation mode ($\rho$), or even its identity model ($\eta_{type}$) based on long-term experience. Potential mechanisms, such as evolutionary algorithms, developmental programs, or high-level reinforcement learning acting on $\theta$, are briefly discussed. However, the chapter emphasizes the profound theoretical and practical challenges involved, including maintaining stability, the long timescales required, computational cost, and the difficulty of credit assignment. While acknowledging its speculative nature, the prospect of architectural plasticity represents a frontier for achieving more radically adaptive artificial intelligence.
	
	\part{Social Cognition and Multi-Agent Dynamics}
	\label{part:social_cognition}
	
	\chapter{Modeling Other Agents within \texorpdfstring{$\Phi$}{Phi} (Theory of Mind)}
\label{chap:ModelingOtherAgents}

\section{Introduction: The Social Manifold}

The preceding Parts have largely focused on the internal cognitive architecture and dynamics of a single agent operating within its semantic state space $\Phi$. We have explored how such an agent forms beliefs, structures knowledge, manages memory, regulates its internal state, learns, and connects thought to action. However, much of intelligent behavior, particularly in humans and sophisticated AI, unfolds within a social context, requiring interaction with, understanding of, and prediction about other agents who possess their own beliefs, goals, and perspectives.

This Part extends the Semantic Manifold framework into the domain of social cognition. This first chapter addresses the fundamental capability often referred to as "Theory of Mind" (ToM): the ability of an agent (Agent A) to represent and reason about the internal mental states (beliefs, desires, intentions, emotions) of another agent (Agent B). Modeling ToM is crucial for explaining and generating complex social behaviors like cooperation, competition, communication, empathy, and deception. We explore potential ways an agent A, equipped with a semantic manifold $\Phi_A$, might represent and infer the state $\phi_B$ of another agent B.

\section{Representational Schemes for Others' Mental States}

How can Agent A represent Agent B's presumed belief state $\phi_B$ within its own semantic space $\Phi_A$? Several representational schemes, potentially varying with the agent's architecture $\theta_A$, are conceivable:

\begin{itemize}
	\item \textbf{Tagged Belief Fragments:} Agent A could store beliefs explicitly attributed to Agent B using tags or meta-data associated with linguistic fragments $\varphi_i$. For example, within $\phi_A$, there might exist $\varphi_{123} =$ "Belief(AgentB, 'The key is under the mat')". These tagged beliefs might reside within A's reflective sector ($\Sigma_{refl}$) or perhaps a dedicated social sector ($\Sigma_{social}$).
	\textbf{Example:} $\phi_A$ contains the fragment $\varphi_{rep\_B} = $ \{"Source: AgentB", "Content: 'It will rain tomorrow'", "Confidence (B): High"\}, likely within $\Sigma_{refl, A}$ or $\Sigma_{social, A}$.
	
	\item \textbf{Dedicated 'Other-Agent' Structures:} Agent A might maintain regions within its $\Phi_A$ dedicated to modeling specific other agents. For example, a state $\phi_{model(B)} \in \Phi_A$ could represent A's current best estimate of B's belief state $\phi_B$, potentially including B's likely goals ($\Gamma_B$), coherence ($\kappa_B$), or even affective state (if Part~\ref{part:semantic_memory} is considered). This allows for more integrated reasoning about B's state.
	\textbf{Example:} $\Phi_A$ contains a dedicated belief ensemble $\phi_{model(B)} = $ \{"Goal(B): Obtain item X", "Belief(B): Item X is at location Y", "Constraint(B): Avoid location Z"\}, representing A's model of B's relevant state.
	
	\item \textbf{Simulation-Based Representations:} Agent A could use its own Embodied Simulation capabilities (Chapter~\ref{chap:EmbodiedSimulation}) to model Agent B. By simulating B's likely perceptions ($X_B(s)$) and internal dynamics (perhaps using a simplified or assumed $\theta_B$), Agent A generates a simulated trajectory $\gamma_{B,sim}(t)$ whose endpoint $\phi_{B,sim}$ serves as a representation of B's likely state. This leverages A's own cognitive machinery for perspective-taking.
	\textbf{Example:} Agent A simulates Agent B seeing event E, applies a simplified version of B's likely assimilation operator $A_B$, resulting in a predicted belief state $\phi_{B,sim}$ within A's simulation space.
\end{itemize}

\begin{figure}[htbp]
	\centering
	\input{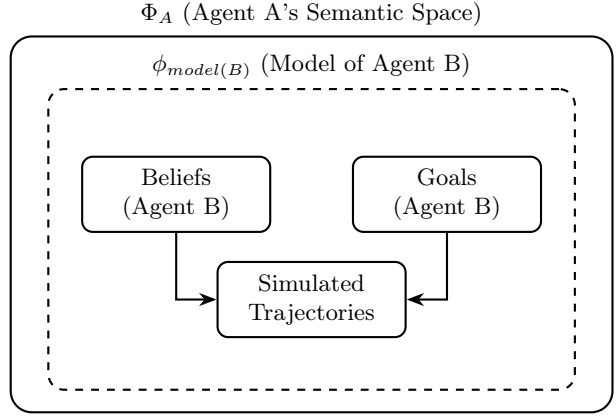}
	\caption{Agent A's semantic space \(\Phi_A\) contains a structured subregion modeling Agent B's presumed belief state \(\phi_{model(B)}\), including inferences about B's goals, beliefs, and emotions.}
	\label{fig:phi_model_b}
\end{figure}

The choice of representation impacts how easily Agent A can update its model of Agent B, reason about B's potential actions, or detect inconsistencies in B's behavior.

\section{Inferring Others' Mental States}

Representing others' states requires mechanisms for inferring those states, as they are not directly observable. Agent A can use several sources of information, processed through its own operators:

\begin{itemize}
	\item \textbf{Inference from Observation ($X$):} Agent A observes Agent B's actions $s_{B,behavior}$. By processing this via its own Observation Encoding $X$ and potentially using internal simulation or inverse planning logic (perhaps within $\Sigma_{plan}$ or $\Sigma_{refl}$), Agent A can infer likely goals or beliefs $\phi_B$ that would explain B's observed behavior.
	\item \textbf{Inference from Communication ($A_{text}$):} Agent B might explicitly communicate its beliefs or goals ($\phi_{B,stated}$). Agent A assimilates this linguistic input via $A_{text}$ (Chapter~\ref{chap:Assimilation}). Forming beliefs about $\phi_B$ from $\phi_{B,stated}$ requires assessing B's reliability, potential deception, and aligning linguistic expressions (related to Gauge Equivalence, Chapter~\ref{chap:SemanticGauge}).
	\item \textbf{Simulation and Perspective-Taking:} Agent A can use its simulation capability (Chapter~\ref{chap:EmbodiedSimulation}) to "put itself in B's shoes." By simulating the world from B's likely perspective (based on beliefs about B's location, perception $O_B$, goals $\Gamma_B$), Agent A can generate hypotheses about $\phi_B$.
	\item \textbf{Meta-Cognitive Insights ($M$):} Agent A might use its own meta-cognitive insights (generated via $M$) about how certain situations affect its own beliefs $\phi_A$ to infer how similar situations might affect $\phi_B$, assuming some shared cognitive principles.
\end{itemize}

\begin{figure}[htbp]
	\centering
	\input{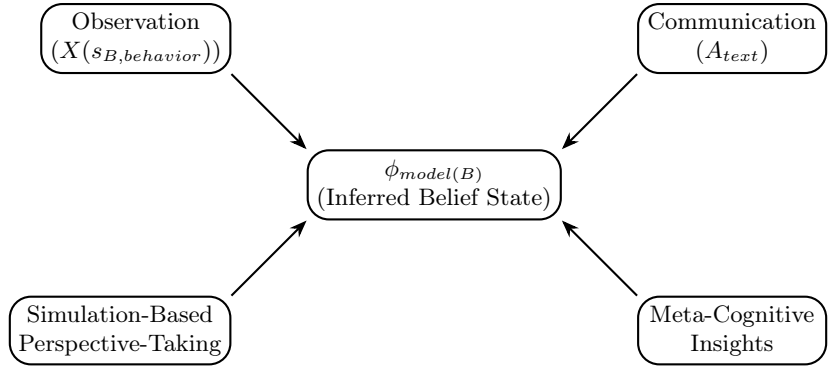}
	\caption{Multiple pathways---observation of behavior, communication, simulation, and meta-cognitive insights---contribute to Agent A's construction of an inferred belief model \(\phi_{model(B)}\) representing Agent B's presumed internal state.}
	\label{fig:inference-other-agent}
\end{figure}

These inference processes update Agent A's representation of $\phi_B$ (whether tagged beliefs or dedicated structures). This representation is inherently uncertain and subject to revision based on new evidence.

\section{Utilizing Models of Other Agents}

The models Agent A forms about Agent B's internal state $\phi_B$ are not merely passive representations; they actively inform A's own cognitive processes and behavior:

\begin{itemize}
	\item \textbf{Strategic Planning ($\Sigma_{plan}$):} When planning actions involving Agent B, Agent A's planning process within $\Sigma_{plan}$ should consult the model of $\phi_B$. This allows anticipating B's likely responses, planning cooperative actions based on shared goals inferred in $\phi_B$, or developing counter-strategies based on conflicting goals.
	\item \textbf{Communication Strategy:} Agent A can tailor its communication (generated via $V$) based on its model of $\phi_B$. It might simplify explanations, provide necessary background, avoid sensitive topics, or frame arguments persuasively based on B's presumed beliefs and goals.
	\item \textbf{Action Execution (Part~\ref{part:embodiment_and_action}):} Predictions about Agent B's behavior, derived from the model $\phi_B$, can directly influence Agent A's action selection and execution (via activation basins $\mathcal{A}_a$), allowing for smoother coordination or avoidance of negative interactions.
	\item \textbf{Social Regulation:} Agent A might regulate its own state ($\kappa, \theta$) or behavior based on its understanding of social norms or B's expectations, as represented in its model of B or broader social knowledge within $\Phi_A$.
\end{itemize}
The ability to accurately model and utilize information about other agents' mental states is thus a powerful extension of the core cognitive capabilities, enabling sophisticated social interaction.

\section{Conclusion: Towards Socially Aware Semantic Agents}

This chapter outlined how the Semantic Manifold framework could be extended to incorporate Theory of Mind capabilities---the representation and inference of other agents' internal states. By leveraging mechanisms like tagged beliefs, dedicated representational structures, internal simulation, and inference from observation and communication, an agent A can build models of other agents B within its own semantic space $\Phi_A$.

These models ($\phi_B$) are crucial for enabling effective social reasoning, communication, and interaction. While implementing robust ToM remains a significant challenge for AI, this framework provides conceptual tools for representing the necessary knowledge and integrating it with the agent's planning, reasoning, and action systems. The next chapter will explore the dynamics of interaction between such socially aware agents, focusing on communication and alignment.


\subsection*{Chapter Summary}
This chapter extends the Semantic Manifold framework into the realm of social cognition, addressing how an agent (Agent A) might implement Theory of Mind (ToM)---the ability to represent and reason about the internal mental states of another agent (Agent B). It explores potential schemes for representing Agent B's presumed belief state ($\phi_B$) within Agent A's own semantic space ($\Phi_A$), including tagged belief fragments, dedicated 'other-agent' model structures ($\phi_{model(B)}$), or leveraging internal simulation capabilities. The chapter discusses how Agent A might infer Agent B's state through various mechanisms, such as interpreting observed behavior, processing communication acts (via assimilation $A$), simulating B's perspective, or using meta-cognitive analogies. These inferred models of other agents are then utilized to inform Agent A's own strategic planning, communication strategies, and action execution, enabling more sophisticated and effective social interaction.
	\chapter{Communication, Alignment, and Shared Belief Spaces}
\label{chap:CommunicationAlignment}

\section{Introduction: From Individual Minds to Interaction Dynamics}

Chapter~\ref{chap:ModelingOtherAgents} explored how an agent operating within the Semantic Manifold framework might represent and reason about the internal states of other agents. This provides the foundation for individual social reasoning (Theory of Mind). However, social intelligence fully manifests in the dynamics of interaction between agents. How do communication acts affect belief states? How do agents achieve mutual understanding or semantic alignment? Can groups of agents develop shared belief structures?

This chapter shifts focus from the individual modeling of others to the processes governing interaction and shared cognition among multiple Semantic Manifold agents (Agent A, Agent B, etc.). We examine how communicative acts can be modeled as operations influencing the respective belief states ($\phi_A, \phi_B$), the crucial challenge of achieving semantic alignment necessary for meaningful communication, and speculate on more advanced concepts like distributed belief states or collective coherence ($\kappa_{collective}$). Understanding these interaction dynamics is essential for designing agents capable of effective collaboration, negotiation, and participation in social systems.

\section{Modeling Communication Acts as Belief State Operations}

Communication fundamentally involves transmitting information with the intent to influence the belief state of another agent. Within the Semantic Manifold framework, communicative acts (speech acts) can be modeled as operations triggered by the sender (Agent A) that generate input for the receiver's (Agent B) assimilation operators ($A$ or $M$).
\begin{itemize}
	\item \textbf{Assertion (`Assert(A, B, $\varphi$)`):} Agent A generates a linguistic expression $\varphi$ (via $V$) intended to be assimilated by Agent B. Agent B processes this via $A_{text}$, potentially updating $\phi_B$. Success depends on B's assessment of A's credibility (part of B's model of A, Chapter~\ref{chap:ModelingOtherAgents}) and coherence with $\phi_B$.
	\item \textbf{Query (`Query(A, B, $?\varphi$)`):} Agent A expresses an information need (a query $\phi_{query}$ formulated perhaps via $Q$). Agent B processes this, potentially triggering its own retrieval ($R$) and generating a response (assertion) via $V$.
	\item \textbf{Command (`Command(A, B, Action)`):} Agent A expresses a directive. Agent B processes this, potentially creating a new goal in its $\Sigma_{plan}$ or triggering an action via its Execution system (Part~\ref{part:embodiment_and_action}), mediated by B's goals and model of A.
	\item \textbf{Expression of Internal State (`Express(A, B, $\psi$)`):} Agent A communicates an internal state (e.g., an affective state $\psi \in \Sigma_{affect, A}$; or a meta-belief $\psi \in \Sigma_{refl, A}$ about its own $\kappa$ or $\lambda$). Agent B assimilates this via $A_{text}$ or $M$ (if interpreting it as meta-information about A), updating its model of A's internal state $\phi_A$.
\end{itemize}

\begin{figure}[htbp]
	\centering
	\input{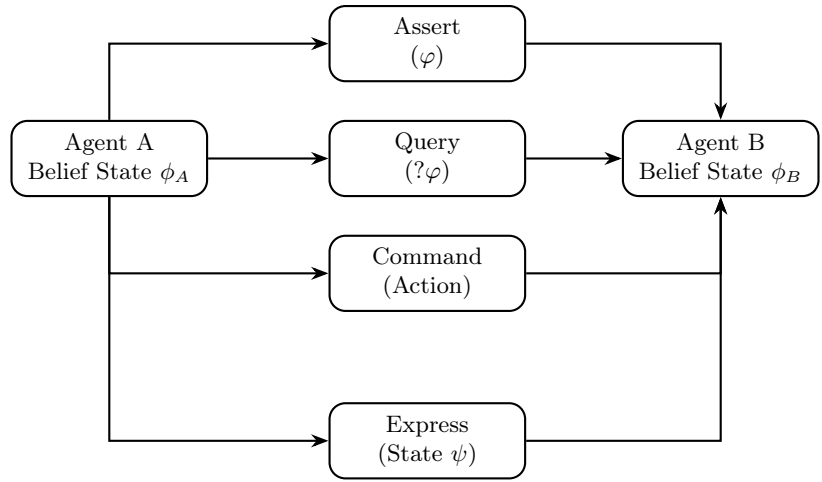}
	\caption{Modeling communication as operations that transform belief states. Agent A produces different communicative acts (Assertion, Query, Command, Expression) intended to influence Agent B's belief state $\phi_B$ through assimilation, meta-assimilation, or retrieval operators.}
	\label{fig:communication-acts}
\end{figure}

Modeling communication requires considering the dialogue context (shared history within interaction), the agents' models of each other (Chapter~\ref{chap:ModelingOtherAgents}), and the pragmatic intent behind utterances. The operators $A, M, Q, V, R$ provide the building blocks for these interactions within each agent's $\Phi$.

\section{Semantic Alignment and Grounding in Interaction}

Meaningful communication requires more than just exchanging symbols; it necessitates some degree of semantic alignment between the interacting agents. How can Agent A be confident that Agent B interprets an expression $\varphi$ in a functionally equivalent way? This involves aligning aspects of their respective semantic manifolds, $\Phi_A$ and $\Phi_B$.
\begin{itemize}
	\item \textbf{Achieving Shared Gauge ($\sim_{gauge}$):} Agents need to converge on using expressions that are functionally equivalent within the context of their interaction (related to Chapter~\ref{chap:SemanticGauge}). This might involve learning mappings between their internal representations or adopting shared linguistic conventions through feedback and clarification dialogues.
	\item \textbf{Aligning Semantic Metrics ($d$):} For communication about similarity or categorization to succeed, agents may need to implicitly or explicitly align their internal semantic distance functions ($d_A, d_B$). This could occur through shared learning experiences or negotiation.
	\item \textbf{Grounding through Joint Action and Observation:} As discussed in Chapter~\ref{chap:GroundingSemanticBelief}, shared interaction with a common environment provides a powerful basis for grounding. When Agent A and Agent B jointly attend to and act upon the same external objects or events ($s \in S$), they can co-calibrate the meaning of the corresponding belief fragments formed via $X$. Pointing, gesturing, and other embodied signals facilitate this joint grounding.
	\item \textbf{Feedback and Repair Mechanisms:} Communication breakdown due to misalignment triggers repair sequences (e.g., requests for clarification, paraphrasing, providing examples via $V$) aimed at re-establishing shared understanding. Meta-cognition ($M$) about communicative success is crucial here.
\end{itemize}

\begin{figure}[htbp]
	\centering
	\input{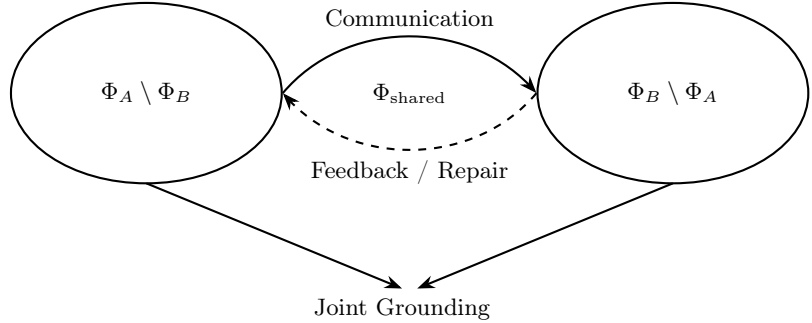}
	\caption{Semantic alignment between two agents. Communication, feedback, and joint grounding experiences dynamically expand the shared belief subspace \(\Phi_{\text{shared}} \subset \Phi_A \cap \Phi_B\), allowing for progressively more reliable mutual understanding.}
	\label{fig:semantic-alignment-process}
\end{figure}

Semantic alignment is an ongoing, dynamic process, facilitated by interaction, feedback, and shared grounding experiences. Perfect alignment may be unattainable, but functional alignment sufficient for the task at hand is often the goal.

\section{Shared or Distributed Belief Spaces (Speculative)}

Can multiple interacting agents develop more integrated forms of shared cognition beyond individual alignment? This section explores \textbf{speculative possibilities} that extend the core framework:
\begin{itemize}
	\item \textbf{Overlapping Semantic Manifolds:} In tightly coupled collaboration, could regions of $\Phi_A$ and $\Phi_B$ become so strongly aligned and synchronized through constant communication and shared experience that they function almost as a single shared space for certain topics? Defining the precise nature and boundaries of such overlap remains a theoretical challenge.
	\item \textbf{Collective Coherence ($\kappa_{collective}$):} Could we define a coherence metric that applies across the belief states of multiple agents engaged in a joint task? How might agents regulate their communication and internal states to optimize this collective coherence? Formalizing $\kappa_{collective}$ in a meaningful way is complex.
	\item \textbf{Distributed Belief States:} Can a complex belief or plan be distributed across multiple agents, with no single agent holding the complete representation? This requires sophisticated protocols for querying, retrieving ($R$), and integrating ($A$) information across agent boundaries, potentially involving mechanisms beyond those defined for single agents.
	\item \textbf{Emergent Multi-Agent Structures:} Could interaction dynamics lead to the emergence of higher-level structures (e.g., shared norms, collective identity $\vec{\eta}_{group}$) that exist at the group level rather than within individual agents? Modeling such emergence rigorously is a significant undertaking.
\end{itemize}

\begin{figure}[htbp]
	\centering
	\input{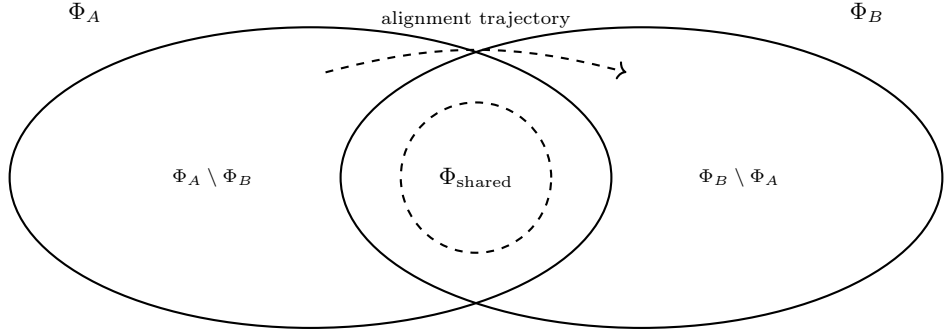}
	\caption{Shared belief geometry between two agents. Each maintains its own semantic state space (\(\Phi_A\), \(\Phi_B\)). Their overlap defines the shared subspace \(\Phi_{\text{shared}} \subset \Phi_A \cap \Phi_B\). Areas outside the overlap correspond to disagreement or divergence. Alignment trajectories aim to expand or reinforce \(\Phi_{\text{shared}}\).}
	\label{fig:shared-belief-space}
\end{figure}

These concepts push the boundaries of the framework, suggesting potential extensions for modeling team cognition, distributed AI systems, or societal-level knowledge structures based on interacting semantic manifolds. However, they remain largely theoretical explorations at this stage.

\section{Challenges: Misalignment, Deception, Complexity}

Modeling multi-agent semantic interaction faces significant challenges:
\begin{itemize}
	\item \textbf{Persistent Misalignment and Communication Failures:} Achieving and maintaining perfect semantic alignment is difficult; ambiguity, differing priors ($\theta_A \neq \theta_B$), noise, and errors in Theory of Mind models (Chapter~\ref{chap:ModelingOtherAgents}) perpetually challenge shared understanding and can lead to communication failures, misunderstandings, or failed coordination. Agents need robust mechanisms for detecting and repairing misalignment.
	\item \textbf{Deception and Trust:} Agents may intentionally misrepresent their beliefs. Modeling the detection of deception and the dynamics of trust based on interaction history and inferred motives (from models of $\phi_B$) is complex and requires sophisticated social reasoning capabilities beyond basic ToM.
	\item \textbf{Computational Complexity:} Explicitly modeling multiple interacting agents, including their recursive models of each other (A models B, B models A, A models B modeling A, etc.), leads to a combinatorial explosion in complexity. Scalable approximations and bounded rationality models are essential for practical implementation.
	\item \textbf{Defining Collective Metrics:} Formalizing concepts like $\kappa_{collective}$ or $\vec{\eta}_{group}$ in a computationally tractable and theoretically sound manner is non-trivial.
\end{itemize}

\section{Conclusion: Towards Interacting Semantic Systems}

This chapter extended the Semantic Manifold framework from individual cognition to the dynamics of multi-agent interaction. By modeling communication acts as operations influencing belief states ($\Phi$) and highlighting the crucial role of semantic alignment (grounded in interaction, shared experience, and potentially shared gauge $\sim_{gauge}$), we can begin to formalize how shared understanding emerges.

While concepts like truly shared or distributed belief spaces remain speculative, the framework provides tools (operators $A, M, Q, V, R$, metrics $d, \kappa$, gauge $\sim_{gauge}$) for analyzing the intricate processes of communication, coordination, and potential collective cognition among agents built on semantic principles. Understanding these multi-agent dynamics, including the challenges of misalignment and complexity, is vital for creating AI systems capable of sophisticated collaboration and social integration. This concludes our exploration of social cognition, leading into the final considerations of the monograph.


\subsection*{Chapter Summary}
This chapter transitions from individual Theory of Mind (Chapter~\ref{chap:ModelingOtherAgents}) to the dynamics of interaction between multiple Semantic Manifold agents. It models communication acts (assertions, queries, commands, expressions) as operations initiated by one agent intended to influence the belief state ($\phi$) of another, processed via operators like Assimilation ($A$) or Meta-Assimilation ($M$). Success hinges on achieving sufficient semantic alignment, which involves processes like converging on shared functional meaning (gauge equivalence, $\sim_{gauge}$), aligning semantic metrics ($d$), grounding concepts through joint action and observation, and employing feedback or repair mechanisms when communication falters. The chapter also speculatively explores more integrated forms of multi-agent cognition, such as overlapping or distributed belief spaces and collective coherence ($\kappa_{collective}$). Key challenges, including managing persistent misalignment, potential deception, and computational complexity, are acknowledged.
	
	\part{Conclusion}
	\label{part:conclusion}
	
	\chapter{Future Outlook}
\label{chap:FutureOutlook}

\section{A Roadmap for Semantic Cognition}

The semantic manifold framework introduced in this monograph lays a foundation for structured belief modeling in artificial agents. It defines a richly geometric, dynamically regulated internal space ($\Phi$) for cognition, inspired by linguistic structure, psychological function, and neuroscientific organization. Yet this is only the beginning. In this chapter, we outline possible directions for expanding, refining, and operationalizing the framework---charting a research agenda toward increasingly capable and coherent cognitive systems based on the principles developed herein.

\section{Metric Learning and Adaptive Geometry}

While the framework defines a semantic state space $\Phi$ equipped with structure-preserving operators, it leaves open the question of how these structures are learned or adapted in practice, as discussed initially in Part~\ref{part:learning_and_adaptation} (Chapter~\ref{chap:LearningSemanticStructures}). A major research frontier involves:
\begin{itemize}
	\item Learning semantic distances $d(\phi_1, \phi_2)$ via contrastive supervision or interaction-based grounding, allowing the agent's notion of similarity to evolve.
	\item Learning abstraction operators $\Lambda$ as neural mappings trained for semantic compression, perhaps optimizing reconstruction via elaboration $V$.
	\item Discovering intrinsic coordinates (e.g., sector axes $\Sigma$, gradient fields $F$, orientation axes $[\omega \rightarrow \omega^{(\infty)}]$) through unsupervised manifold embedding techniques applied to belief state data.
\end{itemize}
This direction suggests an intersection with representation learning, manifold learning, and metric geometry---yielding agents that can reshape their internal geometry $\Phi$ as they learn.

\section{Semantic Dynamics in Learning Agents}

The operators defined over $\Phi$---assimilation $A$, nullification $N_t$, elaboration $V$, orientation policies $\pi_{orient}$, and more (Parts~\ref{part:epistemic_dynamics}, \ref{part:semantic_memory}, \ref{part:regulation_and_control}, \ref{part:meta_cognition})---can potentially be parameterized and adapted via data (as explored in Chapter~\ref{chap:LearningCognitiveOperators}). Future work should explore:
\begin{itemize}
	\item Gradient-based fine-tuning of epistemic operators ($A, \Lambda, V, R, Q, M$), guided by coherence loss ($\kappa$), compression utility, or behavioral reward from interaction.
	\item Curriculum learning over belief space trajectories $\gamma(t)$, progressing from low to high abstraction levels ($\Phi^{(k)}$) or from narrow to broad semantic sectors ($\Sigma$).
	\item Differentiable integration with reinforcement learning, allowing belief evolution (the application and parameters of operators like $A, N_t, K$) to be shaped by outcomes in the external world.
\end{itemize}
This opens the door to agents that not only learn about the world, but learn how to think more effectively within structured semantic space $\Phi$.

\section{Meta-Cognition and Self-Reflective Agents}

Our formalism (Part~\ref{part:meta_cognition}) supports recursive representations of belief, trajectory awareness ($\gamma(t)$), and self-monitoring signals such as cognitive load ($\lambda$) and semantic effort ($\epsilon$). Future advances may include:
\begin{itemize}
	\item Learned control policies ($\pi_{regulate}$, Chapter~\ref{chap:LearningRegulatoryPolicies}) over introspection, enabling agents to selectively reflect (using $M$), summarize, or replan when coherence $\kappa$ drops.
	\item Meta-learning of belief formation strategies (e.g., tuning $A$ or $M$) or effort allocation ($\pi_{effort}$), optimizing epistemic regulation for long-term alignment or task performance.
	\item Epistemic identity ($\vec{\eta}$) preservation, supporting continuity of goals, values, styles, or narrative perspectives across time and tasks by strengthening anchoring ($a_i$) or regulating dynamics affecting $\Sigma_{narr}$ and $\Sigma_{refl}$.
\end{itemize}
These capacities, built upon the meta-cognitive mechanisms detailed in Part~\ref{part:meta_cognition}, are necessary for robust deliberative systems, AI companions, or long-horizon planners with stable self-models.

\section{Multi-Agent Semantic Alignment}

As artificial agents proliferate, they must increasingly interact, cooperate, and align (Part~\ref{part:social_cognition}). The semantic manifold framework provides tools for:
\begin{itemize}
	\item Shared semantic axes (derived from Null Towers, Chapter~\ref{chap:NullTower}, used in Orientation, Chapter~\ref{chap:SemanticOrientation}) and alignment protocols, ensuring agents can coordinate over beliefs even with different internal configurations ($\theta$).
	\item Gauge-aware ($\sim_{gauge}$, Chapter~\ref{chap:SemanticGauge}) communication, where agents learn to translate between gauge-equivalent but structurally distinct belief states, essential for interoperability (Chapter~\ref{chap:CommunicationAlignment}).
	\item Belief fusion and distributed inference, supporting collective reasoning across decentralized semantic agents, potentially involving distributed representations (Chapter~\ref{chap:CommunicationAlignment}).
\end{itemize}
This points toward the design of epistemically interoperable agents---capable of mutual understanding and shared knowledge construction.

\section{Embodiment, Simulation, and Situated Action}

While $\Phi$ represents internal cognitive structure, its connection to the world remains critical (Part~\ref{part:embodiment_and_action}). Semantic execution (Chapter~\ref{chap:SemanticExecution}) and embodied simulation (Chapter~\ref{chap:EmbodiedSimulation}) lay the foundation for:
\begin{itemize}
	\item Closed-loop semantic control, in which beliefs ($\phi$ entering $\mathcal{A}_a$) drive actions and actions reshape beliefs (via $X \rightarrow A$).
	\item Simulated rehearsal in $\Phi$, where plans ($\Sigma_{plan}$) are validated, abstracted ($\Lambda$), or counterfactually revised using internal models before execution.
	\item Grounding via sensorimotor cycles, enriching semantic sectors ($\Sigma$) with embodied regularities derived from interaction (Chapter~\ref{chap:GroundingSemanticBelief}).
\end{itemize}
This enables agents to not only reflect and reason, but to act coherently and safely in physical or interactive environments.

\section{Formal Theory and Mathematical Extensions}

The current framework offers a geometric and operator-theoretic foundation for structured belief. Future mathematical development may include:
\begin{itemize}
	\item Topological convergence analysis of abstraction sequences ($\omega^{(k)} \rightarrow \omega^{(\infty)}$) and belief trajectories $\gamma(t)$, potentially using fixed-point theorems.
	\item Gauge-theoretic models of semantic invariance ($\sim_{gauge}$), leveraging fiber bundle structures or related algebraic topology concepts.
	\item Differential geometric characterizations of epistemic flow ($F$), including curvature of belief evolution and semantic vector fields over $\Phi$.
\end{itemize}
Such formalisms could allow rigorous comparisons of cognitive architectures ($\Phi^{[\theta]}$) and provide new tools for validating coherence ($\kappa$) and safety in high-dimensional reasoning systems.

\section{Toward General Epistemic Systems}

Ultimately, this monograph contributes to the broader endeavor of defining what it means to think---formally, structurally, and semantically. The future of this work lies in the development of:
\begin{itemize}
	\item Unified epistemic systems that couple memory (Part~\ref{part:semantic_memory}), control (Part~\ref{part:regulation_and_control}), abstraction (Part~\ref{part:meta_cognition}), planning (implicated throughout), and identity (Chapter~\ref{chap:EpistemicIdentity}) into coherent cognitive agents.
	\item Open-ended semantic development, where belief structures ($\phi$) grow, fragment, consolidate, and realign without fixed ontologies, driven by learning (Part~\ref{part:learning_and_adaptation}).
	\item Semantic civilizations, where multi-agent belief ecosystems (Part~\ref{part:social_cognition}) develop their own norms of coherence, transmission, and alignment.
\end{itemize}
These long-term directions aim to close the gap between theoretical models of cognition and the creation of general-purpose, self-regulating, epistemically aware agents. The final chapter will distill the core contributions of this monograph and reflect on the broader implications of building machines that think not only symbolically or statistically, but semantically.


\subsection*{Chapter Summary}
This chapter outlines promising future research directions stemming from the Semantic Manifold framework. Key areas for advancement include developing methods for learning the core semantic structures, such as the metric ($d$) and functional sectors ($\Sigma$), potentially adapting the manifold's geometry itself. Further work is envisioned in learning and refining the cognitive operators ($A$, $\Lambda$, $V$, $R$, $Q$, $M$) and the regulatory policies ($\pi_{effort}, \pi_{regulate}$) through techniques like reinforcement learning and representation learning. Enhancing meta-cognitive capabilities, particularly regarding learned self-regulation and robust epistemic identity ($\vec{\eta}$) maintenance, is highlighted. Extending the framework to multi-agent systems focuses on achieving semantic alignment, gauge-aware ($\sim_{gauge}$) communication, and distributed cognition. Strengthening the connection between internal belief structures and embodied action through closed-loop control and simulation is another crucial direction. Finally, the chapter calls for further development of the formal mathematical underpinnings (topology, geometry, gauge theory) and discusses the long-term vision of creating unified, general epistemic systems capable of open-ended development.
	\chapter{Conclusion}
\label{chap:Conclusion}

\section{Summary of Contributions}

This monograph has introduced a theoretical framework for modeling belief as structured evolution within a semantic manifold. Rather than treating cognition as a black-box emergent property or a brittle symbolic construct, we have proposed that belief in intelligent agents arises from---and operates over---a principled internal space: the semantic state space $\Phi$.

We began by constructing the foundational architecture: belief states $\phi$ defined as structured ensembles of linguistic expressions, organized within $\Phi$ by abstraction level ($\Phi^{(k)}$) and functional sector ($\Sigma$). We introduced the Null Tower as a generative scaffold, modeling the emergence of cognition from the epistemic vacuum $\Omega$ via recursive abstraction ($\Lambda$). Semantic scaling ($\Lambda, V$) and sectoral modularity ($\Sigma$) provided the structure; epistemic dynamics such as assimilation ($A$), nullification ($N_t$), annihilation ($K$), and drift ($D$) introduced motion.

Crucially, we formulated mechanisms for reflective regulation. Orientation vectors ($[\omega \rightarrow \omega^{(\infty)}]$), semantic compass structures, gauge equivalence classes ($\sim_{gauge}$), identity-preserving feedback loops ($\vec{\eta}$), and cognitive load/effort metrics ($\lambda, \epsilon$) together enabled agents not only to evolve beliefs, but to monitor, evaluate, and regulate their own epistemic trajectories ($\gamma(t)$). We further connected this internal world to action via semantic execution and activation basins ($\mathcal{A}_a$), and explored internal rehearsal via embodied simulation. The framework was extended to meta-cognition ($M$), learning (Part~\ref{part:learning_and_adaptation}), and multi-agent interactions (Part~\ref{part:social_cognition}).

The primary contributions of this framework, distinguishing it from prior architectures (as detailed in Appendix~\ref{app:Comparisons}), include:
\begin{itemize}
	\item \textbf{Unified Semantic Space ($\Phi$):} Proposing a single, structured, linguistically-grounded space encompassing active beliefs, memory, goals, and meta-cognition, organized by abstraction ($\Phi^{(k)}$) and function ($\Sigma$).
	\item \textbf{Geometric Interpretation:} Conceptualizing cognition as navigation and regulated flow within this geometric manifold, using concepts like distance ($d$), orientation ($\theta$), and trajectories ($\gamma(t)$).
	\item \textbf{Explicit Dynamics and Regulation:} Formalizing core belief lifecycle operators ($A, N_t, K, \Lambda, V$) alongside explicit mechanisms for internal monitoring ($\kappa, \lambda, \epsilon$) and regulation (Orientation, Gauge, Identity $\vec{\eta}$), enabling self-stabilization and coherence maintenance beyond simple error correction or external control.
	\item \textbf{Parameterization ($\Phi^{[\theta]}$):} Providing a meta-architecture allowing diverse agent designs (varying in memory $\mu$, representation $\rho$, identity $\eta_{type}$, etc.) within a unified conceptual scheme.
	\item \textbf{Integration of Concepts:} Synthesizing ideas from symbolic AI (compositionality), connectionism (potential for learned embeddings $\rho$), predictive coding (hierarchy, internal states), and cognitive science (structured memory, meta-cognition, identity).
\end{itemize}
In essence, the framework provides a novel synthesis focused on the structure, dynamics, geometry, and regulation of belief itself.

\section{Architectural Vision}

What emerges from this work is a unified vision of cognition as a form of semantic geometry. Beliefs evolve not in a void, but on a structured manifold; thought is motion in epistemic space. Every aspect of cognition---perception, memory, planning, action, reflection---can be modeled as a transformation within or across structured regions of $\Phi$.

This architecture is inherently modular and recursive. It supports fine-grained representational transparency, operator-based dynamics, sectoral compartmentalization, and higher-order self-monitoring. It enables agents to represent their own thoughts, regulate their coherence ($\kappa$), and realign their beliefs toward epistemic goals. In doing so, it bridges the gap between symbolic reasoning and neural representation, between structured language and embodied intelligence.

\section{Limitations and Open Questions}

While the framework is comprehensive in scope, it remains largely theoretical. Several open challenges remain before its full operationalization:
\begin{itemize}
	\item \textbf{Grounding:} How are belief expressions $\varphi_i$ acquired, contextualized, and robustly grounded in perception and interaction (Chapter~\ref{chap:GroundingSemanticBelief})?
	\item \textbf{Learning:} How can the semantic operators ($A, M, \Lambda, V, R, Q$), structures ($d, \Sigma$), and policies ($\pi_{effort}, \pi_{regulate}$) be effectively learned, parameterized, and adapted in data-driven settings (Part~\ref{part:learning_and_adaptation})?
	\item \textbf{Validation:} What experimental paradigms can be used to empirically test properties of belief trajectories $\gamma(t)$, coherence metrics $\kappa$, or semantic alignment in implemented agents (Appendix~\ref{app:Comparisons})?
	\item \textbf{Scalability:} How can belief structures $\phi$ scale across tasks, domains, or time without losing coherence $\kappa$, tractability (managing $\lambda, \epsilon$), or identity $\vec{\eta}$?
	\item \textbf{Embodiment and Execution:} How do semantic states ($\phi \in \mathcal{A}_a$) drive concrete action in situated environments, and how is semantic structure $\Phi$ shaped by embodied interaction (Part~\ref{part:embodiment_and_action})?
\end{itemize}
Addressing these questions will require advances in both algorithmic modeling and empirical investigation---bridging formal theory with cognitive architectures, machine learning, and human-aligned AI.

\section{Broader Implications}

The development of agents that think semantically---who evolve, evaluate, and reflect upon structured beliefs---holds significant promise. Such agents could offer:
\begin{itemize}
	\item \textbf{Interpretability:} Transparent internal states organized around human-understandable structure ($\phi=\{\varphi_i\}, \Sigma, \Phi^{(k)}$).
	\item \textbf{Coherence:} Consistent reasoning over long horizons, driven by self-regulating belief dynamics ($\kappa$, Orientation, $N_t, K$).
	\item \textbf{Alignment:} Greater capacity for shared meaning (via $\sim_{gauge}$), communication, and potentially value alignment ($\vec{\eta}$, $\Sigma_{refl}$).
	\item \textbf{Adaptivity:} Flexible incorporation of new information ($A$) without collapse or contradiction ($K, \kappa$).
	\item \textbf{Continuity:} The emergence of epistemic identity ($\vec{\eta}$) and introspective development over time ($M, \gamma(t)$).
\end{itemize}
More speculatively, such architectures may serve as the substrate for artificial epistemic cultures: multi-agent systems (Part~\ref{part:social_cognition}) with shared coordinate systems, institutionalized beliefs, and reflexive narratives.

\section{Final Reflection}

To believe is to orient oneself in a space of meaning. To think is to traverse it. The framework developed here proposes that such orientation and traversal can be made mathematically explicit, dynamically stable, and cognitively plausible. It offers not merely a new model for artificial intelligence, but a new way to formalize the structure of understanding itself. The path ahead is open. The geometry of belief has only begun to be charted.
	
	\appendix
	
	\chapter{Glossary of Terms and Notation}
\label{app:GlossaryNotation}

This appendix provides definitions for key terminology and a list of mathematical symbols used throughout the monograph.

\section{Glossary of Key Terms}

\begin{description}[leftmargin=2.5cm, labelindent=0cm, style=nextline]
	\item[Activation Basin ($\mathcal{A}_a$)] A region within the semantic state space $\Phi$ where the semantic preconditions for executing a specific action $a$ are met. Entry into an activation basin enables potential action execution. (See Chapters~\ref{chap:SemanticExecution}, \ref{chap:ActivationBasins})
	
	\item[Annihilation ($K$, $K_{\Sigma}$)] The dynamic operator modeling abrupt, discontinuous erasure of belief structures---either totally ($K: \Phi \rightarrow \Omega$) or within specific sectors ($K_{\Sigma}$). Contrasts with gradual Nullification ($N_t$). (See Chapter~\ref{chap:Annihilation})
	
	\item[Assimilation ($A$)] The dynamic operator responsible for integrating new object-level information (from perception $X(s)$, simulation, communication, or memory retrieval $R$) into the belief state $\phi$, involving coherence checks, correction ($A_{corr}$), and elaboration ($A_{elab}$). (See Chapter~\ref{chap:Assimilation})
	
	\item[Coherence ($\kappa$)] A measure of internal consistency, logical compatibility, and mutual support among belief fragments within a state $\phi$ or region $\Sigma$. (See Chapter~\ref{chap:CognitiveLoad})
	
	\item[Cognitive Load ($L$, $\lambda$)] A measure of the cognitive demands imposed by belief complexity and ongoing tasks. Distinguished from Semantic Effort $\epsilon$. (See Chapter~\ref{chap:CognitiveLoad})
	
	\item[Embodied Simulation] The generation of internal belief trajectories $\gamma_{sim}(t)$ representing hypothetical interactions with the environment, used for planning, prediction, and grounding. (See Chapter~\ref{chap:EmbodiedSimulation})
	
	\item[Epistemic Axis (\mbox{$[\omega \rightarrow \omega^{(\infty)}]$})] A directed semantic trajectory in $\Phi$, abstracting from a null state $\omega$ to its singularity $\omega^{(\infty)}$. (See Chapters~\ref{chap:NullTower}, \ref{chap:SemanticOrientation})
	
	\item[Epistemic Identity ($\vec{\eta}$)] The persistent signature of an agent's beliefs, values, and cognitive traits, distributed within $\Phi$ and especially across $\Sigma_{narr}$ and $\Sigma_{refl}$. (See Chapter~\ref{chap:EpistemicIdentity})
	
	\item[Epistemic Vacuum ($\Omega$)] The null region of belief space $\Phi$ representing semantic emptiness or rest. (See Chapters~\ref{chap:SemanticStateSpace}, \ref{chap:NullTower})
	
	\item[Meta-Assimilation ($M$)] The operator for integrating meta-cognitive information $\phi_{introspective}$ into $\phi$, typically affecting $\Sigma_{refl}$. (See Chapter~\ref{chap:MetaAssimilation})
	
	\item[Meta-Introspection] The agent's capacity to observe, represent, and reason about its own internal state $\phi$, dynamics $A, N_t, K$, and structures $\Phi$, $\Sigma$. (See Chapter~\ref{chap:MetaIntrospection})
	
	\item[Nullification ($N_t$)] The gradual decay operator modeling belief fading due to inactivity or weak anchoring ($a_i$), driving $\phi \rightarrow \Omega$. (See Chapter~\ref{chap:Nullification})
	
	\item[Null Tower] A recursive construction emerging from $\Omega$ via abstraction operators $\Lambda$, producing layers $\Omega^{(k)}$ and culminating in $\omega^{(\infty)}$. (See Chapter~\ref{chap:NullTower})
	
	\item[Semantic Effort ($\epsilon$)] Actively applied finite resource used to execute operations (e.g., $A$, $M$), maintain coherence, or pursue goals. (See Chapter~\ref{chap:SemanticEffort})
	
	\item[Semantic Gauge ($\sim_{gauge}$)] An equivalence relation over belief states indicating cognitive indistinguishability. (See Chapter~\ref{chap:SemanticGauge})
	
	\item[Semantic Geometry] The structural interpretation of $\Phi$ as a manifold with sectors $\Sigma$ and layers $\Phi^{(k)}$, enabling interpretation of belief dynamics as trajectories. (See Chapter~\ref{chap:SemanticGeometry})
	
	\item[Semantic Manifold ($\Phi$, $\Phi^{[\theta]}$)] The structured state space containing all possible belief states $\phi$. Parameterized by architecture configuration $\theta$. (See Chapters~\ref{chap:SemanticStateSpace}, \ref{chap:ParameterizedArchitectures})
	
	\item[Semantic Scaling] A hierarchical organization of $\Phi$ into abstraction levels $\Phi^{(k)}$ with upward $\Lambda$ and downward $V$ operators. (See Chapter~\ref{chap:SemanticScaling})
	
	\item[Semantic Sector ($\Sigma$)] A functionally coherent region of $\Phi$ (e.g., $\Sigma_{perc}$, $\Sigma_{plan}$). Allows modular processing. (See Chapter~\ref{chap:SemanticSectors})
	
	\item[Semantic Singularity ($\omega^{(\infty)}$)] The fixed point of recursive abstraction of $\omega$; represents a limit of semantic form. (See Chapter~\ref{chap:NullTower})
	
	\item[Spontaneous Drift ($D$)] Operator for minimal, internal belief motion from $\Omega$, serving as the cognitive ground state dynamic. (See Chapter~\ref{chap:SpontaneousDrift})
	
	\item[Suppression Surface ($S_a$)] A region of $\Phi$ that inhibits execution of action $a$, despite apparent precondition satisfaction. (See Chapter~\ref{chap:ActivationBasins})
	
	\item[Trajectory Awareness] The capacity to monitor and reason about the belief trajectory $\gamma(t)$ through $\Phi$. (See Chapter~\ref{chap:TrajectoryAwareness})
\end{description}

\clearpage
\section{List of Notation}
\subsection*{Spaces, Structures, and Belief Components}
\begin{itemize}
	\item[$\Phi$] Full semantic state space.
	\item[$\phi$] A specific belief state, $\phi = \{\varphi_1, \varphi_2, \dots\}$.
	\item[$\Phi^{(k)}$] Abstraction layer $k$ of $\Phi$.
	\item[$\Sigma$] Semantic sector (e.g., $\Sigma_{perc}$).
	\item[$\Phi^{[\theta]}$] Architecture-conditioned variant of $\Phi$.
	\item[$\mathcal{A}_a$] Activation basin for action $a$.
	\item[$S_a$] Suppression surface for action $a$.
	\item[$\Omega$] Epistemic vacuum.
	\item[$\mathcal{T}_\Omega$] Null Tower.
	\item[$\omega$] Null state; $\omega \in \Omega$.
	\item[$\omega^{(\infty)}$] Semantic singularity of $\omega$.
\end{itemize}

\subsection*{Operators}
\begin{itemize}
	\item[$A$] Assimilation.
	\item[$M$] Meta-Assimilation.
	\item[$N_t$] Nullification.
	\item[$K$] Total annihilation.
	\item[$K_{\Sigma}$] Sectoral annihilation.
	\item[$D$] Spontaneous drift.
	\item[$\Lambda$] Abstraction operator.
	\item[$V$] Elaboration operator.
	\item[$Q$] Query function.
	\item[$R$] Retrieval function.
\end{itemize}

\subsection*{Metrics and State Properties}
\begin{itemize}
	\item[$\kappa$] Coherence.
	\item[$\lambda$] Cognitive load or activation mass.
	\item[$\epsilon$] Semantic effort.
	\item[$\delta$] Persistence threshold.
	\item[$L$] Load metric (alternative to $\lambda$).
	\item[$a_i$] Anchoring strength of $\varphi_i$.
	\item[$\vec{\eta}$] Epistemic identity vector.
	\item[$\vec{\alpha}(t)$] Sector activation vector.
	\item[$\theta$] Architecture parameter vector.
\end{itemize}

\subsection*{Regulation and Trajectories}
\begin{itemize}
	\item[$\gamma(t)$] Belief trajectory in $\Phi$.
	\item[$F$] Belief vector field.
	\item[$\pi$] Policy function.
	\item[$\pi_{\text{effort}}$] Effort allocation policy.
	\item[$\pi_{\text{regulate}}$] Meta-regulatory policy.
	\item[$\sim_{\text{gauge}}$] Gauge equivalence.
	\item[\mbox{$[\omega \rightarrow \omega^{(\infty)}]$}] Epistemic axis.
\end{itemize}
	\chapter{Formal Operator Definitions}
\label{app:OperatorDefinitions}

This appendix provides more detailed, semi-formal definitions or algorithmic sketches for the core cognitive operators introduced in the main text. These definitions aim to clarify the intended function and inputs/outputs of each operator. Note that concrete implementations will depend heavily on the specific architectural parameters $\theta$, particularly the representation mode $\rho$.

\section{Assimilation (\texorpdfstring{$A$}{A})}

The Assimilation operator integrates object-level input $\phi_{input}$ into the current state $\phi_{current}$.
$$ A : \Phi \times \Phi_{input} \rightarrow \Phi $$
$$ \phi_{new} = A(\phi_{current}, \phi_{input}) $$

A general algorithm sketch involves stages:
\begin{enumerate}
	\item \textbf{Filtering/Salience (Optional):} $\phi'_{input} = \text{Filter}(\phi_{input}, \phi_{current})$ based on relevance/salience.
	\item \textbf{Conflict Detection:} Identify set $C = \{(\varphi_i, \varphi_j) \mid \varphi_i \in \phi_{current}, \varphi_j \in \phi'_{input}, \text{Contradict}(\varphi_i, \varphi_j)\}$.
	\item \textbf{Revision (if $C \neq \emptyset$, corresponds to $A_{corr}$):}
	\begin{itemize}
		\item Determine beliefs to revise/retract from $\phi_{current}$: $\phi_{retract} = R_{\text{rev/ret}}(\phi_{current}, \phi'_{input}, C)$ (based on anchoring $a_i$, coherence $\kappa$, etc.).
		\item Potentially modify input: $\phi''_{input} = \text{ModifyInput}(\phi'_{input}, C, \phi_{retract})$.
		\item Update current state: $\phi_{revised} = \phi_{current} \setminus \phi_{retract}$.
	\end{itemize}
	\item \textbf{Integration:} Combine revised state and input: $\phi_{merged} = \phi_{revised} \cup \phi''_{input}$.
	\item \textbf{Elaboration (corresponds to $A_{elab}$):} Generate elaborations $\phi_{elab} = \text{Elaborate}(\phi_{merged})$. $\phi_{pre\_final} = \phi_{merged} \cup \phi_{elab}$.
	\item \textbf{Abstraction (corresponds to $A_{abs}$, optional):} Identify patterns in $\phi_{pre\_final}$ and generate abstractions $\phi_{abs}$. Potentially remove summarized beliefs. $\phi_{final} = (\phi_{pre\_final} \setminus \phi_{summarized}) \cup \phi_{abs}$.
	\item \textbf{Re-Anchoring:} Update anchoring scores $a_i$ for newly added or reinforced beliefs in $\phi_{final}$.
	\item \textbf{Return:} $\phi_{new} \leftarrow \phi_{final}$.
\end{enumerate}
Specific subtypes emphasize certain steps (e.g., $A_{corr}$ emphasizes step 3, $A_{elab}$ step 5).

\section{Nullification (\texorpdfstring{$N_t$}{N\_t})}

The Nullification operator models gradual decay towards $\Omega$.
$$ N_t : \Phi \rightarrow \Phi $$
$$ \phi_t = N_t(\phi_0) $$
Implemented elementwise based on persistence $d_i(t)$ and anchoring $a_i$:
\begin{enumerate}
	\item For each $\varphi_i \in \phi_0$:
	\begin{itemize}
		\item Retrieve current anchoring $a_i$.
		\item Determine decay rate $\lambda_i = \lambda_0 \cdot f(a_i)$ (where $f$ decreases with $a_i$).
		\item Compute persistence at time $t$ using a decay function, e.g., $d_i(t) = d_i(0) \cdot e^{-\lambda_i t}$.
	\end{itemize}
	\item Define persistence threshold $\delta$.
	\item Construct the nullified state: $\phi_t = \{\varphi_i \in \phi_0 \mid d_i(t) > \delta\}$.
	\item Update internal state to reflect new persistence values $d_i(t)$.
	\item Return $\phi_t$.
\end{enumerate}
The choice of decay function $e^{-\lambda t}$ or alternatives is part of $\delta_{profile}$ in $\theta$.

\section{Annihilation (\texorpdfstring{$K$}{K}, \texorpdfstring{$K_{\Sigma}$}{K\_Sigma})}

Annihilation models abrupt erasure.
\begin{itemize}
	\item \textbf{Total Annihilation ($K$):}
	$$ K : \Phi \rightarrow \Omega $$
	$$ K(\phi) = \phi_{null} \quad \text{where } \phi_{null} \in \Omega $$
	Algorithm: Replace $\phi$ entirely with a pre-defined null state.
	\item \textbf{Sectoral Annihilation ($K_{\Sigma}$):}
	$$ K_{\Sigma} : \Phi \rightarrow \Phi $$
	$$ K_{\Sigma}(\phi) = \phi \setminus \phi|_{\Sigma} $$
	Algorithm: Identify all belief fragments $\varphi_i$ belonging to sector $\Sigma$ (i.e., $\phi|_{\Sigma}$) and remove them from $\phi$.
\end{itemize}

\section{Scaling (\texorpdfstring{$\Lambda$}{Lambda}, \texorpdfstring{$V$}{V})}

These operators manage abstraction levels $\Phi^{(k)}$.
\begin{itemize}
	\item \textbf{Abstraction ($\Lambda$):}
	$$ \Lambda^k : \Phi^{(i)} \rightarrow \Phi^{(i+k)} $$
	Algorithm sketch (depends heavily on $\rho$):
	\begin{enumerate}
		\item Input $\phi^{(i)}$.
		\item Apply compression/generalization function $f_{\Lambda}$ (e.g., autoencoder encoder, summarization model, clustering/prototype extraction).
		\item Output $\phi^{(j)} = f_{\Lambda}(\phi^{(i)})$, ensuring result resides at level $j$.
	\end{enumerate}
	\item \textbf{Elaboration ($V$):}
	$$ V^k : \Phi^{(i)} \rightarrow \Phi^{(i-k)} $$
	Algorithm sketch:
	\begin{enumerate}
		\item Input $\phi^{(j)}$.
		\item Apply instantiation/decompression function $f_{V}$ (e.g., autoencoder decoder, generative model conditioned on $\phi^{(j)}$, rule-based instantiation). $f_V$ may be stochastic.
		\item Output $\phi^{(i)} = f_{V}(\phi^{(j)})$, ensuring result resides at level $i$.
	\end{enumerate}
\end{itemize}
\section{Retrieval (\texorpdfstring{$Q$}{Q}, \texorpdfstring{$R$}{R})}

These operators manage memory access.
\begin{itemize}
	\item \textbf{Query Function ($Q$):} Generates retrieval cue.
	$$ Q : \Phi_{active} \rightarrow \Phi_{query} $$
	Algorithm sketch:
	\begin{enumerate}
		\item Input $\phi_{active}$.
		\item Identify trigger (e.g., active goal, low $\kappa$, associative link).
		\item Extract salient elements, unresolved variables, or context related to the trigger from $\phi_{active}$.
		\item Format these elements into a structured query $\phi_{query}$ (which may be a single $\varphi_i$ or a minimal state $\phi$).
		\item Return $\phi_{query}$.
	\end{enumerate}
	\item \textbf{Retrieval Operator ($R$):} Accesses memory based on cue, returning a belief state composed of relevant fragments.
	$$ R : \mathcal{P}(\Phi_{memory}) \times \Phi_{query} \rightarrow \Phi $$ 
	$$ \phi_{retrieved} = R(\Phi_{memory}, \phi_{query}) $$
	Algorithm sketch:
	\begin{enumerate}
		\item Input $\Phi_{memory}$, $\phi_{query}$.
		\item Define relevance threshold $\tau_{retrieval}$ and potentially search scope (e.g., specific $\Sigma$).
		\item Initialize candidate set $C_{retrieved} = \emptyset$.
		\item For each candidate memory fragment $\{\varphi_k\} \in \Phi_{memory}$ (within scope):
		\begin{itemize}
			\item Calculate relevance score $S(\varphi_k, \phi_{query})$ using semantic distance $d$, link traversal, pattern matching, etc.
			\item Consider persistence $d_k(t)$ or anchoring $a_k$.
			\item If $S(\varphi_k, \phi_{query})$ exceeds $\tau_{retrieval}$ (and potentially $d_k(t) > \delta_{retrieval}$), add $\varphi_k$ to the candidate set $C_{retrieved}$.
		\end{itemize}
		\item If $C_{retrieved} \neq \emptyset$:
		\begin{itemize}
			\item Select/filter/rank candidates from $C_{retrieved}$ as needed.
			\item Construct the output state $\phi_{retrieved}$ as a structured ensemble composed of the final selected fragments $\{\varphi_k\}$.
		\end{itemize}
		\item Else ($C_{retrieved} = \emptyset$):
		\begin{itemize}
			\item Set $\phi_{retrieved} = \phi_{null}$, where $\phi_{null}$ is a designated null state in $\Omega$.
		\end{itemize}
		\item Return $\phi_{retrieved}$. 
	\end{enumerate}
\end{itemize}

\section{Meta-Assimilation (\texorpdfstring{$M$}{M})}

Integrates introspective information $\Phi_{introspective}$.
$$ M : \Phi \times \Phi_{introspective} \rightarrow \Phi $$
$$ \phi_{new} = M(\phi_{current}, \phi_{introspective}) $$
Algorithm sketch:
\begin{enumerate}
	\item Input $\phi_{current}$, $\phi_{introspective}$.
	\item Identify target sector, typically $\Sigma_{refl}$.
	\item Process $\phi_{introspective}$ to form explicit meta-belief expressions $\varphi_{meta}$ (e.g., "$\kappa$ is low", "Load $\lambda$ is high"). This might involve formatting, tagging, or confidence assignment based on the nature of the introspective input.
	\item Integrate $\varphi_{meta}$ into $\phi_{current}|_{\Sigma_{refl}}$, potentially using mechanisms similar to $A$ (checking for consistency with existing meta-beliefs, handling conflicts within $\Sigma_{refl}$).
	\item Update anchoring $a_i$ for new or modified meta-beliefs.
	\item Return $\phi_{new}$.
\end{enumerate}

\section{Spontaneous Drift (\texorpdfstring{$D$}{D})}

Models minimal emergence from $\Omega$.
$$ D : \Phi \rightarrow \Phi $$
Conceptual definition: If $\phi \in \Omega$, $D(\phi)$ produces $\phi'$ which is a minimal semantic perturbation, potentially $\phi' \notin \Omega$. If $\phi \notin \Omega$, $D(\phi)$ might represent minimal background noise or diffusion. The specific mechanism is highly dependent on implementation ($\rho$) and likely involves low-magnitude random changes in the state representation.

\section{Regulatory Functions (Examples)}

These are often components used by operators or policies, rather than top-level operators themselves.
\begin{itemize}
	\item \textbf{Projection ($\pi_{\omega}(\phi)$):} Component of Semantic Orientation (Chapter~\ref{chap:SemanticOrientation}). Given $\phi$ and axis $[\omega \rightarrow \omega^{(\infty)}]$ with vector $\vec{v}_{\omega}$, calculates the projection of $(\phi-\omega)$ onto $\vec{v}_{\omega}$. E.g., $\pi_{\omega}(\phi) = \omega + \frac{\langle \phi-\omega, \vec{v}_{\omega} \rangle}{\|\vec{v}_{\omega}\|^2} \cdot \vec{v}_{\omega}$ (vector space embedding assumed).
	\item \textbf{Angular Deviation ($\theta(\phi, \vec{v}_{\omega})$):} Component of Semantic Orientation. Calculates angle between $(\phi-\omega)$ and $\vec{v}_{\omega}$. E.g., $\theta = \cos^{-1} \left( \frac{\langle \phi-\omega, \vec{v}_{\omega} \rangle}{\|\phi-\omega\| \cdot \|\vec{v}_{\omega}\|} \right)$.
	\item \textbf{Reflective Diagnostic ($\delta_a(\phi_{refl})$):} Component of Semantic Execution gating (Chapter~\ref{chap:SemanticExecution}). Takes reflective state $\phi_{refl} = \phi|_{\Sigma_{refl}}$ and potential action $a$ as input. Outputs 'approve', 'delay', or 'suppress' based on evaluating $a$ against goals, values, constraints represented in $\phi_{refl}$. Mechanism depends on how these are represented.
\end{itemize}
These formal definitions provide a more structured view of the intended behavior of the core cognitive operators within the Semantic Manifold framework.
	\chapter{Detailed Geometric and Topological Considerations}
\label{app:GeometricTopological}

\section{Introduction}

The main text frequently employs geometric and topological language to describe the semantic state space $\Phi$, belief dynamics $\gamma(t)$, and regulatory mechanisms. Terms like manifold, distance ($d$), trajectory, flow ($F$), boundaries ($\partial \mathcal{A}_a$), projection ($\pi_{\omega}$), and gauge symmetry ($\sim_{gauge}$) suggest an underlying mathematical structure. This appendix elaborates on these concepts, providing more formal context and discussing their theoretical implications within the Semantic Manifold framework, while acknowledging that rigorously defining these structures for complex belief spaces remains a significant challenge.

\section{The Semantic Manifold (\texorpdfstring{$\Phi$}{Phi}) as a Structured Space}

We conceptualize $\Phi$ as a "manifold"---a space that locally resembles a simpler Euclidean space but may have a complex global structure. This is primarily a conceptual tool, as the exact dimensionality and structure are highly dependent on the representation mode $\rho$ chosen in the agent's architecture $\theta$.
\begin{itemize}
	\item \textbf{High Dimensionality:} If $\rho$ involves embeddings or complex symbolic structures, $\Phi$ is likely very high-dimensional.
	\item \textbf{Stratification:} As discussed in Chapters \ref{chap:SemanticScaling} and \ref{chap:SemanticSectors}, $\Phi$ is structured by abstraction levels ($\Phi = \bigcup_k \Phi^{(k)}$) and functional sectors ($\Phi = \bigcup_{\Sigma} \Sigma$). This suggests $\Phi$ might be viewed as a stratified space or perhaps a product space (e.g., locally $\approx \mathbb{R}^N \times \mathcal{S} \times \mathbb{Z}_{\ge 0}$ where $\mathcal{S}$ represents sectors).
	\item \textbf{Non-Standard Manifold Properties:} Depending on $\rho$, $\Phi$ might lack properties of smooth differential manifolds (e.g., if based on discrete graphs or symbolic structures). The geometric intuition remains useful even if formal manifold properties do not strictly hold everywhere.
\end{itemize}
The key idea is that $\Phi$ possesses sufficient structure to support notions of proximity, paths, and transformations that can be analyzed geometrically.

\section{Metric Structure (\texorpdfstring{$d$}{d})}

A crucial element for geometric reasoning is the semantic distance metric $d(\phi_1, \phi_2)$. As postulated in Chapter~\ref{chap:SemanticGeometry}, this function should ideally satisfy the standard metric axioms:
\begin{enumerate}
	\item Non-negativity: $d(\phi_1, \phi_2) \ge 0$
	\item Identity of Indiscernibles: $d(\phi_1, \phi_2) = 0 \iff \phi_1 = \phi_2$
	\item Symmetry: $d(\phi_1, \phi_2) = d(\phi_2, \phi_1)$
	\item Triangle Inequality: $d(\phi_1, \phi_3) \le d(\phi_1, \phi_2) + d(\phi_2, \phi_3)$
\end{enumerate}
The challenge lies in defining and learning a $d$ that meaningfully reflects semantic similarity over complex belief structures.
\begin{itemize}
	\item \textbf{Embedding Spaces ($\rho \approx$ Embedding):} $d$ could be based on Euclidean distance, cosine distance, or learned Mahalanobis distances in the embedding space. Learning $d$ equates to learning the embedding function (Chapter~\ref{chap:LearningSemanticStructures}, Appendix~\ref{app:ImplementationExamples}).
	\item \textbf{Symbolic Spaces ($\rho \approx$ Symbolic/Graph):} $d$ might involve weighted edit distances, graph kernel similarities, or distances based on logical entailment or semantic feature overlap. Defining these to satisfy metric axioms and capture semantics well is non-trivial.
\end{itemize}
A well-defined metric $d$ grounds notions of proximity, enables similarity-based retrieval ($R$), allows defining neighborhoods for local processing, and provides a basis for quantifying coherence ($\kappa$) or trajectory properties (e.g., path length). The difficulty in defining a perfect, universal $d$ underscores why some framework components (like $R$) might rely on multiple similarity heuristics rather than a single metric.

\subsection{Illustrative Metric Formulations}

The following are simplified, conceptual examples showing how $d$ might be formulated depending on $\rho$. These are illustrative, not definitive.

\begin{itemize}
	\item \textbf{Embedding Example ($\rho$ = Embedding):} Let $\text{embed}(\phi)$ be a vector representation (e.g., average embedding of its constituent $\varphi_i$). Possible metrics include:
	\begin{itemize}
		\item Cosine Distance: $d(\phi_1, \phi_2) = 1 - \frac{\text{embed}(\phi_1) \cdot \text{embed}(\phi_2)}{\|\text{embed}(\phi_1)\| \|\text{embed}(\phi_2)\|}$
		\item Euclidean Distance: $d(\phi_1, \phi_2) = \| \text{embed}(\phi_1) - \text{embed}(\phi_2) \|_2$
	\end{itemize}
	Learning involves optimizing the `embed` function itself (Chapter~\ref{chap:LearningSemanticStructures}).
	
	\item \textbf{Symbolic/Graph Example ($\rho$ = Graph):} Let $\text{graph}(\phi)$ be a graph representation. A conceptual metric might be:
	$$ d(\phi_1, \phi_2) \approx w_1 \cdot \text{GraphEditDistance}(\text{graph}(\phi_1), \text{graph}(\phi_2)) + w_2 \cdot \Delta_{\text{NodeSemantics}}(\phi_1, \phi_2) $$
	where the first term measures structural difference and the second measures semantic differences between corresponding nodes, with learned weights $w_1, w_2$. Defining these components rigorously is complex.
\end{itemize}

\section{Topological Considerations} 

Beyond metric properties, basic topological concepts are relevant:
\begin{itemize}
	\item \textbf{Neighborhoods:} The metric $d$ induces a topology on $\Phi$, where open balls $B(\phi, \epsilon) = \{\phi' \mid d(\phi, \phi') < \epsilon\}$ define neighborhoods. This formalizes the notion of "nearby" belief states.
	\item \textbf{Continuity:} Ideally, cognitive operators ($A, N_t, \Lambda, V, M$) should exhibit some form of continuity: small changes in input $\phi$ lead to small changes in output $O(\phi)$, except at specific points of discontinuity (like those potentially involved in $A_{corr}$ or $K$). This ensures smooth belief evolution in most cases. Proving continuity depends heavily on the specific implementation of $\Phi$ and the operators.
	\item \textbf{Connectedness:} Are activation basins $\mathcal{A}_a$ connected regions? Can any two points within $\mathcal{A}_a$ be joined by a path entirely within $\mathcal{A}_a$? Connectedness impacts how easily a trajectory can enter or leave a basin. Fragmentation of basins could represent complex disjunctive conditions for action readiness.
	\item \textbf{Convergence ($\omega^{(\infty)}$):} The definition of the semantic singularity $\omega^{(\infty)} = \lim_{k\rightarrow\infty} \Lambda^k(\omega)$ requires the sequence of abstractions to converge in the topology induced by $d$. This implies that repeated abstraction eventually leads to diminishing changes, settling towards a fixed point. Conditions for convergence might involve the abstraction operators $\Lambda$ being contractive mappings under $d$.
\end{itemize}
These topological notions help formalize intuitions about smooth evolution, regions of readiness, and the limiting behavior of core processes like abstraction.

\section{Semantic Flow Fields (\texorpdfstring{$F$}{F})}

Chapter~\ref{chap:SemanticGeometry} introduced the idea of modeling belief evolution $\gamma(t)$ via a semantic flow field $F: \Phi \rightarrow T\Phi$ (where $T\Phi$ is the tangent space, representing possible directions of change), such that $\frac{d\gamma}{dt} = F(\gamma(t))$.
\begin{itemize}
	\item \textbf{Vector Field Concept:} $F(\phi)$ specifies the instantaneous velocity (direction and magnitude) of belief change at state $\phi$. This field encapsulates the combined effects of internal dynamics ($N_t, D$, background $A, M$) and potentially control signals ($\pi_{regulate}$).
	\item \textbf{Challenges:} Defining and computing $F$ globally over high-dimensional $\Phi$ is extremely challenging. $F$ is likely non-stationary (changing based on context, goals, external input $X(s)$) and highly complex. In practice, $F$ serves more as a conceptual tool for understanding local dynamics and attractor landscapes.
	\item \textbf{Fixed Points:} States $\phi^*$ where $F(\phi^*) = 0$ are fixed points. Stable fixed points act as attractors (representing coherent schemas, memories, goals), while unstable ones act as repellers (e.g., points of high contradiction). The landscape of these fixed points shapes the overall dynamics.
	\item \textbf{Relation to Operators:} Specific operators correspond to specific components or modifications of $F$. Assimilation $A$ injects change, $N_t$ provides a drift towards $\Omega$, orientation control adjusts $F$ to align with axes $\vec{v}_{\omega}$.
\end{itemize}
Viewing dynamics through flow fields provides intuition about stability, convergence, and the forces shaping belief trajectories, even if $F$ itself is rarely explicitly computed.

\subsection{Illustrative Coherence Formulation (\texorpdfstring{$\kappa$}{kappa})}

Coherence ($\kappa$, Chapter~\ref{chap:CognitiveLoad}) measures internal consistency. Example formulations depend heavily on $\rho$:

\begin{itemize}
	\item \textbf{Symbolic Example ($\rho$ = Logic Fragments):} Let $\text{Contradict}(\varphi_i, \varphi_j)$ be a predicate checking for logical contradiction. A simple measure could be:
	$$ \kappa(\phi) = 1 - \frac{ |\{ (\varphi_i, \varphi_j) \in \phi \times \phi \mid \text{Contradict}(\varphi_i, \varphi_j) \}| }{ |\phi|^2 } $$
	(Normalized count of contradictions, where 1 is perfect coherence). More sophisticated measures might weight contradictions based on belief strength or relevance.
	\item \textbf{Embedding Example ($\rho$ = Embedding):} Coherence might relate to the compactness or variance of the embeddings of belief fragments $\varphi_i \in \phi$. Low variance might indicate high coherence.
	$$ \kappa(\phi) \approx \frac{1}{1 + \text{Variance}(\{\text{embed}(\varphi_i) \mid \varphi_i \in \phi\})} $$
	Alternatively, it could be the output of a learned classifier trained to predict consistency, or derived from the potential function of an energy-based model trained on coherent belief states.
\end{itemize}
These illustrate how $\kappa$ aims to capture internal consistency, but its practical calculation varies greatly.

\subsection{Illustrative Cognitive Load Formulation (\texorpdfstring{$\lambda$}{lambda} or \texorpdfstring{$L$}{L})}

Cognitive Load ($L$, Chapter~\ref{chap:CognitiveLoad}), often proxied by $\lambda$, represents the demand on processing resources. A conceptual formula might combine static and dynamic factors:
$$ L(\phi(t)) \approx c_1 \cdot \text{Complexity}(\phi(t)) + c_2 \cdot \sum_{\Sigma} \alpha_{\Sigma}(t) \cdot \text{Cost}(\Sigma) + c_3 \cdot \text{Rate}(\text{Ops}, \Delta t) $$
where:
\begin{itemize}
	\item $\text{Complexity}(\phi(t))$: Measures static complexity (e.g., number of active $\varphi_i$, structural complexity like graph density).
	\item $\sum_{\Sigma} \alpha_{\Sigma}(t) \cdot \text{Cost}(\Sigma)$: Weighted sum of active sectors, where $\text{Cost}(\Sigma)$ reflects the inherent processing demand of sector $\Sigma$.
	\item $\text{Rate}(\text{Ops}, \Delta t)$: Measures recent dynamic intensity (e.g., number/complexity of $A, M, R$ operations in the last time window $\Delta t$).
\end{itemize}
The coefficients $c_1, c_2, c_3$ are weights. This formula is purely illustrative; actual measures would need calibration based on the specific architecture $\theta$ and implementation.

\section{Gauge Symmetry (\texorpdfstring{$\sim_{\text{gauge}}$}{sim\_gauge})}

Semantic gauge equivalence $\phi_1 \sim_{gauge} \phi_2$ (Chapter~\ref{chap:SemanticGauge}) relates states that are functionally indistinguishable. Mathematically:
\begin{itemize}
	\item \textbf{Equivalence Relation:} $\sim_{gauge}$ partitions $\Phi$ into disjoint equivalence classes $[\phi]_{gauge}$.
	\item \textbf{Quotient Space:} The space of functionally distinct states can be thought of as the quotient space $\Phi / \sim_{gauge}$, where each point represents an entire gauge class. Cognitive dynamics that respect gauge symmetry effectively operate on this quotient space.
	\item \textbf{Fiber Bundle Analogy (Speculative):} In physics, gauge theories often involve fiber bundles, where the base space represents the physically relevant degrees of freedom and the fibers represent gauge-equivalent internal configurations. Analogously, $\Phi$ could potentially be viewed as a bundle where the base space is $\Phi / \sim_{gauge}$ (functional states) and the fibers are the gauge classes $[\phi]_{gauge}$ (representational variations). Operators respecting gauge symmetry act primarily on the base space, while gauge transformations move within fibers. This is a highly speculative analogy requiring rigorous development.
\end{itemize}
The mathematical perspective highlights gauge symmetry as a fundamental structural property reducing the effective complexity of the state space by identifying functionally redundant representations.

\section{Conclusion} 

Applying concepts from geometry and topology provides powerful tools for analyzing the structure and dynamics of the semantic state space $\Phi$. Viewing $\Phi$ as a structured manifold, potentially equipped with a metric $d$ and supporting flow fields $F$, allows us to model belief evolution $\gamma(t)$ in intuitive ways. Concepts like neighborhoods, continuity, connectedness, convergence, fixed points, and partitions under gauge symmetry ($\sim_{gauge}$) offer a richer language for describing cognitive processes. The illustrative formulations provided for $d$, $\kappa$, and $\lambda$ demonstrate how these abstract concepts might be concretized depending on representation $\rho$, although their precise, robust definition remains a key challenge. While rigorously formalizing these structures for complex belief representations is ongoing work, the geometric perspective itself provides valuable insights into the principles governing structured belief, regulation, and intelligent navigation within the internal landscape of meaning.
	\chapter{Implementation Strategies and Examples}
\label{app:ImplementationExamples}

\section{Introduction}

The Semantic Manifold framework provides a high-level functional architecture for structured belief, regulation, and dynamics. Realizing this architecture in a practical computational system requires bridging the abstract definitions of $\Phi$, its operators ($A, M, N_t, K, \Lambda, V, R, Q$), structures ($\Sigma, k$), and metrics ($d, \kappa, \lambda, \epsilon$) with concrete implementation paradigms.

This appendix explores potential strategies for implementing agents based on the Semantic Manifold principles, explicitly linking these strategies to the representation mode parameter $\rho$ defined within the architectural configuration $\theta$ (Chapter~\ref{chap:ParameterizedArchitectures}). We then present a simplified, illustrative example simulation run to demonstrate how these components might interact dynamically.

\section{Implementation Paradigms (Relating to \texorpdfstring{$\rho$}{rho})}

The choice of representation mode $\rho$ fundamentally shapes how the semantic manifold $\Phi$ and its associated operators are implemented.

\subsection{Symbolic Implementations ($\rho \approx$ Symbolic)}
\begin{itemize}
	\item \textbf{Representation:} Belief states $\phi$ are represented using symbolic structures like knowledge graphs (e.g., RDF, property graphs), logical propositions (e.g., first-order logic, description logics), structured databases, or frame systems. Linguistic expressions $\varphi_i$ are nodes or structured entries.
	\item \textbf{Operators:}
	\begin{itemize}
		\item $A, M$: Implemented via graph update algorithms, rule-based inference engines (adding/retracting facts/rules), constraint satisfaction solvers (for $A_{corr}$).
		\item $\Lambda, V$: Implemented using symbolic generalization techniques (e.g., inductive logic programming, conceptual clustering) or predefined hierarchical taxonomies/ontologies.
		\item $N_t$: Implemented by rules that decrease activation weights or remove nodes/edges based on time-stamps or usage counters (reflecting $a_i$).
		\item $K, K_{\Sigma}$: Implemented by direct node/edge deletion operations.
		\item $Q, R$: Implemented using graph traversal algorithms, database queries, or logical inference to find matching patterns or related concepts based on $\phi_{query}$.
	\end{itemize}
	\item \textbf{Pros/Cons:} Strong potential for interpretability, formal verification, precise logical reasoning. May struggle with ambiguity, scalability, grounding, and learning from unstructured data.
\end{itemize}

\subsection{Connectionist/Embedding Implementations (\texorpdfstring{$\rho \approx$}{rho approximately} Embedding)}

\begin{itemize}
	\item \textbf{Representation:} Belief states $\phi$ or fragments $\varphi_i$ are represented as vectors in a high-dimensional embedding space (learned via e.g., language models, graph embedding techniques). Structure might be implicit in the vector space geometry or explicitly represented via attention mechanisms or pointer networks.
	\item \textbf{Operators:} Often implemented as specialized neural network modules:
	\begin{itemize}
		\item $\Lambda, V$: Could be encoder/decoder components of autoencoders or sequence-to-sequence models.
		\item $A, M$: Might involve attention mechanisms to combine embeddings of $\phi_{current}$ and $\phi_{input}$, potentially using gated recurrent units or transformers to update the state representation. $A_{corr}$ might involve learning to identify and down-weight conflicting vector components.
		\item $N_t$: Could be modeled by multiplicative decay applied to activation vectors, potentially modulated by learned gating functions representing anchoring $a_i$.
		\item $K$: Might involve setting state vectors to zero or a null embedding. $K_{\Sigma}$ requires mechanisms to zero out specific subspace components representing sectors.
		\item $Q, R$: Implemented using similarity search (e.g., nearest neighbor search based on learned metric $d$) in the embedding space based on the query vector derived from $\phi_{active}$.
	\end{itemize}
	\item \textbf{Pros/Cons:} Naturally handles ambiguity, similarity, and learning from large datasets. Can struggle with interpretability, precise logical inference, and catastrophic forgetting. Structure is often less explicit.
\end{itemize}

\subsection{Hybrid Implementations ($\rho \approx$ Hybrid)}
\begin{itemize}
	\item \textbf{Representation:} Combines symbolic structures (e.g., a knowledge graph for core facts and relations) with learned embeddings (e.g., for representing nuanced meanings or handling unstructured input). This is often the focus of neuro-symbolic AI.
	\item \textbf{Operators:} Use a mix of techniques. Symbolic rules might handle high-level reasoning or coherence checks ($A_{corr}$), while neural modules handle perception ($X$), elaboration ($A_{elab}, V$), or similarity-based retrieval ($R$). Operators need well-defined interfaces to interact across representational modes.
	\item \textbf{Pros/Cons:} Aims to combine the strengths of both paradigms---interpretability and precision with learning and generalization. Integration complexity is a major challenge.
\end{itemize}

\subsection{Role of Large Language Models (LLMs)}
\begin{itemize}
	\item \textbf{LLMs as Components, Not the Architecture:} LLMs can serve as powerful components within a Semantic Manifold architecture ($\Phi^{[\theta]}$), particularly for implementing specific operators, rather than being the entire architecture itself. The Semantic Manifold framework defines the overall structure, dynamics, and regulation ($\theta$), which an LLM alone typically lacks (e.g., explicit persistent state $\phi$, sectors $\Sigma$, operators $N_t, K$, regulation $\kappa, \lambda, \epsilon, \theta$).
	\item \textbf{Potential Implementations using LLMs:}
	\begin{itemize}
		\item $X$: Using an LLM to parse unstructured input $s$ into initial belief fragments $\varphi_i$.
		\item $A_{elab}, V$: Using prompted generation to elaborate on input/retrieved memories or instantiate abstract concepts.
		\item $M$: Prompting an LLM with state information ($\kappa, \lambda$) and $\Phi_{introspective}$ to generate reflective meta-beliefs.
		\item $Q$: Prompting an LLM based on $\phi_{active}$ to formulate appropriate retrieval queries $\phi_{query}$.
		\item $R$: Using retrieval-augmented generation (RAG), where the LLM's generation is conditioned on documents retrieved based on $\phi_{query}$.
	\end{itemize}
	\item \textbf{Challenges:} Ensuring the LLM component respects the constraints and dynamics of the broader architecture (e.g., coherence $\kappa$, gauge $\sim_{gauge}$), managing state consistency, controlling hallucination, and integrating LLM outputs with other symbolic or numeric components remain key challenges.
\end{itemize}

\section{Illustrative Simulation Example}

This highly simplified example illustrates the interplay of operators over a few time steps. Assume a simple symbolic representation ($\rho \approx$ Symbolic) and basic sectors $\Sigma_{perc}, \Sigma_{task}, \Sigma_{mem}$. Let the persistence threshold $\delta = 0.1$. Anchoring $a_i$ defaults to 1 unless specified.

\begin{enumerate}
	\item \textbf{Time $t=0$:} Initial state near vacuum. $\phi_0 \approx \Omega$. Agent receives input $s$= "Task: Check Sensor Alpha status."
	\item \textbf{Observation Encoding ($X$):} $X(s)$ generates $\phi_{input} = \{\varphi_1: \text{"Goal: Check Sensor Alpha status"}\}$. Assign high anchor $a_1 = 10$.
	\item \textbf{Assimilation ($A$):} $\phi_1 = A(\phi_0, \phi_{input})$. State becomes $\phi_1 = \{\varphi_1\}$. Assume $A$ routes this to $\Sigma_{task}$. $d_1(0)=1$.
	\item \textbf{Query Generation ($Q$):} Goal $\varphi_1$ triggers query. $Q(\phi_1)$ generates $$\phi_{query} = \{\text{"Query: Current status of Sensor Alpha?"}\}.$$
	\item \textbf{Retrieval ($R$):} $R$ searches $\Phi_{memory}$ using $\phi_{query}$. Assume it finds a relevant, slightly decayed memory: $\phi_{retrieved} = \{\varphi_2: \text{"Sensor Alpha reported OK at T-5min"}\}$ with current persistence $d_2(t_R) = 0.6$.
	\item \textbf{Integration ($A$):} $\phi_2 = A(\phi_1, \phi_{retrieved})$. Operator $A$ integrates $\varphi_2$ (likely into $\Sigma_{mem}$ or $\Sigma_{perc}$) and potentially adds elaboration via $A_{elab}$. Let $$\phi_2 = \{\varphi_1, \varphi_2, \varphi_3: \text{"Sensor Alpha status likely OK (based on recent memory)"}\}.$$ Operator $A$ also re-anchors $\varphi_2$, increasing $a_2$ (e.g., $a_2=5$) and resetting its persistence $d_2(0)=1$. State persists: $\phi_2 = \{\varphi_1(a=10), \varphi_2(a=5), \varphi_3(a=1)\}$.
	\item \textbf{Time $t=20$ (Nullification $N_t$):} Assume no further input related to these beliefs for 20 time units. Let decay $\lambda_i \approx 0.02 / (1+a_i)$.
	\begin{itemize}
		\item $\varphi_1$: $\lambda_1 \approx 0.02/11 \approx 0.0018$. $d_1(20) \approx e^{-0.0018 \times 20} \approx 0.96 > \delta$. Belief persists.
		\item $\varphi_2$: $\lambda_2 \approx 0.02/6 \approx 0.0033$. $d_2(20) \approx e^{-0.0033 \times 20} \approx 0.94 > \delta$. Belief persists.
		\item $\varphi_3$: $\lambda_3 \approx 0.02/2 = 0.01$. $d_3(20) \approx e^{-0.01 \times 20} \approx 0.82 > \delta$. Belief persists.
	\end{itemize}
	The state remains $\phi_3 = \phi_2$, but internal persistence values have decreased.
	\item \textbf{Time $t=100$ (Nullification $N_t$):} Assume still no related input.
	\begin{itemize}
		\item $\varphi_1$: $d_1(100) \approx e^{-0.0018 \times 100} \approx 0.84 > \delta$. Persists.
		\item $\varphi_2$: $d_2(100) \approx e^{-0.0033 \times 100} \approx 0.72 > \delta$. Persists.
		\item $\varphi_3$: $d_3(100) \approx e^{-0.01 \times 100} \approx 0.37 > \delta$. Persists.
	\end{itemize}
	State is still $\phi_4 = \phi_2$.
	\item \textbf{Time $t=250$ (Nullification $N_t$):}
	\begin{itemize}
		\item $\varphi_1$: $d_1(250) \approx e^{-0.0018 \times 250} \approx 0.64 > \delta$. Persists.
		\item $\varphi_2$: $d_2(250) \approx e^{-0.0033 \times 250} \approx 0.44 > \delta$. Persists.
		\item $\varphi_3$: $d_3(250) \approx e^{-0.01 \times 250} \approx 0.08 < \delta$. Decays!
	\end{itemize}
	State becomes $\phi_5 = \{\varphi_1, \varphi_2\}$. The weakly anchored elaboration $\varphi_3$ has faded due to $N_t$. The goal $\varphi_1$ and the re-anchored memory $\varphi_2$ remain.
\end{enumerate}
This simplified trace illustrates how input ($X$), integration ($A$), memory access ($Q, R$), and decay ($N_t$) interact, mediated by anchoring ($a_i$), to shape the belief state over time.

\section{Challenges and Considerations}

Implementing the Semantic Manifold framework involves several practical challenges:
\begin{itemize}
	\item \textbf{Scalability:} Managing potentially vast belief spaces $\Phi$ and complex operator interactions efficiently requires careful architectural design and optimization, regardless of $\rho$.
	\item \textbf{Integration Across Paradigms (Hybrid $\rho$):} Ensuring seamless interaction and semantic consistency between symbolic and embedding-based components is non-trivial.
	\item \textbf{Operator Definition Detail:} The abstract definitions must be translated into concrete, robust algorithms with well-defined behavior for edge cases and complex inputs.
	\item \textbf{Parameter Tuning ($\theta$):} Even if not learning $\theta$ itself (Chapter~\ref{chap:ArchitecturalAdaptation}), selecting appropriate architectural parameters (memory size $\mu$, decay profiles $\delta_{profile}$, etc.) for a given application is crucial.
	\item \textbf{Debugging and Validation:} Understanding and debugging the behavior of agents with complex internal dynamics and structures requires sophisticated logging, visualization (Chapter~\ref{chap:SemanticGeometry}), and validation techniques.
\end{itemize}

\section{Conclusion}

The Semantic Manifold framework offers a flexible blueprint adaptable to various implementation strategies, determined largely by the choice of representation mode $\rho$ within the architectural parameter vector $\theta$. Symbolic, connectionist/embedding, and hybrid approaches each offer pathways for realizing the structured belief space $\Phi$ and its core cognitive operators, with different trade-offs regarding interpretability, learning capacity, and formal precision. Large Language Models can serve as powerful components for specific operators within this broader architecture. The illustrative example highlights the dynamic interplay between assimilation, memory retrieval, and nullification in shaping belief over time. While practical implementation faces challenges, these strategies provide concrete starting points for building agents capable of the structured, dynamic, and regulated cognition envisioned by this framework.
	\chapter{Comparison with Related Cognitive Architectures and Frameworks}
\label{app:Comparisons}

\section{Introduction}

The Semantic Manifold framework presented in this monograph offers a distinct perspective on modeling structured belief and cognition. To better understand its positioning and potential contributions, this appendix compares its core tenets with several established cognitive architectures and related concepts in artificial intelligence. We focus on key dimensions such as representation, dynamics, learning, regulation, and overall architectural philosophy, drawing parallels and highlighting differences. This comparison aims to clarify the unique aspects of the Semantic Manifold approach, as summarized in Chapter~\ref{chap:Conclusion}.

\section{Comparison with Symbolic Cognitive Architectures (e.g., SOAR, ACT-R)}

Symbolic architectures like SOAR and ACT-R have significantly advanced our understanding of rule-based reasoning, procedural skill acquisition, and modeling human performance data.
\begin{itemize}
	\item \textbf{Representation:} SOAR relies on production rules operating on working memory. ACT-R uses declarative memory chunks and production rules. The Semantic Manifold ($\Phi$) differs by positing a unified, linguistically-grounded belief space structured by geometry ($\Sigma, k$), where states are ensembles $\{\varphi_i\}$ rather than primarily chunk- or rule-based. While chunks might map to anchored beliefs $a_i$, the emphasis on linguistic structure, continuous geometric notions ($d, F$), and explicit sectors is distinct.
	\item \textbf{Dynamics:} Symbolic architectures focus on rule firing, chunk retrieval/activation dynamics, and conflict resolution. The Semantic Manifold introduces explicit operators for belief lifecycle ($A, N_t, K, D$) and navigation ($\Lambda, V, R, Q, M$) operating directly on structured belief states, alongside potentially continuous flow dynamics ($F$). $N_t$ (gradual decay) differs significantly from simple chunk decay in ACT-R by incorporating anchoring $a_i$ and structure sensitivity. Annihilation $K$ provides a more drastic reset mechanism.
	\item \textbf{Learning:} ACT-R has sophisticated mechanisms for learning chunks and tuning rule utilities. SOAR learns via chunking. The Semantic Manifold framework (Part~\ref{part:learning_and_adaptation}) accommodates learning at multiple levels (structures $d, \Sigma$; operators $A, \Lambda, \dots$; policies $\pi_{effort}, \pi_{regulate}$), potentially integrating RL and representation learning, but these are perhaps less developed currently than the learning theories in mature symbolic architectures.
	\item \textbf{Regulation:} While ACT-R includes base-level activation and utility calculations, the Semantic Manifold places stronger emphasis on explicit meta-cognitive regulation through dedicated mechanisms like Semantic Orientation ($\theta, r$), coherence monitoring ($\kappa$), load/effort management ($\lambda, \epsilon, \pi_{effort}$), and identity maintenance ($\vec{\eta}$), often operating within a reflective sector $\Sigma_{refl}$ and using operator $M$.
\end{itemize}
In summary, the Semantic Manifold offers a more geometrically inspired, linguistically grounded space with a different suite of explicit dynamic operators and a stronger focus on intrinsic regulation compared to classic symbolic architectures.

\section{Comparison with Connectionist and Hybrid Architectures (e.g., CLARION)}

Connectionist approaches emphasize distributed representations and learning, while hybrid architectures like CLARION attempt to bridge symbolic and sub-symbolic processing.
\begin{itemize}
	\item \textbf{Representation:} The Semantic Manifold allows for embedding-based representations ($\rho \approx$ Embedding) or hybrid models ($\rho \approx$ Hybrid), sharing similarities with connectionist and hybrid approaches. However, even with embeddings, the framework imposes explicit high-level structure (Sectors $\Sigma$, Abstraction $\Phi^{(k)}$) and assumes interpretable linguistic grounding ($\phi = \{\varphi_i\}$) at some level, which may be less central in purely connectionist models. CLARION explicitly separates implicit (subsymbolic) and explicit (symbolic) knowledge; the Semantic Manifold integrates various belief types within the unified space $\Phi$, distinguishing them perhaps by sector ($\Sigma$) or abstraction ($k$) rather than architectural dichotomy.
	\item \textbf{Learning:} Connectionist models excel at learning from data. The Semantic Manifold framework incorporates learning (Part~\ref{part:learning_and_adaptation}) but emphasizes learning specific operators ($\Lambda, V, A, \dots$) and structures ($d, \Sigma$) that align with its predefined architectural principles, potentially differing from end-to-end gradient descent across an entire network.
	\item \textbf{Dynamics and Regulation:} The explicit dynamic operators ($N_t, K, M$, etc.) and regulatory mechanisms (Orientation, Gauge, Load/Effort) are defining features of the Semantic Manifold, providing structured control often absent or emergent in purely connectionist systems. CLARION includes drives and meta-cognitive processes, but the specific geometric regulation (Orientation) and belief lifecycle dynamics ($N_t, K$) proposed here are distinct.
\end{itemize}
The Semantic Manifold aims to integrate the learning power suggested by connectionism within a more explicitly structured and regulated cognitive architecture grounded in linguistic semantics.

\section{Comparison with Emergentist/Embodied Frameworks (e.g., Sigma, Active Inference)}

Frameworks like Sigma (focused on integrating graphical models and cognitive science principles) or Active Inference (based on Bayesian principles and free energy minimization) share interests in integration and grounding but differ architecturally.
\begin{itemize}
	\item \textbf{Representation:} Sigma utilizes factor graphs and messages; Active Inference uses generative models and variational densities. The Semantic Manifold uses structured linguistic ensembles ($\phi=\{\varphi_i\}$) within a geometric space ($\Phi, \Sigma, k$). While mappings might exist, the core representational formalisms differ.
	\item \textbf{Dynamics:} Active Inference posits dynamics driven by minimizing prediction error (free energy). Sigma uses message passing. The Semantic Manifold defines explicit operators ($A, N_t, K, \Lambda, V, M, R, Q, D$) with distinct functional roles related to belief construction, decay, abstraction, memory, and meta-cognition.
	\item \textbf{Regulation:} Active Inference frames regulation as part of the free energy minimization process (e.g., selecting actions or adjusting priors to minimize surprise). The Semantic Manifold proposes specific regulatory mechanisms tied to its geometry and structure (Orientation based on Null Tower axes, coherence $\kappa$, load $\lambda$, effort $\epsilon$, identity $\vec{\eta}$).
	\item \textbf{Emphasis:} The Semantic Manifold places strong emphasis on the internal structure and geometry of belief, the lifecycle dynamics ($N_t, K$), and explicit meta-cognitive operators ($M$) and regulatory constructs (Orientation, Gauge, Identity), offering a different explanatory vocabulary compared to probabilistic inference or message passing frameworks.
\end{itemize}
While sharing goals of unified cognition, the Semantic Manifold proposes a unique set of architectural primitives centered on geometric structure and explicit dynamic/regulatory operators over linguistic beliefs.

\section{Relation to Modern AI Concepts (LLMs, Knowledge Graphs)}

It is crucial to distinguish the Semantic Manifold framework from simply using components like Large Language Models (LLMs) or Knowledge Graphs (KGs).
\begin{itemize}
	\item \textbf{Framework vs. Component:} LLMs offer powerful capabilities for processing and generating language; KGs provide structured knowledge storage. The Semantic Manifold is an architecture that can potentially leverage LLMs or KGs as implementation components (as discussed in Appendix~\ref{app:ImplementationExamples}, e.g., for operators $A, V, R, Q$ or for representing parts of $\Phi$) but provides an overarching structure and set of dynamics they inherently lack. An LLM alone is not typically considered a full cognitive architecture with persistent belief states ($\phi$), explicit decay ($N_t$), reflective regulation ($\Sigma_{refl}, M$), or stable identity ($\vec{\eta}$) in the way defined here.
	\item \textbf{Structured Dynamics and Regulation:} The Semantic Manifold defines specific operators ($N_t, K, \Lambda, V, M, D$) and regulatory mechanisms ($\kappa, \lambda, \epsilon$, Orientation, Gauge, Identity) that govern the evolution and stability of belief. These are generally absent from standard LLM or KG formalisms, which focus on generation/inference or knowledge representation, respectively.
	\item \textbf{Parameterization ($\Phi^{[\theta]}$):} The framework allows specifying diverse architectures ($\theta$) that might use LLMs/KGs internally but differ significantly in their designed memory ($\mu$), decay ($\delta_{profile}$), identity ($\eta_{type}$), etc.
\end{itemize}
The Semantic Manifold framework aims to provide the architectural context---the structured space, dynamics, and regulation---within which components like LLMs or KGs can be integrated to build more coherent, stable, and self-aware cognitive systems.

\section{Positioning and Unique Contributions}

The Semantic Manifold framework, especially in its parameterized form $\Phi^{[\theta]}$, offers a synthesis and extension of previous approaches:
\begin{itemize}
	\item From symbolic AI, it inherits compositional structure and the potential for interpretable reasoning.
	\item From connectionism, it can adopt continuous representations ($\rho$) and leverage powerful learning mechanisms for implementation.
	\item From predictive coding, it borrows concepts of hierarchical abstraction and internal regulation based on internal states.
	\item From cognitive science, it integrates models of structured memory, introspection, coherence maintenance, and identity.
\end{itemize}
Its key novelty lies in the explicit formalization of cognition as a regulated trajectory ($\gamma(t)$) within a structured, agent-specific semantic space ($\Phi^{[\theta]}$), governed by defined operators ($A, N_t, K, \Lambda, V, M, R, Q, D$) and internal regulatory mechanisms (Orientation, Gauge, coherence $\kappa$, load $\lambda$, effort $\epsilon$, identity $\vec{\eta}$). It proposes a distinct cognitive architecture capable of coherence, drift, recovery, and self-alignment over time, providing a unique geometric and dynamic perspective on structured belief.
	\chapter{Extended Philosophical and Cognitive Motivations}
\label{app:ExtendedMotivations}

\section{Introduction}

Chapters \ref{chap:PhilosophicalMotivations}, \ref{chap:PsychologicalMotivations}, and \ref{chap:NeuroscientificMotivations} briefly outlined the grounding of the Semantic Manifold framework in various disciplines. This appendix offers optional, extended reflections on some of these foundational connections for readers interested in the deeper conceptual underpinnings. It does not introduce new core components but elaborates on the philosophical and cognitive context surrounding key architectural choices.

\section{On the Linguistic Basis of Belief (\texorpdfstring{$\phi = \{\varphi_i\}$}{phi = {varphi\_i}})}

The framework's central assumption that belief states $\phi$ are structured ensembles of linguistic expressions $\varphi_i$ aligns with, but is not identical to, philosophical traditions like Fodor's Language of Thought (LoT).
\begin{itemize}
	\item \textbf{Compositionality and Productivity:} Like LoT, this assumption naturally accounts for the compositional nature of thought (complex beliefs built from simpler parts) and its productivity (the ability to generate novel beliefs). Representing beliefs linguistically provides a combinatorial syntax and semantics.
	\item \textbf{Interpretability:} Grounding $\phi$ in language makes internal states potentially interpretable, both for external observers/designers and for the agent itself via meta-cognitive processes ($M$, operating on $\Sigma_{refl}$). This contrasts with purely sub-symbolic representations where meaning is often opaque.
	\item \textbf{Differences from Classical LoT:} Unlike some interpretations of LoT focusing on purely abstract, innate symbols, the $\varphi_i$ here are framed as natural language expressions. Their meaning arises not just from formal structure but also from grounding (Chapter~\ref{chap:GroundingSemanticBelief}) via sensorimotor input ($X$), linguistic context ($A_{text}$ from corpora), simulation, and social interaction. Furthermore, the embedding within a dynamic, geometric manifold $\Phi$ (governed by $A, N_t, K, \Lambda, V, F$, etc.) moves beyond the purely symbolic manipulation often associated with classical LoT.
	\item \textbf{Scope Limitation?:} Does grounding belief in language limit the framework to language-using agents? While the current formalization emphasizes linguistic $\varphi_i$, the underlying principles of a structured state space $\Phi$, operators, geometry, and regulation might potentially be generalized to non-linguistic representations (e.g., grounded sensorimotor schemas, affective states), though this remains a significant extension.
\end{itemize}
The linguistic basis provides a powerful blend of structure, interpretability, and generativity, serving as a productive, though not necessarily exclusive, foundation for modeling complex belief.

\section{On Semantic Geometry and Conceptual Spaces}

The proposal to view $\Phi$ geometrically (Chapter~\ref{chap:SemanticGeometry}), structured by sectors $\Sigma$ and abstraction levels $\Phi^{(k)}$ and potentially endowed with a metric $d$, resonates with theories of conceptual spaces (e.g., Gardenfors).
\begin{itemize}
	\item \textbf{Geometry as Meaning:} Conceptual space theories propose that concepts are represented as regions in multi-dimensional quality spaces, and meaning arises from geometric relationships (distance, convexity, neighborhood). Our Semantic Geometry shares this core idea, defining meaning partially through position $(\Sigma, k)$ and proximity ($d$) within $\Phi$.
	\item \textbf{Structure Beyond Quality Dimensions:} While conceptual spaces often focus on low-level perceptual dimensions, the Semantic Manifold's geometry incorporates functional sectors ($\Sigma$) and abstraction hierarchies ($\Phi^{(k)}$) as primary organizing axes, providing structure relevant to higher-level cognition (planning, reflection, narrative) beyond direct perception.
	\item \textbf{Dynamic Geometry:} Crucially, the Semantic Manifold emphasizes the dynamics within this space---belief evolution as trajectories $\gamma(t)$, governed by flow fields $F$ and operators $A, N_t, K$, etc. It is not just a static map but an active landscape. The metric $d$ and even the sectoral structure $\Sigma$ might themselves be learned and adapted (Chapter~\ref{chap:LearningSemanticStructures}).
\end{itemize}
The geometric perspective provides a unifying language for structure and dynamics, connecting low-level representation to high-level reasoning through concepts of position, distance, and motion within the belief space.

\section{On the Null Tower and Generative Abstraction}

The Null Tower (Chapter~\ref{chap:NullTower}) offers a specific, generative account of how structured cognitive potential can emerge from the epistemic vacuum $\Omega$.
\begin{itemize}
	\item \textbf{Emergence from Potentiality:} Starting from irreducible null states $\omega \in \Omega^{(0)}$ (representing pure potential or minimal architectural bias), the recursive application of abstraction $\Lambda$ constructs layered representations $\omega^{(k)}$. This models cognition building upon itself, generating complexity from minimal origins, echoing constructivist or developmental perspectives.
	\item \textbf{Abstraction as Compression and Form Extraction:} The $\Lambda$ operator is viewed not just as discarding information, but as extracting invariant structures or essential forms that persist across levels. The trajectory towards the singularity $\omega^{(\infty)}$ represents a path towards pure semantic form, devoid of specific content but capturing the essence of the abstraction process itself.
	\item \textbf{Grounding Structure Internally:} Unlike approaches relying solely on external input ($X$) to structure $\Phi$, the Null Tower provides an internal source of structure and orientation (via epistemic axes $[\omega \rightarrow \omega^{(\infty)}]$ used in Chapter~\ref{chap:SemanticOrientation}), potentially explaining innate cognitive biases or foundational conceptual frameworks.
	\item \textbf{Relation to Mathematical Concepts:} The iterative process $\omega^{(k+1)} = \Lambda(\omega^{(k)})$ and the limit $\omega^{(\infty)}$ conceptually link to dynamical systems theory (fixed points, attractors) and potentially category theory (limits, colimits, functorial relationships between layers $\Phi^{(k)}$ if $\Lambda$ has suitable properties).
\end{itemize}
The Null Tower provides a principled, though abstract, mechanism for the genesis of cognitive structure and internal reference frames.

\section{On Gauge Equivalence and Functionalism}

Semantic Gauge Equivalence ($\sim_{gauge}$, Chapter~\ref{chap:SemanticGauge}) formally captures the idea that functional role, not specific representational form, is primary for certain aspects of cognition.
\begin{itemize}
	\item \textbf{Alignment with Functionalism:} This resonates strongly with functionalist philosophies of mind, where mental states are defined by their causal relations to inputs, outputs, and other mental states, rather than their physical or structural realization. Gauge equivalence defines epistemic states by their behavior under the framework's operators ($A, N_t, K, \pi, \dots$).
	\item \textbf{Representational Flexibility:} It allows agents significant internal flexibility. They can reformat beliefs, use different phrasings $\varphi_i$, or even different representational modes $\rho$ (to some extent), as long as functional equivalence is preserved at the level relevant for interaction and core dynamics defined by $\theta$.
	\item \textbf{Identity and Continuity:} As discussed in Chapter~\ref{chap:EpistemicIdentity}, gauge theory suggests that epistemic identity $\vec{\eta}$ might persist even if the underlying belief structures $\phi$ change significantly, provided the agent remains within the same functional equivalence class $[\phi]_{gauge}$ relevant to its identity-defining operations. Identity becomes functional continuity.
	\item \textbf{Limitations:} Gauge equivalence depends on the set of operators and contexts considered. Two states might be gauge-equivalent for simple operations but distinguishable by more sensitive introspective processes ($M$) or finer-grained actions. The scope of $\approx_{behav}$ in the definition is crucial.
\end{itemize}
Gauge theory provides a formal way to handle the "many-to-one" mapping between internal representation and external function.

\section{On Memory, Dynamics, and Constructivism}

The framework's approach to belief dynamics (Part~\ref{part:epistemic_dynamics}) and memory (Part~\ref{part:semantic_memory}) aligns with constructivist views of cognition.
\begin{itemize}
	\item \textbf{Active Construction ($A$):} Assimilation is not passive storage but an active process of integrating input into existing structures $\phi$, involving elaboration and potentially revision ($A_{corr}$), mirroring Bartlett's ideas of memory as reconstruction based on schemas.
	\item \textbf{Dynamic Memory:} Memory is not a static store but part of the dynamic space $\Phi$, subject to decay ($N_t$) based on anchoring ($a_i$) and reinforcement through retrieval and re-assimilation ($Q \rightarrow R \rightarrow A$), aligning with usage-based theories of memory strength.
	\item \textbf{Forgetting as Functional ($N_t, K$):} Both gradual nullification ($N_t$) and abrupt annihilation ($K$) are presented not just as failures but as potentially functional aspects of managing cognitive load ($\lambda$) and maintaining coherence ($\kappa$), reflecting adaptive forgetting concepts in psychology.
	\item \textbf{Emergence from Dynamics:} Structures like stable beliefs (high $a_i$), attractor states (stable points of $F$), or even identity $\vec{\eta}$ emerge from the ongoing interplay of dynamic operators rather than being solely pre-specified.
\end{itemize}
This dynamic, constructivist perspective contrasts with models treating memory as a separate, static database or relying solely on retrieval strength without explicit decay/revision operators.

\section{On Meta-Cognition and Self-Models}

The emphasis on meta-cognition (Part~\ref{part:meta_cognition}: $M$, Introspection, Trajectory Awareness, Load $\lambda$, Effort $\epsilon$, Allocation $\pi_{effort}$) aligns with research highlighting the importance of self-monitoring and self-control for higher intelligence.
\begin{itemize}
	\item \textbf{Explicit Self-Representation ($\Sigma_{refl}$):} Dedicating a sector to reflective processes and meta-beliefs allows for explicit reasoning about the self and internal states.
	\item \textbf{Internal Sensing ($I_j \rightarrow M$):} Formalizing introspection as internal sensing feeding into Meta-Assimilation provides a mechanism for self-awareness to influence subsequent processing.
	\item \textbf{Resource Management ($\lambda, \epsilon, \pi_{effort}$):} Explicitly modeling load and effort acknowledges cognitive resource limits and enables the modeling of attentional control and strategic resource allocation, crucial aspects often abstracted away in simpler models.
	\item \textbf{Recursive Potential:} The framework naturally allows for recursion (beliefs about beliefs, monitoring of monitoring processes), supporting potentially deep levels of self-awareness, limited primarily by complexity and computational resources.
\end{itemize}
By incorporating these meta-cognitive components explicitly, the framework aims to model agents capable of more robust, adaptive, and potentially insightful self-regulation compared to purely object-level reasoners.
	
	\backmatter
	\bibliographystyle{amsplain}
	\bibliography{references}
	\nocite{*}
	
\end{document}